\documentclass[twocolumn,runningheads]{svjour3}

\usepackage{amssymb}
\usepackage{graphicx}

\usepackage{amsmath,amssymb,amscd} 
\def\bbbr{{\mathbb R}} 
\def\bbbz{{\mathbb Z}}

\usepackage{url}

\usepackage{xcolor}
\usepackage{natbib}
\usepackage{graphicx}
\usepackage{subcaption} 
\usepackage{float}
\usepackage{comment} 

\def\final{\scriptsize\mbox{final}}
\def\centre{\scriptsize\mbox{centre}}
\def\maximum{\scriptsize\mbox{max}}
\def\avg{\scriptsize\mbox{average}}
\def\logsumexp{\scriptsize\mbox{logsumexp}}
\def\spatmax{\scriptsize\mbox{spat-max}}
\def\sample{\scriptsize\mbox{sample}}
\def\integrated{\scriptsize\mbox{int}}
\def\norm{\scriptsize\mbox{norm}}

\journalname{arXiv preprint manuscript No.}

\begin{document}


\title{Scale generalisation properties of extended scale-covariant and scale-invariant Gaussian derivative networks on image datasets with spatial scaling variations%
\thanks{The support from the Swedish Research Council (contracts 2018-03586 and 2022-02969) is gratefully acknowledged.}}

\titlerunning{Scale generalisation properties of extended Gaussian derivative networks}

\author{Andrzej Perzanowski and Tony Lindeberg}

\institute{Andrzej Perzanowski (corresponding author)\\ amper@kth.se \medskip \\           
                Tony Lindeberg\\ tony@kth.se \medskip \\
                Computational Brain Science Lab,
                Division of Computational Science and Technology,
                KTH Royal Institute of Technology,
                SE-100 44 Stockholm, Sweden.}

              \date{}
              
\maketitle

\begin{abstract}
Due to the variabilities in image structures caused by perspective scaling transformations, it is essential for deep networks to have an ability to generalise to scales not seen during training. This paper presents an in-depth analysis of the scale generalisation properties of the scale-covariant and scale-invariant Gaussian derivative networks, complemented with both conceptual and algorithmic extensions. For this purpose, Gaussian derivative networks (GaussDerNets) are evaluated on new rescaled versions of the Fashion-MNIST and the CIFAR-10 datasets, with spatial scaling variations over a factor of 4 in the testing data, that are not present in the training data. Additionally, evaluations on the previously existing STIR datasets show that the GaussDerNets achieve better scale generalisation than previously reported for these datasets for other types of deep networks.

We first experimentally demonstrate that the GaussDerNets have quite good scale generalisation properties on the new datasets, and that average pooling of feature responses over scales may sometimes also lead to better results than the previously used approach of max pooling over scales.
Then, we demonstrate that using a spatial max pooling mechanism after the final layer enables localisation of non-centred objects in image domain, with maintained scale generalisation properties. We also show that regularisation during training, by applying dropout across the scale channels, referred to as scale-channel dropout, improves both the performance and the scale generalisation.

In additional ablation studies, we show that, for the rescaled CIFAR-10 dataset, basing the layers in the GaussDerNets on derivatives up to order three leads to better performance and scale generalisation for coarser scales, whereas networks based on derivatives up to order two achieve better scale generalisation for finer scales. Moreover, we demonstrate that discretisations of GaussDerNets based on the discrete analogue of the Gaussian kernel in combination with central difference operators perform best or among the best, compared to a set of other discrete approximations of the Gaussian derivative kernels. Furthermore, we show that the improvement in performance obtained by learning the scale values of the Gaussian derivatives, as opposed to using the previously proposed choice of a fixed logarithmic distribution of the scale levels, is usually only minor, thus supporting the previously postulated choice of using a logarithmic distribution as a very reasonable prior.

Finally, by visualising the activation maps and the learned receptive fields, we demonstrate that the GaussDerNets have very good explainability properties.

\end{abstract}

\section{Introduction}
Differences in scale are common in natural image data, given (i)~the many possible ways by which objects in the world can be seen from different distances by a visual observer, resulting in spatial scaling transformations when mapped to the 2-D image domain, as well as due to the fact that (ii)~similarly looking objects may also be of different 3-D size in the world. This fact has prompted extensive research into constructing deep learning architectures that are covariant (also referred to as equivariant) to variations in scale, as developed by Worrall and Welling (\citeyear{WorWel19-NeuroIPS}), Bekkers (\citeyear{Bek20-ICLR}), Sosnovik et al. (\citeyear{SosSzmSme20-ICLR}, \citeyear{SosMosSme21-BMVC}),  Zhu et al. (\citeyear{ZhuQiuCalSapChe22-JMLR}), Jansson and Lindeberg (\citeyear{JanLin22-JMIV}), Lindeberg (\citeyear{Lin22-JMIV}), Zhan et al. (\citeyear{ZhaSunLi22-ICCRE}) and Wimmer et al. (\citeyear{WimGolDa23-arXiv}).
Because of the scale-covariant properties of these networks, image structures of different size in the image domain are treated in a structurally similar manner, which enables better processing of objects at varying scales, compared to models that are not equipped with such covariance properties.

Additionally, a fundamental problem, that needs to be addressed in the context of processing real-world image data, is endowing deep learning architectures with the ability to perform testing at scales that have not been seen during training, referred to as \textit{scale generalisation}. In general, if a regular, not scale-covariant or scale-invariant, deep network is subject to testing data with variabilities that are not spanned by the training data, the performance can be very poor. A theoretical explanation for this is that training of deep networks by stochastic gradient descent merely corresponds to interpolation between the training data, with very poor abilities to perform extrapolation, see Domingos (\citeyear{Dom20-arXiv}) and Courtois et al. (\citeyear{CurMorAri23-arXiv}) for a detailed analysis. 

\subsection{Using scale priors to achieve scale generalisation in deep networks}

By adding prior information to the deep networks concerning the variabilities of the image data under natural image transformations, in our case spatial scaling transformations, we can obtain provable scale extrapolation or scale generalisation properties, that may substantially improve the performance, when deep networks are subject to testing data outside the domain spanned by the training data.

In initial work, Lindeberg (\citeyear{Lin20-JMIV}) proposed a general family of hierarchical deep networks, based on scale-normalised scale-space operations coupled in cascade, that were shown to be provably scale-covariant, and experimentally demonstrated that, a particular subfamily of these networks, constructed to mimic some of the qualitative properties of complex cells in biological vision, had a capability to perform classification at scales not spanned by the training data, and in this way, thus demonstrating basic scale generalisation properties (see Figures~15 and~16 in Lindeberg (\citeyear{Lin20-JMIV})).
This idea of scale generalisation was then furthered by Jansson and Lindeberg (\citeyear{JanLin22-JMIV}) in terms of scale-covariant foveated networks and by Lindeberg (\citeyear{Lin22-JMIV}) in terms of scale-covariant Gaussian derivative networks (GaussDerNets). This was in turn followed by closely related work by Sangalli et al. (\citeyear{SanBluVelAng22-BMVC}) and Yang et al. (\citeyear{YanDasMah23-arXiv}) on scale-covariant U-Nets, as well as by Barisin et al. (\citeyear{BarAngSchRed24-SIIMS, BarSchRed24-JMIV}) based on scale-invariant Riesz networks. For other closely related work aiming at scale generalisation, see Altstidl et al. (\citeyear{AltNguSchKoeMutEskZan23-IJCNN}) and Velasco-Forero (\citeyear{Velasco2023-ICGSI}).

\subsection{Gaussian derivative networks}

In this paper, we will follow the framework of scale-covariant and scale-invariant GaussDerNets in Lindeberg (\citeyear{Lin22-JMIV}), where the layers are parameterised arbitrary linear combinations of scale-normalised Gaussian derivatives, a structure inspired by scale-space theory, see Koenderink and van Doorn (\citeyear{KoeDoo92-PAMI}) and Lindeberg (\citeyear{Lin93-Dis, Lin21-Heliyon}) for theoretical background information. These layers are coupled in cascade, with point-wise non-linearities inserted in between, and the model itself consists of multiple scale channels with shared parameters. The underlying idea of this architecture is to provide complementary a priori knowledge, in terms of scale covariance and scale invariance, which has been shown to enable scale generalisation. Gaussian derivatives have also been used as primitives in deep networks by Jacobsen et al. (\citeyear{JacGemLouSme16-CVPR}), Pintea et al. (\citeyear{PinTomGoeLooGem21-IP}) and Penaud-Polge et al. (\citeyear{PenVelAng22-ICIP}), although not explicitly aiming at scale covariance or scale invariance, see also Gavilima-Pilataxi and Ibarra-Fiallo (\citeyear{GavIva23-ICPRS}). More generally, by applying clustering to the learned receptive fields in depth-separable CNNs, Babaiee et al. (\citeyear{BabKiaRusGro24-ICLR}) have demonstrated that a substantial fraction of those learned receptive fields have spatial patterns that agree quite well with receptive field shapes that can be generated within the Gaussian derivative paradigm.

\subsection{Related deep learning approaches}

Other types of deep networks based on mathematical models of the filter weights have been presented, with models by Luan et al. (\citeyear{LuaCheZhaHanLiu18-PAMI}) and Yuan et al. (\citeyear{YuWaZh22-PR}) based on Gabor functions, by Sangalli et al. (\citeyear{SangaliBluVel22--ICIP}) in terms of differential invariants defined from Gaussian derivatives, by Paz et al. (\citeyear{Paz23-CG}) based on sine waves, and by Penaud-Polge et al. (\citeyear{PenVelAn23-SSVM}) based on elementary symmetric polynomials over Gaussian derivatives. Structured models of deep networks have also been defined in terms of solutions of partial differential equations by Ruthotto and Haber (\citeyear{RutHab20-JMIV}), Shen et al. (\citeyear{SheHeLinMa20-ICML}), Smets et al. (\citeyear{SmePorBekDui23-JMIV}) and Bellaard et al. (\citeyear{BellSak-2025-JMIV}). In that context, it should be noted that the Gaussian smoothing operation, that we make use of in the GaussDerNets, corresponds to solutions of the diffusion equation.

There are also structurally very close relationships to the area of geometric deep learning, see Bronstein et al. (\citeyear{BroBruCohVel21-arXiv}) and Gerken et al. (\citeyear{GerAroCarLinOhlPetPer23-AIRev}), where similar types of symmetry properties are applied to deep networks as constituting foundations for the generalised axiomatic scale-space framework developed by Lindeberg (\citeyear{Lin13-BICY, Lin13-ImPhys, Lin21-Heliyon, Lin24-arXiv-UnifiedJointCovProps}), that underlies the formulation of the scale-covariant and scale-invariant Gaussian derivative networks studied in this treatment.

Concerning deep networks, that in different ways handle scale information, there also exist approaches that are not necessarily based on a formalism of covariant image operations. Early work aiming towards scale-invariant deep networks was done by Xu et al. (\citeyear{XuXiaZhaYanZha14-arXiv}), Kanazawa et al. (\citeyear{KanShaJac14-arXiv}), Marcos et al. (\citeyear{MarKelLobTui18-arXiv}) and Ghosh and Gupta (\citeyear{GhoGup19-arXiv}). The notion of combining Gaussian filters with learned filters to optimise and adapt the receptive fields in the network was proposed by Shelhamer et al. (\citeyear{SheWanDar19-arXiv}). Chen et al. (\citeyear{CheFanXuYanKalRohYanFen19-ICCV}) proposed to reduce the redundancy in CNNs by octave convolution. Xu et al. (\citeyear{XuVenSun20-CVPR}) proposed the notion of blur integrated gradients to determine at what scale a network recognises an object. For the task of object detection, Wang et al. (\citeyear{WanZhaYuFenZha20-CVPR}) proposed to use scale-equalising pyramid convolution, while Singh et al. (\citeyear{SinNajShaDav22-PAMI}) made use of scale-normalised image pyramids. Fu et al. (\citeyear{FuZhaLiuRonWu22-CircSystVidTech}) developed a Scale-Net, to reduce scale differences for large scale image matching.

\subsection{Scope of this paper}
Previous work introducing the GaussDerNets (Lindeberg \citeyear{Lin22-JMIV}) constituted a proof of concept, demonstrating that scale covariance and scale invariance can be achieved by coupling scale-space operations in cascade in a deep network. This is a result that builds on the scale-covariant properties of the Gaussian scale-space, previously explored for constructing scale-covariant and scale-invariant feature detectors in classical computer vision, see Lindeberg (\citeyear{Lin98-IJCV, Lin97-IJCV,Lin21-EncCompVis}), Bretzner and Lindeberg (\citeyear{BL97-CVIU}), Lowe (\citeyear{Low04-IJCV}), and Bay et al. (\citeyear{BayEssTuyGoo08-CVIU}) for examples of such approaches. 
The ability of the model to generalise to scales not seen in the training data was confirmed experimentally on the MNIST Large Scale dataset (Jansson and Lindeberg \citeyear{JanLin20-MNISTLargeScale}).

What remains to be explored scientifically with regard to these previous results, however, is how such scale generalisation properties would transfer to image data sets that contain more complex image structures. Currently, there is a lack of good data sets for such purposes, to comprise scaling variations over sufficiently large spans of spatial scaling factors. For this reason, we will in this paper introduce new rescaled versions of the Fashion-MNIST and the CIFAR-10 datasets, comprising spatial scaling factors that span a variability over a factor of 4, in order to be able to evaluate the scale generalisation properties of deep networks to systematic scaling variations, over reasonably wide ranges of spatial scaling factors. For the Fashion-MNIST dataset, we also introduce a second variant of the dataset, that in addition to the aforementioned spatial scaling, contains spatial translations of the objects within the images, to be able to simultaneously evaluate the scale generalisation properties of the model, as well as how the model localises image structures in a larger search area in the image domain.

With regard to the formulation of scale-covariant and scale-invariant GaussDerNets, for which a proof-of-concept was presented of their ability to perform scale generalisation, to make it possible to apply these networks at spatial scales not spanned by the training data, we will also present a number of conceptual and algorithmic extensions to this family of deep networks, to enable better performance on image data sets that contain more complex types of image structures, than used in the previously reported experimental results on the MNIST Large Scale dataset.

First of all, we will complement the previous use of max pooling over the scale channels, as the scale selection mechanism for combining information in the different scale channels, with average pooling over the scale channels as the main scale selection mechanism, which will be demonstrated to sometimes lead to better accuracy and scale generalisation for the networks, by making more effective use of the information in the different scale channels, than basing the classification on the information from a single scale channel only.

Then, to be able to handle datasets, for which the objects are not a priori centered in the image domain, we will introduce a spatial max pooling step over the image domain, to perform spatial selection in each image, in close combination with the previously described scale selection mechanism over the scale channels. Additionally, we will make use of a number of complementary regularisation mechanisms in the training stage, which will also turn out to be rather helpful, for improving the performance of the GaussDerNets on more complex datasets. 

After these extensions to the GaussDerNets have been introduced, we will then apply the resulting extended Gaussian derivative networks to the new rescaled Fashion-MNIST datasets (with or without random object translation) and to the new rescaled CIFAR-10 dataset, as well as to the recently introduced STIR dataset by Altstidl et al. (\citeyear{AltNguSchKoeMutEskZan23-IJCNN}), also comprising scaling variations, while for a substantially lower number of images in that dataset than in the number of images in the rescaled Fashion-MNIST dataset or in the rescaled CIFAR-10 dataset. The main goal of these experiments, to be reported below, is to evaluate the performance of the GaussDerNets, when exposed to the scale generalisation task of performing classification into object classes for testing data for different sizes in the image domain than are spanned by the training data.

In these ways, we will in this work demonstrate how it is possible to improve the performance of deep networks on image data subjected to natural image transformations, by integrating provable scale-covariant mechanisms into the network architecture, to be able to handle testing data that is sampled from a distribution that is not spanned by the training data.

\subsection{Contributions and novelty}
Beyond extending the applicability of scale generalisation of Gaussian derivative networks to more complex datasets, and adding a spatial max pooling mechanism to such networks, this work will also present a number of other technical contributions to GaussDerNets, to increase their performance, as well as be accompanied with visualisations, to demonstrate the explainability properties of these networks. In summary, the main contributions of the paper are as follows:

\begin{itemize}    
    \item[\textbullet] We complement the use of max pooling over scales with average pooling over scales, which is shown to enable use of information over multiple scales in a more efficient manner than selecting image information from a single scale only, as done with max pooling over scales. We show that GaussDerNets using either of these methods have similar scale selection properties to classical methods based on scale-normalised derivatives. \\

    \item[\textbullet] We extend the training process of the GaussDerNets with data augmentation and specialised regularisation techniques, in terms of scale-channel dropout and cutout. \\

    \item[\textbullet] We complement the previous use of Gaussian derivatives up to order 2 by Gaussian derivatives up to order 3, and demonstrate that Gaussian derivatives up to order 2 perform very well in comparison to Gaussian derivatives up to order 3.\\

    \item[\textbullet] In the initial experiments with GaussDerNets, the studied objects were centered in the image domain, implying that it was sufficient to select the central pixel as an object localisation method. In this work, we complement the central pixel extraction with a (global) spatial max pooling stage, for the purpose of handling varying spatial locations of objects. More generally, if there may be multiple objects in the same image, this step could, in turn, be replaced by an object localisation step, based on multiple spatial maxima. \\

    \item[\textbullet] We investigate the scale-covariant properties of the GaussDerNet, and show that the scale transformation properties of the network enable evaluation of image data at scales not seen in the training data. 
    Additionally, in Appendix~A.1 of the supplementary material, we show that because of weight sharing, these scale transformation properties enable the multi-scale-channel network to be based on weights transferred from a trained single-scale-channel network. However, we demonstrate that much better scale generalisation is achieved by optimising the entire multi-scale-channel network during the training stage. \\
    
    \item[\textbullet] We demonstrate that the GaussDerNet can achieve good performance when constructed using different discretisation methods, namely by approximating the Gaussian derivatives by either (i)~the discrete analogue of the Gaussian kernel combined with central difference operators, (ii)~sampled or integrated Gaussian derivatives, or (iii)~the integrated or sampled or normalised sampled Gaussian kernel, combined with central difference operators. \\ 

    \item[\textbullet] We report experimental results of learning the scale levels in the GaussDerNet during training, as opposed to the original approach of using fixed scale levels according to a logarithmic distribution, demonstrating that (i)~the original approach of distributing the scale levels according to a logarithmic distribution performs quite well, (ii)~in some cases learning of the scale levels may lead to a distribution of scale levels very similar to a logarithmic distribution, and (iii)~in some cases a minor improvement in performance can be obtained by performing explicit learning of the scale levels.
 
\end{itemize}
We introduce new versions of the Fashion-MNIST and CIFAR-10 datasets subject to scaling variations over factors between $1/2$ and 2 and experimentally investigate the effects of our proposed network mechanisms on these new datasets. Scale generalisation of the GaussDerNets on the STIR datasets is also evaluated. Additionally, we visualise how the GaussDerNets function, by inspecting its learned filters and the activation maps. Because of our deliberate choice of not including any spatial subsampling in the GaussDerNets, we demonstrate that the activation maps are visually very interpretable, in this way demonstrating the explainability properties of these networks.

\section{Review of Gaussian derivative networks}
\label{sec-enhanced-gaussdernets}

In this section, we will first give an overview of the formal definition of scale-covariant and scale-invariant GaussDerNets, complemented with a few technical details that were not included in the original publication (Lindeberg \citeyear{Lin22-JMIV}). Then, in
the following Section~\ref{sec-alg-improv} we will introduce a set of extensions of this notion, which lead to better performance on more challenging data sets than used in the initial proof-of-concept work.


Essentially, the notion of Gaussian derivative networks constitutes a hybrid approach between deep learning and scale-space theory, by defining the layers in a deep network from linear combinations of Gaussian derivatives. Specifically, based on the scale-covariant properties of the underlying Gaussian kernels, the GaussDerNets do therefore become provably scale-covariant, and can also be made scale invariant, by performing max pooling or average pooling over scale.

\subsection{Gaussian derivative layer}

Given input data $f$, which could be either an input image
$f: \mathbb{R}^{2} \to \mathbb{R}$ or the output from
a previous layer, its scale-space representation
$L: \mathbb{R}^{2} \times \mathbb{R}_{+} \to \mathbb{R}$
is defined by convolving (smoothing) the input with the Gaussian kernel
$g: \mathbb{R}^{2} \times \mathbb{R}_{+} \to \mathbb{R}$
with scale parameter $\sigma \in \mathbb{R}_+$:
\begin{equation} 
\label{eq:scsp-representation}
    L(\cdot, \cdot; \ \sigma) = g(\cdot,\cdot; \ \sigma) * f(\cdot, \cdot).
\end{equation}
Here, the Gaussian kernel $g(x, y;\; \sigma)$ in two dimensions is given by 
\begin{equation}
\label{eq:gaussian-kernel}
g(x, y;\; \sigma) = \frac{1}{2 \pi \sigma^2} e^{-(x^2 + y^2)/2\sigma^2}, 
\end{equation}
from which Gaussian derivatives are defined as
\begin{equation}
\label{eq:gaussian-derivative}
g_{x^{\alpha} y^{\beta}}(x, y;\; \sigma)
= \partial_{x^{\alpha} y^{\beta}} g(x, y;\; \sigma).
\end{equation}
Scale-normalised derivatives are then defined according to Lindeberg (\citeyear{Lin98-IJCV, Lin97-IJCV}) as
\begin{equation} 
\label{eq:scale-normalised-derivatives}
\begin{aligned}
    \partial_{\xi} = \sigma^{\gamma} \, \partial_{x}, \\
    \partial_{\eta} = \sigma^{\gamma} \, \partial_{y},
\end{aligned}
\end{equation}
where we henceforth for simplicity choose the most scale-invariant choice of setting the scale normalisation parameter $\gamma = 1$. Based on this notion, a Gaussian derivative layer of order 2 is then defined as a linear combination of such scale-normalised Gaussian derivatives up to order 2 (Lindeberg \citeyear{Lin22-JMIV} Equation~(6))
\begin{multline}
  \label{eq:2-Jet}
  J_{2,\sigma}(f(\cdot, \cdot)) = C_0 + C_x \, L_{\xi}(\cdot, \cdot;\; \sigma) + C_y \, L_{\eta}(\cdot, \cdot;\; \sigma) \\
              + \frac{1}{2} (C_{xx} \, L_{\xi\xi}(\cdot, \cdot;\; \sigma) + 2 C_{xy} \, L_{\xi\eta}(\cdot, \cdot;\; \sigma) + C_{yy} \, L_{\eta\eta}(\cdot, \cdot;\; \sigma)),
\end{multline}
where $L_{\xi}$, $L_{\eta}$, $L_{\xi\xi}$, $L_{\xi\eta}$ and $L_{\eta\eta}$ denote the scale-normal\-ised Gaussian derivatives computed from image information $f$ from the previous layer, and depend on the scale parameter $\sigma$. The parameters
$C_0 \in \mathbb{R}$, $C_{x}  \in \mathbb{R}$, $C_{y}  \in \mathbb{R}$,
$C_{xx}  \in \mathbb{R}$, $C_{xy}  \in \mathbb{R}$ and $C_{yy}  \in \mathbb{R}$
are the trainable weights, to be learned from the training data. These filter weights will additionally depend upon the feature channels in the network, and will then be different for each layer.

This construction is closely related to the notion of a Taylor expansion of the local image structure, proposed as a model for local visual operations by Koenderink and van Doorn (\citeyear{KoeDoo92-PAMI}), and also used for constructing deep networks by Jacobsen et al. (\citeyear{JacGemLouSme16-CVPR}). Note, however, that no explicit dependency on any zero-order term in terms of image intensity is used here, since the addition of such a term would make the output of the network dependent on the DC-component of the image intensities,\footnote{Strictly speaking, these arguments about invariance to additive intensity transformations of the form $f \mapsto f + f_0$ for intensity offset $f_0$ may possibly only be applied regarding the first layer, and could possibly be relaxed in the higher layers. We leave it as a topic for further research to investigate how a reformulation of the second-order $N$-jet layer according to (5) into a redefinition as  
  $J_{2,\sigma}(f(\cdot, \cdot)) = C_0 \, L (\cdot, \cdot;\, \sigma)
  + C_x \, L_{\xi}(\cdot, \cdot;\, \sigma)
  + C_y \, L_{\eta}(\cdot, \cdot;\, \sigma)
  + \frac{1}{2} (C_{xx}  \, L_{\xi\xi}(\cdot, \cdot;\, \sigma)
  + 2 C_{xy} \, L_{\xi\eta}(\cdot, \cdot;\, \sigma)
  + C_{yy} \, L_{\eta\eta}(\cdot, \cdot;\, \sigma))$
for the higher layers could affect the properties of the GaussDerNets.}
making the output of the network sensitive to illumination variations. The parameter $C_0$ in Equation~(\ref{eq:2-Jet}) thus only acts as a bias term.

The kernels defined in Equation~(\ref{eq:2-Jet}) span all possible linear combinations of first- and second-order directional derivatives, because directional derivatives in any direction $\varphi \in [-\pi, \pi]$ can be expressed as linear combinations of partial derivatives, where for the first two derivative orders we have
\begin{align}
  \begin{split}
    \label{eq-1st-order-dir-der}
    L_{\varphi} 
     = \cos \varphi \, L_{x} + \sin \varphi \, L_{y},
  \end{split}\\ 
  \begin{split}
    \label{eq-2nd-order-dir-der}
     L_{\varphi\varphi} 
     = \cos^2 \varphi \, L_{xx} + 2 \cos \varphi \, \sin \varphi \, L_{xy} + \sin^2 \varphi \, L_{yy}.
  \end{split}
\end{align}
This approach is overall in line with the classical computer vision approaches for scale-invariant feature detection, based on possibly non-linear scale-normalised homogeneous polynomial differential expressions of the form $\mathcal{D}_{\gamma - \norm}L$ (Lindeberg \citeyear{Lin97-IJCV} Equations~(21) and~(23)), with the restriction that the $N$-jet layer, defined according to Equation~(\ref{eq:2-Jet}), is here a purely linear operator.

\subsection{Single-scale-channel Gaussian derivative networks}
A single-scale-channel GaussDerNet is created by coupling Gaussian derivative layers in cascade, with pointwise non-linearities (ReLU) and batch normalisation operations between the layers, together with a spatial selection operation applied to the output of the final layer. 

This network is parametrised by an initial scale value $\sigma_{0}$, which determines the scale at which the network operates, and defines the scale parameter $\sigma_{k}$ in layer $k$ according to a geometric series
\begin{equation} 
\label{eq:sigma-geometric}
    \sigma_{k} = r^{k-1} \, \sigma_{0},
\end{equation} 
for some fixed parameter $r > 1$ and some initial
scale level $\sigma_0 \in \mathbb{R}_+$ for the network hierarchy.
The scale parameter $\sigma_{k}$ determines%
\footnote{If we would remove the non-linearities between the layers and also
  replace the Gaussian derivative kernels by pure zero-order kernels
  with the same values of the scale parameters, then the effective scale
  parameter in layer $k$ to represent the total amount of smoothing would be
  $\sigma_{\text{eff},k} = \sqrt{\sigma_1^2 + \dots + \sigma_k^2}$.
  This property follows from the additive properties of variances
  under convolution of non-negative kernels, and more specifically
  from the semi-group property of the Gaussian kernel, where the
  scale parameters are additive in units of $s = \sigma^2$. When instead using
  Gaussian derivative kernels in the layers with non-linearities in between,
  the situation becomes more complicated. Still, however, the
  effective scale parameter in layer $k$
  can be determined as a function of the the scale parameter $\sigma_k$, if
  complemented with the value index $k$ of the scale level and the
  scale parameter ratio $r$.}
the size of the receptive field of a layer, which steadily increases with the depth of the network. Going from lower to higher layers will therefore correspond to a gradual integration of structural information, from smaller to progressively larger image regions. The first layer takes as input the input image $f$ and uses the scale parameter $\sigma_{1}$ to define its filters.

A general expression for the output of the layer $k+1>1$ in such a network can be defined recursively as
\begin{equation}
  \label{def-gaussdernet-layer-k+1}
  F_{k+1}^{c_{out}} (\cdot, \cdot;\; \sigma_{k+1}) = 
  \sum_{c_{in} \in [1, N_k]} J_{2,\sigma_{k+1}}^{k+1,c_{out},c_{in}} (f^{c_{in}}_{k}(\cdot, \cdot;\; \sigma_k)),
\end{equation}
with the input from the previous layer defined as
\begin{equation}
  \label{def-gaussdernet-layer-k-input}
  f^{c_{in}}_{k}(\cdot, \cdot;\; \sigma_k) =  
  \theta_k^{c_{in}}(\operatorname{BatchNorm}(F_k^{c_{in}}(\cdot, \cdot;\; \sigma_k))),
\end{equation}
except for the first layer for which instead
\begin{equation}
  \label{def-gaussdernet-layer-1}
  F_{1}^{c_{out}} (\cdot, \cdot;\; \sigma_{1}) = 
  \sum_{c_{in} \in [1, N_0]} J_{2,\sigma_{1}}^{1,c_{out},c_{in}} (f^{c_{in}}(\cdot, \cdot;))
\end{equation}
with $f^{c_{in}}(\cdot, \cdot)$ denoting the colour channel with index
$c_{in} \in \{ 1, 2, 3 \}$ for the case of colour input data.
In Equations~(\ref{def-gaussdernet-layer-k+1}) and~(\ref{def-gaussdernet-layer-k-input}), the feature channels are
indexed by $c_{in} \in \mathbb{Z}_+$ and $c_{out} \in \mathbb{Z}_+$,
where $N_{k}$ denotes  the number of feature channels in layer $k$, and
$\theta_k^{c_{in}} \colon \mathbb{R} \rightarrow \mathbb{R}$ represents a pointwise non-linearity. 

The output of the final layer is also subjected to a pointwise non-linearity and a batch normalisation stage, the result of which, for a network with $M$ layers, we denote from now on as
\begin{equation}
\label{eq:final-layer-output}
    F_{\final}^{c}(x, y;\ \sigma_{M}) = f_{M}^{c}(x,y;\; \sigma_{M}).
\end{equation} 
The output feature channel size $c$ in the last layer is set equal to the number of classes of the given dataset.

The final part of the network does not use any fully connected layer; instead a spatial selection operation is performed. In our experiments, we will often use central pixel extraction as the spatial selection operation, where the value of the final layer output at the central pixel position $(x_{\centre}, y_{\centre})$ is selected as the output of the network. For a single-scale-channel GaussDerNet with $M$ layers and the initial scale value $\sigma_{0}$, this is denoted as
\begin{equation}
\label{eq:central-pixel-extraction}
    F_{\centre}^{c}(\sigma_{0}) = F_{\final}^{c}(x_{\centre},y_{\centre};\ \sigma_{M}).
\end{equation}
The choice to extract the central pixel preserves scale covariance and is well motivated when the objects contained in input images are centred. For an even input image size, central pixel extraction is approximated by extracting an average of the 2$\times$2 central square pixel patch instead. To handle non-centered objects, we will later in Section~\ref{sec-spatial-maxpool} replace the central pixel extraction stage with a spatial max-pooling stage.

In this work, we will denote the resulting single-scale-channel GaussDerNet, complemented by spatial selection and parametrised by the initial scale $\sigma_{0}$, by $\Gamma_{\sigma_{0}}$, which for a network with $M$ layers, the relative scale ratio $r$ and any given input image $f \in V$ is the mapping $\Gamma_{\sigma_{0}} \colon V \rightarrow \mathbb{R}^{N_{M}}$.  Combining the definitions in Equations~(\ref{eq:sigma-geometric})--(\ref{eq:central-pixel-extraction}),
and denoting the spatial selection operator by $\operatorname{SpatSel}$,
the effect of this network $\Gamma_{\sigma_{0}}$ on the input image $f$ can be expressed as
\begin{equation}
\label{eq:func-comp-gamma-expression}
\Gamma_{\sigma_{0}} (f(\cdot, \cdot))
  = \operatorname{SpatSel}(f_M(\cdot, \cdot;\; \sigma_M)).
\end{equation}
As will be described in more detail in Section~\ref{sec-scale-covariance-properties},
when complemented with a variability over the initial scale level $\sigma_0$,
this entire structure
is provably scale-covariant,
\footnote{This scale covariance property holds in the sense that if the input image $f$ is rescaled by spatial scaling factor $S \in \bbbr_+$, then provided that the initial scale level $\sigma_0$ used for constructing the hierarchy of Gaussian derivative layers is multiplied by the same spatial scaling factor, the output from any layer in the resulting hierarchical network of the rescaled input image can be perfectly matched to the output of the corresponding Gaussian derivative layer applied to the image data, provided that the output layers are also rescaled by a corresponding spatial scaling transformations. To apply this scale covariance property in situations when the spatial scaling factor $S$ is not known beforehand, we handle this problem by expanding the image representation over multiple scale channels, and do then apply a pooling mechanism over the multiple scale levels to achieve scale-invariant image classification.}
due to the scale-covariant building blocks and total avoidance of scale covariance breaking operations, such as spatial pooling or fully connected layers.

\begin{figure}[h]
  \begin{center}
    \begin{tabular}{cc}
      \hspace{3mm}
      \includegraphics[width=0.14\textwidth]{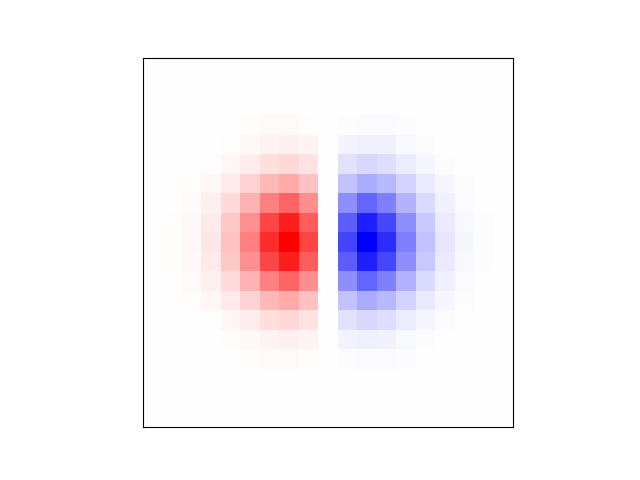} &
      \hspace{-10mm}
      \includegraphics[width=0.14\textwidth]{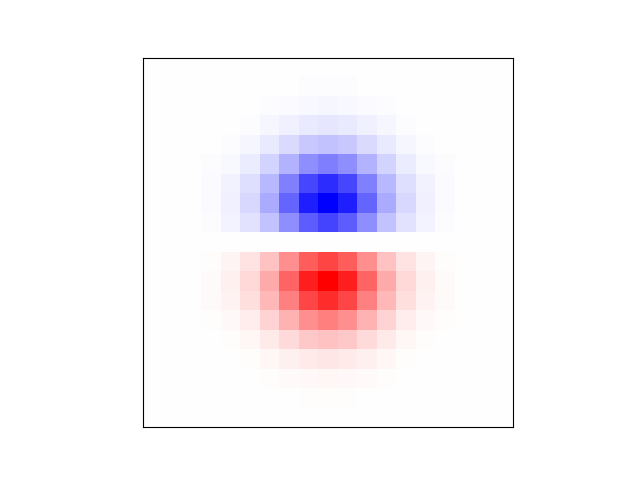} \\
    \end{tabular} 
  \end{center}

  \vspace{-5mm}

  \begin{center}
    \begin{tabular}{ccc}
      \hspace{3mm}
      \includegraphics[width=0.14\textwidth]{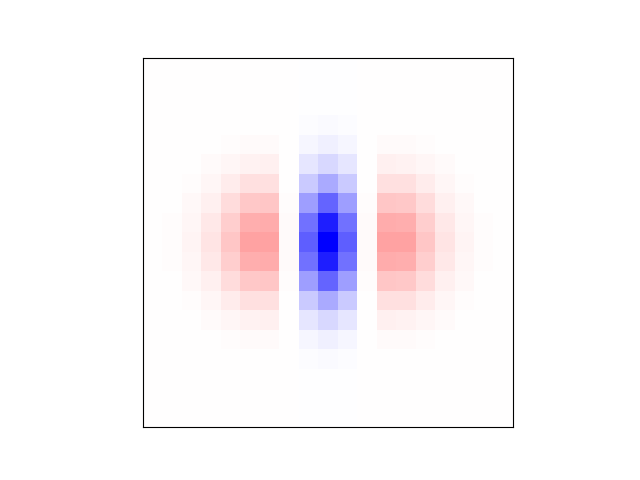} &
      \hspace{-10mm}
      \includegraphics[width=0.14\textwidth]{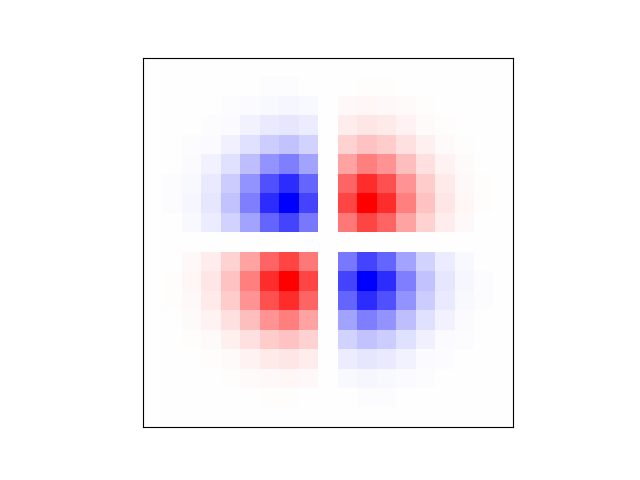} &
      \hspace{-10mm}
      \includegraphics[width=0.14\textwidth]{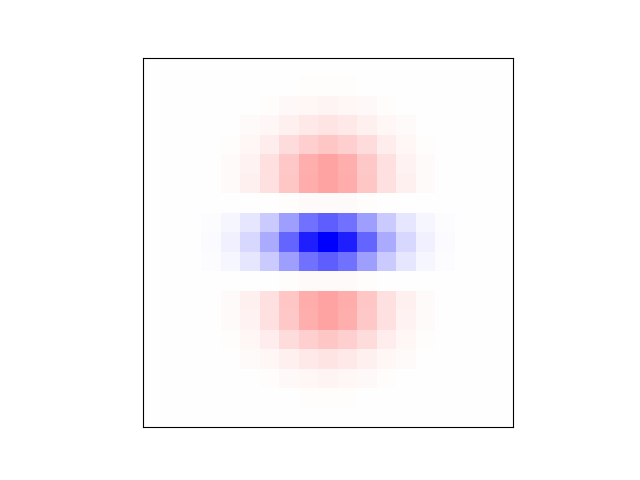} \\
    \end{tabular} 
  \end{center}

  \vspace{-5mm}

  \begin{center}
    \begin{tabular}{cccc}
      \includegraphics[width=0.14\textwidth]{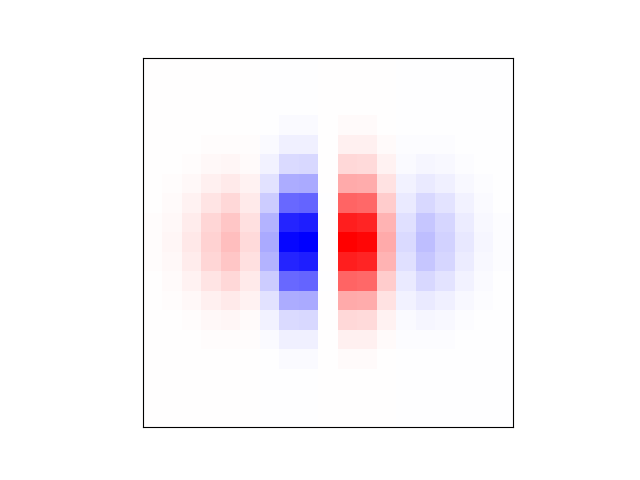} &
      \hspace{-10mm}
      \includegraphics[width=0.14\textwidth]{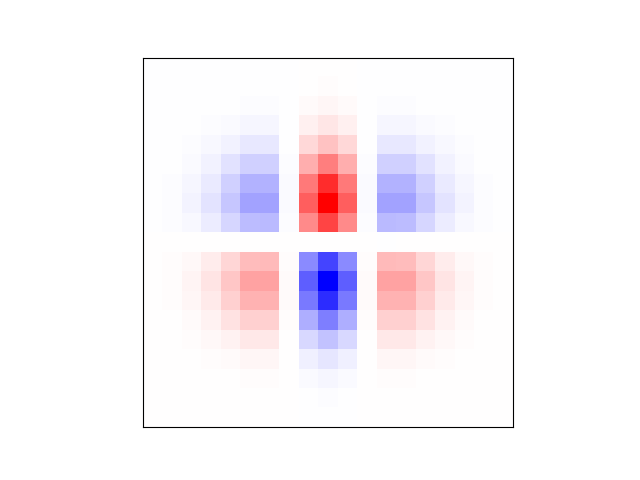} &
      \hspace{-10mm}
      \includegraphics[width=0.14\textwidth]{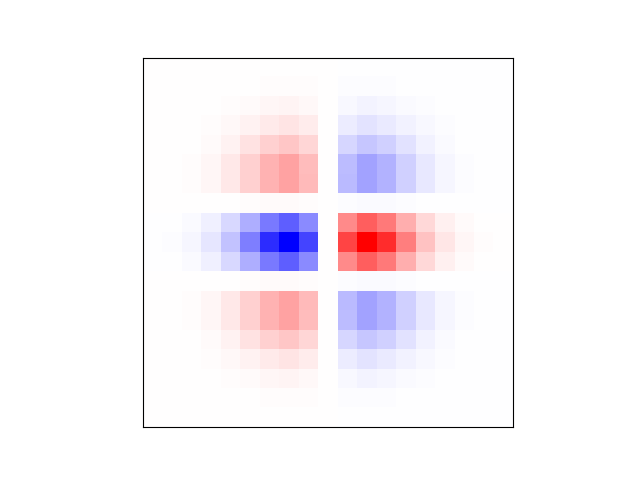} &
      \hspace{-10mm}
      \includegraphics[width=0.14\textwidth]{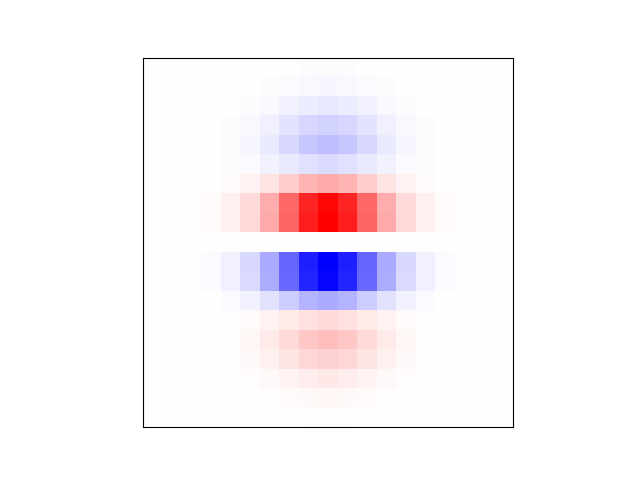} \\
    \end{tabular} 
  \end{center}
  \caption{Equivalent receptive fields that represent the equivalent effect of first smoothing the input image with the discrete analogue of the Gaussian kernel according to Equation~(\ref{eq:discrete-gaussian-kernel}) for $\sigma = 3$, and then applying central difference operators of the forms shown in Equations~(\ref{eq:central-difference-operators-1}) and~(\ref{eq:central-difference-operators-2}) to the smoothed image. (Top row) Discrete approximations of the first-order Gaussian derivatives $g_x$ and $g_y$. (Middle row) Discrete approximations of the second-order Gaussian derivatives $g_{xx}$, $g_{xy}$ and $g_{yy}$. (Bottom row) Discrete approximations of the third-order Gaussian derivatives $g_{xxx}$, $g_{xxy}$, $g_{xyy}$ and $g_{yyy}$.}
  \label{fig-gauss-ders-disgauss-central-diff}
\end{figure}

\subsubsection{Discrete implementation}
In practice, the images are not continuous, which requires a discrete implementation of the Gaussian derivative kernels in the Gaussian derivative network. Our default implementation of discrete approximations of Gaussian derivatives is based on first smoothing the input data with the discrete analogue of the Gaussian kernel (Lindeberg \citeyear{Lin90-PAMI})
\begin{equation}
\label{eq:discrete-gaussian-kernel}
  T(n;\; s) = e^{-s} I_n(s),
\end{equation}
for $s = \sigma^2$, where $I_n$ denote the modified Bessel functions of integer order, and then applying discrete central difference operators of the forms
\begin{align}
  \begin{split}
  \label{eq:central-difference-operators-1}
    \delta_x & = (-1/2, 0, 1/2),
  \end{split} \\
  \begin{split}
  \label{eq:central-difference-operators-2}
    \delta_{xx} & = (1, -2, 1),
  \end{split}
\end{align}
to the spatially smoothed image data. Equivalent discrete derivative approximation kernels corresponding to this way of discretising the continuous Gaussian derivative kernels are shown in Figure~\ref{fig-gauss-ders-disgauss-central-diff}.

As described in more detail in (Lindeberg \citeyear{Lin24-JMIV-discgaussder}), this method for discrete approximation has substantially better properties at very fine scale levels than {\em e.g.\/} convolving the discrete image data by sampled Gaussian derivative kernels. This method of computing discrete derivative approximations, from a set of central difference operators of small support applied to the spatially smoothed image data, is also computationally much more efficient, compared to separate smoothing with a corresponding set of sampled Gaussian derivative kernels, for which the spatial smoothing operation cannot be shared between the convolution operations required for computing discrete derivative approximations of different orders.

\subsection{Multi-scale-channel Gaussian derivative networks}
\label{sec-multi-gaussdernets-theory}
A multi-scale-channel GaussDerNet is essentially a set of $N$ copies of a single-scale-channel GaussDerNet $\Gamma_{\sigma_{0}}$, referred to as scale channels, but with each copy being based on a different value of $\sigma_{0}$,
for a multi-scale-channel network denoted by $\sigma_{n,0}$.
In the discrete case, the $\sigma_{n,0}$-values are chosen to cover a certain discrete range of scales, that are separated by some factor $\lambda$, typically chosen as $\lambda = \sqrt{2}$. The value $\sigma_{n,0}$ that parametrises the $n$:th single-scale-channel GaussDerNet $\Gamma_{\sigma_{n,0}}$ can therefore be expressed as 
\begin{equation}
\label{eq:sigma0-multichan}
    \sigma_{n,0} = \lambda^{(n-1)} \cdot \sigma_{1,0}, 
\end{equation}
where the first number in the subscript represents which scale channel the initial scale value belongs to. In this way, the scale channels process the same input image in parallel, but each using receptive fields of different size corresponding to a certain scale determined by $\sigma_{n,0}$. Equation~(\ref{eq:sigma-geometric}) can then be generalised to express the scale parameter at layer $k$ of the scale channel $\Gamma_{\sigma_{n,0}}$ as
\begin{equation} 
\label{eq:sigma-geometric-multi}
    \sigma_{n,k} = r^{k-1} \sigma_{n,0}.
\end{equation} 
This can be interpreted as each scale channel being a copy of each other, but with the $\sigma$-values in every layer of the scale channel being scaled according to a self-similar distribution. In a multi-scale-channel network, all the scale channels share the same value of the relative scale ratio $r$. 

To preserve scale covariance of this multi-scale-channel construction, the final prediction in the classification stage does not use any fully connected layer. Instead, a permutation-invariant pooling operation over scales is applied across the outputs of these scale channels. In this work, we mainly make use of either max pooling over scales or average pooling over scales.

\begin{figure}[hbt]
  \begin{center}
    \begin{tabular}{c}
       \includegraphics[width=0.49\textwidth]{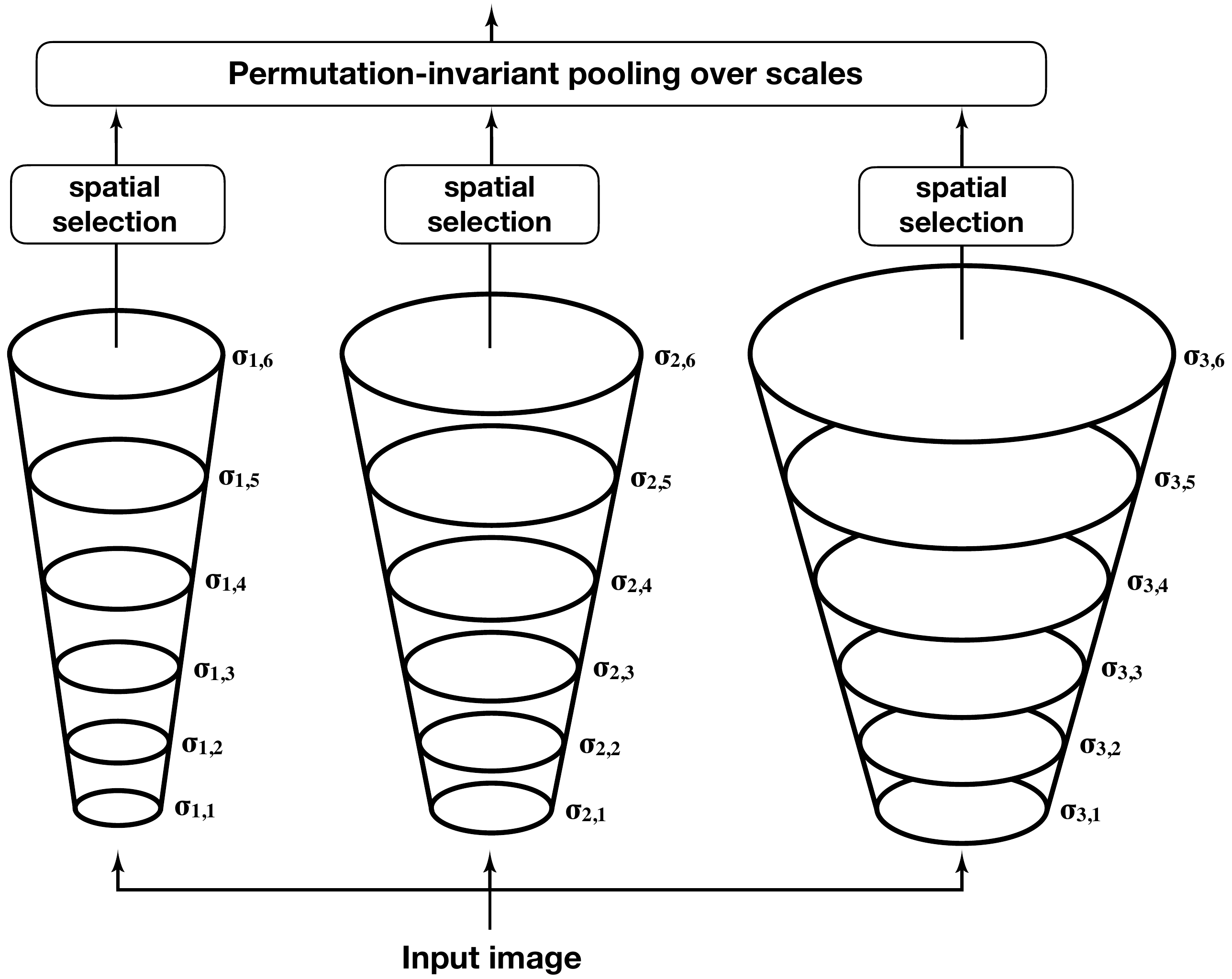}
    \end{tabular}
  \end{center}
\caption{Schematic illustration of a multi-scale-channel GaussDerNet, with 6 layers and 3 parallel scale channels: $\Gamma_{\sigma_{1,0}}$, $\Gamma_{\sigma_{2,0}}$ and $\Gamma_{\sigma_{3,0}}$. Each scale channel is based on a different initial scale value $\sigma_{i,0}$, with the scale spacing in the discrete case being determined by Equation~(\ref{eq:sigma0-multichan}). The scale parameter of the receptive field in each layer is determined according to Equation~(\ref{eq:sigma-geometric-multi}), with the scale levels becoming coarser with depth, as depicted in the diagram. All the scale channels share the same weights. As permutation-invariant pooling operations over scale, we do in this paper consider max pooling over scale or average pooling over scale. As spatial selection methods, we consider either central pixel extraction, for datasets where the objects are centered, or spatial max pooling, for datasets where the objects are not centered.}
\label{fig-architecture}
\end{figure}

For a Gaussian derivative model with $N$ scale channels and $M$ layers, a max pooling operation over scales is thus used to produce the final output $F_{\maximum}$ of the whole network. Assuming that a central pixel extraction operation has already been performed for each scale channel $\Gamma_{\sigma_{i,0}}$, the max pooling over scales is defined as
\begin{equation}
\label{eq:maxpooling-expression}
    F_{\maximum}^{c} = \max_{i \in [1,2,...,N]} F_{\centre}^{c}(\sigma_{i,0}),
\end{equation}
given an input image $f$, with $c$ representing the class index. In the absence of a selection of a specific image point, we could also consider performing max pooling over scale at every image point according to
\begin{equation}
\label{eq:maxpooling-expression-xy}
    F_{\maximum}^{c}(x, y) = \max_{i \in [1,2,...,N]} F_{\final}^{c}(x, y;\ \sigma_{i,M}).
\end{equation}
These types of max pooling over scale-channel outputs specifically allow for detection of maxima over scale, resembling scale selection approaches based on local extrema over scales of scale-normalised derivatives (Lindeberg \citeyear{Lin97-IJCV,Lin21-EncCompVis}). The complete architecture is visualised in Figure~\ref{fig-architecture}.

Since the scale channels $\Gamma_{\sigma_{i,0}}$ are copies of each other over scales, this means that they share the same weights. The weight sharing implies that if a rescaled version of an image with some pattern learned at a certain scale is used as input into the model,\footnote{In Jansson and Lindeberg (\citeyear{JanLin22-JMIV}) a closely
  related dual approach to defining scale-channel networks is considered,
  where the same deep network is applied to a set of rescaled input images.
  For continuous image data, such a dual approach based on rescaling the
  input data is formally equivalent to the approach of rescaling filter kernels
  considered in this work. For discrete image data, the two types of
  approaches may, however, differ, depending on discretisation issues.}
then this will cause a strong response at a scale channel $\Gamma_{\sigma_{j,0}}$ with $\sigma_{j,0}$ corresponding to this new scale, even if the pattern was never seen at this scale during training. 

\subsubsection{Handling a limited number of scale channels}
In practice, the deep network model needs to be discretised, meaning that only a finite set of scale channels can be used, which introduces a risk for picking up incorrect types of image structures, when generalising from training data over a narrow scale range only. This is because responses corresponding to scales beyond the range spanned by the actual training scales may move into the operational scale range, when performing testing at new, not previously seen, scales far away from the training scales. By adding extra scale channels at the boundaries of the scale interval, such problems can be substantially reduced, since the network then has the ability to adjust the weights (in the entire network) in such a way that responses further away in scale range should not lead to misclassifications. This boundary handling approach has specifically been demonstrated to lead to very good scale generalisation properties on the MNIST Large Scale dataset (Jansson and Lindeberg \citeyear{JanLin20-MNISTLargeScale}, Jansson and Lindeberg \citeyear{JanLin22-JMIV}, Lindeberg \citeyear{Lin22-JMIV}).

\subsection{Scale covariance of Gaussian derivative networks}
\label{sec-scale-covariance-properties}
Under natural image transformations, images are subject to spatial scaling transformations, which constitute a group of dimensionality 1. In this section, we will describe how the scale covariance property of GaussDerNets manifests itself in terms of explicit transformation properties between the scale channels under such spatial scaling transformations.

Generally, the notion of covariance, often referred to as equivariance in deep learning literature, refers to the property where applying a symmetry transformation to the output from the network can be matched to a corresponding transformation of the input of the network, thus constraining its representation space in a way that improves generalisation.

Formally, an operator $\Phi$, which for deep learning typically refers to a network or a layer that maps one representation to another, is covariant with respect to a group $G$, if for any input $f \in \Omega$ it satisfies
\begin{equation} 
\label{eq:equivariance-equation}
    \Phi(\mathcal{T}_{g} \, f) = \mathcal{T}'_{g} \, \Phi(f) \qquad \forall g \in G, f \in \Omega,
\end{equation}
where $\mathcal{T}_{g}$ is the group action\footnote{An action of a group $G$ on a set $\Omega$ is defined as a group homomorphism of $G$ into the symmetric group $\operatorname{Sym}(\Omega)$, which is the group of bijections from $\Omega$ to itself.} of the group $G$.
This definition implies that the map $\mathcal{T}_{g}:\Omega \times G \rightarrow \Omega$ is a family of transformations with a group structure, and that $\mathcal{T}_{g}$ and $\mathcal{T}'_{g}$ need not be the same.
Thus, Equation~(\ref{eq:equivariance-equation}) means that the order between applying the covariant network $\Phi$ and some corresponding group action does not matter.
A network $\Phi$ that obeys such properties can also be said to be equivariant with respect to $G$.

The special case when the transformation $\mathcal{T}'_{g}$ in Equation~(\ref{eq:equivariance-equation}) is the identity transformation is referred to as invariance, which then means that applying the operator $\Phi$ provides the same result, for all possible transformations ${\cal T}_g$ of the objects $f$ in the group, including the identity transformation ${\cal T}_g = {\cal I}$, that leaves the object $f$ unchanged. Thus, the representation is invariant to all the transformations ${\cal T}_g$ in the group for all the objects $f$ in the domain $\Omega$ by the group action.

The GaussDerNet architecture is scale covariant, with respect to the group of scaling transformations $\mathcal{S}$ on functions $f(x,y)$ in the image domain.
In the case where the network is modelled as continuous, and assumed to be made up of an infinite number of scale channels, the Gaussian scale-space primitives used to define each layer of the continuous GaussDerNet and the use of weight sharing ensures provable scale covariance of the entire Gaussian derivative network architecture. 

\begin{figure}[ht]
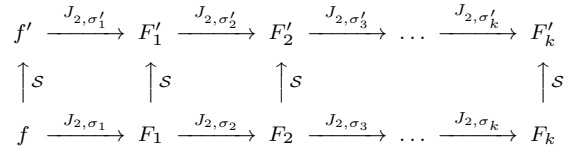

\
\begin{center}
    \[
      \hspace{7mm}\begin{CD} 
       f' @>{J_{2,\sigma'_{1}}}>> F_{1}' @>{J_{2,\sigma'_{2}}}>>  F_{2}' @>{J_{2,\sigma'_{3}}}>> \hdots @>{J_{2,\sigma'_{k}}}>> F_{k}' \\     
       @AA{\mathcal{S}}A @AA{\mathcal{S}}A @AA{\mathcal{S}}A @. @AA{\mathcal{S}}A \\
       f @>{J_{2,\sigma_{1}}}>> F_{1} @>{J_{2,\sigma_{2}}}>> F_{2} @>{J_{2,\sigma_{3}}}>> \hdots @>{J_{2,\sigma_{k}}}>> F_{k}
    \end{CD}
    \]
  \end{center}
  \caption{Commutative diagram showing the scale-covariant properties of continuous GaussDerNets based on second-order scale-normalised Gaussian derivatives, under a scaling transformation $\mathcal{S}$ of the scale parameter and the image domain, defined as $(x', y';\; \sigma')=(S \, x, S \, y;\; S \, \sigma)$, for a positive spatial scaling factor $S$. When this definition is extended to the layers, it follows that due to the coupling of the layers in cascade in the Gaussian derivative network, that under a scaling transformation $\mathcal{S}$ the layers $F'_{k}$ and $F_{k}$ can be matched according to $F_{i}^{c_{out}}(x, y;\; \sigma_{i})$ = ${F'}_{i}^{c_{out}}(S \, x, S \, y;\; S \, \sigma_{i})$, the full relation being expressed in Equation~(\ref{eq:general-scaling-tranform-layers}). For discrete scale channels, the spacing factor between neighbouring scale channels is set to be the empirically chosen constant $\lambda=\sqrt{2}$, which for continuous images makes it possible to perfectly match the output from the scale channels for spatial scaling factors $S$ that are integer powers of the scale ratio $\lambda = \sqrt{2}$, whereas the corresponding relations will instead only be approximate for other spatial scaling factors.}
  \label{fig-comm-diag-hier-network}
\end{figure}

With regard to the scale-normalised Gaussian derivative operators in Equation~(\ref{eq:gaussian-derivative}), which are used as the basic primitives for expressing the filter weights of the form defined in Equation~(\ref{eq:2-Jet}) in terms of linear combinations of these primitives, the action of the scaling group ${\cal S}$ on these Gaussian derivatives corresponds to the following transformation property
\begin{equation}
  L_{{x'}^{\alpha}{y'}^{\beta}}(x';\; \sigma') = L_{{x}^{\alpha}{y}^{\beta}}(x;\; \sigma)
\end{equation}
for the spatial Gaussian derivative responses to the original image $f(x, y)$ and the transformed image $f'(x', y')$ according to
\begin{align}
  \begin{split}
     L_{{x}^{\alpha}{y}^{\beta}}(x;\; \sigma) 
    = g_{{x}^{\alpha}{y}^{\beta}}(x;\; \sigma) * f(x, y)
   \end{split}\\
  \begin{split}
     L_{{x'}^{\alpha}{y'}^{\beta}}(x';\; \sigma') 
    = g_{{x'}^{\alpha}{y'}^{\beta}}(x;\; \sigma) * f'(x', y')
   \end{split}
\end{align}
under any spatial scaling transformation with spatial scaling factor
$S \in \mathbb{R}_+$ of the form
\begin{equation}
  \label{eq-spat-sc-transf}
   (x', y') = (S \, x, S \, y)
\end{equation}
for matching values of the spatial scale parameters according to (see Lindeberg \citeyear{Lin97-IJCV} Section~4.1)
\begin{equation}
 \label{eq-sigma-sc-transf}
  \sigma' = S \, \sigma.
\end{equation}
Then, with regard to the resulting scale channels in the GaussDerNet constructed from these primitives, it holds that under spatial scaling transformation of the form given in Equations~(\ref{eq-spat-sc-transf}) and (\ref{eq-sigma-sc-transf}), the responses of the Gaussian derivative layer of order 2 to the input image $f$ and its rescaled counterpart $f'$ are scale covariant (see Lindeberg \citeyear{Lin22-JMIV} Equation~(17)):
\begin{equation}
 \label{eq:N-jet-scale-transform}
  J_{2,\sigma'}(f'(\cdot, \cdot)) 
 = J_{2,\sigma}(f(\cdot, \cdot)). 
\end{equation}
From Equation~(\ref{eq:N-jet-scale-transform}), it follows that the layers $F_{k+1}$ and ${F'}_{k+1}$ defined according to Equation~(\ref{def-gaussdernet-layer-k+1}) are related by a scaling transformation according to (Lindeberg \citeyear{Lin22-JMIV} Equation (19))
\begin{align}
  \label{eq:general-scaling-tranform-layers}
  \begin{split}
   {F'}_{k+1}^{c_{out}}(x', y';\; \sigma'_{k+1}) 
  \end{split}\nonumber\\
  \begin{split}
  = \sum_{c_{in} \in [1, N_k]}
         J_{2,\sigma'_{k+1}}^{k+1,c_{out},c_{in}}(\theta_k^{c_{in}}({F'}_k^{c_{in}}(\cdot, \cdot;\; \sigma'_k)))(x', y';\; \sigma'_{k+1}) 
  \end{split}\nonumber\\
  \begin{split}
  =  \sum_{c_{in} \in [1, N_k]} J_{2,\sigma_{k+1}}^{k+1,c_{out},c_{in}}(\theta_k^{c_{in}}(F_k^{c_{in}}(\cdot, \cdot;\; \sigma_k)))(x, y;\; \sigma_{k+1})
   \end{split}\nonumber\\
  \begin{split}
    = F_{k+1}^{c_{out}}(x, y;\; \sigma_{k+1}),
  \end{split}
\end{align}
provided that the scale parameters $\sigma$ and $\sigma'$ over the two image domains, over which the mutually rescaled input images $f$ and $f'$ respectively are defined, are related according to Equation (\ref{eq-sigma-sc-transf}). The relation in Equation~(\ref{eq:general-scaling-tranform-layers}) is illustrated in Figure~\ref{fig-comm-diag-hier-network}, and also holds for GaussDerNets based on higher-order derivatives.

With regard to the specific implementations of GaussDerNets that we use in this work, where the scale parameters in the different scale channels are powers of the empirically chosen suitable constant\footnote{For the class of multi-scale representations known as pyramids, a scale ratio of 2 between adjacent scale levels was a common choice in the earliest works (Burt and Adelson \citeyear{BA83-COM}, Crowley and Parker \citeyear{Cro84-peaks}), as well as in some of the early work on Gaussian scale-space representations. An empirical observation, however, was that a denser scale sampling of the scale levels by a factor of $\sqrt{2}$ or lower can lead to substantially better results, both regarding pyramids and regarding classical scale-space algorithms (Lindeberg \citeyear{Lin93-Dis}, Lindeberg and Bretzner \citeyear{LinBre03-ScSp}). Regarding deep networks based on multiple scale channels, an experimental investigation in (Jansson and Lindeberg \citeyear{JanLin22-JMIV}, see Figure 8) showed that a substantial increase in scale generalisation performance could be obtained by decreasing the scale sampling ratio from 2 to $\sqrt{2}$, while only a minor improvement was obtained by decreasing the scale sampling ratio further to $\sqrt[4]{2}$. Based on these analogies, we therefore throughout make use of a scale sampling ratio of $\sqrt{2}$ in this work.} $\lambda = \sqrt{2}$, this result thus means that, provided that the spatial scaling factor $S \in \mathbb{R}$ in the spatial scaling transformation between the two images $f$ and $f'$ is an integer power of this constant
\begin{equation}
   \label{eq-S-pow-sqrt-2}
   S = \lambda^{n} = \sqrt{2}^{n}, \qquad \text{where } n \in \mathbb{Z},
\end{equation}
then the outputs from the different scale channels in the GaussDerNet can be perfectly matched under such spatial scaling transformations. For other values of the spatial scaling factor $S$, a corresponding matching would, however, only be approximate.

This matching property does, in turn imply that, if we compute the maximum value over an infinite number of scale channels, or apply some other permutation invariant pooling operation, such as average pooling over scales, over such an infinite set of scale channels, then the result of such permutation invariant pooling mechanisms will be provably invariant under uniform spatial scaling transformations for spatial scaling factors according to Equation~(\ref{eq-S-pow-sqrt-2}).

When using a finite number of scale channels, the corresponding results will, however, only be approximate, and will then depend on the relationship between possibly ``missing'' scale channels in the testing stage in relation to the training stage. 

When implementing these networks on discrete data, there is yet another source to approximate relations, when the continuous Gaussian kernels used for deriving the scale-covariant properties, underlying the construction of the truly scale-invariant network, are to be replaced by discrete approximations of these kernels. That topic will be treated in Section~\ref{sec-der-gauss-kernels}.

\section{Architectural and algorithmic extensions and modifications of Gaussian derivative networks}
\label{sec-alg-improv}

For applying the GaussDerNets to more complex datasets than reported in the original work (Lindeberg \citeyear{Lin22-JMIV}), we propose the following conceptual and algorithmic extensions and improvements:

\subsection{Average pooling across scales}
\label{sec-avgpool}
Although general permutation invariant pooling mechanisms across the different scale channels were proposed as compatible with the original formulation of scale-covariant and scale-invariant GaussDerNets, the experimental results in the original publication were only reported for max pooling over scale, defined in Equation~(\ref{eq:maxpooling-expression}). In this work, we additionally perform average pooling over scale. 

If we assume that central pixel extraction has already been performed for each scale channel $\Gamma_{\sigma_{i,0}}$, then average pooling over scale for a Gaussian derivative model with $N$ scale channels is performed according to
\begin{equation}
\label{eq:avgpooling-expression}
    F_{\avg}^{c} = \frac{1}{N} \sum_{i=1}^{N} F_{\centre}^{c}(\sigma_{i,0}).
\end{equation}
In the absence of a spatial selection step, average pooling could also be performed at every image pixel according to
\begin{equation}
\label{eq:avgpooling-expression-xy}
    F_{\avg}^{c}(x,y) = \frac{1}{N} \sum_{i=1}^{N} F_{\final}^{c}(x,y;\ \sigma_{i,M}).
\end{equation}
These types of average pooling over scale-channel outputs are related to scale selection methods based on weighted averaging of feature responses over scale levels (Lindeberg \citeyear{Lin12-JMIV}). With regard to the experiments that will be presented later in Sections~\ref{sec-experiments} and~\ref{sec-ablation}, average pooling over scales typically leads to a more stable training process compared to max pooling over scales.

For datasets with more complex patterns and textures, average pooling over the scale channels notably has the potential benefit of making it possible for the deep network to learn how to combine information from multiple scales, when making a classification with respect to an object class, in contrast to max pooling that will only have the ability to make use of the information from a single scale channel.

\subsection{Spatial max pooling as a spatial location mechanism}
\label{sec-spatial-maxpool}
For datasets where the objects are not always centered in the image, we replace the central pixel extraction defined in Equation~(\ref{eq:central-pixel-extraction}), that was used in the original implementation of the GaussDerNets, with a spatial max pooling stage, which then serves as a basic region-of-interest mechanism. This region-of-interest mechanism chooses the maximal activation in the final layer, no matter what its spatial location, thus allowing for off-centre activations to be more influential, while still preserving the scale-covariant and scale-invariant properties of the model. For a scale channel $\Gamma_{\sigma_{0}}$ with $M$ layers and initial scale value $\sigma_{0}$, given an input image $f$, this operation can be defined as
\begin{equation}
\label{eq:spatial-maxpooling-expression}
    F_{\spatmax}^{c}(\sigma_{0}) = \max_{x,y} F_{\final}^{c}(x,y;\ \sigma_{M}).
\end{equation}
Specifically, the combination of spatial max pooling with max pooling over scales generalises the notion of scale-space extrema from previous application to scale-invariant interest point detection in classical computer vision (Lindeberg \citeyear{Lin97-IJCV,Lin12-JMIV,Lin15-JMIV}), to the application to joint object detection and object localisation in deep networks. These conceptual similarities will be demonstrated in experiments in Sections~\ref{sec-spatial-max-pooling} and~\ref{sec-stir-experiments}.

Spatial average pooling could possibly also be considered as a method to make the responses more robust to variations in the spatial locations of the objects, but such a method would have the potential drawback of interference between multiple objects or background structures within the support region of the image, and will therefore not be considered here.

\subsection{Data augmentation with respect to symmetry transformations and regularisation}
\label{sec-training-enhancements}
In the training process, we perform random horizontal image flipping, and for RGB datasets random color jitter, which involves changing the brightness, contrast, saturation and hue of the images. These data augmentation techniques effectively increase the size of the training set and introduce additional variability into the data. Other training data augmentation techniques could be used to aid the training process, such as rotations or random cropping, however, these are not considered in this work.

In this work, training of the GaussDerNet is performed using the AdamW optimiser (Loshchilov and Hutter \citeyear{LosHut19-ICLR}) instead of the standard Adam optimiser (Kingma and Ba \citeyear{KinBa15-ICLR}), because it typically yields improved generalisation performance, due to its implementation of a decoupled weight decay.

Furthermore, we investigate the use of regularisation by cutout (DeVries and Taylor \citeyear{DevTay17-arXiv}), which involves masking out randomly selected square regions of input images during training and has been shown to benefit models designed to handle scaling variations (Sosnovik et al. \citeyear{SosSzmSme20-ICLR}).

\subsection{Scale-channel dropout} 
\label{sec-scale-dropout}
To make the classification results more robust to perturbations, we make use of a form of scale dropout, where dropout is applied across the outputs from the scale channels before performing the permutation-invariant scale pooling during the training stage. This teaches the multi-scale-channel GaussDerNet to not overly rely on information at specific scales when making a prediction, which is especially useful when using max pooling over scales. The GaussDerNets based on average pooling over scales can also benefit from such regularisation to some extent, since during training they get to learn from different combinations of information across multiple scales, which teaches the network to consider different sets of image patterns. The effect of applying scale-channel dropout to the output of a scale channel $\Gamma_{\sigma_{i,0}}$ with $M$ layers, after spatial selection has already been performed, is defined as 
\begin{multline}
\label{eq:scale-channel-dropout}
    \operatorname{ScaleChannelDropout}_{p}(F_{\centre}^{c}(\sigma_{i,0})) = \\
    X(c,i) \, F_{\centre}^{c}(\sigma_{i,0}),
\end{multline}
where $X(c, i) \sim \operatorname{Bernoulli}(p)$, with the probability of an element being equal to 1 being $p \in [0, 1]$ and $c$ being the class index. In our experiments, we use dropout factors $(1 - p) = q \in [0,0.2,0.3]$, where datasets with RGB images use $q = 0.3$ and grayscale datasets use $q = 0.2$, while our ablation studies investigate the effects of using no scale-channel dropout, with $q = 0$.
The outputs are scaled by a factor of $1/p$ during training, like typically done in standard dropout.
This approach to dropping scale information during training shares some similarities with the notion of scale dropout introduced in Sangalli et al. (\citeyear{SanBluVelAng22-BMVC}).

\subsection{Extensions from the 2-jet to the 3-jet}
\label{sec-3-jet}
The experiments on the original implementation of scale-covariant and scale-invariant GaussDerNets were based on Gaussian derivatives up to order 2. In classical computer vision, adding third-order Gaussian derivatives is sometimes helpful to effectively model more complex image structures. Therefore, in this work we also report experiments based on Gaussian derivatives up to order 3. Analogously to Equation~(\ref{eq:2-Jet}), a Gaussian derivative layer of order 3 is defined as a linear combination of scale-normalised Gaussian derivatives up to order 3:
\begin{multline}
  \label{eq:3-Jet}
  J_{3,\sigma}(f(\cdot,\cdot)) = C_0
  +   C_x \, L_{\xi}(\cdot,\cdot;\; \sigma)
  +   C_y \, L_{\eta}(\cdot,\cdot;\; \sigma) \\
  + \frac{1}{2!} (C_{xx} \, L_{\xi\xi}(\cdot,\cdot;\; \sigma)
  +    2 C_{xy} \, L_{\xi\eta}(\cdot,\cdot;\; \sigma)
  +    C_{yy} \, L_{\eta\eta}(\cdot,\cdot;\; \sigma))\\
  + \frac{1}{3!} (C_{xxx} \, L_{\xi\xi\xi}(\cdot,\cdot;\; \sigma)
    + 3 C_{xxy} \, L_{\xi\xi\eta}(\cdot,\cdot;\; \sigma) \\
    + 3 C_{xyy} \, L_{\xi\eta\eta}(\cdot,\cdot;\; \sigma)
    + C_{yyy} \, L_{\eta\eta\eta}(\cdot,\cdot;\; \sigma)).
\end{multline}

\subsection{Choices of discrete derivative approximation kernels}
\label{sec-der-gauss-kernels}
There exist several different approaches to discretising continuous Gaussian derivative kernels, that can be considered when implementing GaussDerNets on discrete data. Beyond the discrete analogue of the Gaussian kernel complemented with small-support central difference operators, as described in Equations~(\ref{eq:discrete-gaussian-kernel})--(\ref{eq:central-difference-operators-2}), we will also consider discretisation in terms of pure spatial sampling, meaning defining sampled Gaussian derivative kernels for derivatives of order $\alpha$ as (Lindeberg \citeyear{Lin24-JMIV-discgaussder} Equation~(53))
\begin{equation}
    \label{eq:sampled-Gaussian-kernel}
    T_{\sample,x^{\alpha}}(n;\ s) = g_{x^{\alpha}}(n;\ s),
\end{equation} 
where $n \in \mathbb{Z}$. Additionally, we will also consider the normalised sampled Gaussian kernel, obtained by using the discrete $l_{1}$-norm of the sampled Gaussian kernel to normalise it, in combination with central differences. Furthermore, we will consider discretisation in terms of local integration of Gaussian derivative kernels over each pixel support region, meaning defining integrated Gaussian derivative kernels for derivatives of order $\alpha$ as (Lindeberg \citeyear{Lin24-JMIV-discgaussder} Equation~(54))
\begin{equation}
    \label{eq:integrated-Gaussian-kernel}
    T_{\integrated,x^{\alpha}}(n;\ s) = \int_{x=n-1/2}^{n+1/2}g_{x^{\alpha}}(x;\ s) \, dx.
\end{equation}
Out of all these discrete approximation methods for Gaussian derivatives considered in this work, the approach based on the discrete analogue of the Gaussian kernel has been previously proposed as the best numerical approximation for very low values of the scale parameter (Lindeberg \citeyear{Lin24-JMIV-discgaussder}). The approach for computing multiple Gaussian derivative responses for different orders of spatial differentiation with the discrete kernels is also computationally efficient, based on the associated central difference operators of small spatial support. However, implementations equivalent to convolution with sampled Gaussian derivative kernels may constitute the most common discretisation choice by other authors who implement deep networks involving Gaussian derivative operators.

As described in Lindeberg (\citeyear{Lin24-JMIV-discgaussder}), the integrated Gaussian derivative kernel degenerates less severely compared to the sampled Gaussian derivatives kernel at very fine scale levels. For larger scale values, however, this discretisation approach does introduce a certain scale offset $\Delta s_{\integrated}$, resulting from the use of box integration. This entails a difference in the spatial variance of the discrete kernel compared to the spatial variance of the continuous Gaussian kernel.\footnote{While one could consider compensating for the scale offset for the integrated Gaussian derivative kernels, we have not explored such a path for the fixed-scale networks studied in this work.}

The pros and cons of implementing each of these Gaussian derivative discretisation methods in the GaussDerNets will be empirically investigated in our comparative experiments in Section~\ref{sec-new-der-kernels}.

\subsection{Learning of the scale values}
\label{sec-gaussdernet-sig-lr}
In the original work on scale-covariant and scale-invariant GaussDerNets, the scale values across the layers were distributed according to a logarithmic distribution, motivated by self-similarity over scales. In this work, we also report the results of learning the scale levels, as has also been previously done by Pintea et al. (\citeyear{PinTomGoeLooGem21-IP}), Saldanha et al. (\citeyear{SalPinGeTo21-BMVC}), Penaud-Polge et al. (\citeyear{PenVelAng22-ICIP}), Yang et al. (\citeyear{YanDasMah23-arXiv, YangDasMah23-rot-arXiv}), and Basting et al. (\citeyear{Basting-2024-IJCCV}).

Since the modified Bessel functions do not exist as full-fledged built-in functions in PyTorch, when learning the scale levels from training data, we will instead approximate the continuous Gaussian derivative operators by sampled or integrated Gaussian derivatives.

The learning of the scale values is implemented by defining the scale parameters as trainable parameters with their own learning rate. During training, the scale parameters may specifically need to be clamped within a certain range.

\begin{figure*}[htbp]
    \begin{center}
    \textit{Samples from the rescaled Fashion-MNIST dataset}
    \end{center}
    \vspace{-2mm}
    \begin{tabular}{ccccccccccc}
      \includegraphics[width=0.083\textwidth]{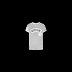} &
      \hspace{-2.7mm}
      \includegraphics[width=0.083\textwidth]{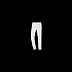} &
      \hspace{-2.7mm}
      \includegraphics[width=0.083\textwidth]{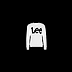} &
      \hspace{-2.7mm}
      \includegraphics[width=0.083\textwidth]{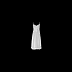} &
      \hspace{-2.7mm}
      \includegraphics[width=0.083\textwidth]{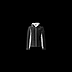} &
      \hspace{-2.7mm}
      \includegraphics[width=0.083\textwidth]{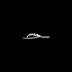} &
      \hspace{-2.7mm}
      \includegraphics[width=0.083\textwidth]{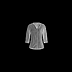} &
      \hspace{-2.7mm}
      \includegraphics[width=0.083\textwidth]{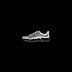} &
      \hspace{-2.7mm}
      \includegraphics[width=0.083\textwidth]{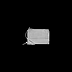} &
      \hspace{-2.7mm}
      \includegraphics[width=0.083\textwidth]{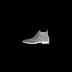} &
      \\
    \end{tabular} 

  \vspace{0.5mm}
  
    \begin{tabular}{cccccccccc}
      \includegraphics[width=0.083\textwidth]{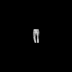} &
      \hspace{-2.7mm}
      \includegraphics[width=0.083\textwidth]{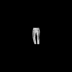} &
      \hspace{-2.7mm}
      \includegraphics[width=0.083\textwidth]{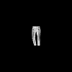} &
      \hspace{-2.7mm}
      \includegraphics[width=0.083\textwidth]{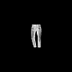} &
      \hspace{-2.7mm}
      \includegraphics[width=0.083\textwidth]{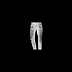} &
      \hspace{-2.7mm}
      \includegraphics[width=0.083\textwidth]{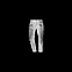} &
      \hspace{-2.7mm}
      \includegraphics[width=0.083\textwidth]{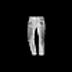} &
      \hspace{-2.7mm}
      \includegraphics[width=0.083\textwidth]{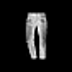} &
      \hspace{-2.7mm}
      \includegraphics[width=0.083\textwidth]{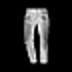} &
      \\
    \end{tabular} 

  \vspace{2mm}

    \begin{center}
    \textit{Samples from the rescaled Fashion-MNIST with translations dataset}
    \end{center}
    \vspace{-2mm}
    \begin{tabular}{ccccccccccc}
      \includegraphics[width=0.083\textwidth]{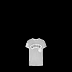} &
      \hspace{-2.7mm}
      \includegraphics[width=0.083\textwidth]{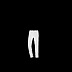} &
      \hspace{-2.7mm}
      \includegraphics[width=0.083\textwidth]{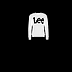} &
      \hspace{-2.7mm}
      \includegraphics[width=0.083\textwidth]{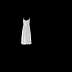} &
      \hspace{-2.7mm}
      \includegraphics[width=0.083\textwidth]{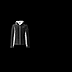} &
      \hspace{-2.7mm}
      \includegraphics[width=0.083\textwidth]{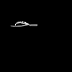} &
      \hspace{-2.7mm}
      \includegraphics[width=0.083\textwidth]{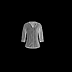} &
      \hspace{-2.7mm}
      \includegraphics[width=0.083\textwidth]{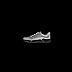} &
      \hspace{-2.7mm}
      \includegraphics[width=0.083\textwidth]{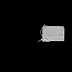} &
      \hspace{-2.7mm}
      \includegraphics[width=0.083\textwidth]{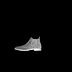} &
      \\
    \end{tabular} 

  \vspace{0.5mm}
  
    \begin{tabular}{cccccccccc}
      \includegraphics[width=0.083\textwidth]{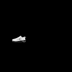} &
      \hspace{-2.7mm}
      \includegraphics[width=0.083\textwidth]{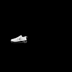} &
      \hspace{-2.7mm}
      \includegraphics[width=0.083\textwidth]{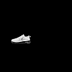} &
      \hspace{-2.7mm}
      \includegraphics[width=0.083\textwidth]{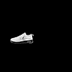} &
      \hspace{-2.7mm}
      \includegraphics[width=0.083\textwidth]{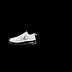} &
      \hspace{-2.7mm}
      \includegraphics[width=0.083\textwidth]{fashion_tr_img39_sc1189.png} &
      \hspace{-2.7mm}
      \includegraphics[width=0.083\textwidth]{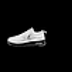} &
      \hspace{-2.7mm}
      \includegraphics[width=0.083\textwidth]{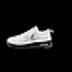} &
      \hspace{-2.7mm}
      \includegraphics[width=0.083\textwidth]{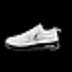} &
      \\
    \end{tabular}
  \caption{Sample images from the rescaled (top) Fashion-MNIST and (bottom) Fashion-MNIST with translations datasets. The top rows in both the subfigures depict examples from size factor 1 test set, for each of the 10 classes: ``T-shirt/top'', ``trouser'', ``pullover'', ``dress'', ``coat'', ``sandal'', ``shirt'', ``sneaker'', ``bag'' and ``ankle boot''. The bottom rows in both the subfigures show a single test sample for all scale variations between 1/2 and 2. Zero padding is used to obtain the final 72$\times$72 image size.}
  \label{fig-fashion_collages}
\end{figure*}

\begin{figure*}[htbp]
    \begin{tabular}{ccccccccccc}
      \includegraphics[width=0.083\textwidth]{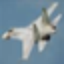} &
      \hspace{-2.7mm}
      \includegraphics[width=0.083\textwidth]{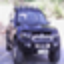} &
      \hspace{-2.7mm}
      \includegraphics[width=0.083\textwidth]{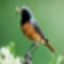} &
      \hspace{-2.7mm}
      \includegraphics[width=0.083\textwidth]{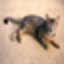} &
      \hspace{-2.7mm}
      \includegraphics[width=0.083\textwidth]{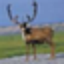} &
      \hspace{-2.7mm}
      \includegraphics[width=0.083\textwidth]{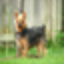} &
      \hspace{-2.7mm}
      \includegraphics[width=0.083\textwidth]{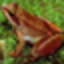} &
      \hspace{-2.7mm}
      \includegraphics[width=0.083\textwidth]{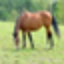} &
      \hspace{-2.7mm}
      \includegraphics[width=0.083\textwidth]{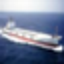} &
      \hspace{-2.7mm}
      \includegraphics[width=0.083\textwidth]{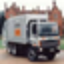} &
      \\
    \end{tabular} 

  \vspace{0.5mm}
  
    \begin{tabular}{cccccccccc}
      \includegraphics[width=0.083\textwidth]{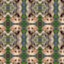} &
      \hspace{-2.7mm}
      \includegraphics[width=0.083\textwidth]{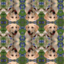} &
      \hspace{-2.7mm}
      \includegraphics[width=0.083\textwidth]{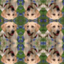} &
      \hspace{-2.7mm}
      \includegraphics[width=0.083\textwidth]{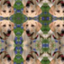} &
      \hspace{-2.7mm}
      \includegraphics[width=0.083\textwidth]{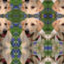} &
      \hspace{-2.7mm}
      \includegraphics[width=0.083\textwidth]{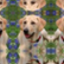} &
      \hspace{-2.7mm}
      \includegraphics[width=0.083\textwidth]{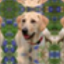} &
      \hspace{-2.7mm}
      \includegraphics[width=0.083\textwidth]{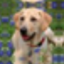} &
      \hspace{-2.7mm}
      \includegraphics[width=0.083\textwidth]{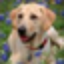} &
      \\
    \end{tabular} 
  \caption{Sample images from the rescaled CIFAR-10 dataset. The top row depicts examples from size factor 2 test set, for each of the 10 classes: ``airplane'', ``automobile'', ``bird'', ``cat'', ``deer'', ``dog'', ``frog'', ``horse'', ``ship'' and ``truck''. The bottom row shows a single test sample for all scaling factors between 1/2 and 2. Mirror extension is used to achieve an image size of $64 \times 64$ pixels for scaling factors $<$ 2, including the size factor 1.}
  \label{fig-cifar10_collage}
\end{figure*}

\section{Datasets with scale variations} 
\label{sec-generating-datasets}
In order to evaluate the scale generalisation properties of the GaussDerNets in more complex settings than provided by the MNIST Large Scale dataset, we have extended the Fashion-MNIST and the CIFAR-10 datasets with systematic spatial scale variations. In our scale generalisation experiments, we also investigate the STIR dataset.

\subsection{Fashion-MNIST with scale variations} 
The Fashion-MNIST dataset (Xiao et al. \citeyear{ZiaRasVol-arXiv}) consists of size 28$\times$28 gray-scale images of clothes from the 10 different classes: ``T-shirt/top'', ``trouser'', ``pullover'', ``dress'', ``coat'', ``sandal'', ``shirt'', ``sneaker'', ``bag'' and ``ankle boot''. All the images have a black background and the object in the frame is always centred. The dataset contains very little scale variations between objects within the same class, making it well suited for creating a dataset with well-defined scale variations.

The rescaled Fashion-MNIST dataset is generated by rescaling the original Fashion-MNIST images, using bicubic interpolation with default anti-aliasing in Matlab, with the considered image scaling factors $S$ relative to the original image being $\{S : 1/2 \leq S \leq 2 \text{, } S = \sqrt[\leftroot{0}\uproot{1}4]{2}^{n} \text{ for } n \in \mathbb{Z}\}$. All the images are zero padded to size 72$\times$72, to ensure an ample distance between the image boundary and the object. While the original images have a very small pixel resolution, they contain very rudimentary textures, where the prediction relies mainly on shapes, thus making this a suitable dataset. The top part of Figure~\ref{fig-fashion_collages} shows some representative images from this dataset.

The dataset is split into 50~000 training samples, 10~000 validation samples and 10~000 testing samples.

\subsubsection{Translated Fashion-MNIST with scale variations}
To investigate the spatial max pooling approach for handling image data where the objects are not centred, a Fashion-MNIST dataset with combined scale variations and translated objects is also created. It is generated in exactly the same way as the rescaled Fashion-MNIST dataset, with the only difference being that the objects have also been randomly shifted in the spatial domain, up to 4 pixels away from the image boundary. See the bottom part of Figure~\ref{fig-fashion_collages} for example images from this dataset.

The dataset is split into 50~000 training samples, 10~000 validation samples and 10~000 testing samples.

\subsection{CIFAR-10 with scale variations} 
The CIFAR-10 dataset (Krizhevsky and Hinton \citeyear{KriHin09-CIFAR}) consists of size 32$\times$32 tightly cropped real-world RGB images of animals and vehicles from the 10 distinct classes: ``airplane'', ``automobile'', ``bird'', ``cat'', ``deer'', ``dog'', ``frog'', ``horse'', ``ship'' and ``truck''. This implies that the dataset comprises a significantly larger variety of natural textures and patterns.

The generation of the new CIFAR-10 with scale variations dataset also uses bicubic interpolation with default anti-aliasing in Matlab to rescale the original CIFAR-10 images with scaling factors $\{S : 1/2 \leq S \leq 2 \text{, } S = \sqrt[\leftroot{0}\uproot{1}4]{2}^{n} \text{ for } n \in \mathbb{Z}\}$. For scaling factors less than 2, the images are first extended by mirroring at the image boundaries, and then, after the interpolation stage, they are cropped to size 64$\times$64. Figure~\ref{fig-cifar10_collage} shows a few images from the dataset.

The dataset is split into 40~000 training samples, 10~000 validation samples and 10~000 testing samples.

It should be noted, however, that in contrast to the input images for generating the rescaled FashionMNIST data set, which all contain objects of the same size, the input images used for generating the rescaled CIFAR-10 do comprise a notable variability in the sizes of the objects. Hence, the scale selection properties for the rescaled CIFAR-10 data set may be less distinct from what can be achieved with the rescaled FashionMNIST data set.

\subsection{The STIR traffic sign and the STIR aerial datasets}
\label{sec-stir}
For evaluating the scale generalisation properties of GaussDerNets, beyond the new datasets comprising spatial scaling variations introduced above, we will also perform experiments on the already existing Scaled and Translated Image Recognition (STIR) dataset, introduced in Altstidl et al. (\citeyear{AltNguSchKoeMutEskZan23-IJCNN}). 

We limit our investigation to the aerial and traffic sign subdivisions of this dataset. These two sub-datasets are generated from images containing objects within bounding boxes that are at least 64$\times$64. These objects are then downscaled to varying sizes, with the new bounding box lengths in the range $l \in [17,64]$, and randomly placed in images of size 64$\times$64. 

The traffic sign dataset, based on the Mapillary Traffic Sign Dataset by Ertler et al. (\citeyear{ertler2020-ECCV}), consists of RGB images of traffic signs from 16 classes, with 25 instances per class. It is split into 400 training samples, 400 validation samples and 400 testing samples, at each scale.

The aerial dataset, based on the Dataset for Object Detection in Aerial Images by Xia et al. (\citeyear{xia2018-CVPR}), consists of RGB images of aerial objects from 9 classes, with 25 instances per class. It is split into 225 training samples, 225 validation samples and 225 testing samples, at each scale.

\begin{table*}[hbtp]
  \centering
  \begin{tabular}{lcccc}
       \hline
       \textbf{Model Parameters} & \textbf{Symbol} & \textbf{rescaled Fashion-MNIST} & \textbf{rescaled CIFAR-10} & \textbf{STIR}\\ 
       \hline
       Number of scale channels & $N$ & 7 & 6  & 6\\ 
       Initial scale value range & [$\sigma_{i,0}$] & [$\sqrt{2}^{-3},2\sqrt{2}$] & [$\sqrt{2}^{-3},2$] & [$\sqrt{2}^{-3},2$] \\ 
       Relative scale ratio & $r$ & 1.28 & 1.45 & 1.45 \\ 
       Number of layers & $M$ & 6 & 6 & 6 \\ 
       Intermediate feature channels & $c$ & 32-48-64-96-128 & 64-96-128-160-192 &  32-48-64-96-128\\ 
       Spatial selection method & - & \text{central pixel} & \text{central pixel} & \text{spatial max pooling}\\ 
       Number of network parameters & - & 146k & 430k & 146k \\ 
       Input image size & - & 72$\times$72 & 64$\times$64 & 64$\times$64\\ \hline
  \end{tabular}
  \caption{Core multi-scale-channel model parameter values and methods used in our experiments, for each dataset. The same parameters are used for the rescaled Fashion-MNIST with translations dataset as for the rescaled Fashion-MNIST dataset, with the exception of using spatial max pooling as the spatial selection method.}
  \label{tab:model-parameters}
\end{table*}

\begin{table*}[hbtp]
  \centering
  \begin{tabular}{lccc}
       \hline
       \textbf{Training Parameters} & \textbf{rescaled Fashion-MNIST} & \textbf{rescaled CIFAR-10} & \textbf{\enspace \enspace STIR \enspace \quad}\\ 
       \hline
       Initial learning rate & 0.01 & 0.01 & 0.01\\ 
       Batch size & 32 & 30 & 32\\ 
       Number of training epochs & 32 & 60 & 24\\
       Weight decay & 0.05 & 0.025 & 0.025 \\  \hline
  \end{tabular}
  \caption{Training details for multi-scale-channel model used in our experiments, for each dataset. The same training procedure is used for the rescaled Fashion-MNIST with translations dataset as for the rescaled Fashion-MNIST dataset.}
  \label{tab:training-parameters}
\end{table*}

\section{Network architecture and training configurations}
\label{sec-train-proced}

\subsection{Network architecture and structural hyperparameters}
In the experiments to be presented later in Sections~\ref{sec-experiments} and~\ref{sec-ablation}, we will evaluate the performance of the GaussDerNets for each one of the datasets described in Section~\ref{sec-generating-datasets}, using the core model parameters summarised in Table~\ref{tab:model-parameters}.

The datasets used for investigating the influence of scaling variations in this work are based on test sets with size factors in the range [1/2, \,2]. Therefore, the initial scale values $\sigma_{i,0}$ of the multi-scale-channel networks have been manually selected to cover that range of scales, distributed using a scale spacing factor of $\sqrt{2}$. For training on the Fashion-MNIST dataset, we used the scale channels with the initial scale values $\sigma_{i,0} \in \{1/(2\sqrt{2}), 1/2, 1/\sqrt{2}, 1, \sqrt{2}, 2, 2\sqrt{2} \}$. For training on the CIFAR-10 dataset and the STIR datasets, the boundary scale channel with initial scale value $2\sqrt{2}$ was excluded, meaning we used the scale channels with initial scale values $\sigma_{i,0} \in \{1/(2\sqrt{2}), 1/2, 1/\sqrt{2}, 1, \sqrt{2}, 2\}$. Depending on the original image size, different values of the relative scale factor $r$ were used for the different datasets, as specified in Table 1.

Single-scale-channel networks are defined and trained in a similar way and using similar parameter settings as the multi-scale-channel networks, however, using only one scale channel in each deep network.

Table~\ref{tab:model-parameters} does also give a specification for how many feature channels were used in the intermediate layers of the deep networks used for the experiments, as well as details regarding the image size for each dataset and the use of central pixel extraction or spatial max pooling for spatial selection in the model.

The GaussDerNets considered in our experiments use either central pixel extraction or spatial max pooling stage for spatial selection after the final network layer, while regarding permutation-invariant scale pooling either average pooling or max pooling over scales is used. Which combination is used for a given experiment will be specified in Sections~\ref{sec-experiments}--\ref{sec-network-function-visualisation}.

\begin{table*}[hbtp]
  \centering
  \begin{tabular}{ll}
       \hline
       \textbf{Section} & \textbf{Topic}\\ 
       \hline
       \ref{sec-fashion-experiments} & \textbf{Experiments with GaussDerNets on the rescaled Fashion-MNIST dataset}\\ 
       \ref{sec-fashion-expm} & Scale generalisation properties of single- and multi-scale-channel GaussDerNets\\ 
       \ref{sec-fashion-scale-sel-hist} & Scale selection properties of GaussDerNets based on different permutation invariant pooling methods\\ \hline
       \ref{sec-fashion-translation-experiments} & \textbf{Experiments with GaussDerNets on the rescaled Fashion-MNIST with translations dataset}\\ 
       \ref{sec-spatial-max-pooling} & Scale generalisation properties of GaussDerNets based on spatial max pooling over scales\\ \hline
       \ref{sec-cifar10-experiments} & \textbf{Experiments with GaussDerNets on the rescaled CIFAR-10 dataset}\\ 
       \ref{sec-cifar10-expm} & Scale generalisation properties of single- and multi-scale-channel GaussDerNets \\ 
       \ref{sec-cifar10-scale-sel-hist} & Scale selection properties of GaussDerNets based on different permutation invariant pooling methods\\ \hline 
       \ref{sec-stir-experiments} & \textbf{Experiments with GaussDerNets on the STIR datasets} \\ 
       \ref{sec-stir-expm} & Scale generalisation properties of GaussDerNets compared to other deep network architectures \\ \hline
  \end{tabular}
  \caption{Overview of the content covered by the subsections in Section~\ref{sec-experiments}, which present the key experimental results of this paper. These subsections are organised by dataset, with each detailing experiments conducted on a specific dataset. The aim of these experiments is to demonstrate scale generalisation and scale selection properties of GaussDerNets, on different datasets.}
  \label{tab:experiment-section-overview}
\end{table*}

\subsection{Training procedure and hyperparameters}
For each of the datasets, the predetermined data splits provided in Section~\ref{sec-generating-datasets} were used, with the validation sets used to find parameter settings and training schedule. In our experiments, the training and the validation sets were then combined to provide as much training data as possible, except for the STIR datasets.

All the networks were trained using the AdamW optimiser with the weight decay value adapted to the particular dataset. An initial learning rate was used that decayed with a cosine learning rate schedule towards a minimum learning rate of $10^{-5}$. No further training was done to avoid the risk of overfitting.
The batch size was adjusted based on model size, image size and the GPU capacity, with the number of training iterations adjusted accordingly. The settings used in our experiments are summarised in Table~\ref{tab:training-parameters}. The models were initialised using a uniform He initialisation and trained by minimising a cross-entropy loss.

When training the networks, scale-channel dropout was applied as described in Section~\ref{sec-scale-dropout}, with the dropout values specified for each experiment in Sections~\ref{sec-experiments},~\ref{sec-ablation} and~\ref{sec-network-function-visualisation}. 
We also used random horizontal flipping of the images with 50\% probability during training, except when training on the STIR datasets, to artificially increase the amount of samples. Random color jittering during training was only used for the CIFAR-10 dataset, using the inbuilt color jitter PyTorch transform with the settings (brightness, contrast, saturation, hue) = (0.2, 0.2, 0.2, 0.05). In one of our ablation studies, we also explored applying cutout during training, implemented with a square of size 16.

\section{Experiments on scale generalisation}
\label{sec-experiments}
In this section, we will show the results of applying the extended GaussDerNets to the new rescaled versions of the Fashion-MNIST and CIFAR-10 datasets, as well as to the previously created STIR datasets.
For the experimental evaluation, we will focus on (i)~the scale generalisation properties of the Gaussian derivative networks, in that training is only performed at a single scale, and then tested on a set of previously unseen scales. Additionally, we will (ii)~compare the effect different permutation-invariant scale pooling methods have on scale selection properties of the multi-scale-channel networks, and also (iii)~compare the performance of multi-scale-channel networks to the performance of single-scale-channel networks. Furthermore, we report the results of (iv)~performing spatial max pooling to handle objects that are not centered in the images. An overview of how these experiments are organised in this section is given in Table~\ref{tab:experiment-section-overview}.

\subsection{Experiments on the rescaled Fashion-MNIST dataset}
\label{sec-fashion-experiments}
\subsubsection{Single-scale-channel vs. multi-scale-channel networks}
\label{sec-fashion-expm}
For the rescaled Fashion-MNIST dataset, we trained multi-scale-channel GaussDerNets as described in Section~\ref{sec-train-proced}, using scale channels with $\sigma_{i,0} \in \{1/(2\sqrt{2}), 1/2$, $1/\sqrt{2}, 1, \sqrt{2}, 2, 2\sqrt{2} \}$.
Since the objects in the Fashion-MNIST dataset are centered, central pixel extraction was used. A single-scale-channel network was also trained in the same way as the multi-scale networks, with $\sigma_{0} = 1$. The multi-scale-channel model using max pooling was trained with scale-channel dropout with $q=0.2$, while the average pooling model was trained without scale-channel dropout. We will investigate effects of scale-channel drop\-out in Section~\ref{sec-ablation-regularisation}.

\begin{figure}[ht]
  \begin{center}
       \textit{\quad \quad Scale generalisation on rescaled Fashion-MNIST}
       \includegraphics[width=0.48\textwidth]{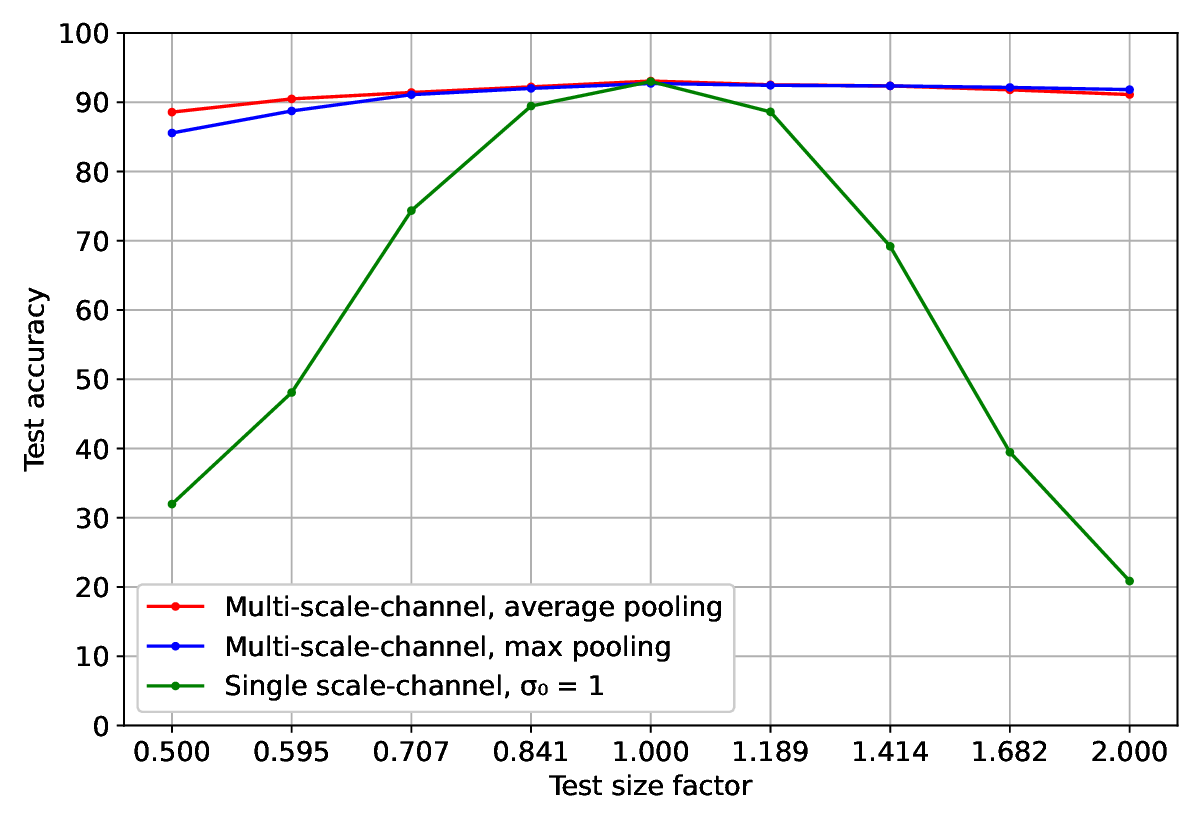}
  \end{center}
\caption{Scale generalisation curves for a single-scale-channel network, and two multi-scale-channel networks using either max pooling over scale or average pooling over scale, applied to the Fashion-MNIST dataset subject to scale variations. In this experiment, all the models were trained for the single size factor 1 in the dataset, and then tested for all the size factors between 1/2 and 2. For test data with the same size factor 1 as the training data, all the models achieved an accuracy of $\sim$93\%. As can be seen from the graphs, the multi-scale-channel networks have much better scale generalisation properties compared to the single-scale-channel network, in agreement with the theoretical formulation of the scale-covariant and scale-invariant GaussDerNets.}
\label{fig-fashion-scale-gen}
\end{figure}

Figure~\ref{fig-fashion-scale-gen} shows the results for this experiment, with training performed at the single size factor 1 and testing performed for all size factors between 1/2 and 2, and using either max pooling over scale or average pooling over scale for the multi-scale channel networks. As can be seen from the graphs, the testing performance is rather flat over scales for the multi-scale-channel network, while the testing performance decreases rapidly with deviations between the testing scales and the training scale for the single-scale-channel network. 

This result is overall consistent with predictions from the theory, in that the multi-scale-channel network is in the ideal continuous case provably scale-invariant, while the single-scale-channel network does not have any such scale-invariant properties.

\begin{figure*}[hbt]
  \begin{center}
    \begin{subfigure}{0.47\textwidth}
        \centering
        \includegraphics[width=\linewidth]{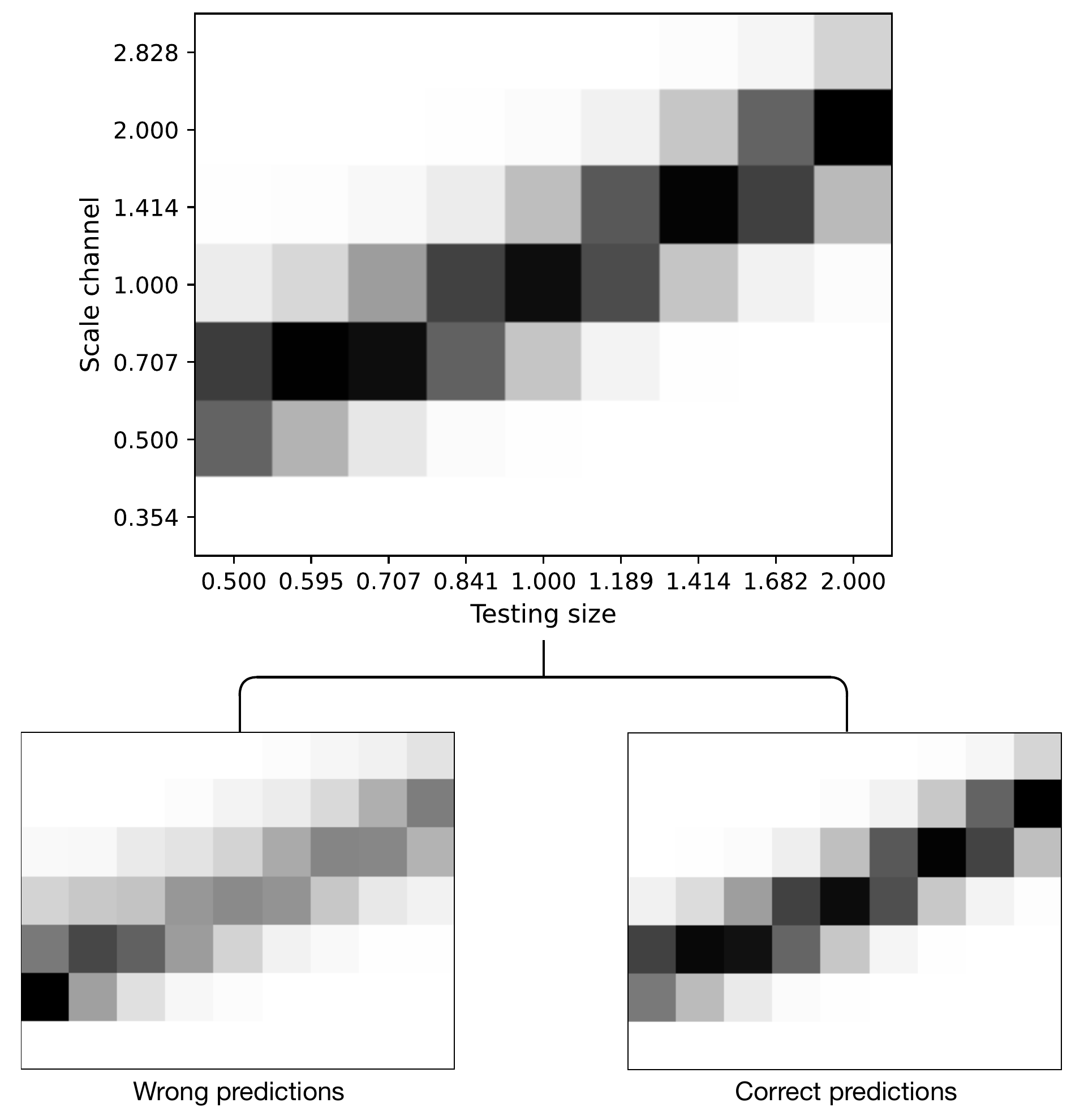}
        \caption{Max pooling model}
        \label{fig:subfigb-1}
    \end{subfigure}
    \hfill
    \begin{subfigure}{0.47\textwidth}
        \centering
        \includegraphics[width=\linewidth]{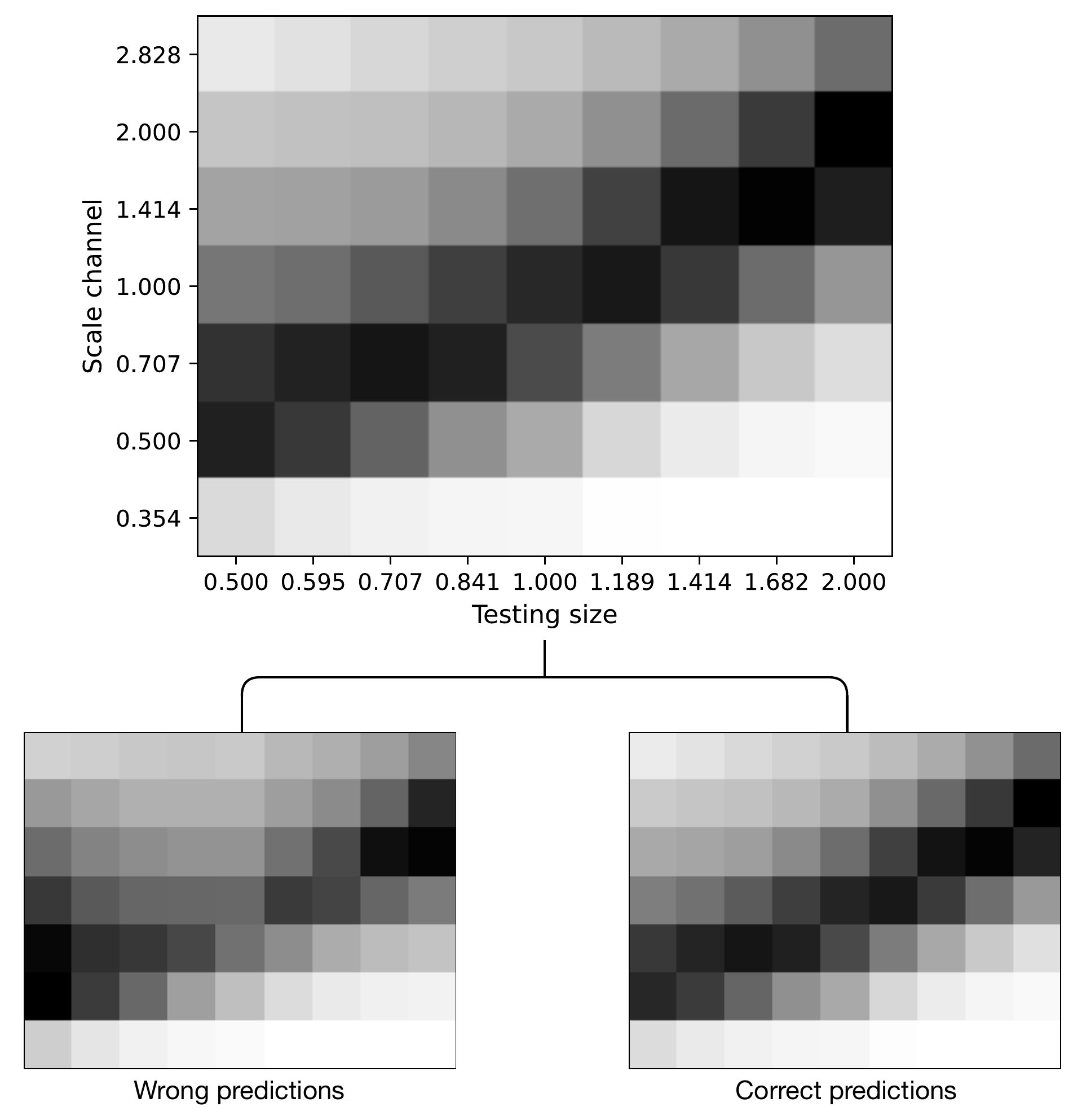}
        \caption{Average pooling model}
        \label{fig:subfiga-1}
    \end{subfigure}
  \end{center}
\caption{Scale selection histograms for (a) max pooling over scales and (b) average pooling over scales for the multi-scale-channel networks applied to the Fashion-MNIST dataset. These scale selection histograms visualise which scale channels contribute to the final prediction for either the max pooling operation over scales or the average pooling operation over scales. In the top part of the figure, the selection histograms depict the total histograms over all the test data samples, whereas in the bottom part a decomposition has been made, depending on whether a test sample was correctly or incorrectly classified. A clear linear trend can be seen in these scale selection histograms, in that the selected scale levels, as reflected by the scale channels, are proportional to the size in the testing data, this way reflecting the scale-covariant property of the GaussDerNets.}
\label{fig-fashion-scale-select}
\end{figure*}

For the multi-scale-channel networks, there is, however, a certain gradual drop in performance, of about 4-7~ppt between the size factor 1 and the size factor 1/2, likely caused by sampling effects and loss of information, when reducing the image size by a factor of 2. Towards larger size factors up to 2, the performance is, however, almost the same as for the size factor 1, at which the training was performed. Concerning the choice between average pooling or max pooling over scales, average pooling does here lead to somewhat higher performance than max pooling for size factors $<$ 1, whereas max pooling leads to marginally better performance than average pooling for the size factor 2.

\subsubsection{Scale selection histograms}
\label{sec-fashion-scale-sel-hist}

In an ideal trained multi-scale-channel network, one would expect that the
classification of an input with a certain scale is determined from the
corresponding scale-channel(s) with (roughly) equal value(s) of the
scale-parameter(s) as the scale in the input. Specifically, if varying
the size of the image structures in the input data by rescaling the
input images with a uniform scaling factor, we would expect the
scale levels from which the classification is based to increase
linearly with the spatial rescaling factor of the input data.
Experimentally, we can thus qualitatively inspect this property
by accumulating scale-selection histograms, that reflect which scale
channels contribute to the classification for each spatial scaling factor
in a scale generalisation experiment.

In Figure~\ref{fig-fashion-scale-select}, we visualise such scale selection properties of the average pooling and max pooling multi-scale-channel models using scale selection histograms. The upper part of Figure~\ref{fig-fashion-scale-select} shows scale selection histograms accumulated for the scale generalisation experiment applied to the rescaled Fashion-MNIST dataset. For each size factor of the data, a bin counter for the max pooling approach has been increased for the scale channel at which the maximum response over scale was assumed. For the average pooling approach, a corresponding bin increment has instead been accumulated, that reflects the relative contribution to the average pooling result from each scale channel.

As can be seen from these graphs, for both the max pooling approach and the average pooling approach, there is a clear linear trend in that the dominant response over the scale channels is proportional to the size of the image structures in the testing data, indicating that the discrete implementation of the network approximates the scale covariance and scale channel matching properties described in Section~\ref{sec-scale-covariance-properties} quite well.
In this respect, both approaches show structural similarities to scale selection approaches based on either local extrema over scale of scale-normalised derivatives (Lindeberg \citeyear{Lin97-IJCV}) or weighed averaging of scale-normalised feature responses over scale (Lindeberg \citeyear{Lin12-JMIV}), for which the selected scale levels can be formally shown to be proportional to the inherent scales in the image data.

For the max pooling approach, the linear trend of the selected scale channels is, however, sharper than for the average pooling approach. The average pooling approach does instead accumulate information from a wider span of scale channels, thus giving this scale pooling method a better ability to form its decisions based on multiple scale cues, as opposed to a single cue as for the max pooling approach.

In the lower part of Figure~\ref{fig-fashion-scale-select}, we have additionally made a separation between the contributions to the scale selection histogram that originate from correct vs. incorrect classifications. As can be seen from these results, for both the max pooling approach and the average pooling approach, there are strong peaks in the wrong classifications for the scale channels corresponding to the finest scales, possibly caused by a loss of information when rescaling images to a substantially smaller size. Additionally, for the average pooling approach, there is also a peak in the scale selection histogram of wrong predictions for the largest size factor, possibly caused by interference with the outer scale of the image.

\subsection{Experiments on the rescaled Fashion-MNIST with translations dataset}
\label{sec-fashion-translation-experiments}

\subsubsection{Robustness of spatial max pooling under spatial translations}
\label{sec-spatial-max-pooling}
For the Fashion-MNIST dataset with scale variations and translations, we trained multi-scale-channel networks with central pixel extraction substituted by spatial max pooling, defined by Equation~(\ref{eq:spatial-maxpooling-expression}), to investigate its applicability as a spatial selection method. 

The networks used either average pooling over scales or max pooling over scales, and were trained using the same parameters and methods as for the rescaled Fashion-MNIST dataset presented in Section~\ref{sec-train-proced}, using scale-channel dropout with $q=0.2$ applied during training. The scale channels $\sigma_{i,0} \in \{1/(2\sqrt{2}), 1/2, 1/\sqrt{2}, 1, \sqrt{2}, 2, 2\sqrt{2}\}$ were used for these networks.

\begin{figure}[ht]
  \begin{center}
       \includegraphics[width=0.48\textwidth]{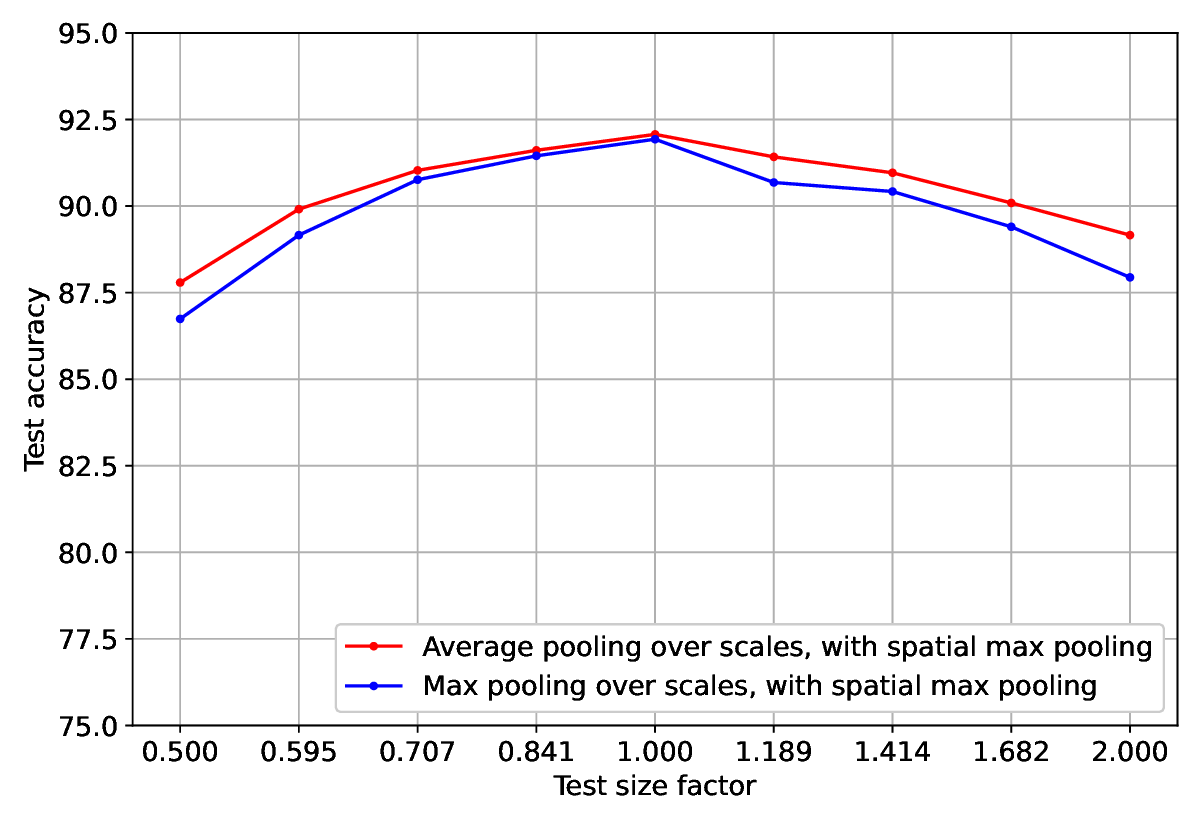}
  \end{center}
\caption{Scale generalisation performance on the rescaled Fashion-MNIST with translations dataset, for multi-scale-channel GaussDerNets with {\em spatial max pooling\/} as the spatial selection mechanism, and using either average pooling over scales or max pooling over scales as the scale selection mechanism. In this experiment, the networks were trained on the training data for size factor 1, and then evaluated on test data for all the size factors between 1/2 and 2. As can be seen from the graph, rather flat scale generalisation curves are obtained, confirming that spatial max pooling can be used as a mechanism to perform combined spatial and scale selection in Gaussian derivative networks.}
\label{fig-fashion-translated-scale-gen}
\end{figure}

Figure~\ref{fig-fashion-translated-scale-gen} shows the results of this experiment, with the networks trained on image data for the size factor 1 and tested on image data for all the size factors ranging from 1/2 to 2. As can be seen from the graph, incorporation of a spatial max pooling mechanism into the GaussDerNet makes it possible for the network to handle objects with different positions in the image domain, while maintaining the scale generalisation properties of the network to a reasonable extent. The network using average pooling over scales is found to achieve a slightly flatter scale generalisation curve compared to the network using max pooling over scales.
The scale generalisation curves are rather similar to the scale generalisation results shown in Figure~\ref{fig-fashion-scale-gen}, for the network using central pixel extraction and trained on the regular rescaled Fashion-MNIST dataset, which only contains centred objects. This demonstrates that a Gaussian derivative network using spatial max pooling as a spatial selection mechanism can effectively learn to detect maxima at different spatial locations and perform scale selection on this simple dataset. 

Using a network with central pixel extraction as the spatial selection method for training on this dataset is clearly not suitable, due to the off-centre nature of the objects in the dataset.

\subsection{Experiments on the rescaled CIFAR-10 dataset}
\label{sec-cifar10-experiments}

\subsubsection{Single-scale-channel vs. multi-scale-channel networks}
\label{sec-cifar10-expm}
For the rescaled CIFAR-10 dataset, we trained multi-scale-channel GaussDerNets as described in Section~\ref{sec-train-proced}, using scale channels with $\sigma_{i,0} \in \{1/(2\sqrt{2}), 1/2, 1/\sqrt{2}, 1$, $\sqrt{2}, 2\}$. Central pixel extraction was used for all the models, due to the objects in the dataset generally being well centred. A single-scale-channel network was also trained in the same way as the multi-scale networks, with $\sigma_{0} = 1$. The multi-scale-channel models were trained using scale-channel dropout with $q=0.3$.

\begin{figure}[ht]
  \begin{center}
       \textit{\quad \quad Scale generalisation on rescaled CIFAR-10}
       \includegraphics[width=0.48\textwidth]{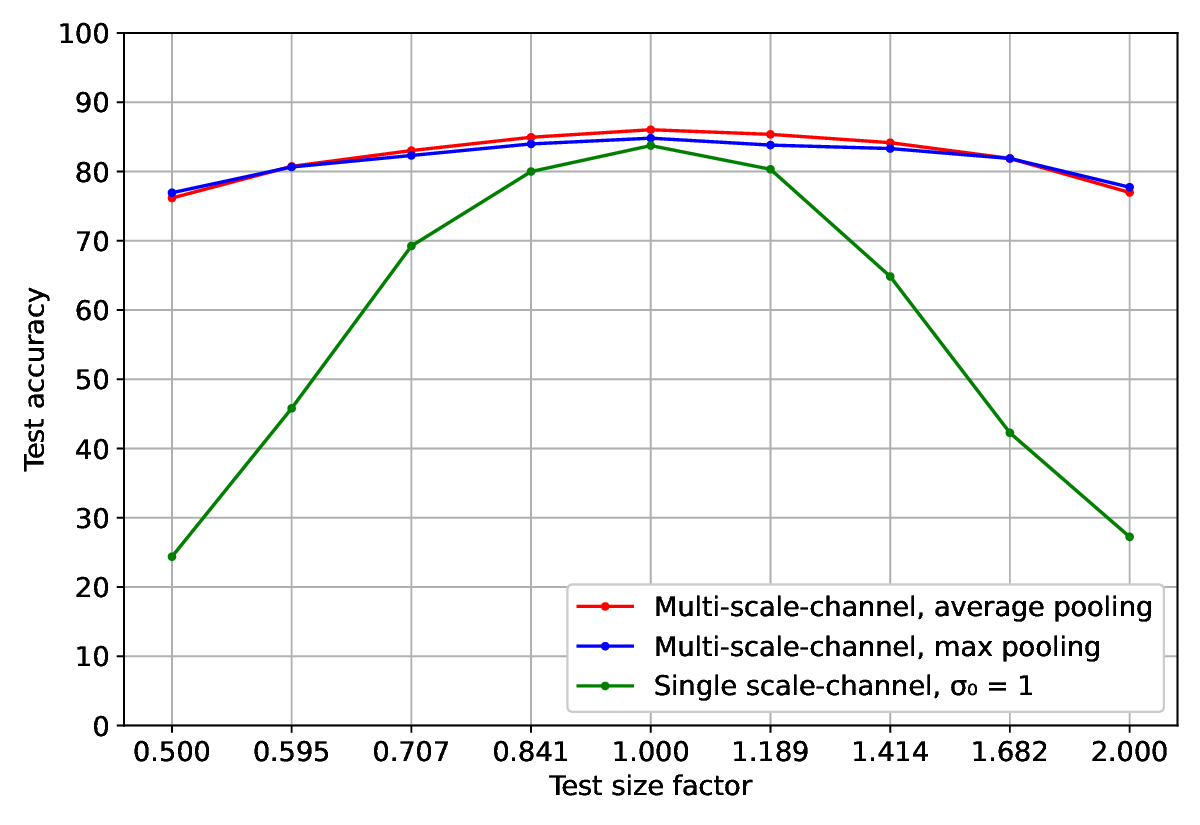}
  \end{center}
\caption{Scale generalisation curves for a single-scale-channel network and two multi-scale-channel networks applied to the CIFAR-10 dataset subject to scale variations, using either max pooling over scale or average pooling over scale for the multi-scale-channel networks. In this experiment, all the models were trained on the training set (corresponding to scale factor 1) and then evaluated on test data for all the size factors between 1/2 and 2. For test data with size factor 1, the same as the training data, the multi-scale-channel models achieved an accuracy of 85-86\%. As can be seen from the graphs, the multi-scale-channel networks have much better scale generalisation properties compared to the single-scale-network, similar to the findings obtained for the rescaled Fashion-MNIST dataset.}
\label{fig-cifar10-scale-gen}
\end{figure}

The results from this experiment are presented in Figure~\ref{fig-cifar10-scale-gen}. Just as for the rescaled Fashion-MNIST dataset, training was performed at the single size factor 1 and testing performed for all the size factors between 1/2 and 2, and using either max pooling over scale or average pooling over scale for the multi-scale channel networks. The results presented in the graph show similar behaviour as observed for the rescaled Fashion-MNIST dataset, with the multi-scale-channel networks having a much better ability to generalise to previously unseen scales, thus resulting in a relatively flat testing performance over scales, while the testing performance of the single-scale channel network forms a peak that decreases rapidly when moving towards testing scales further away from the training scale. 

The single-scale-channel model actually under-performs slightly on size factor 1 compared to the multi-scale-channel models, possibly because of the benefit of using scale-channel dropout during the training of multi-scale-channel networks.

\begin{figure*}[hbt]
  \begin{center}
    \begin{subfigure}{0.47\textwidth}
        \centering
        \includegraphics[width=\linewidth]{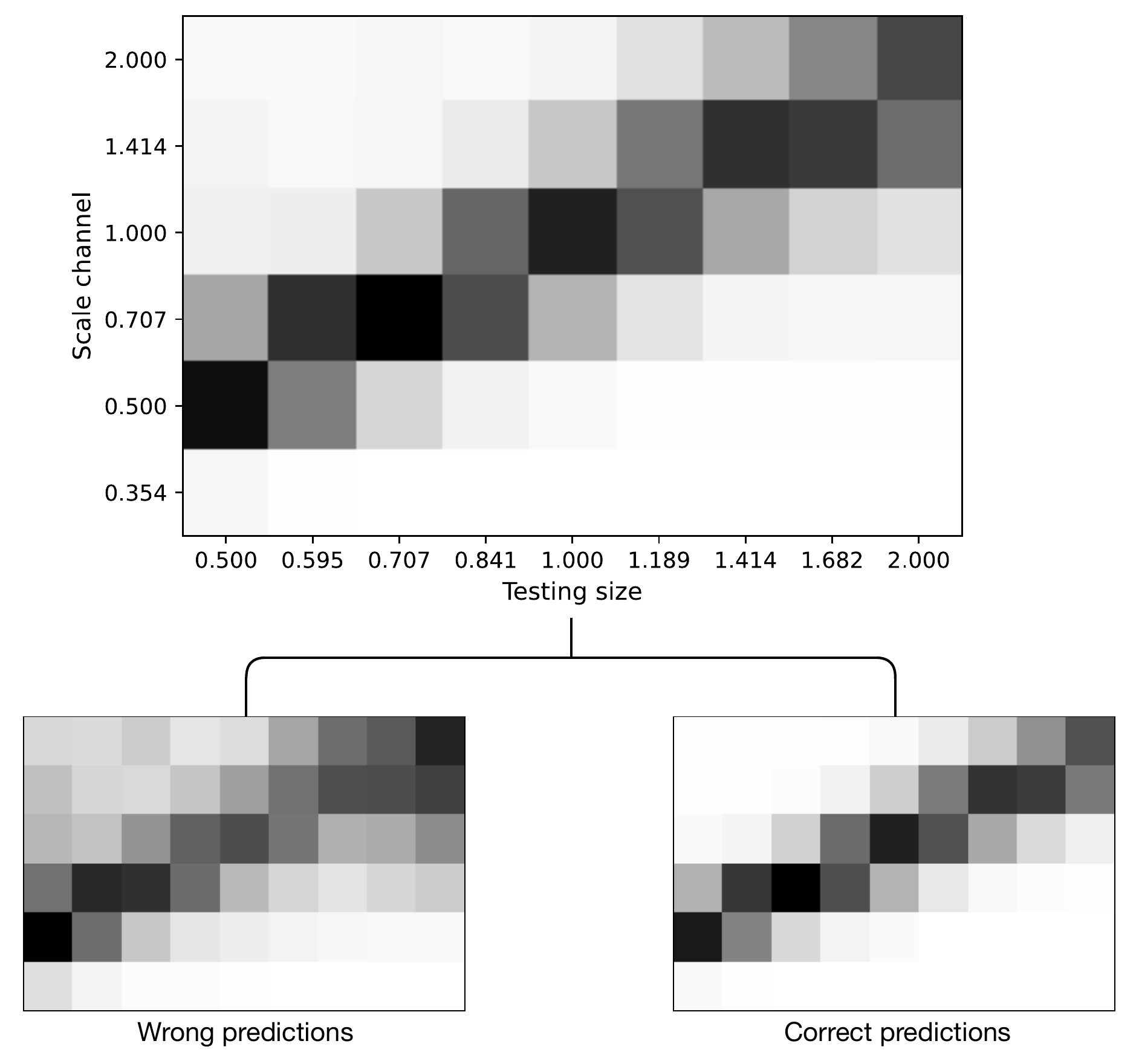}
        \caption{Max pooling model}
        \label{fig:subfigb-2}
    \end{subfigure}
    \hfill
    \begin{subfigure}{0.47\textwidth}
        \centering
        \includegraphics[width=\linewidth]{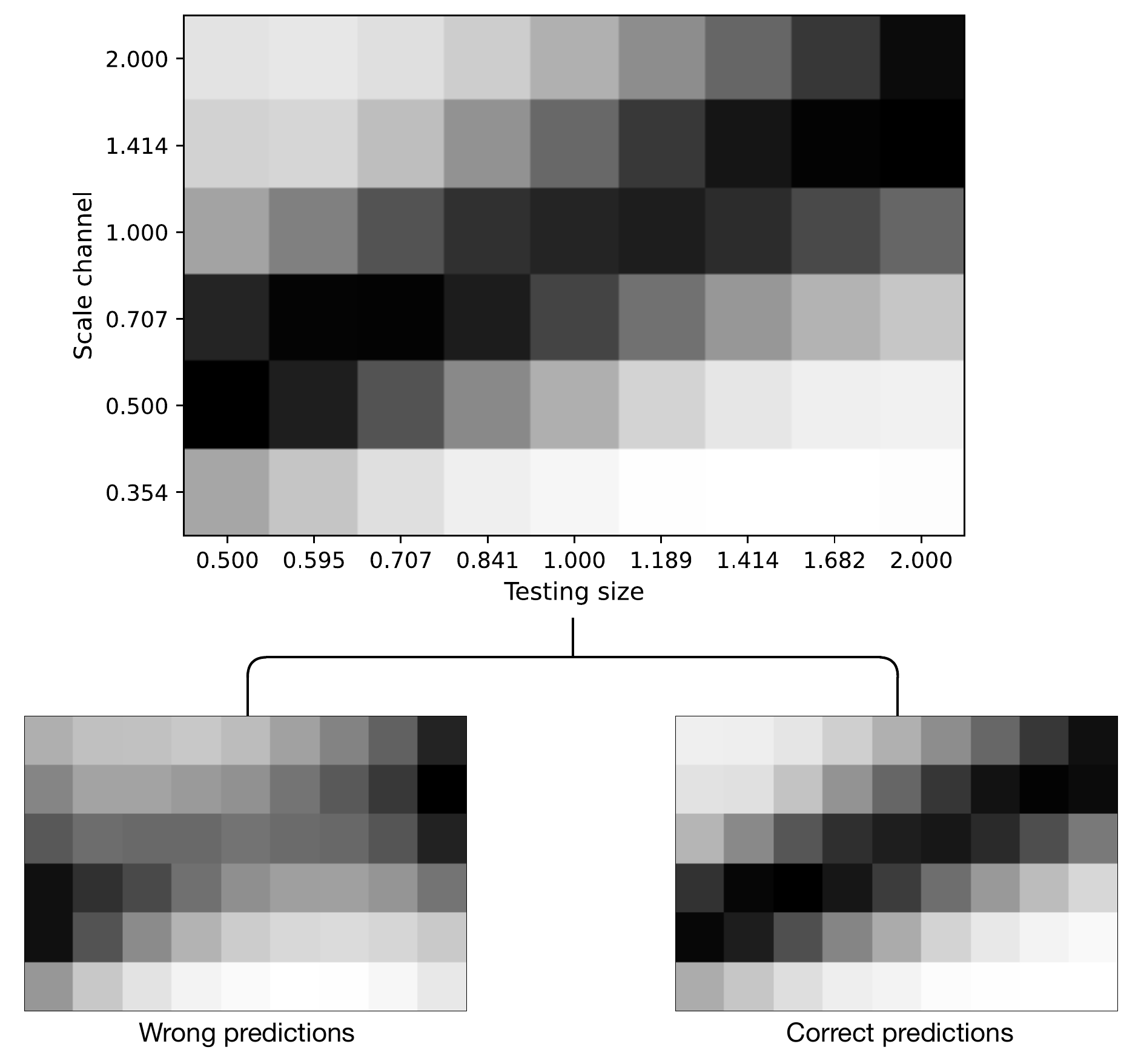}
        \caption{Average pooling model}
        \label{fig:subfiga-2}
    \end{subfigure}
  \end{center}
\caption{Scale selection histograms for (a) max pooling over scales and (b) average pooling over scales for the multi-scale-channel networks applied to the CIFAR-10 dataset. These scale selection histograms visualise which scale channels contribute to the final prediction for either the max pooling operation over scales or the average pooling operation over scales. In the top part of the figure, the selection histograms depict the total histograms over all the test data samples, whereas in the bottom part a decomposition has been made, depending on whether a test sample was correctly or incorrectly classified. As can be seen from these scale selection histograms, there is a clear linear trend in that the selected scale levels, as reflected by the scale channels, are proportional to the size in the testing data, similar to the findings obtained for the rescaled Fashion-MNIST dataset. This again reflects the scale-covariant property of the GaussDerNets.}
\label{fig-cifar10-scale-select}
\end{figure*}

The scale generalisation curves obtained from the multi-scale-channel networks are not as flat as for the rescaled Fashion-MNIST dataset; there is a gradual drop in performance of about 8-10~ppt between the size factor 1 and the size factor 1/2, likely caused by sampling effects and loss of information when reducing the image size by a factor of 2. Towards larger size factors, the performance also gradually drops, by 7-9~ppt between the size factor 1 and the size factor 2. These scale generalisation results are therefore not as clear as for the previously studied rescaled Fashion-MNIST dataset, possibly caused by the mirroring process used to extend the images, the existence of background textures, or other properties of the dataset. 

The scale generalisation performance for scaling factors less than 1 is indeed rather close to previously reported results for foveated scale-channel networks (see Figure 15 in Jansson and Lindeberg \citeyear{JanLin22-JMIV}). The results for scaling factors greater than 1 are, however, not comparable, because their image data were not increased in image size for scaling factors greater than 1.

In Appendix~A.1 in the supplementary material, we further demonstrate the scale-covariant properties of the GaussDerNet architecture, where we (i)~evaluate single-scale-channel networks using initial scale values other than the ones they were trained with, and (ii)~show that optimising the entire multi-scale-channel network during training gives significantly better scale generalisation properties compared to basing the multi-scale-channel network on weights transferred from a trained single-scale-channel network, which is also possible due to the weight sharing.

\subsubsection{Scale selection histograms}
\label{sec-cifar10-scale-sel-hist}
The upper part of Figure~\ref{fig-cifar10-scale-select} shows scale selection histograms accumulated for the scale generalisation experiment applied to the rescaled CIFAR-10 dataset. For each size factor of the data, a bin counter for the max pooling approach has been increased for the scale channel with the global maximum over scale. For the average pooling approach, a corresponding bin increment has instead been accumulated, that reflects the relative contribution to the average pooling result from each scale channel.

The results shown in the graph are manifestly similar to scale selection histograms obtained for the rescaled Fashion-MNIST dataset in Figure~\ref{fig-fashion-scale-select}. A linear trend between the dominant responses over the scale channels and the image scale is found for both the average and the max pooling approaches, resembling the responses expected from classical methods for scale selection, and being consistent with the scale-covariant properties of the GaussDerNets described in Section~\ref{sec-scale-covariance-properties}. The average pooling approach once again results in a less localised linear trend than for max pooling, due to its use of cues across multiple scales when computing the predictions.

The lower part of Figure~\ref{fig-cifar10-scale-select} shows a split between the contributions to the scale selection histogram into correct and incorrect classifications. As can be seen from these results, for both the max pooling approach and the average pooling approach, there are more pronounced peaks in both wrong and correct classifications for the scale channels corresponding to the finest and coarsest scales, likely due to the lower performance at the corresponding size factors.

\begin{figure*}[hbt]
  \centering
  \textit{\hspace{0.2cm} Training with small object sizes \hspace{1.4cm} Training with medium object sizes \hspace{1.4cm} Training with large object sizes}
  \includegraphics[width=0.32\textwidth]{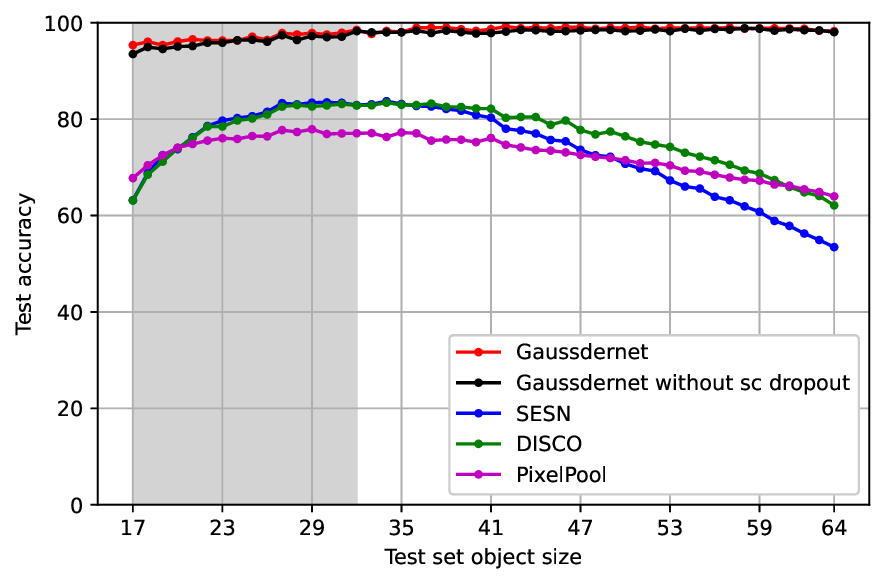} 
  \hspace{1mm}
  \includegraphics[width=0.32\textwidth]{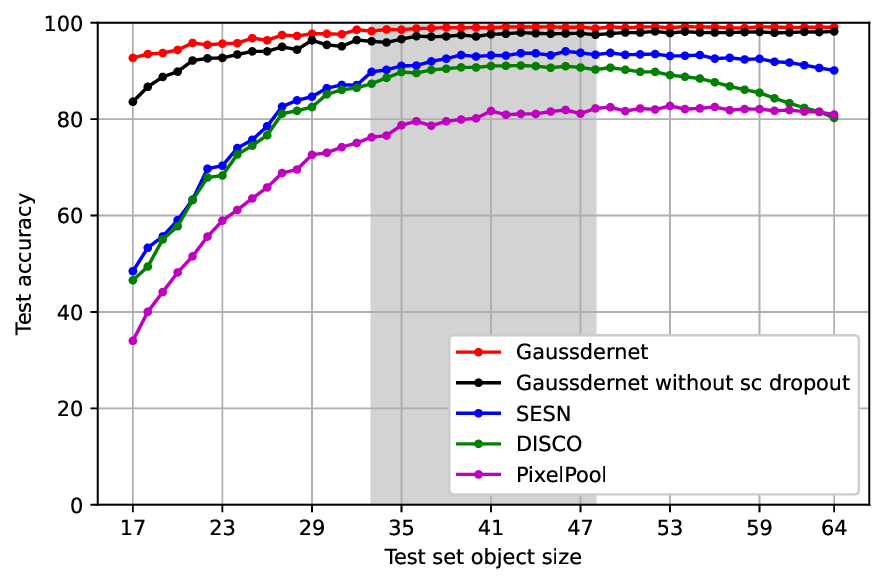} 
  \hspace{1mm}
  \includegraphics[width=0.32\textwidth]{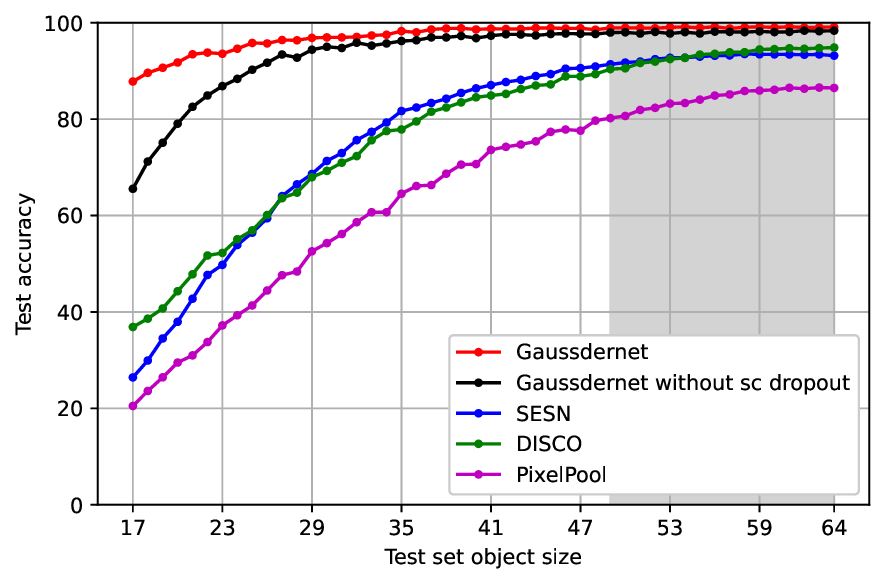} 
  \caption{Scale generalisation curves for the STIR traffic sign dataset, obtained using (i) the GaussDerNet with average pooling over scales, (ii) the scale-equivariant steerable (SESN) network (Sosnovik et al. \citeyear{SosSzmSme20-ICLR}), also referred to as the Hermite model by Altstidl et al. (\citeyear{AltNguSchKoeMutEskZan23-IJCNN}), (iii) the discrete scale convolutions (DISCO) network (Sosnovik et al. \citeyear{SosMosSme21-BMVC}), and (iv) the PixelPool network (Altstidl et al. \citeyear{AltNguSchKoeMutEskZan23-IJCNN}). (Left) Models trained on {\em smaller-size\/} training data, with object bounding box length $l \in [17,32]$. (Middle) Models trained on {\em middle-size\/} training data, with object bounding box length $l \in [33,48]$. (Right) Models trained on {\em larger-size\/} training data, with object bounding box length $l \in [49,64]$. All the models are evaluated on test data at all the scales, $l \in [17,64]$, with the experimental results for the GaussDerNet computed as the mean over 5 runs. The experimental results for the SESN, DISCO and PixelPool models have been averaged over 25 runs, based on experiments performed by (Altstidl et al. \citeyear{AltNguSchKoeMutEskZan23-IJCNN}). The object sizes seen during training are shaded in grey.} 
  \label{fig-stir-sign-plot}
\end{figure*}

\begin{figure*}[hbt]
  \centering
  \textit{\hspace{0.2cm} Training with small object sizes \hspace{1.4cm} Training with medium object sizes \hspace{1.4cm} Training with large object sizes}
  \includegraphics[width=0.32\textwidth]{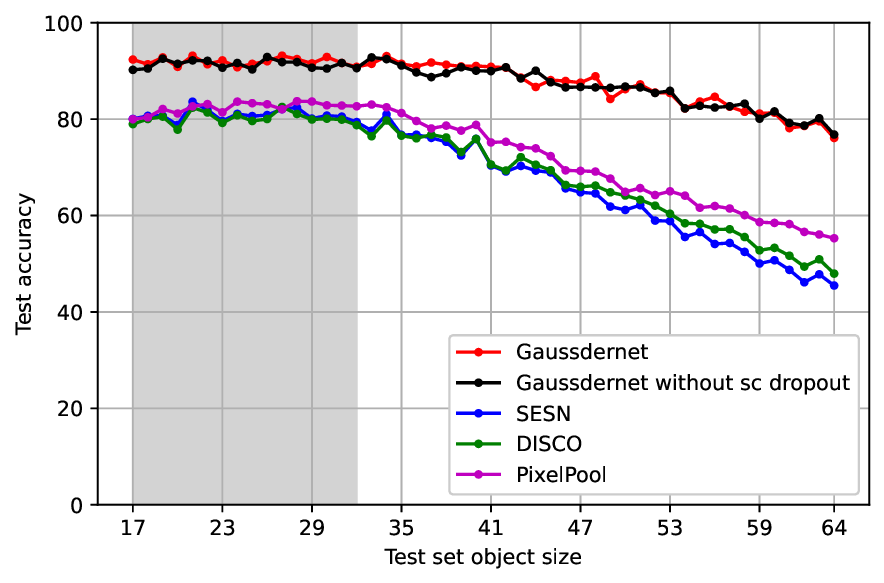} 
  \hspace{1mm}
  \includegraphics[width=0.32\textwidth]{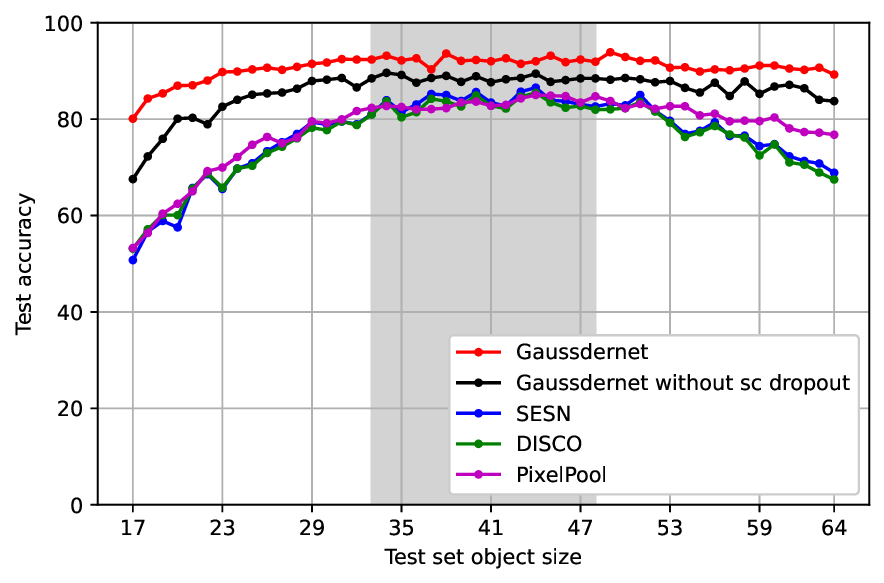} 
  \hspace{1mm}
  \includegraphics[width=0.32\textwidth]{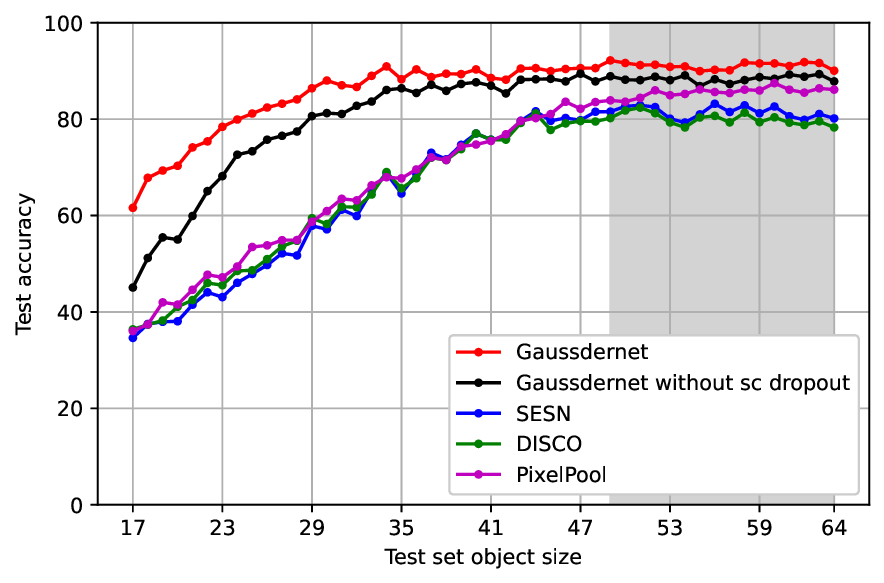} 
  \caption{Scale generalisation curves for the STIR aerial dataset, obtained using (i) the GaussDerNet with average pooling over scales, (ii) the scale-equivariant steerable (SESN) network (Sosnovik et al. \citeyear{SosSzmSme20-ICLR}), also referred to as the Hermite model by Altstidl et al. (\citeyear{AltNguSchKoeMutEskZan23-IJCNN}), (iii) the discrete scale convolutions (DISCO) network (Sosnovik et al. \citeyear{SosMosSme21-BMVC}), and (iv) the PixelPool network (Altstidl et al. \citeyear{AltNguSchKoeMutEskZan23-IJCNN}). (Left) Models trained on {\em smaller-size\/} training data, with object bounding box length $l \in [17,32]$. (Middle) Models trained on {\em middle-size\/} training data, with object bounding box length $l \in [33,48]$. (Right) Models trained on {\em larger-size\/} training data, with object bounding box length $l \in [49,64]$. All the models are evaluated on test data at all the scales, $l \in [17,64]$, with the experimental results for the GaussDerNet computed as the mean over 5 runs. The experimental results for the SESN, DISCO and PixelPool models have been averaged over 25 runs, based on experiments performed by (Altstidl et al. \citeyear{AltNguSchKoeMutEskZan23-IJCNN}). The object sizes seen during training are shaded in grey.}
  \label{fig-stir-aerial-plot}
\end{figure*}

\begin{figure*}[htbp]
  \begin{center}
    \begin{subfigure}{0.95\textwidth}
        \centering
        \includegraphics[width=0.337\textwidth]{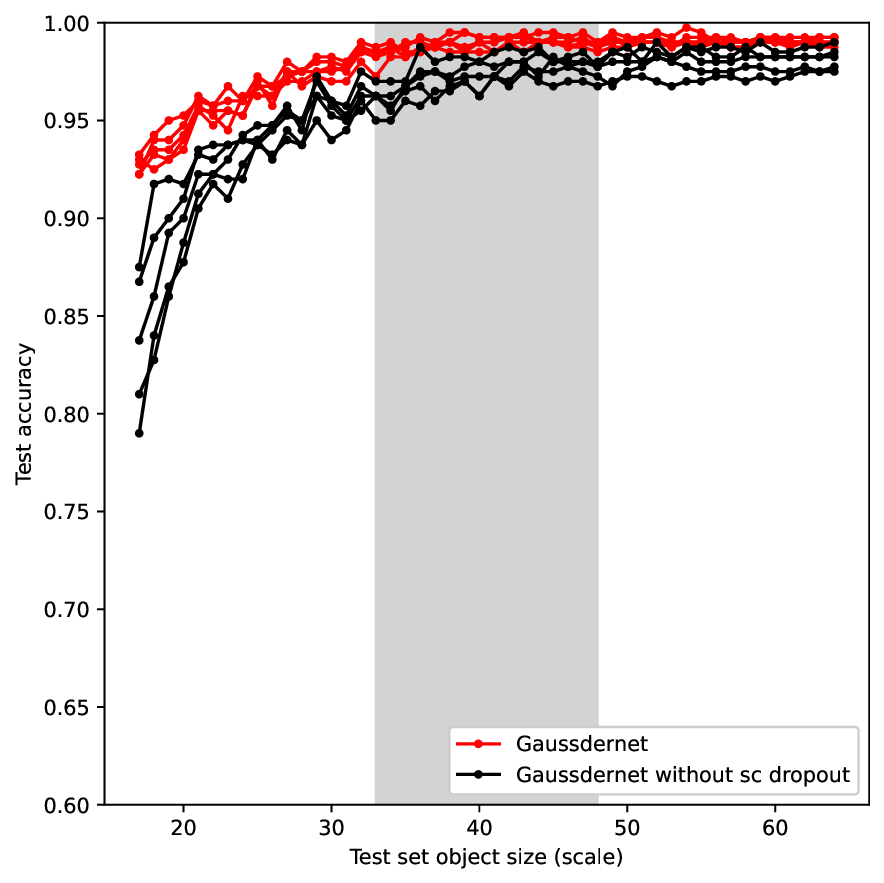}  
        \hspace{2cm}
        \raisebox{0.53mm}{\includegraphics[width=0.35\textwidth]{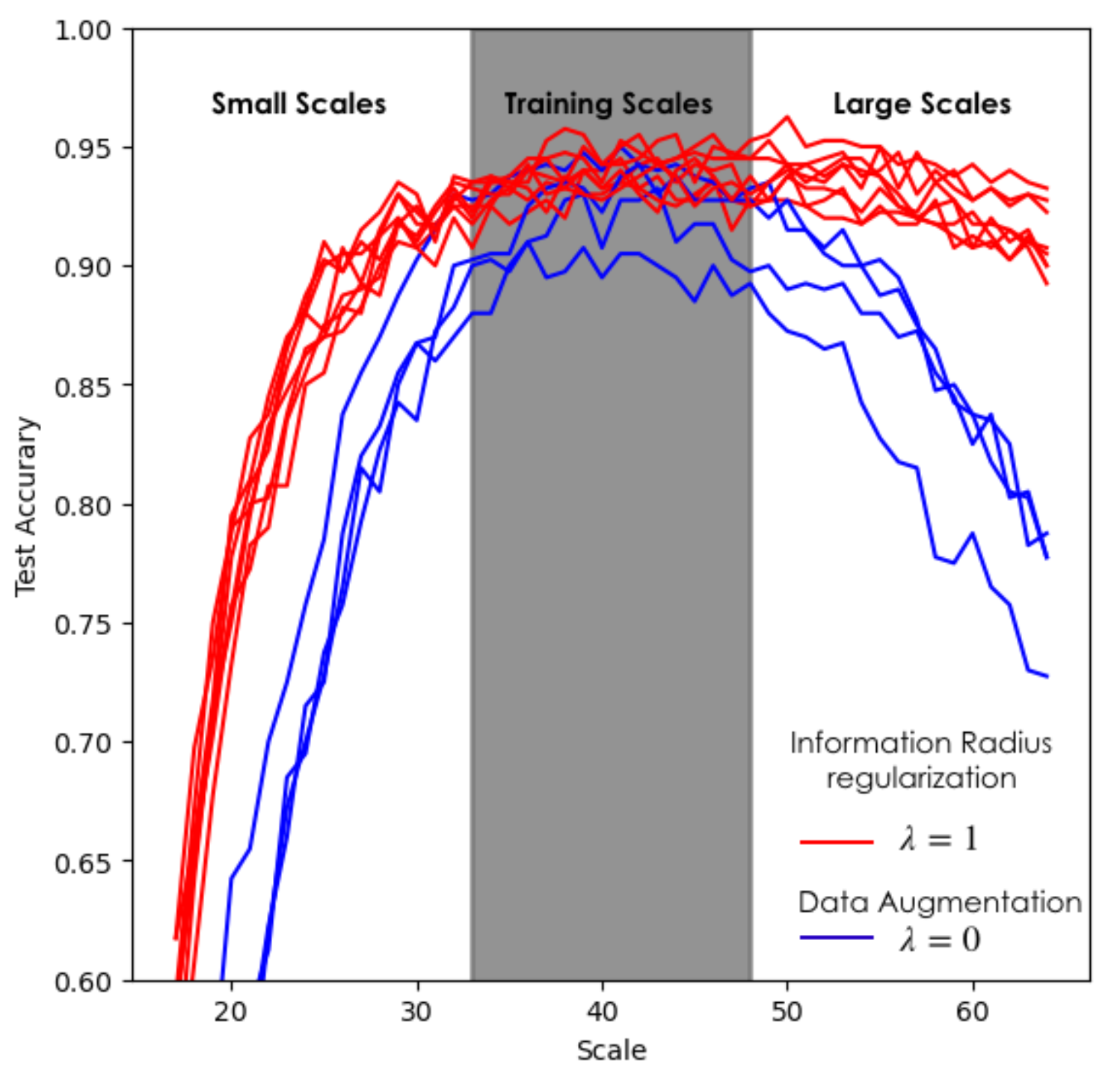}}
        \caption{Scale generalisation curves for the STIR traffic sign dataset, obtained using (left) our GaussDerNet, (right) the GD-CNN network by Velasco-Forero (\citeyear{Velasco2023-ICGSI}).}
        \label{fig:subfig-stir-sign-mid-comparison}
    \end{subfigure}
    
    \vspace{0.5cm} 
    
    \begin{subfigure}{0.95\textwidth}
        \centering
        \includegraphics[width=0.337\textwidth]{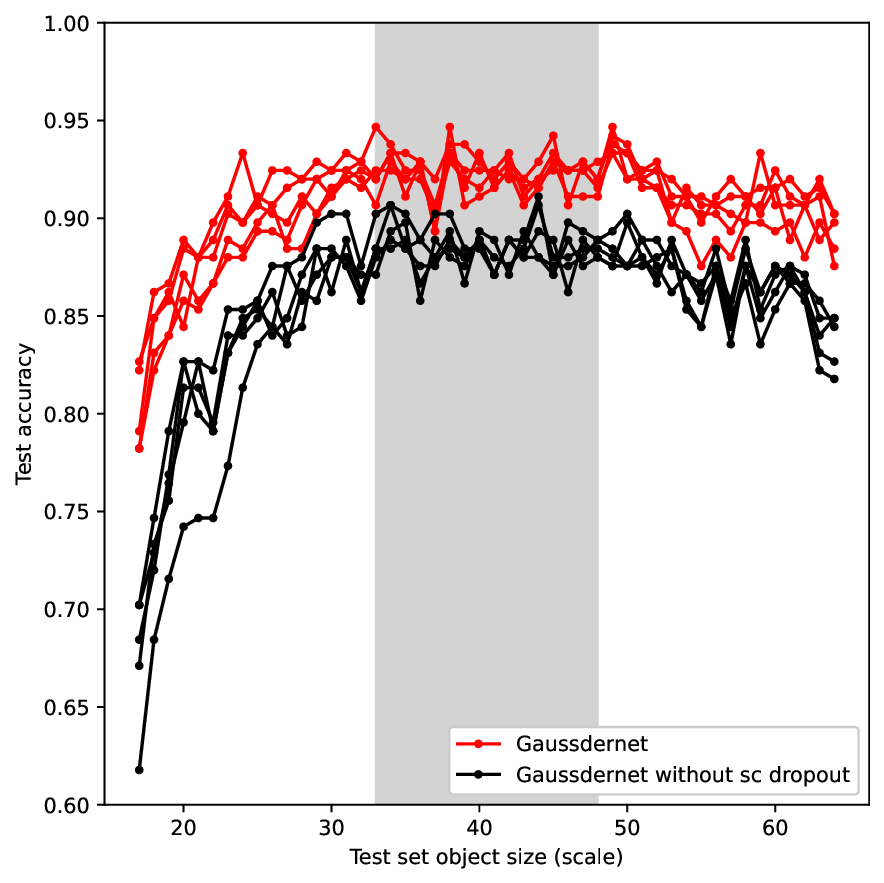}  
        \hspace{2cm}
        \raisebox{0.53mm}{\includegraphics[width=0.35\textwidth]{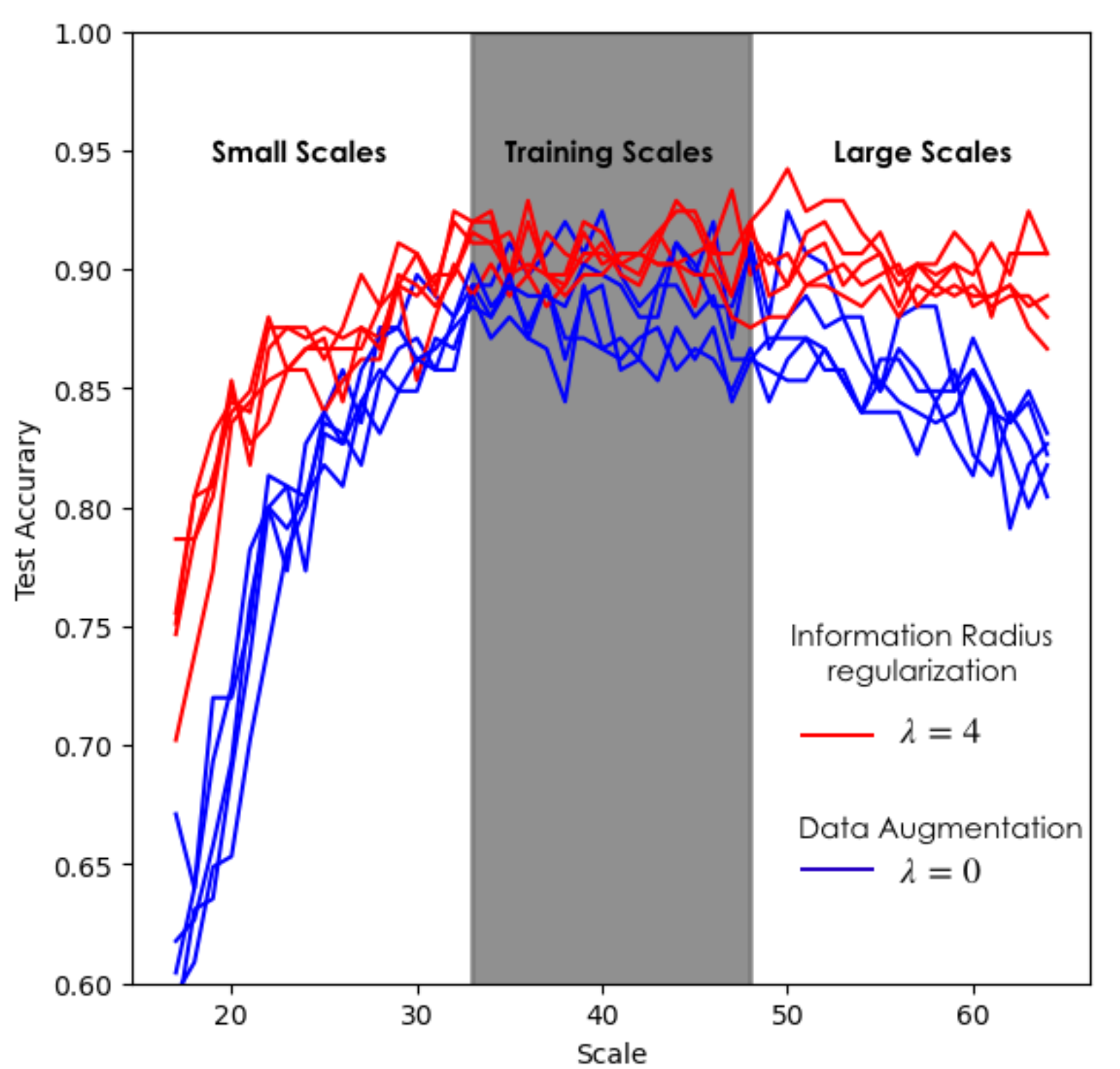}}
        \caption{Scale generalisation curves for the STIR aerial dataset, obtained using (left) our GaussDerNet, (right) the GD-CNN network by Velasco-Forero (\citeyear{Velasco2023-ICGSI}).}
        \label{fig:subfig-stir-aerial-mid-comparison}
    \end{subfigure}
  \end{center}
\caption{Scale generalisation curves for (a) the STIR traffic sign dataset, and (b) the STIR aerial dataset, obtained using (left column) a GaussDerNet with average pooling over scales, trained with or without scale-channel dropout, and (right column) the two-layer convolutional network (GD-CNN) by Velasco-Forero (\citeyear{Velasco2023-ICGSI}), with invariance inducing regularisation ($\lambda = 4$) or classical data augmentation. Training of all the networks was done for the {\em middle-size\/} training data, with object bounding boxes of length $l \in [33,48]$. These scales seen during training are shaded in grey in the graphs. All the models are evaluated on the test data for all the object sizes, $l \in [17,64]$, with the experimental results shown for each of the 5 runs. The graphs on the right have been reproduced from Velasco-Forero (\citeyear{Velasco2023-LNCS}), for which only the results for training on the {\em middle-size\/} are available. As can be seen from the graphs, GaussDerNets have clearly better scale generalisation properties compared to the GD-CNN network.}
\label{fig-stir-velasco-mid-comparison}
\end{figure*}

\subsection{Experiments on the STIR datasets}
\label{sec-stir-experiments}

\subsubsection{Scale generalisation of the Gaussian derivative networks compared to other deep networks}
\label{sec-stir-expm}

For the STIR traffic sign dataset and the STIR aerial dataset, we trained multi-scale-channel GaussDerNets with average pooling over scales as described in Section~\ref{sec-train-proced}, using scale channels with $\sigma_{i,0} \in \{1/(2\sqrt{2}), 1/2$, $1/\sqrt{2}, 1, \sqrt{2}, 2\}$, and either with or without applying scale-channel dropout with $q=0.3$ during training, however without any data augmentations being used. Spatial max pooling was used for all the models, since the objects are not centred in the images. 

We considered three distinct scenarios for training; in each of these, the networks were trained on a different subset of the full training dataset. The networks were trained on either training data with small object sizes (object bounding box length $l \in [17,32]$), training data with medium object sizes (object bounding box length $l \in [33,48]$), or training data with large object sizes (object bounding box length $l \in [49,64]$). The trained networks were then evaluated on test data at all scales, $l \in [17,64]$. In this way, we study the capacity of the multi-scale-channel GaussDerNet to generalise to any previously unseen scale, when trained on training data covering different ranges of object sizes.

The scale generalisation performance of the trained GaussDerNets was compared to four of some of the best performing models\footnote{In their investigation of scale generalisation results for the STIR dataset, Altstidl et al. (\citeyear{AltNguSchKoeMutEskZan23-IJCNN}) show the results of evaluating 11 different types of deep networks. For the purpose of simplicity of comparison in our study, we have selected the three best methods for comparison, for the subset of use cases that we consider in our study. Specifically, the experimental results, that we report, have been obtained from the GitHub repository at https://github.com/taltstidl/scale-equivariant-cnn/blob/main/plots/generalization.csv, and have not been obtained from the original authors, for the cases where they have applied methods developed not by themselves.} on these datasets, the scale-equivariant steerable network (SESN) developed by Sosnovik et al. (\citeyear{SosSzmSme20-ICLR}), also referred to as a Hermite network, the DISCO network developed by Sosnovik et al. (\citeyear{SosMosSme21-BMVC}), and the PixelPool network developed by Altstidl et al. (\citeyear{AltNguSchKoeMutEskZan23-IJCNN}). In a separate comparison, we will also compare our results to previously reported experimental results for the generalised divergence induced invariant CNN (GD-CNN) developed by Velasco-Forero (\citeyear{Velasco2023-ICGSI}).

The results from this experiment are presented in Figures~\ref{fig-stir-sign-plot},~\ref{fig-stir-aerial-plot} and~\ref{fig-stir-velasco-mid-comparison}.
From the scale generalisation results for the traffic sign dataset shown in Figure~\ref{fig-stir-sign-plot}, we can see that the scale generalisation performance of the GaussDerNet is very good for the STIR traffic sign dataset, with almost flat curves, as also previously obtained for the MNIST Large Scale dataset (Lindeberg \citeyear{Lin22-JMIV}). 
In comparison with the scale generalisation results for the SESN, DISCO and PixelPool models, as experimentally obtained by Altstidl et al. (\citeyear{AltNguSchKoeMutEskZan23-IJCNN}), the GaussDerNet does also have clearly better scale generalisation performance than those networks. The scale generalisation curves for the GaussDerNet are significantly flatter, especially for small object sizes.

From the scale generalisation results for the aerial image dataset shown in Figure~\ref{fig-stir-aerial-plot}, we can see that the scale generalisation performance for the GaussDerNet is rather good for the STIR aerial dataset, although not as flat as for the traffic sign dataset.
In comparison with the scale generalisation results for the SESN, DISCO and PixelPool networks, as again experimentally obtained by Altstidl et al. (\citeyear{AltNguSchKoeMutEskZan23-IJCNN}), the GaussDerNet does also have clearly better scale generalisation performance.

We can also see from these experimental results, that for both of the STIR datasets it is strongly beneficial to use scale-channel dropout when training the GaussDerNet. For the experiments on the STIR datasets, the scale-channel dropout both improves the test accuracy and results in flatter scale generalisation curves, here especially when generalising to smaller object sizes.

\begin{table*}[hbtp]
  \centering
  \begin{tabular}{ll}
       \hline
       \textbf{Section} & \textbf{Topic}\\ 
       \hline
    \ref{sec-ablation-regularisation} &
    Influence of different regularisation mechanisms \\ 
    \ref{sec-jet-comparison} & 
    Choice of 2-jet vs. 3-jet layers\\ 
    \ref{sec-permute-invariant} & 
    Alternative permutation invariant pooling methods\\ 
    \ref{sec-new-der-kernels} & 
    Choice of discretisation method for the Gaussian derivative operators\\
    \ref{sec-sigma-learning} & 
    Influence of learning of the scale levels during training\\ \hline
  \end{tabular}
  \caption{Overview of the content covered by the subsections in Section~\ref{sec-ablation}, which present ablation and comparative studies. These subsections are organised by the kind of experiment they investigate, with each detailing an experiment conducted on both the rescaled Fashion-MNIST dataset and the rescaled CIFAR-10 dataset. These studies aim to explore the properties of GaussDerNets under various conditions, architectures and training methods.}
  \label{tab:ablation-section-overview}
\end{table*}

From the scale generalisation results for the traffic sign dataset shown in Figure~\ref{fig-stir-velasco-mid-comparison}(\subref{fig:subfig-stir-sign-mid-comparison}), we can see that the GaussDerNet has significantly better scale generalisation properties compared to the GD-CNN network, with substantially better performance for small object sizes. The test accuracy of the GaussDerNet at the training scales is also roughly 5~ppt higher compared to the GD-CNN network.

From the scale generalisation results for the aerial dataset shown in Figure~\ref{fig-stir-velasco-mid-comparison}(\subref{fig:subfig-stir-aerial-mid-comparison}), we can see that for the GaussDerNet there is a trend suggesting approximately 2~ppt higher test accuracy at the training scales compared to the GD-CNN network. Additionally, the scale generalisation is clearly better for small object sizes, and the scale generalisation for large object sizes is rather flat for both the network architectures.

These results also demonstrate that the GaussDerNets are capable of achieving good scale generalisation even with very limited training data.

\section{Ablation and comparative studies}
\label{sec-ablation}
In this section, we will perform ablation and comparative studies on GaussDerNets, for the Fashion-MNIST and CIFAR-10 datasets. We will present the results of (i)~studies on different training regularisation methods, (ii)~compare networks based on Gaussian derivatives up to order 2 to extended networks based on Gaussian derivatives up to order 3, and (iii)~examine the applicability of GaussDerNets based on alternative kinds of permutation invariant pooling over scales. Moreover, we will (iv)~investigate the influence of using different types of discrete approximations of the Gaussian derivative operators, and (v)~explore introducing learning of the scale levels in the networks.
An overview of the experiments covered in this section is given in Table~\ref{tab:ablation-section-overview}.

\subsection{Benefits of the regularisation mechanisms}
\label{sec-ablation-regularisation}
For the rescaled Fashion-MNIST and CIFAR-10 datasets, we performed ablation experiments to investigate the impact of incorporating scale-channel dropout or cutout in the training process for the GaussDerNet. 

For each dataset, we trained two sets of multi-scale-channel networks, one set that uses scale-channel dropout during training, and another set without any scale-channel dropout. This was done both for models that use max pooling over scales and average pooling over scales. For models that use scale-channel dropout, a dropout value of $q=0.2$ was used for training on the Fashion-MNIST dataset, and $q=0.3$ for the CIFAR-10 dataset. Additionally, for the CIFAR-10 dataset we trained a single model using max pooling over scales, trained with cutout, which uses a single square of size 16 in the image. Central pixel extraction was used in all the models, and training for each dataset was performed as described in Section~\ref{sec-train-proced}. The models trained on the Fashion-MNIST dataset used the scale channels with $\sigma_{i,0} \in \{1/(2\sqrt{2}), 1/2, 1/\sqrt{2}, 1, \sqrt{2}, 2, 2\sqrt{2}\}$, while the models \\
trained on the CIFAR-10 dataset used $\sigma_{i,0} \in \{1/(2\sqrt{2})$, $1/2, 1/\sqrt{2}, 1, \sqrt{2}, 2\}$.

\begin{figure}[hbt]
  \begin{center}
    \begin{subfigure}{0.48\textwidth}
        \centering
        \includegraphics[width=\linewidth]{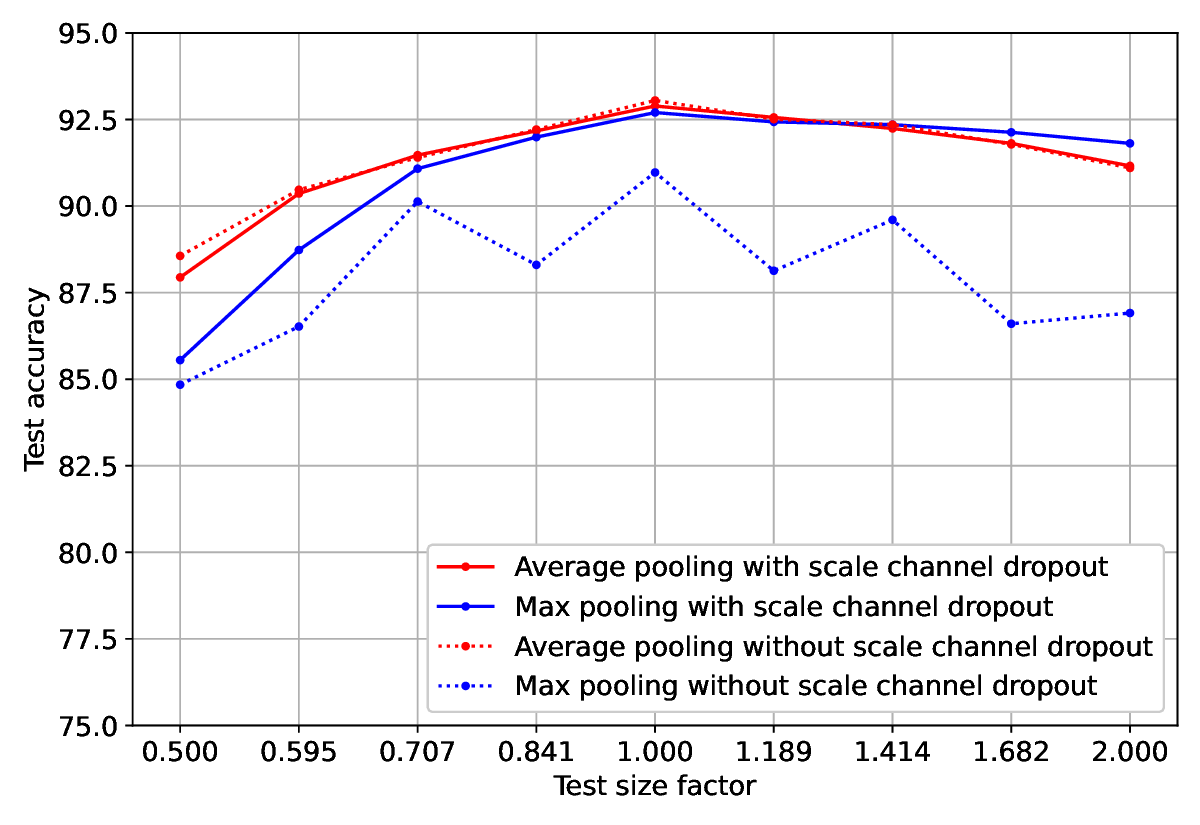}
        \label{fig:subfiga-3}
    \end{subfigure}
  \end{center}
\caption{Scale generalisation curves for multi-scale-channel networks, trained with scale-channel dropout or no additional regularisation method at all, using either max pooling over scales or average pooling over scales, for the rescaled Fashion-MNIST dataset. In this experiment, all the GaussDerNets were tested on all the size factors between 1/2 and 2, after being trained on the training set for size factor 1. The graphs show that using scale-channel dropout during training results in better scale generalisation, compared to training without any regularisation.}
\label{fig-scale-drop-ablation}
\end{figure}

\begin{figure}[hbt]
  \begin{center}
    \begin{subfigure}{0.48\textwidth}
        \centering
        \textit{\hspace{0.5cm} Networks with average pooling over scales}
        \includegraphics[width=\linewidth]{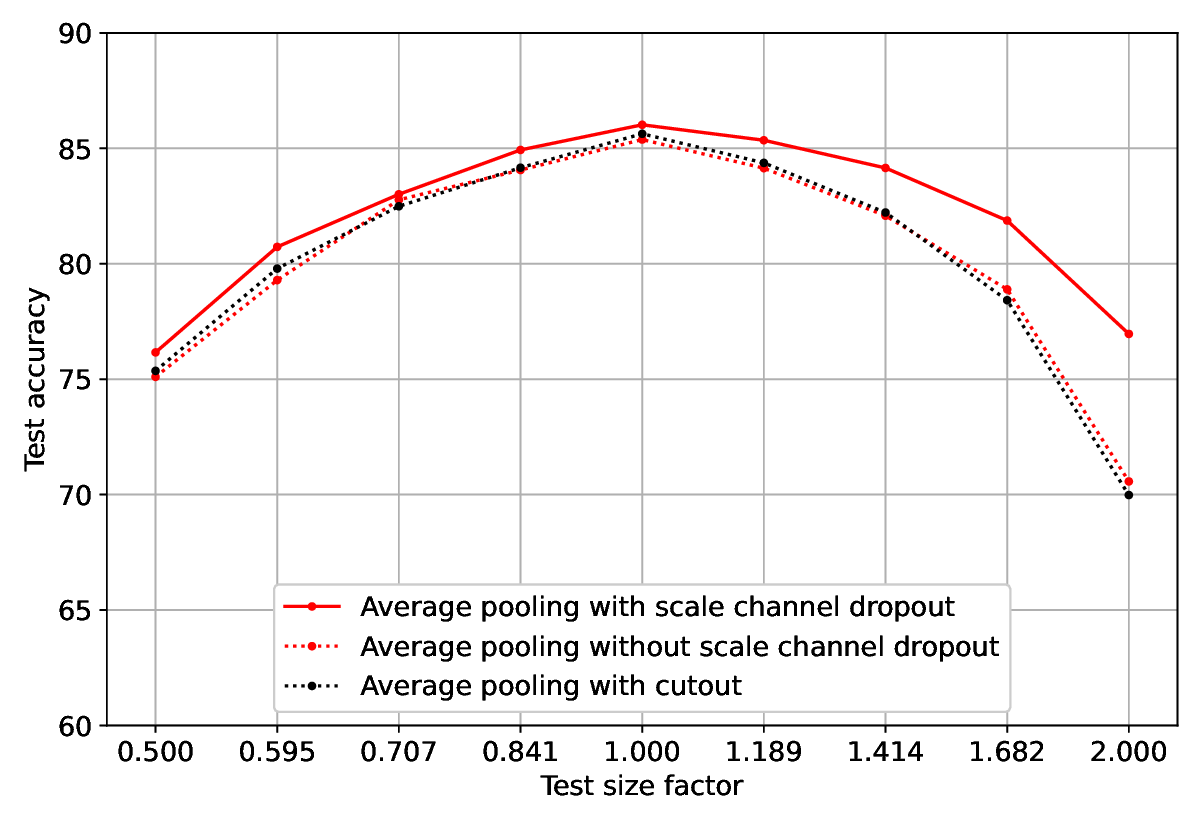}
        \label{fig:subfigb-3}
    \end{subfigure}
    \begin{subfigure}{0.48\textwidth}
        \centering
        \textit{\hspace{0.5cm} Networks with max pooling over scales}
        \includegraphics[width=\linewidth]{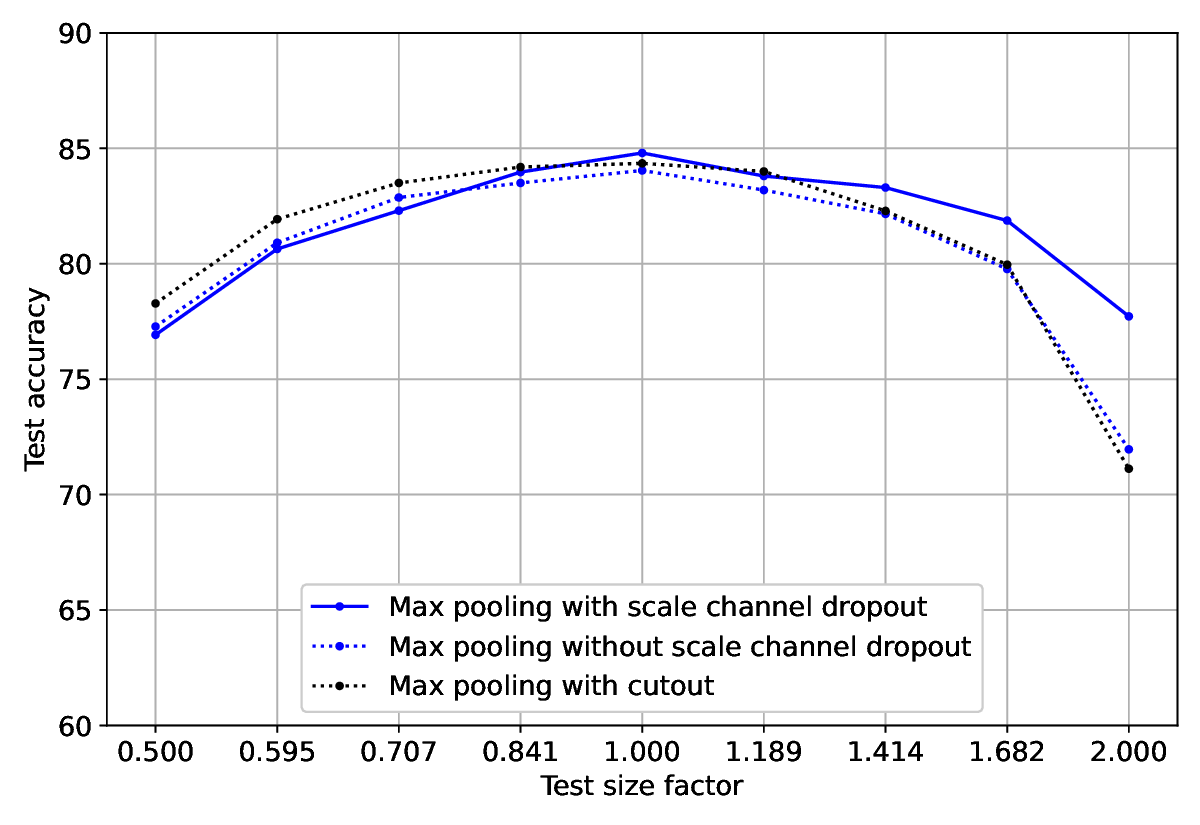}
        \label{fig:subfiga-4}
    \end{subfigure}
  \end{center}
\caption{Scale generalisation curves for multi-scale-channel networks, trained with scale-channel dropout, cutout, or no additional regularisation method at all, using either (top) average pooling over scales or (bottom) max pooling over scales, for the rescaled CIFAR-10 dataset. In this experiment, all the GaussDerNets were tested on all the size factors between 1/2 and 2, after being trained on the training data for size factor 1. The graphs show that using scale-channel dropout during training results in better scale generalisation, compared to training without any regularisation. Using cutout results in only minor improvement on small size factors for the rescaled CIFAR-10 dataset.}
\label{fig-scale-drop-ablation-cifar}
\end{figure}

Figures~\ref{fig-scale-drop-ablation} and~\ref{fig-scale-drop-ablation-cifar} show the results of these ablation studies, with the models trained on the training set for size factor 1, using either max pooling over scales or average pooling over scales, and tested on all the size factors between 1/2 and 2. As we can see from Figure~\ref{fig-scale-drop-ablation}, using scale-channel dropout during training on the Fashion-MNIST dataset, for a model that uses average pooling over scales, results in a negligible change in performance. For the model trained using max pooling over scales, however, we find that not using scale-channel dropout during training can sometimes result in the training converging to a suboptimal solution, with a scale generalisation curve that is not smooth.

Figure~\ref{fig-scale-drop-ablation-cifar} demonstrates that for the CIFAR-10 dataset, training with scale-channel dropout results in improved scale generalisation performance for size factors $>$ 1. These findings are consistent with the notion that using scale-channel dropout during training can help the network to learn from information at different scales more effectively, proving especially beneficial for datasets that include complex shapes and textures. Similar benefits have been observed in other scale dropout methods proposed in previous work by Sangalli et al. (\citeyear{SanBluVelAng22-BMVC}).

Training with cutout instead of scale-channel dropout only results in minor increase in performance for size factors $<$ 1, compared to models trained without any additional regularisation method.

\subsection{Gaussian derivative networks based on 2-jet vs. 3-jet layers}
\label{sec-jet-comparison}

For the rescaled Fashion-MNIST and CIFAR-10 datasets, we performed comparative studies on the effect of basing the GaussDerNets on Gaussian derivatives up to second or third order, with the layers defined according to Equations~(\ref{eq:2-Jet}) or~(\ref{eq:3-Jet}). 

For each dataset, we trained two multi-scale-channel networks, one based on 2-jet layers, and a second based on 3-jet layers. The models used average pooling over scales, with central pixel extraction, and were trained using the parameters described in Section~\ref{sec-train-proced} for each respective dataset. Scale-channel dropout was applied during training of each network, once again using $q=0.2$ for training on the Fashion-MNIST dataset and $q=0.3$ for the CIFAR-10 dataset. The models trained on the Fashion-MNIST dataset used the scale channels with $\sigma_{i,0} \in \{1/(2\sqrt{2}), 1/2, 1/\sqrt{2}, 1, \sqrt{2}, 2, 2\sqrt{2}\}$, and the models trained on the CIFAR-10 dataset used $\sigma_{i,0} \in \{1/(2\sqrt{2})$, $1/2, 1/\sqrt{2}$, $1, \sqrt{2}, 2\}$.

\begin{figure}[hbt]
  \begin{center}
    \begin{subfigure}{0.48\textwidth}
        \centering
        \textit{\hspace{0.6cm} N:th-order networks on rescaled Fashion-MNIST}
        \includegraphics[width=\linewidth]{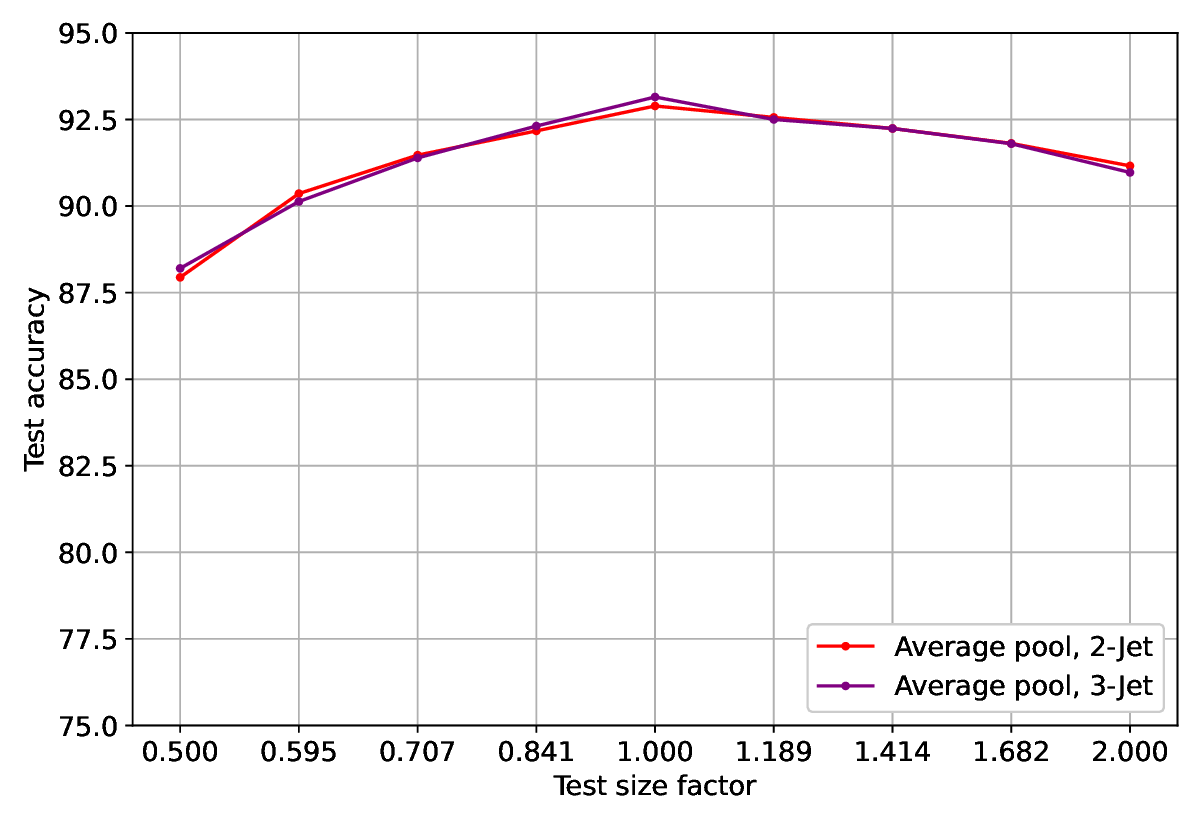}
        \label{fig:subfigb-4}
    \end{subfigure}
    
    \vspace{0.5 cm}
    
    \begin{subfigure}{0.48\textwidth}
        \centering
        \textit{\hspace{0.5cm} N:th-order networks on rescaled CIFAR-10}
        \includegraphics[width=\linewidth]{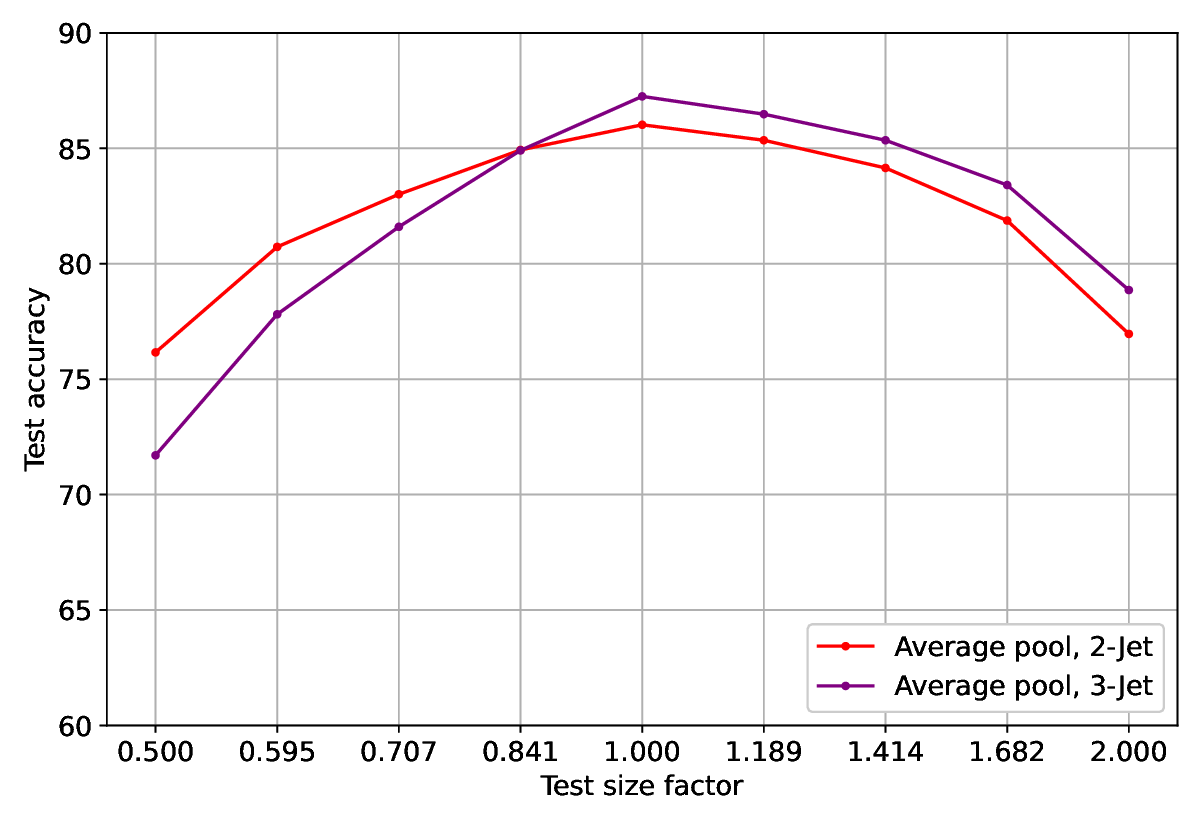}
        \label{fig:subfiga-5}
    \end{subfigure}
  \end{center}
\caption{Comparison between the scale generalisation performance of second-order networks and third-order networks on the rescaled Fashion-MNIST dataset and the rescaled CIFAR-10 dataset, based on Gaussian derivative layers that are based on either the Gaussian 2-jet or the Gaussian 3-jet, respectively, and using averaging pooling over scales for scale selection. As in the previous experiments, each network was trained on training data for size factor 1, and then evaluated on test data for all the size factors ranging from 1/2 to 2. As can be seen from the graphs, for the Fashion-MNIST dataset there is only a minor difference in scale generalisation performance, when using either 2-jet layers or 3-jet layers. For the CIFAR-10 dataset, the network based on 3-jet layers is able to achieve better scale generalisation performance for size factors $\geq$ 1, while the network based on 2-layers leads to better scale generalisation performance for lower size factors.}
\label{fig-jet-comparison}
\end{figure}

The results of these experiments are shown in Figure~\ref{fig-jet-comparison}, where the networks were trained at the single size factor 1 and tested for all the size factors between 1/2 and 2. The upper graph shows the results for training on the Fashion-MNIST dataset, where we can see that using a layer based on third-order Gaussian derivatives results in small but notable improvement on the test data for the size factor 1. Despite this minor improvement in performance on the training scale, the scale generalisation performance was comparable to a network based on the 2-jet, meaning that basing the layers in the network on higher-order derivatives does not lead to any meaningful improvement in the scale generalisation properties.

For the networks trained on the CIFAR-10 dataset, shown in the bottom graph of Figure~\ref{fig-jet-comparison}, we can see an improvement in performance for size factors $\geq$ 1, suggesting that third-order Gaussian derivatives provided better representations of the image structures. On the other hand, for size factors $<$ 1 the performance drops at a faster rate compared to the model based on the 2-jet, showing a reduced ability to effectively generalise to previously not seen scales. This is likely caused by the approximations of the Gaussian derivatives becoming less accurate at finer scales, with the transformation properties of higher-order derivatives being more sensitive to sampling problems caused by low resolution, and in that way affecting the scale-covariant properties in the discrete implementation of the network.

\subsection{Gaussian derivative networks based on alternative permutation invariant pooling methods}
\label{sec-permute-invariant}

\begin{figure}[hbt]
  \begin{center}
    \begin{subfigure}{0.48\textwidth}
        \centering
        \textit{\hspace{0.6cm} Different pooling methods on rescaled Fashion-MNIST}
        \includegraphics[width=\linewidth]{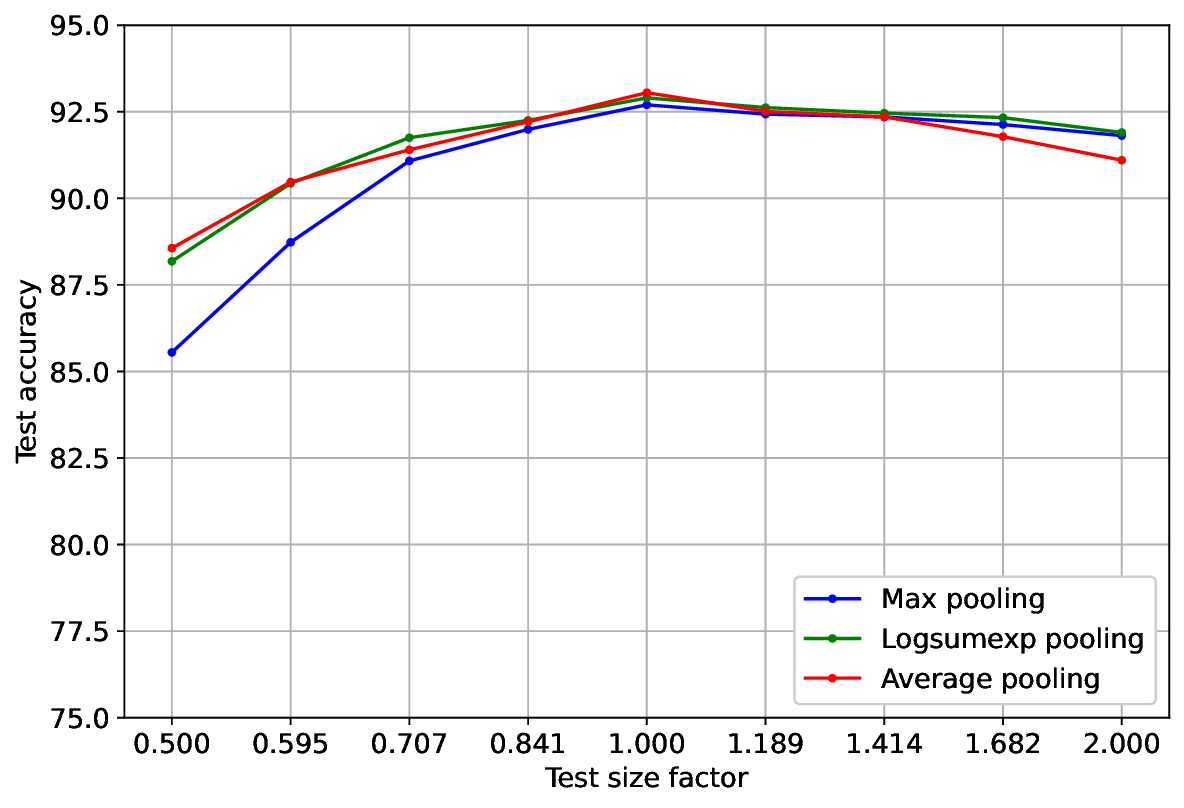}
        \label{fig:subfiga-logsumexp}
    \end{subfigure}
    
    \vspace{0.5 cm}
    
    \begin{subfigure}{0.48\textwidth}
        \centering
        \textit{\hspace{0.5cm} Different pooling methods on rescaled CIFAR-10}
        \includegraphics[width=\linewidth]{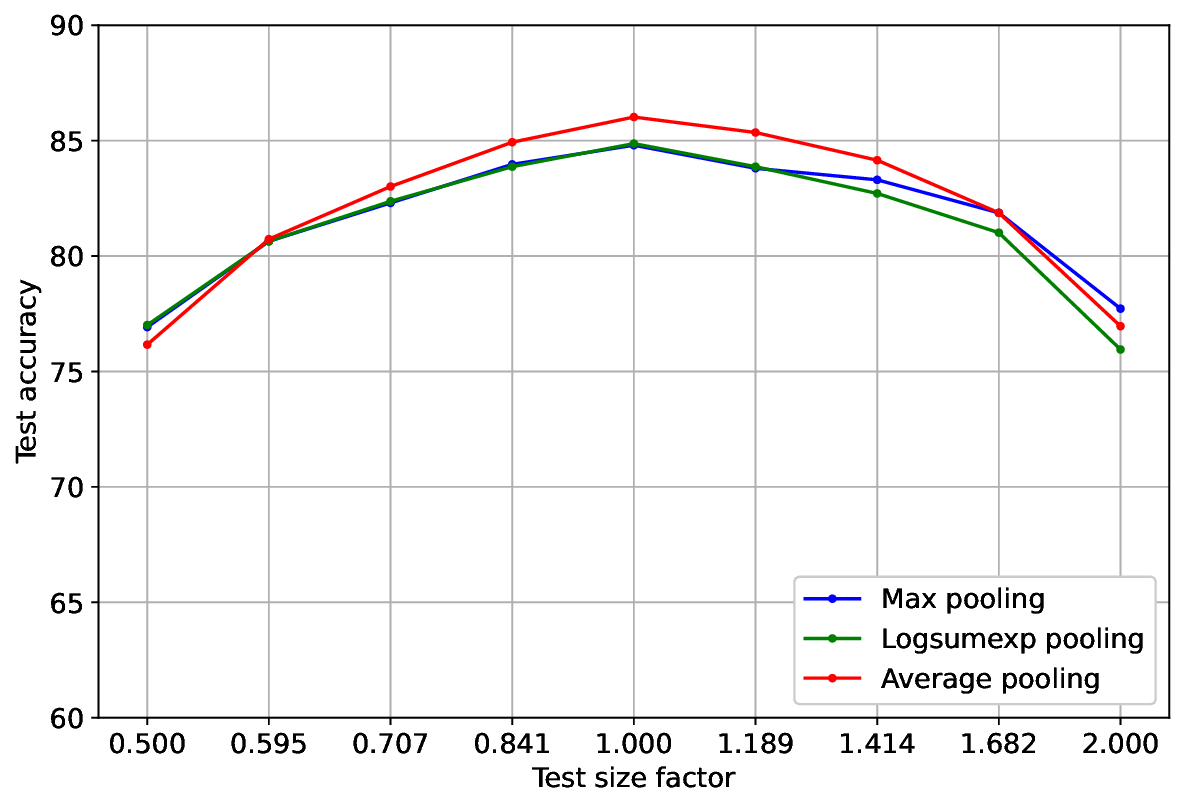}
        \label{fig:subfigb-logsumexp}
    \end{subfigure}
  \end{center}
\caption{Comparison between the scale generalisation performance of Gaussian derivative networks based on different permutation invariant pooling over scales approaches, on the rescaled Fashion-MNIST dataset and the rescaled CIFAR-10 dataset, with the networks based on either max, smooth max (logsumexp) or average pooling over scales for scale selection. Each network was trained on training data for size factor 1, and then evaluated on test data for all the size factors ranging from 1/2 to 2. The graphs show that for the Fashion-MNIST dataset, using logsumexp pooling over scales results in a flatter scale generalisation curve, compared to max and average pooling based networks. For the CIFAR-10 dataset, max and average pooling based Gaussian derivative networks achieve better scale generalisation performance on large size factors than logsumexp pooling based networks.}
\label{fig-jet-comparison}
\label{fig-logsumexp-comparison}
\end{figure}

For the rescaled Fashion-MNIST and CIFAR-10 datasets, we performed comparative studies on the effect of basing the GaussDerNets on an alternative method for permutation invariant pooling over scales using a smooth approximation of the maximum operator, compared to the max or average pooling over scales approaches.

We considered the logsumexp pooling method, defined by taking the logarithm of summed exponentials of scale channel outputs, defined as:
\begin{equation}
\label{eq:logsumexp-pooling-expression}
    F_{\logsumexp}^{c} = \log{\sum_{i=1}^{N} \exp\left(F_{\centre}^{c}(\sigma_{i,0})\right)}
\end{equation}
and which in the context of machine learning has the desirable property of promoting more stable training.

For each dataset, we trained three multi-scale-channel networks, one for each type of pooling method. The models used central pixel extraction, and were once again trained using the parameters described in Section~\ref{sec-train-proced} for each respective dataset. Scale-channel dropout was applied during training of each network, using $q=0.2$ for training on the Fashion-MNIST dataset and $q=0.3$ for the CIFAR-10 dataset. When training on the Fashion-MNIST dataset, the models used the scale channels with $\sigma_{i,0} \in \{1/(2\sqrt{2}), 1/2, 1/\sqrt{2}, 1, \sqrt{2}, 2, 2\sqrt{2}\}$, while the models trained on the CIFAR-10 dataset used $\sigma_{i,0} \in \{1/(2\sqrt{2}), 1/2, 1/\sqrt{2}, 1, \sqrt{2}, 2\}$.

The results of these experiments are shown in Figure~\ref{fig-logsumexp-comparison}, where training is done at size factor 1 and testing at all the size factors between 1/2 and 2. The results for training GaussDerNets on the Fashion-MNIST dataset are shown in the upper graph, where we can see that using the smooth approximation of the maximum, logsumexp, leads to a flatter curve than for both max and average pooling over scales. This might suggest that basing GaussDerNets on logsumexp pooling over scales provides additional stabilisation to the training, since we saw in Section~\ref{sec-ablation-regularisation} that for the Fashion-MNIST dataset, GaussDerNets based on max pooling over scales had a tendency to converge to a suboptimal solution during training, if no additional regularisation was used.\footnote{Additional experiments not reported here, however, suggest that the difference in scale generalisation performance between logsumexp pooling and max pooling can be reduced if better training parameters are found.} The results for training GaussDerNets on the CIFAR-10 dataset are shown in the bottom graph of Figure~\ref{fig-logsumexp-comparison}, where we can see that in this case using max or average pooling over scales leads to better performance on large size factors, compared to the logsumexp pooling. These findings suggest that using logsumexp pooling over scales can sometimes be beneficial, however this property may also be dataset dependent.

\begin{table*}[htbp]
  \centering
  \begin{tabular}{lccc}
       \hline
       \textbf{Type of kernel} & $\sigma_{0} = 1/2$ & $\sigma_{0} = 1/\sqrt{2}$ & $\sigma_{0} = 1$ \\ 
       \hline 
       Discrete Gaussian + central differences & $93.76 \pm~0.10$ & $93.55 \pm~0.19$ & $93.08 \pm~0.05$ \\ 
       Sampled Gaussian derivatives & $93.74 \pm~0.10$ & $93.45 \pm~0.12$ & $92.99 \pm~0.18$ \\ 
       Integrated Gaussian derivatives & $93.48 \pm~0.15$ & $93.22 \pm~0.09$ & $92.77 \pm~0.07$ \\
       \noalign{\vskip 0.8mm}
       \hline
       \noalign{\vskip 0.9mm}
       Sampled Gaussian + central differences & $93.60 \pm~0.13$ & $93.23 \pm~0.13$ & $92.84 \pm~0.08$ \\ 
       Normalised sampled Gaussian + central differences & $93.49 \pm~0.09$ & $93.23 \pm~0.18$ & $92.73 \pm~0.05$ \\
       Integrated Gaussian + central differences & $93.32 \pm~0.18$ & $93.09 \pm~0.09$ & $92.67 \pm~0.21$ \\ \hline
  \end{tabular}
  \caption{Influence of the choice of discrete derivative approximation method of the Gaussian derivative operators on the mean and unbiased standard deviation of accuracy over 5 runs (and in \%) of GaussDerNets, applied to the {\em regular\/} Fashion-MNIST dataset (with image size $28 \times 28$ pixels), for different initial values $\sigma_0 \in \{1/2, 1/\sqrt{2}, 1\}$ of the GaussDerNets. The relative scale ratio is correspondingly set to be $r \in \{1.47, 1.37 ,1.28\}$, which has been chosen in such a way that the standard deviation of the Gaussian derivatives in the final layer is the same for each network, that is $\sigma_k = \sigma_0 \, r^{k-1} \approx D_{1}$, where $D_{1}$ is a constant and $k$ represents the number of layers in the model.}
  \label{tab:new_der_fashion_table}
\end{table*}

\begin{table*}[htbp]
  \centering
  \begin{tabular}{lccc}
       \hline
       \textbf{Type of kernel} & $\sigma_{0} = 1/2$ & $\sigma_{0} = 1/\sqrt{2}$ & $\sigma_{0} = 1$ \\ 
       \hline 
       Discrete Gaussian + central differences & $86.72 \pm~0.23$ & $85.56 \pm~0.09$ & $83.40 \pm~0.15$ \\ 
       Sampled Gaussian derivatives & $86.92 \pm~0.27$ & $85.20 \pm~0.17$ & $82.92 \pm~0.41$ \\
       Integrated Gaussian derivatives & $86.12 \pm~0.08$ & $84.87 \pm~0.21$ & $82.70 \pm~0.32$ \\ 
       \noalign{\vskip 0.8mm}
       \hline
       \noalign{\vskip 0.9mm}
       Sampled Gaussian + central differences & $85.58 \pm~0.32$ & $84.27 \pm~0.30$ & $82.00 \pm~0.17$ \\
       Normalised sampled Gaussian + central differences & $85.82 \pm~0.08$ & $84.35 \pm~0.28$ & $82.29 \pm~0.10$ \\
       Integrated Gaussian + central differences & $85.23 \pm~0.25$ & $83.91 \pm~0.14$ & $81.81 \pm~0.28$ \\ \hline
  \end{tabular}
  \caption{Influence of the choice of discrete derivative approximation method of the Gaussian derivative operators on the mean and unbiased standard deviation of accuracy over 5 runs (and in \%) of GaussDerNets, applied to the {\em regular\/} CIFAR-10 dataset (with image size $32 \times 32$ pixels), for different initial values $\sigma_0 \in \{1/2, 1/\sqrt{2}, 1\}$ of the GaussDerNets. The relative scale ratio is correspondingly set to be $r \in \{1.67, 1.56, 1.45\}$, which has been chosen in such a way that the standard deviation of the Gaussian derivatives in the final layer is the same for each network, that is $\sigma_k = \sigma_0 \, r^{k-1} \approx D_{2}$, where $D_{2}$ is a constant and $k$ represents the number of layers in the model.}
  \label{tab:new_der_cifar_table}
\end{table*}

\subsection{Choice of discrete approximations for the Gaussian derivative kernels}
\label{sec-new-der-kernels}
The original definition of the GaussDerNet is formulated for continuous image data, while a numerical implementation will depend on a discrete approximation of the Gaussian derivative operators. To investigate how the choice of discretisation method for the Gaussian derivative operators may affect the performance of a resulting discrete model implementation, we applied different discretisation methods to the Gaussian derivative networks and trained on the regular Fashion-MNIST and CIFAR-10 datasets (without additional scaling transformations).

The main discretisation methods considered were: (i) the default choice of using the discrete analogue of the Gaussian kernel complemented by central difference operators to obtain the discrete derivative approximations, (ii) sampled Gaussian derivative operators, defined according to Equation~(\ref{eq:sampled-Gaussian-kernel}), and (iii) integrated Gaussian derivative operators, defined according to Equation~(\ref{eq:integrated-Gaussian-kernel}). For comparison, we also added the following alternative approaches: (iv) the sampled Gaussian kernel, (v) the normalised sampled Gaussian kernel, and (vi) the integrated Gaussian kernel, with each one of these three methods complemented by central difference operators to obtain the discrete derivative approximations.

The motivation for studying this latter set of kernels (iv)--(vi), is that the corresponding discrete derivative approximations can be computed more efficiently, compared to explicit convolutions with sets of either sampled Gaussian derivatives or integrated Gaussian derivatives, and thus using essentially the same amount of computational work as the computation of discrete derivative approximations by convolutions with the discrete analogue of the Gaussian kernel followed by central differences (Lindeberg \citeyear{Lin25-FrontSignProc}).

To investigate the effect of using different initial scale values of $\sigma_0$ for the networks,\footnote{Concerning the comparisons between the different discretisation methods, it should, however, be noted that for a given value of the scale parameter $\sigma$, the different discretisation methods will give rise to different types of discrete kernels with different discrete spatial spread measures as quantifying their spatial extent, see Figure~10 in Lindeberg (\citeyear{Lin24-JMIV-discgaussder}) and Figure~2 in Lindeberg (\citeyear{Lin25-FrontSignProc}) for graphs. For a given value of the scale parameter $\sigma$, the amount of discrete spatial smoothing in the discrete derivative approximations can thus be different, depending on the choice of discrete derivative approximation method for the Gaussian derivative operators. For the comparisons between the GaussDerNets here, we do, however, for simplicity, disregard this effect, while noting that the comparisons between the different discretisation methods for the GaussDerNets based on learned scale levels to be performed later, in Tables~\ref{tab:sigma_learning_performance_fashion} and~\ref{tab:sigma_learning_performance_cifar}, can be expected to automatically compensate for such effects.} we additionally performed the experiments for the different values of $\sigma_0 \in \{ 1/2, 1/\sqrt{2}, 1 \}$. The relative scale ratio $r$ was adjusted for each model in such a way that the standard deviations of the Gaussian derivative kernels in the last layer of the network were the same for all the networks, for a given dataset. Apart from these modifications, all the models were trained as described in Section~\ref{sec-train-proced} for the rescaled versions of the Fashion-MNIST and CIFAR-10 datasets, using the same parameters as for the default choice of discretisation method.

Tables~\ref{tab:new_der_fashion_table} and~\ref{tab:new_der_cifar_table} show the results of the experiments\footnote{These performance values are obtained with convolutions using reflection padding outside the image boundaries for the regular CIFAR-10 dataset, and would actually become better with zero padding outside of the image boundaries. In fact, we get 89.33\% mean test accuracy across 5 runs, for the GaussDerNet based on the discrete analogue of Gaussian kernel, when trained with zero padding at scale $\sigma_{0} = 1/2$. Here, however, we report the results of reflection padding, to be as close as possible to the experimental situation in the previously treated case for the rescaled CIFAR-10 dataset.}, as the average value as well as the unbiased standard deviation of the test accuracy over 5 runs. By directly comparing the average performance, from Table~\ref{tab:new_der_fashion_table}, we can see that for the Fashion-MNIST dataset, within this range of scale values using a lower value of $\sigma_0$ generally results in minor improvements in the mean test accuracy.
The discrete analogue of the Gaussian kernel complemented by central differences, the sampled Gaussian derivative operator, and the sampled Gaussian kernel complemented by central differences approaches achieved the highest test accuracies, while the integrated Gaussian complemented by central differences achieved somewhat lower performance.

From Table~\ref{tab:new_der_cifar_table}, we can see that there is a substantial difference in performance for the networks trained on the CIFAR-10 dataset, depending on the choice of the initial scale level $\sigma_0$. In fact, the network with $\sigma_0 = 1/2$ on average gives $\sim$ 4~ppt better performance than the network with $\sigma_0 = 1$. A possible explanation for this is that fine-scale textures in the image domain may be particularly important for classifying the image data in the CIFAR-10 dataset. The discrete analogue of the Gaussian kernel complemented by central differences results in the best performance out of all methods when $\sigma_0 = 1/\sqrt{2}$ or $\sigma_0 = 1$, whereas the sampled Gaussian derivative kernel gives the best results when $\sigma_0 = 1/2$.

The results in both of these tables also show that the discretisation methods based on central difference operators applied in combination with either the sampled Gaussian kernel or the integrated Gaussian kernel mostly lead to lower accuracy compared to the more direct approximation methods in terms of sampled Gaussian derivatives or integrated Gaussian derivatives, with these differences being larger for the CIFAR-10 dataset than for the Fashion-MNIST dataset.

Concerning the interpretation of the results obtained using the integrated Gaussian derivative approximation of the Gaussian derivative operators, it should be noted that the box integration used in this concept implies an additional amount of spatial smoothing in the discrete derivative approximation, which is, however, not compensated for in the current implementation. Thus, the effective scale levels will be somewhat coarser. In view of the trend for the  Fashion-MNIST and the CIFAR-10 datasets, that better results are obtained at finer scale levels, this coarsening of the scale level may constitute an explanation why the integrated Gaussian kernels and the integrated Gaussian derivative kernels lead to somewhat lower performance in these experiments.

In Appendix~A.3.3 in the supplementary material, a complementary statistical analysis is given of these results, aimed at answering which of the differences in performance between the different discretisation methods can be regarded as statistically significant.

We leave the issue of investigating the influence of the choice of derivative approximation method on the scale generalisation performance for future work.

\begin{table*}[htbp]
  \centering
  \begin{tabular}{lccc}
       \hline
       \textbf{Type of kernel} & $\eta_{\sigma} = 0.0001$ & $\eta_{\sigma} = 0.001$ & $\eta_{\sigma} = 0.01$ \\ 
       \hline 
       Sampled Gaussian derivatives & $93.47 \pm~0.10$ & $93.50 \pm~0.12$ & $93.71 \pm~0.05$ \\ 
       Integrated Gaussian derivatives & $93.59 \pm~0.10$ & $93.61 \pm~0.08$ & $93.55 \pm~0.14$ \\ 
       Sampled Gaussian derivatives + no BN final layer & $93.12 \pm~0.12$ & $93.35 \pm~0.15$ & $93.45 \pm~0.13$ \\ 
       Integrated Gaussian derivatives + no BN final layer & $93.09 \pm~0.09$ & $93.22 \pm~0.05$ & $93.32 \pm~0.09$ \\
       \noalign{\vskip 0.8mm}
       \hline
       \noalign{\vskip 0.9mm}
       Sampled Gaussian + central differences & $93.47 \pm~0.14$ & $93.68 \pm~0.19$ & $93.55 \pm~0.08$ \\
       Integrated Gaussian + central differences & $93.54 \pm~0.11$ & $93.52 \pm~0.10$ & $93.52 \pm~0.10$ \\
       Sampled Gaussian + central differences and no BN final layer & $93.17 \pm~0.09$ & $93.31 \pm~0.12$ & $93.40 \pm~0.04$ \\
       Integrated Gaussian + central differences and no BN final layer & $93.23 \pm~0.12$ & $93.19 \pm~0.13$ & $93.35 \pm~0.05$ \\ \hline
  \end{tabular}
  \caption{Mean and unbiased standard deviation of accuracy over 5 runs (and in \%), obtained with {\em learning of the scale-parameters\/} for the different layers in single-scale-channel networks, applied to the {\em regular\/} Fashion-MNIST dataset (with image size $28 \times 28$ pixels). Here, the networks have been trained for different learning rates $\eta_{\sigma} \in \{0.0001, 0.001, 0.01 \}$ for the scale parameters $\sigma_k$ in the layers, while keeping the learning rate for the weights $C_0$, $C_x$, $C_y$, $C_{xx}$, $C_{xy}$ and $C_{xx}$ fixed to $\eta_C = 0.01$. The scale parameters $\sigma_k$ in the layers have, in turn, been initialised to a similar geometric distribution according to Equation~(\ref{eq:sigma-geometric}), for $\sigma_0 = 1$ and $r = 1.28$, as used for corresponding fixed-scale networks.}
  \label{tab:sigma_learning_performance_fashion}
\end{table*}

\begin{table*}[htbp]
  \centering
  \begin{tabular}{lccc}
       \hline
       \textbf{Type of kernel} & $\eta_{\sigma} = 0.0001$ & $\eta_{\sigma} = 0.001$ & $\eta_{\sigma} = 0.01$ \\ 
       \hline 
       Sampled Gaussian derivatives & $87.71 \pm~0.20$ & $87.52 \pm~0.29$ & $87.33 \pm~0.31$ \\ 
       Integrated Gaussian derivatives & $86.93 \pm~0.14$ & $86.52 \pm~0.36$ & $86.64 \pm~0.22$ \\
       Sampled Gaussian derivatives + no BN final layer & $85.63 \pm~0.39$ & $86.75 \pm~0.23$ & $86.52 \pm~0.18$ \\
       Integrated Gaussian derivatives + no BN final layer & $85.33 \pm~0.21$ & $86.06 \pm~0.17$ & $85.96^{*} \pm~0.23$ \\
       \noalign{\vskip 0.8mm}
       \hline
       \noalign{\vskip 0.9mm}
       Sampled Gaussian + central differences & $86.90 \pm~0.28$ & $86.61 \pm~0.29$ & $86.81 \pm~0.20$ \\
       Integrated Gaussian + central differences & $86.74 \pm~0.26$ & $86.66 \pm~0.40$ & $86.79 \pm~0.29$ \\
       Sampled Gaussian + central differences and no BN final layer & $85.25 \pm~0.13$ & $86.03 \pm~0.32$ & $85.83 \pm~0.33$ \\
       Integrated Gaussian + central differences and no BN final layer & $85.34 \pm~0.30$ & $85.93 \pm~0.15$ & $86.17 \pm~0.39$ \\ \hline
  \end{tabular}
  \caption{Mean and unbiased standard deviation of accuracy over 5 runs (and in \%), obtained with {\em learning of the scale-parameters\/} for the different layers in single-scale-channel networks, applied to the {\em regular\/} CIFAR-10 dataset (with image size $32 \times 32$ pixels). Here, the networks have been trained for different learning rates $\eta_{\sigma} \in \{0.0001, 0.001, 0.01 \}$ for the scale parameters $\sigma_k$ in the layers, while keeping the learning rate for the weights $C_0$, $C_x$, $C_y$, $C_{xx}$, $C_{xy}$ and $C_{xx}$ fixed to $\eta_C = 0.01$. The scale parameters $\sigma_k$ in the layers have, in turn, been initialised to a similar geometric distribution according to Equation~(\ref{eq:sigma-geometric}), for $\sigma_0 = 1$ and $r = 1.4$, similar to values used for corresponding fixed-scale networks. The result marked with a * has been initialised using $r = 1.48$ instead, and had the final layer scale parameter fixed during training, to prevent the corresponding kernel from exceeding the image size.}
  \label{tab:sigma_learning_performance_cifar}
\end{table*}

\begin{figure*}[hbtp]
  \begin{center}
    \begin{subfigure}{0.47\textwidth}
        \centering
        \textit{\hspace{0.5cm} Learned scale values for regular Fashion-MNIST}
        \includegraphics[width=\linewidth]{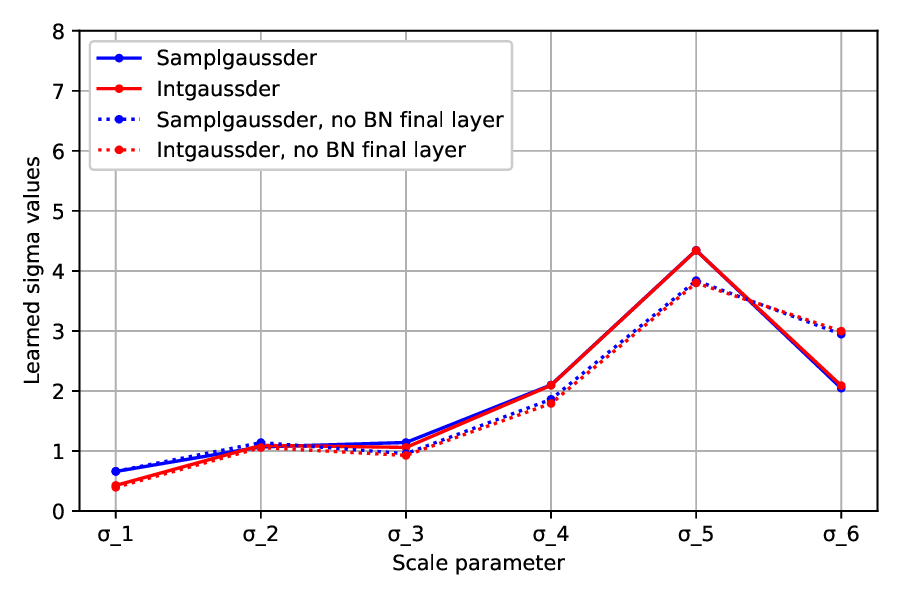}
        \label{fig:subfigb-5}
    \end{subfigure}
    \begin{subfigure}{0.47\textwidth}
        \centering
        \textit{\hspace{0.5cm} Learned scale values for regular CIFAR-10}
        \includegraphics[width=\linewidth]{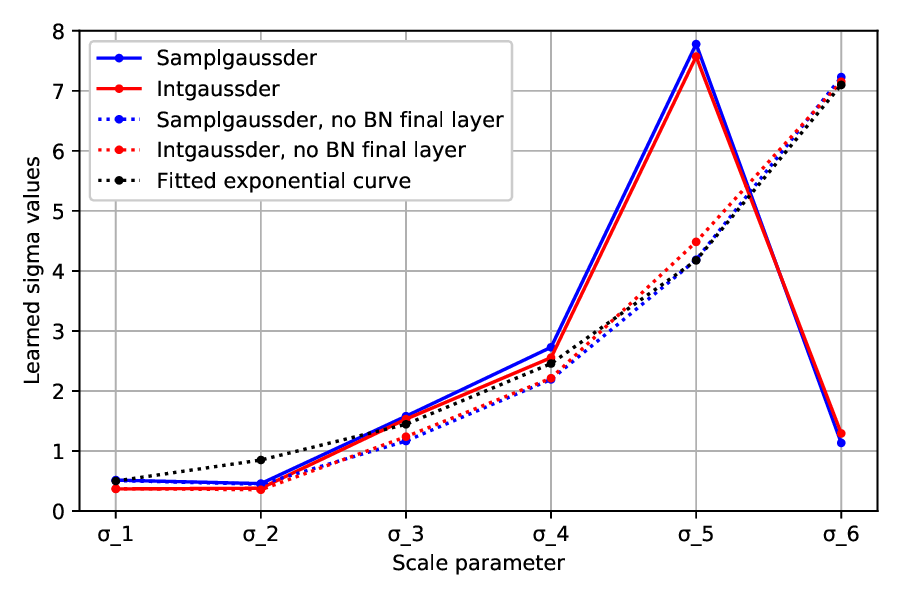}
        \label{fig:subfiga-6}
    \end{subfigure}
  \end{center}
  
  \vspace{-5mm}
  
\caption{Visualisation of the learned scale parameters for the different layers in 6-layer single-scale-channel GaussDerNets, with the continuous Gaussian derivatives approximated by either sampled Gaussian derivatives or integrated Gaussian derivatives, and then applied to either (left) the regular Fashion-MNIST dataset (with images of size $28 \times 28$ pixels) or (right) the regular CIFAR-10 dataset (with images of size $32 \times 32$ pixels). For each discrete derivative approximation method, results are shown both when using a batch normalisation layer after the last Gaussian derivative layer in the network (the solid blue and the solid red curves) or when not using any batch normalisation layer after the last Gaussian derivative layer (the dashed blue and the dashed red curves, which mostly overlap). Additionally, we show a fitted exponential curve defined by the expression $\frac{1}{2} \times 1.7^{k-1}$, with $k$ denoting the layer number, which approximates the geometrically distributed scale parameters. In all the cases, the scale learning was initialised with a geometric distribution of the scale values according to Equation~(\ref{eq:sigma-geometric}), and then trained with the scale learning rate $\eta_{\sigma} = 0.001$. As can be seen from these graphs, the learned scale parameters do increase from lower to higher layers, and with a resulting distribution that shares some qualitative similarities to a geometric distribution of the scale levels. For the last layer, there is, however, a drop in the scale values, which is also markedly stronger when batch normalisation is used after the last Gaussian derivative layer.}
\label{fig-sigma_learning_comparison}
\end{figure*}

\subsection{Learning of the scale parameters}
\label{sec-sigma-learning}
In this section, we will investigate how learning of the scale levels during training affects the performance and the distribution of the scale parameters of the single-scale-channel GaussDerNet, in order to determine how this approach differs from training using the fixed logarithmic spacing of scales between the layers, that the default network is defined with, on the regular Fashion-MNIST dataset and the regular CIFAR-10 dataset (without additional scaling transformations).

We trained single-scale-channel GaussDerNets on each of the datasets, with the learning of scale parameters implemented as outlined in Section~\ref{sec-gaussdernet-sig-lr}, with no coupling of the scale parameters between adjacent layers. We compared the performance of networks using different discretisation methods, to investigate what effect different choices of discretisation will have on the learned scale parameters.

The discretisation methods considered for this experiment were: (i)~the sampled Gaussian derivative operators, given by Equation~(\ref{eq:sampled-Gaussian-kernel}), and (ii)~the integrated Gaussian derivative operators, given by Equation~(\ref{eq:integrated-Gaussian-kernel}). For comparison, as in the previous section, we also considered the hybrid, meaning central difference based, discretisation approaches that require much less computational work than methods (i) and (ii): (iii)~the sampled Gaussian kernel, and (iv)~the integrated Gaussian kernel, with both of these methods complemented by central difference operators to obtain discrete derivative approximations. 

The training was done as described in Section~\ref{sec-train-proced}, for the rescaled versions of the Fashion-MNIST and the CIFAR-10 datasets respectively, with the initialisation using the initial scale value $\sigma_0 = 1$ and the relative scale ratio $r = 1.28$ for the regular Fashion-MNIST dataset, and the initial scale value $\sigma_0 = 1$ and the relative scale ratio $r = 1.4$ for the regular CIFAR-10 dataset. Central pixel extraction was used in each network, with the scale parameters initialised using a geometric distribution given by Equation~(\ref{eq:sigma-geometric}). Importantly, the training was performed with the scale parameters $\sigma$ having their own learning rate $\eta_{\sigma} \in \{0.0001, 0.001, 0.01 \}$, separate from the learning rate for the weights $\eta_{C}$, to study how different relative learning rates affect the distribution of the learned scale parameters, something that might not have been addressed in other works dealing with learning of the scale parameters. The scale parameters also had their own separate weight decay value, set to 0.025.
The experiments were performed for two versions of the GaussDerNet, one with a batch normalisation stage after the final layer, and one without, to see how such an architectural choice affects the learned scale parameter values.

Tables~\ref{tab:sigma_learning_performance_fashion} and~\ref{tab:sigma_learning_performance_cifar} show the results of these experiments,\footnote{Once again, we report the results for networks using reflection padding here, to be as close as possible to the experimental situation to the previously treated case for the rescaled CIFAR-10 dataset. We get 90.27\% average test accuracy across 5 runs, for the GaussDerNet based on the sampled Gaussian derivative operators, when trained with zero padding and learning of the scale parameters with the learning rate $\eta_{\sigma} = 0.0001$.} as the average value as well as the unbiased standard deviation of the test accuracy over 5 runs. By directly comparing the average accuracies, we can see that for the Fashion-MNIST dataset, the learning rate $\eta_{\sigma} = 0.01$ generally leads to the best results, while for the CIFAR-10 dataset using a somewhat smaller scale learning rate compared to $\eta_{C}$ often results in good performance, with networks trained using $\eta_{\sigma}=0.001$ and $\eta_{\sigma}=0.0001$ achieving the best performance for several of the discretisation methods, however, the optimal $\eta_{\sigma}$ value varies somewhat between the different discretisation approaches and datasets.
The value of the scale learning rate has a strong influence on how the scale parameters converge, and therefore needs to be chosen appropriately, to allow a balanced and stable training process. In our experiments, we had no need to constrain the scale parameters, except for one case in Table~\ref{tab:sigma_learning_performance_cifar}, where for the largest learning rate $\eta_{\sigma} = 0.01$ the scale parameter in the final layer was fixed, for the filter size to not exceeded the size of the image domain.

From Table~\ref{tab:sigma_learning_performance_fashion}, we can see that for the regular Fashion-MNIST dataset, the networks with learned scale parameters achieve comparable performance to the best performing networks with fixed scale parameters, shown in Table~\ref{tab:new_der_fashion_table}, while from Table~\ref{tab:sigma_learning_performance_cifar} we can see that even for the regular CIFAR-10 dataset, which contains natural images, learning of the scale parameters leads to similar, or even in some cases also somewhat better, results compared to the corresponding best performing networks trained using fixed scale parameters, shown in Table~\ref{tab:new_der_cifar_table}. This observed improvement in performance, compared to fixed scale parameter models for the different discrete derivative approximation method, is partly caused by the learned scale value in final layer being fairly huge, unlike for fixed parameter networks which are defined to use comparatively much smaller scale values in the higher layers. Additionally, some of the networks with learned scale parameters based on central difference discretisation methods even achieve comparable performance to networks based on either sampled Gaussian derivatives or integrated Gaussian derivatives.

From Figure~\ref{fig-sigma_learning_comparison}, we can see that for both of the datasets, the learned scale parameter values are small in the first layers, then gradually increase for the higher layers. The fact that the network gravitates towards such small values of the scale parameter in the early layers likely signifies that the model prefers to learn to perform the computations based on fine-scale image information, focusing on the local textures in the image, which is particularly important for the CIFAR-10 dataset. This kind of behaviour is not unexpected, as deep vision networks typically capture fine-grained image structures in the early layers, and a gradual increase in the learned scale parameter values in the higher layers has also been observed by Pintea et al. (\citeyear{PinTomGoeLooGem21-IP}), when training networks together with scale parameter tuning.

When using a network with a batch normalisation stage after the final layer, there is often a clear drop in the value of the scale parameter in the final layer. This phenomenon is not observed for the network trained on the regular CIFAR-10 dataset without a batch normalisation stage after the final layer, as can be seen in the right side of Figure~\ref{fig-sigma_learning_comparison}. In fact, we can see that for that network, the learned distributions of the scale parameters can for the most part be fitted with an exponential curve, in line with the previously assumed prior of a geometric distribution.

There is no significant difference between the final learned distributions of the scale parameters for the networks based on either the sampled Gaussian derivatives or the integrated Gaussian derivatives, the main difference between these being that the integrated GaussDerNet tends to learn slightly smaller value of $\sigma_{1}$ in the first layer. This is likely due to the integrated Gaussian approach introducing a certain scale offset $\Delta s_{\integrated}$ at finer scales, as discussed in Section~\ref{sec-der-gauss-kernels}.

We leave the investigation of scale parameter learning in a multi-scale-channel network, including its effect on scale generalisation performance, for future work. 

In Appendix~A.2 in the supplementary material, we compare the scale values learned by the scale learning methodology, and the spatial extent for the discrete kernels that they lead to, for the different types of discretisation methods for the Gaussian derivative operators. In Appendix~A.3.4 in the supplementary material, we additionally perform a statistical analysis of the results in this section, in order to assess the statistical significance of the differences in performance for the different ways of performing learning of the scale values.

\section{Visualisations of Gaussian derivative networks}
\label{sec-network-function-visualisation}
In this section, we will show visualisations of how the multi-scale-channel GaussDerNet functions, in order to better understand the behaviour of the model, and to provide explainability for what the network has learned. We will specifically inspect (i)~the learned effective filters of multi-scale-channel GaussDerNets based on either the 2-jet or the 3-jet, trained on the rescaled CIFAR-10 dataset, and (ii)~the activation maps in the final layer of a multi-scale-channel GaussDerNet trained on the rescaled Fashion-MNIST with translations dataset, and the rescaled CIFAR-10 dataset.  

\subsection{Activation maps}
\label{sec-activation-maps}
For the theoretically motivated experimental investigations in this paper, we have deliberately not introduced any spatial subsampling factor in the network, in order to avoid additional discretisation issues and to work with networks that as closely as possible reproduce the computational functions of the Gaussian derivative operators. For this reason, it is therefore possible to visualise what image structures the different scale channels respond to, when applied to different image data, by mere inspection of the feature maps, in the following referred to as activation maps.

In this section, we will visualise the ability of the multi-scale-channel GaussDerNets to operate at different scales, by generating and then inspecting such activation maps in the final layer of the model, for either the rescaled Fashion-MNIST with translations dataset or the rescaled CIFAR-10 dataset. 

For the rescaled Fashion-MNIST with translations dataset, the activation maps were generated using the multi-scale-channel network trained according to Section~\ref{sec-spatial-max-pooling}, specifically the model using max pooling over scales and spatial max pooling. For the rescaled CIFAR-10 dataset, the activation maps were generated using the multi-scale-channel network trained according to Section~\ref{sec-cifar10-experiments}, specifically the model with average pooling over scales and central pixel extraction. The network for the rescaled Fashion-MNIST with translations dataset uses the relative scale ratio $r=1.28$ and the scale channels with the initial scale values $\sigma_{i,0} \in \{1/(2\sqrt{2}), 1/2$, $ 1/\sqrt{2}, 1, \sqrt{2}, 2$, $2\sqrt{2} \}$, while the network for the rescaled CIFAR-10 dataset uses the relative scale ratio $r=1.45$ and the scale channels with the initial scale values $\sigma_{i,0} \in \{1/(2\sqrt{2})$, $1/2, 1/\sqrt{2}, 1, \sqrt{2}, 2\}$.

For the rescaled Fashion-MNIST with translations dataset, twelve illustrative images correctly predicted by the corresponding network were chosen for generating activation maps. Specifically, we picked one image per class for the classes ``trouser'', ``bag'', ``sandal'' and ``T-shirt/top'', respectively, from the dataset for the size factor 1, as well as corresponding rescaled images for the same objects from the datasets for the size factors 1/2 and 2. As the model uses max pooling over scales, the activation maps were directly obtained by extracting the output of the final layer in the winning scale channel, for the feature channel corresponding to the correct class of the input image.

For the rescaled CIFAR-10 dataset, five illustrative images correctly predicted by the corresponding network were chosen for generating activation maps. We picked one image per class from the classes ``airplane'', ``cat'', ``deer'', ``frog'' and ``boat'', from the dataset for the size factor 2, because these samples do not contain mirror extensions. As the model here uses average pooling over scales, the activation maps were obtained by extracting the output of the final layer for each one of the scale channels with the initial scale values $\sigma_{0} = \{1/2,1,2\}$, for the feature channel corresponding to the correct class of the input image, in order to study how the responses at different scales contribute to the final prediction.

\begin{figure*}[htbp]
  \begin{center}
    \begin{subfigure}{0.47\textwidth}      
        \centering
        \raggedright
        \textit{Size factor 1/2 \hspace{0.80cm} Size factor 1 \hspace{0.95cm} Size factor 2}
        \includegraphics[width=0.24\textwidth]{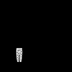}
        \hspace{6.7mm}
        \includegraphics[width=0.24\textwidth]{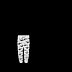}
        \hspace{6.7mm}
        \includegraphics[width=0.24\textwidth]{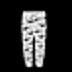}
        \\
        \includegraphics[trim=60 30 65 30, clip, width=0.305\textwidth]{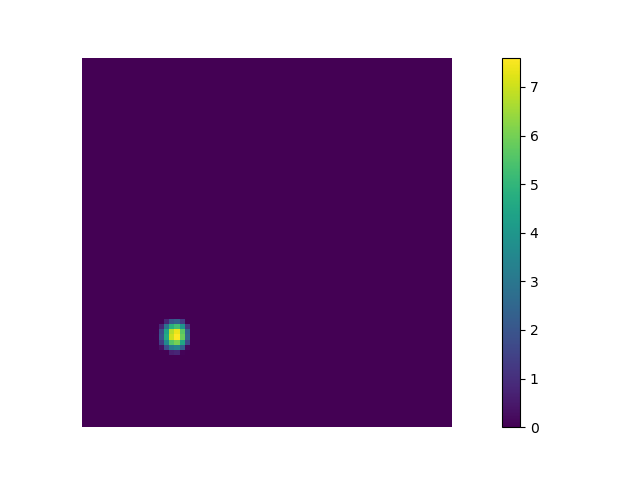}
        \hspace{1.5mm}
        \includegraphics[trim=60 30 65 30, clip, width=0.305\textwidth]{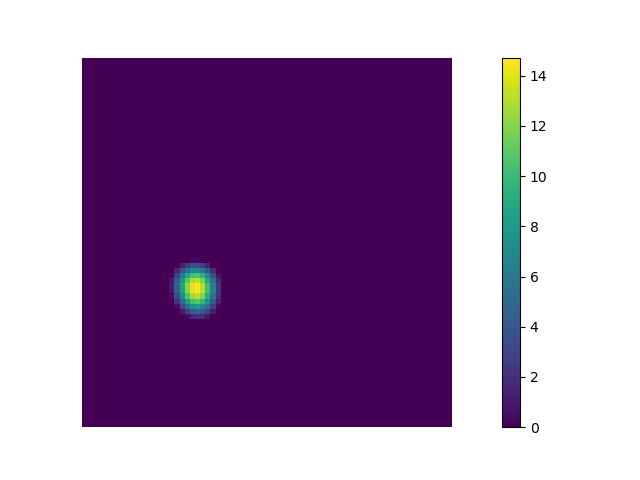}
        \hspace{1.5mm}
        \includegraphics[trim=60 30 65 30, clip, width=0.305\textwidth]{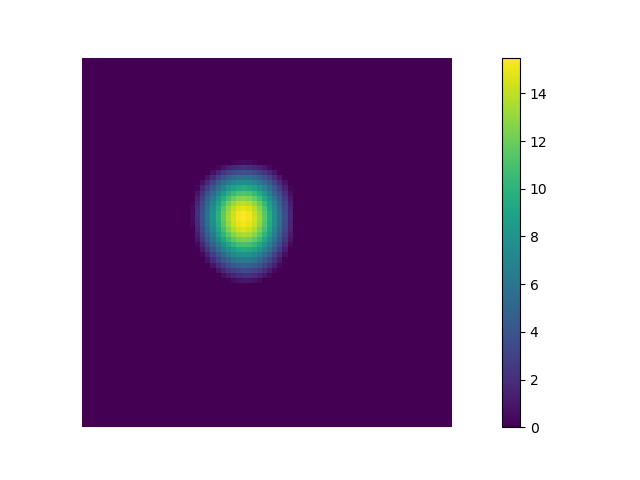}       
        \caption{Activation maps for rescaled versions of an image from the class ``trouser''. The activations are shown for the selected scale channels for the respective mutually rescaled image data, these corresponding winning scale channels having the initial scale values $\sigma_0$ (from left to right) $\sigma_{0}= \frac{1}{\sqrt{2}}$, $\sigma_{0}=1$ and $\sigma_{0}=2$, respectively.}
        \label{fig:subfig-activation-pants}
    \end{subfigure} 
    \hfill
    \begin{subfigure}{0.47\textwidth}
        \centering
        \raggedright
        \textit{Size factor 1/2 \hspace{0.80cm} Size factor 1 \hspace{0.95cm} Size factor 2}
        \includegraphics[width=0.24\textwidth]{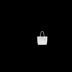}
        \hspace{6.7mm}
        \includegraphics[width=0.24\textwidth]{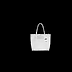}
        \hspace{6.7mm}
        \includegraphics[width=0.24\textwidth]{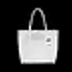}
        \\
        \includegraphics[trim=60 30 65 30, clip, width=0.305\textwidth]{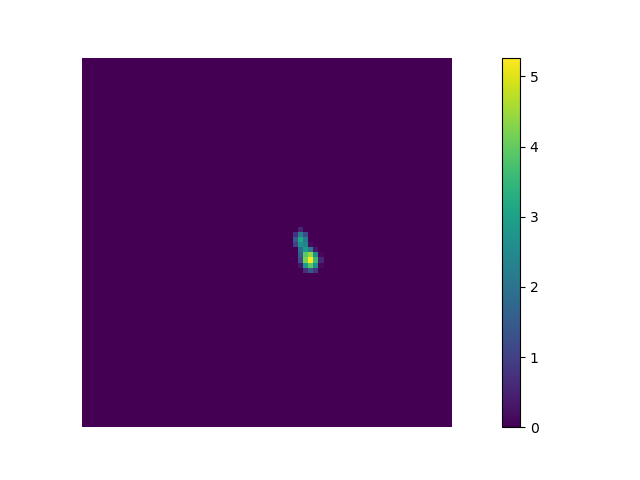}
        \hspace{1.5mm}
        \includegraphics[trim=60 30 65 30, clip, width=0.305\textwidth]{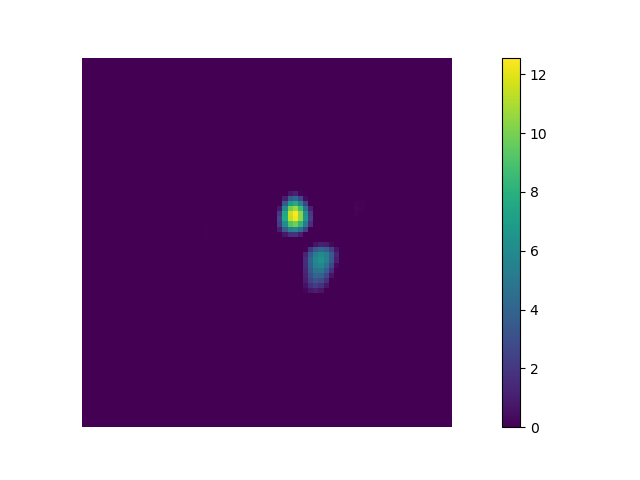}
        \hspace{1.5mm}
        \includegraphics[trim=60 30 65 30, clip, width=0.305\textwidth]{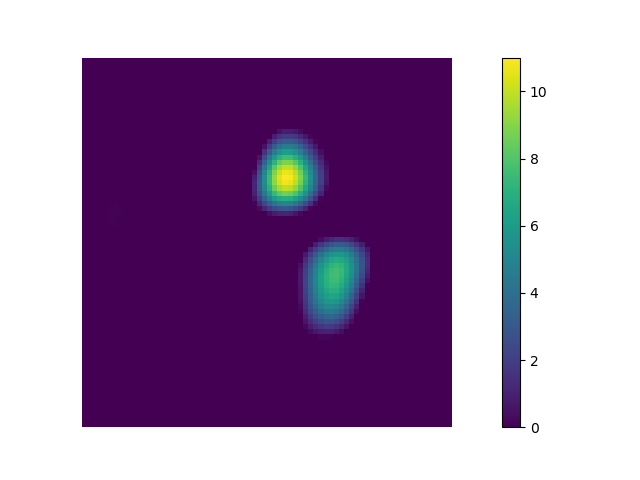} 
        \caption{Activation maps for rescaled versions of an image from the class ``bag''. The activations are shown for the selected scale channels for the respective mutually rescaled image data, these corresponding winning scale channels having the initial scale values $\sigma_0$ (from left to right) $\sigma_{0}= \frac{1}{2}$, $\sigma_{0}=\frac{1}{\sqrt{2}}$ and $\sigma_{0}=\sqrt{2}$, respectively.}
        \label{fig:subfig-activation-bag}
    \end{subfigure} 
    
    \vspace{0.5cm} 
    
    \begin{subfigure}{0.47\textwidth}
        \centering
        \raggedright
        \textit{Size factor 1/2 \hspace{0.80cm} Size factor 1 \hspace{0.95cm} Size factor 2}
        \includegraphics[width=0.24\textwidth]{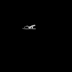}
        \hspace{6.7mm}
        \includegraphics[width=0.24\textwidth]{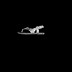}
        \hspace{6.7mm}
        \includegraphics[width=0.24\textwidth]{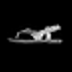}
        \\
        \includegraphics[trim=60 30 65 30, clip, width=0.305\textwidth]{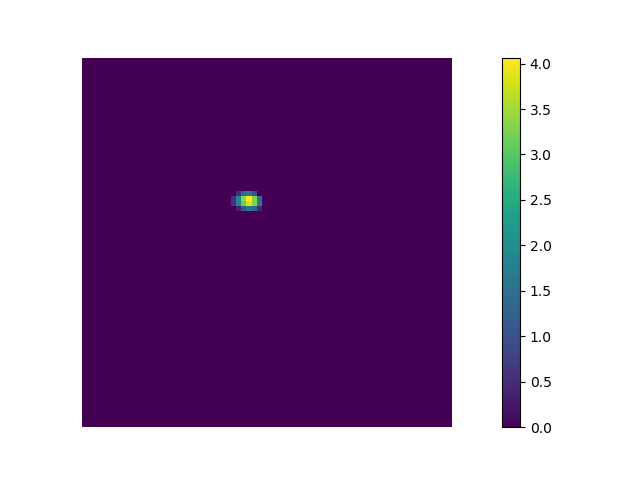}
        \hspace{1.5mm}
        \includegraphics[trim=60 30 65 30, clip, width=0.305\textwidth]{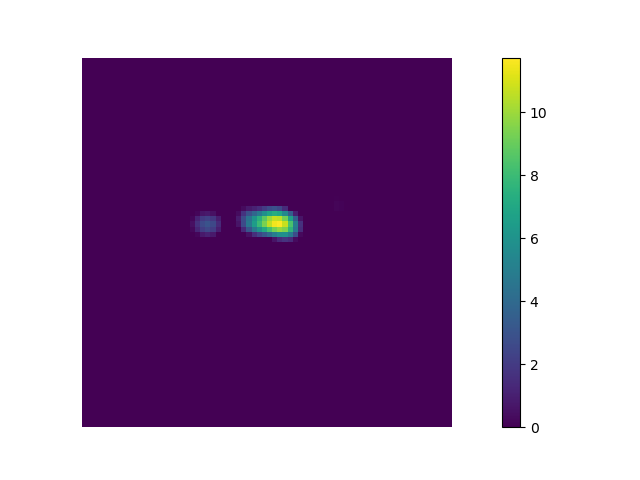}
        \hspace{1.5mm}
        \includegraphics[trim=60 30 65 30, clip, width=0.305\textwidth]{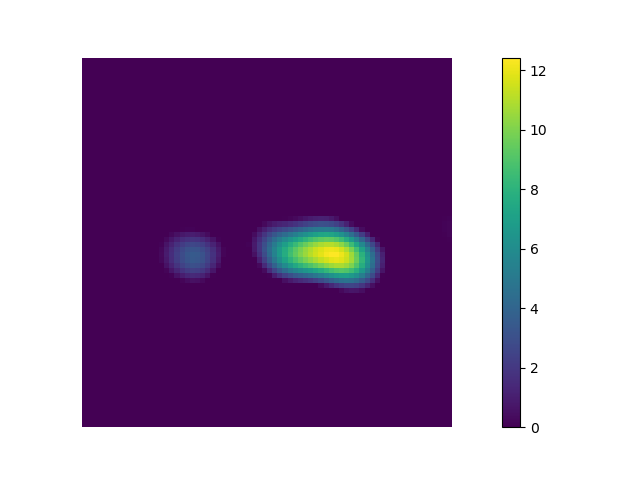} 
        \caption{Activation maps for rescaled versions of an image from the class ``sandal''. The activations are shown for the selected scale channels for the respective mutually rescaled image data, these corresponding winning scale channels having the initial scale values $\sigma_0$ (from left to right) $\sigma_{0}= \frac{1}{2}$, $\sigma_{0}=\frac{1}{\sqrt{2}}$ and $\sigma_{0}=\sqrt{2}$, respectively.}
        \label{fig:subfig-activation-shoes}
    \end{subfigure}
    \hfill
    \begin{subfigure}{0.47\textwidth}
        \centering
        \raggedright
        \textit{Size factor 1/2 \hspace{0.80cm} Size factor 1 \hspace{0.95cm} Size factor 2}
        \includegraphics[width=0.24\textwidth]{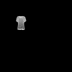}
        \hspace{6.7mm}
        \includegraphics[width=0.24\textwidth]{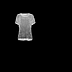}
        \hspace{6.7mm}
        \includegraphics[width=0.24\textwidth]{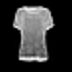}
        \\
        \includegraphics[trim=60 30 65 30, clip, width=0.305\textwidth]{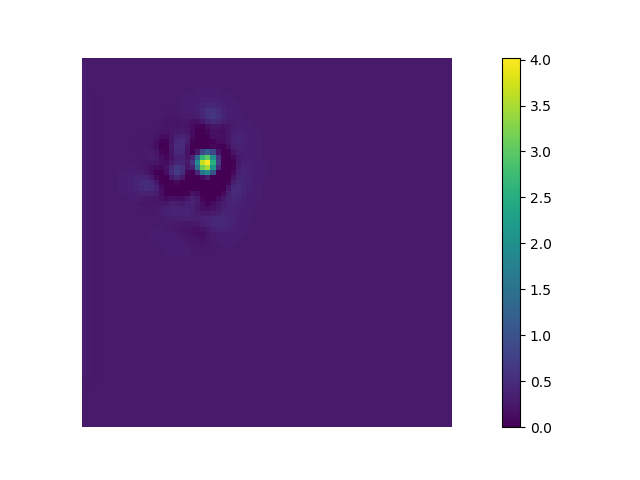}
        \hspace{1.5mm}
        \includegraphics[trim=60 30 65 30, clip, width=0.305\textwidth]{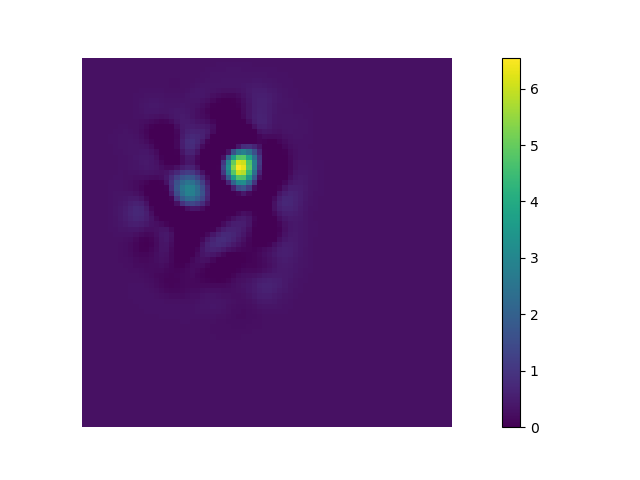}
        \hspace{1.5mm}
        \includegraphics[trim=60 30 65 30, clip, width=0.305\textwidth]{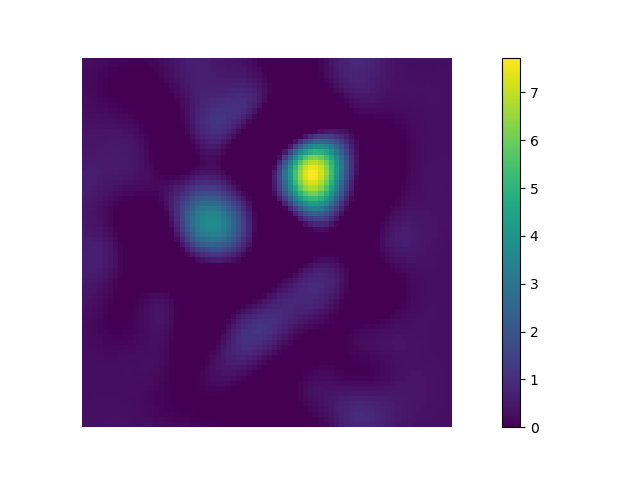} 
        \caption{Activation maps for rescaled versions of an image from the class ``T-shirt/top''. The activations are shown for the selected scale channels for the respective mutually rescaled image data, these corresponding winning scale channels having the initial scale values $\sigma_0$ (from left to right) $\sigma_{0}= \frac{1}{2}$, $\sigma_{0}=\frac{1}{\sqrt{2}}$ and $\sigma_{0}=\sqrt{2}$, respectively.}
        \label{fig:subfig-activation-shirt}
    \end{subfigure}
  \end{center}
\caption{Activation maps generated from the final layer of a multi-scale-channel GaussDerNet, based on combined spatial max pooling and max pooling over scales, for correctly classified sample images at different scales from the Fashion-MNIST with translations dataset. For each input image, the activation map from the correct feature channel of the winning scale channel is shown. These activation maps thus visualise the spatial position of the response region in the feature channel corresponding to the correct class, in the final layer of the automatically selected scale channel, for each one of the given input images. These images have, in turn, been selected from the test sets, for each one of the size factors 1/2, 1 and 2, for the four classes: (a)~''trouser'', (b)~''bag'', (c)~''sandal'', and (d)~''T-shirt/top''. As can be seen from the activation maps, the model can reliably localise the spatial position of objects in these images, and identify subsets of image structures crucial for a correct classification of each one of the objects, such as the handles of the bag, the sleeves of the T-shirt, or the soles of the sandal.}
\label{fig-activations-new-fashion}
\end{figure*}

\begin{figure*}[htbp]
  \begin{center}
    \begin{subfigure}{0.9\textwidth}      
        \centering
        \raggedright
        \textit{ \hspace{0.4cm} Input image \hspace{1.3cm} Scale channel $\sigma_{0}=1/2$ \hspace{0.9cm} Scale channel $\sigma_{0}=1$ \hspace{1.1cm} Scale channel $\sigma_{0}=2$}
        \raisebox{0.915mm}{\includegraphics[width=0.178\textwidth]{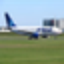}}
        \hspace{2.3mm}
        \includegraphics[trim=30 30 30 30, clip, width=0.27\textwidth]{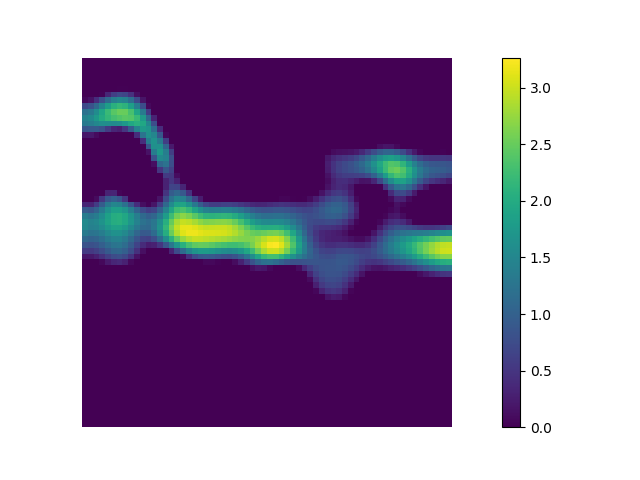}
        \hspace{-3.1mm}
        \includegraphics[trim=30 30 30 30, clip, width=0.27\textwidth]{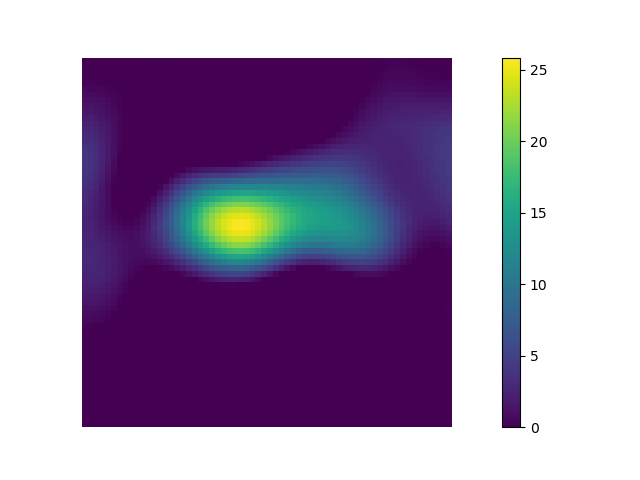}
        \hspace{-3.1mm}
        \includegraphics[trim=30 30 30 30, clip, width=0.27\textwidth]{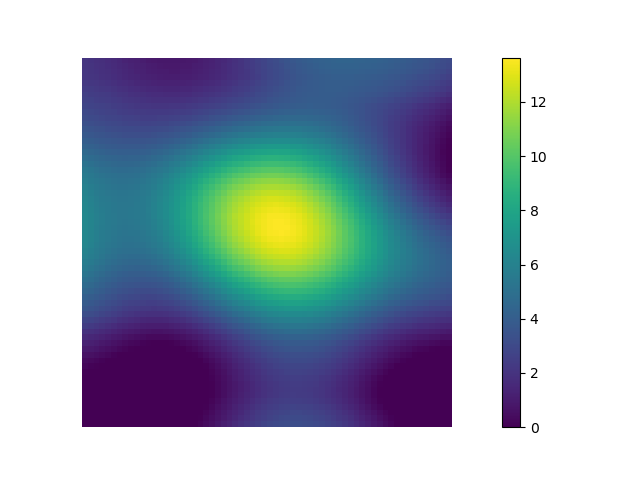}
        \caption{Activation maps for a sample image from the class ``airplane''}
        \label{fig:subfig-activation-airplane}
    \end{subfigure} 
    
    \vspace{0.4cm} 
    
    \begin{subfigure}{0.9\textwidth}      
        \centering
        \raggedright
        \textit{ \hspace{0.4cm} Input image \hspace{1.3cm} Scale channel $\sigma_{0}=1/2$ \hspace{0.9cm} Scale channel $\sigma_{0}=1$ \hspace{1.1cm} Scale channel $\sigma_{0}=2$}
        \raisebox{0.915mm}{\includegraphics[width=0.178\textwidth]{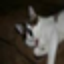}}
        \hspace{2.3mm}
        \includegraphics[trim=30 30 30 30, clip, width=0.27\textwidth]{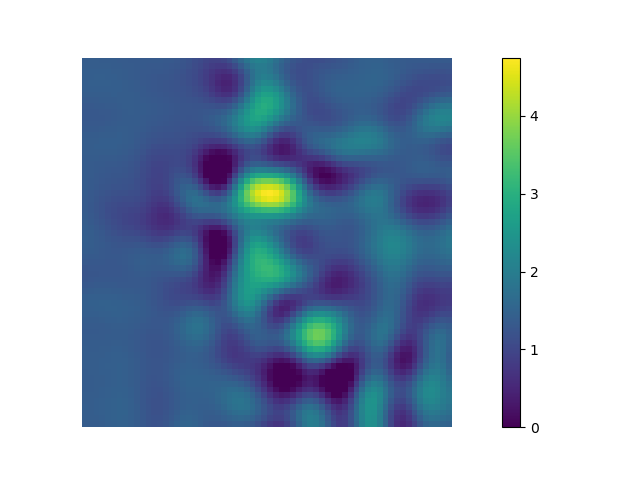}
        \hspace{-3.1mm}
        \includegraphics[trim=30 30 30 30, clip, width=0.27\textwidth]{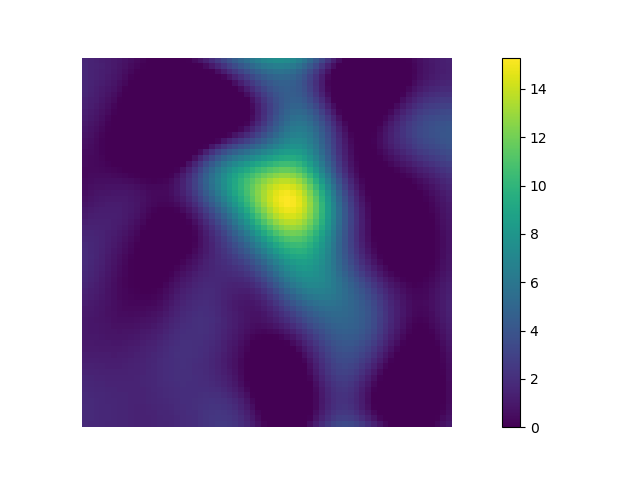}
        \hspace{-3.1mm}
        \includegraphics[trim=30 30 30 30, clip, width=0.27\textwidth]{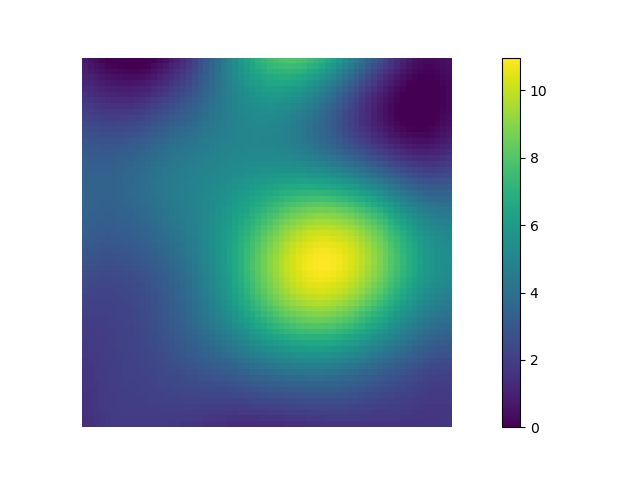}
        \caption{Activation maps for a sample image from the class ``cat''}
        \label{fig:subfig-activation-cat}
    \end{subfigure} 
    
    \vspace{0.4cm} 
    
    \begin{subfigure}{0.9\textwidth}      
        \centering
        \raggedright
        \textit{ \hspace{0.4cm} Input image \hspace{1.3cm} Scale channel $\sigma_{0}=1/2$ \hspace{0.9cm} Scale channel $\sigma_{0}=1$ \hspace{1.1cm} Scale channel $\sigma_{0}=2$}
        \raisebox{0.915mm}{\includegraphics[width=0.178\textwidth]{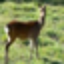}}
        \hspace{2.3mm}
        \includegraphics[trim=30 30 30 30, clip, width=0.27\textwidth]{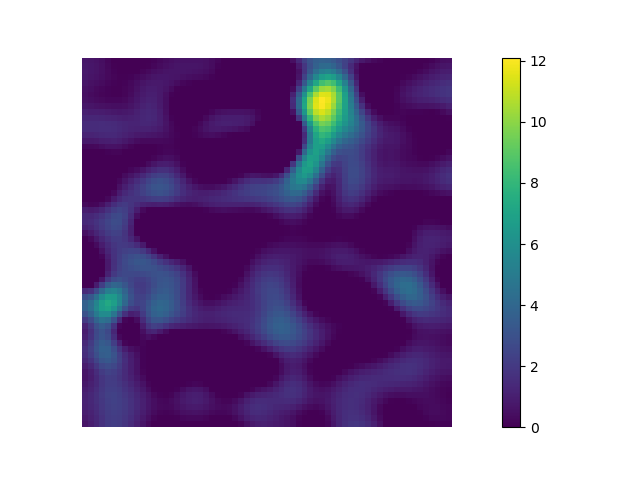}
        \hspace{-3.1mm}
        \includegraphics[trim=30 30 30 30, clip, width=0.27\textwidth]{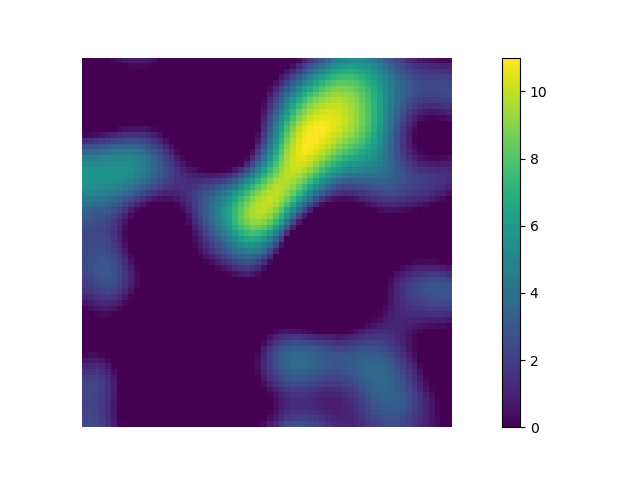}
        \hspace{-3.1mm}
        \includegraphics[trim=30 30 30 30, clip, width=0.27\textwidth]{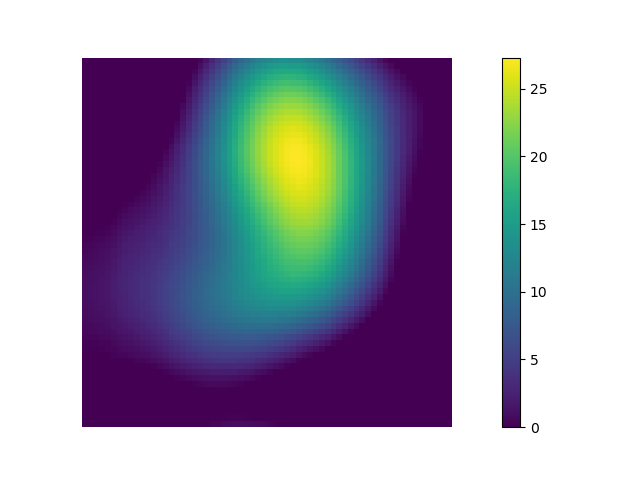}
        \caption{Activation maps for a sample image from the class ``deer''}
        \label{fig:subfig-activation-deer}
    \end{subfigure} 
  
    \vspace{0.4cm} 
    
    \begin{subfigure}{0.9\textwidth}      
        \centering
        \raggedright
        \textit{ \hspace{0.4cm} Input image \hspace{1.3cm} Scale channel $\sigma_{0}=1/2$ \hspace{0.9cm} Scale channel $\sigma_{0}=1$ \hspace{1.1cm} Scale channel $\sigma_{0}=2$}
        \raisebox{0.915mm}{\includegraphics[width=0.178\textwidth]{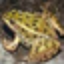}}
        \hspace{2.3mm}
        \includegraphics[trim=30 30 30 30, clip, width=0.27\textwidth]{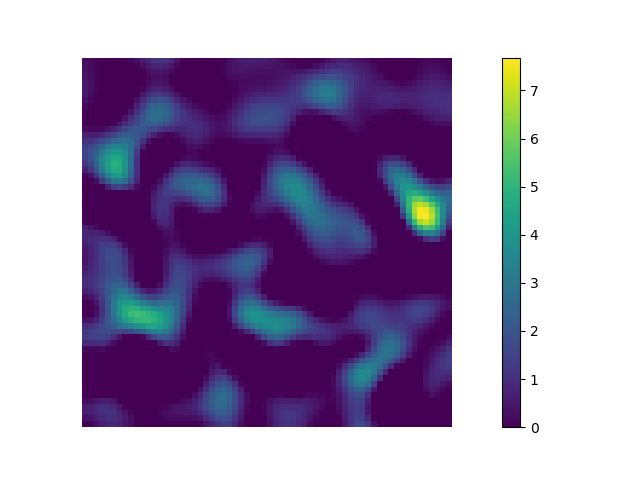}
        \hspace{-3.1mm}
        \includegraphics[trim=30 30 30 30, clip, width=0.27\textwidth]{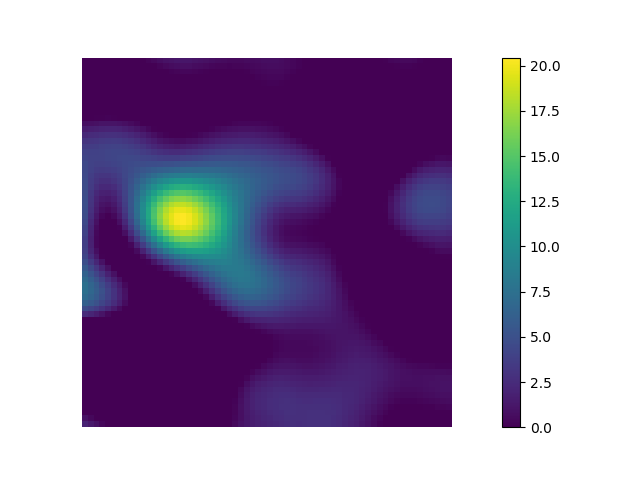}
        \hspace{-3.1mm}
        \includegraphics[trim=30 30 30 30, clip, width=0.27\textwidth]{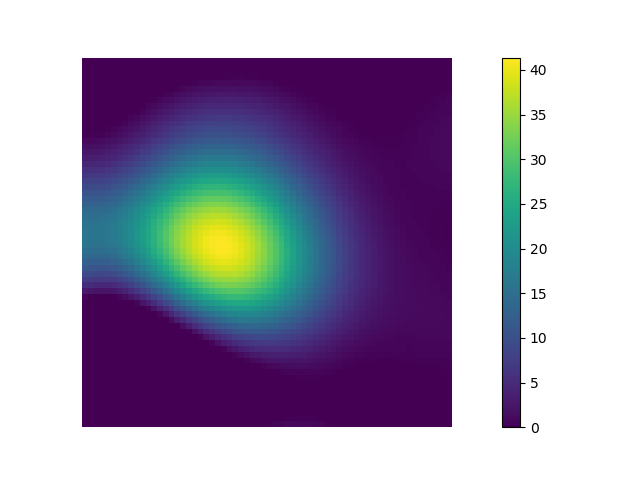}
        \caption{Activation maps for a sample image from the class ``frog''}
        \label{fig:subfig-activation-frog}
    \end{subfigure} 
    
    \vspace{0.4cm} 
    
    \begin{subfigure}{0.9\textwidth}      
        \centering
        \raggedright
        \textit{ \hspace{0.4cm} Input image \hspace{1.3cm} Scale channel $\sigma_{0}=1/2$ \hspace{0.9cm} Scale channel $\sigma_{0}=1$ \hspace{1.1cm} Scale channel $\sigma_{0}=2$}
        \raisebox{0.915mm}{\includegraphics[width=0.178\textwidth]{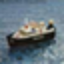}}
        \hspace{2.3mm}
        \includegraphics[trim=30 30 30 30, clip, width=0.27\textwidth]{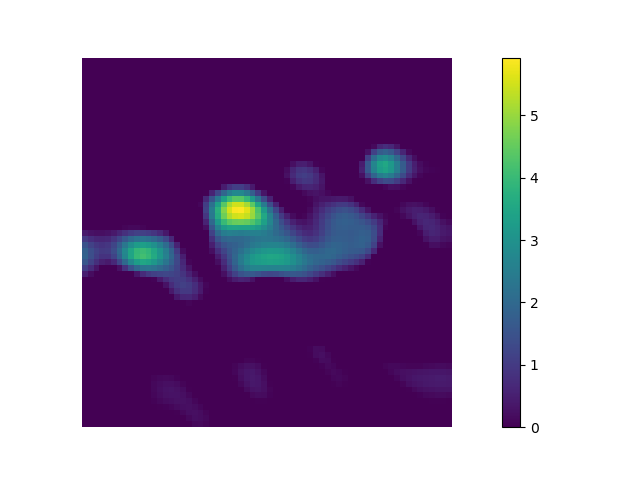}
        \hspace{-3.1mm}
        \includegraphics[trim=30 30 30 30, clip, width=0.27\textwidth]{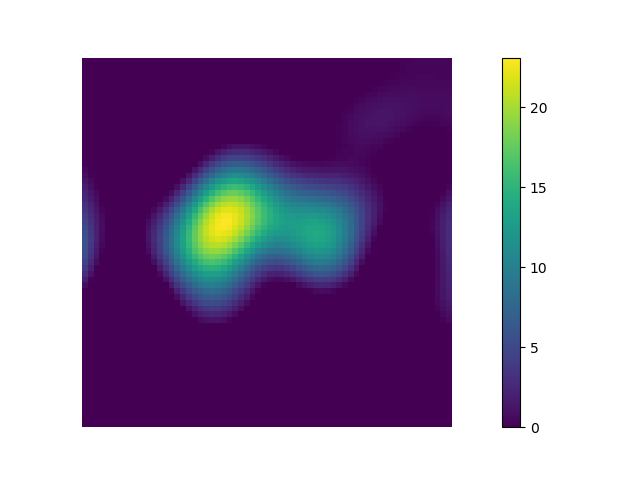}
        \hspace{-3.1mm}
        \includegraphics[trim=30 30 30 30, clip, width=0.27\textwidth]{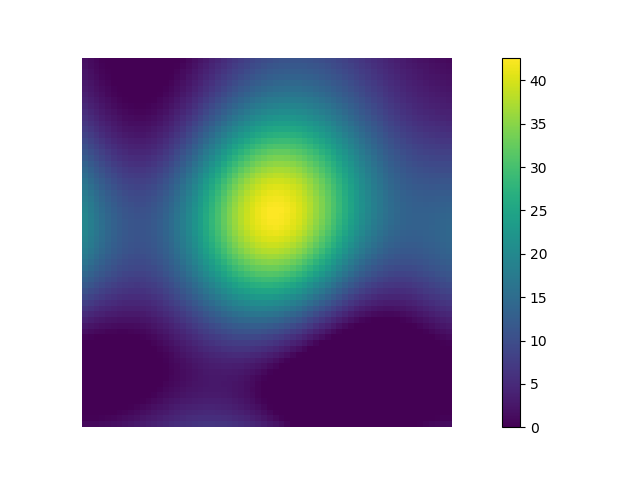}
        \caption{Activation maps for a sample image from the class ``boat''}
        \label{fig:subfig-activation-boat}
    \end{subfigure} 
  \end{center}
\caption{Activation maps generated from the final layer of a multi-scale-channel network, using central pixel extraction for spatial selection and average pooling over scales for scale selection, for correctly classified sample images from the CIFAR-10 dataset. For each input image, the activation maps of the correct feature channel from the scale channels with initial scale values $\sigma_{0} = \{1/2, 1, 2\}$ are shown. These activation maps visualise the spatial positions of the response regions in the feature channel corresponding to the correct class, in the final layer of the here manually chosen subsets of scale channels, for each input image. The images have been selected from the test set for the size factor 2, for the five classes: (a)~``airplane'', (b)~``cat'', (c)~``deer'', (d)~``frog'' and (e)~``boat''. As can be seen from the results, the different scale channels respond to different type of information at the different scales, with the finer-scale scale channels responding to image details of smaller size, whereas the coarser-scale scale channels, for which both the amount of smoothing and the size receptive field is larger, respond to the entire objects as a whole.}
\label{fig-activations-new-cifar}
\end{figure*}

Figures~\ref{fig-activations-new-fashion} and~\ref{fig-activations-new-cifar} show the resulting activation maps. From Figure~\ref{fig-activations-new-fashion}, we can see that for the rescaled Fashion-MNIST with translations dataset, the network using spatial max pooling combined with max pooling over scales can reliably localise the spatial positions of objects, also when the objects are not centred. The network also demonstrates scale-covariant properties, as it can automatically select the scale channel with its initial scale value $\sigma_{0}$ approximately proportional to the size of the object; the multiplicative relations between the selected scale levels being in agreement with the transformation property between the scale levels in Equation~(\ref{eq-sigma-sc-transf}) at coarse scales, but holding only approximately for fine scale levels, where the discretisation effects begin to affect model performance.
This demonstrates the ability of the GaussDerNets to operate at different scales, in agreement with both the previously reported results in Section~\ref{sec-spatial-max-pooling} and the scale-covariant properties of the network architecture, according to Section~\ref{sec-scale-covariance-properties}.

The interesting image structures, that the GaussDerNet is trained to respond to, give rise to responses in terms of local peaks in the activation maps in corresponding scale channels.
We can also see that, despite relatively small and rounded receptive field responses in the final layer, the model can correctly classify images by detecting distinct structures crucial for correct classification of the objects, such as the handles for the bag in Figure~\ref{fig-activations-new-fashion}(\subref{fig:subfig-activation-bag}), or the sleeves for the T-shirt in Figure~\ref{fig-activations-new-fashion}(\subref{fig:subfig-activation-shirt}). 
In particular, Figure~\ref{fig-activations-new-fashion}(\subref{fig:subfig-activation-shoes}), which shows the activations for sandals of different sizes, demonstrates that the model focuses on the heel sole, the ankle and the frontal strap, which are the main characteristics for the class ``sandal''. The increased intensity of the activation in the regions around the heel and the ankle is important, because these are among the key areas for differentiating between the shoe classes.

Figure~\ref{fig-activations-new-cifar} shows that, for the rescaled CIFAR-10 dataset, the different scale channels detect information at different scales. The activation maps for the finer-scale scale channels can be seen to respond to image structures of smaller size, such as small shapes and local changes in pixel intensity, reacting most intensely to distinct structures that uniquely identify the objects, such as heads, faces, limbs, or parts with appropriate textures. The coarser-scale scale channels are, in turn, able to localise larger shapes, due to increased amounts of smoothing by the kernels and receptive fields of larger size, with the largest scale channels focusing on entire objects.

For some of the images, different scale channels can be seen to give rise to multiple spatial maxima at different spatial locations in the activation maps, demonstrating that the average pooling operation over scales can combine information from multiple scales and from different parts of the image, as also would be expected from the theory.  

For a given input image, the contributions to the predictions in the final layer of the network may differ in strength between the scale channels; only the most dominant activations will influence the decisions in the spatial selection and the scale selection stages, where in each scale channel the extrema are detected and selected. The magnitude of these contributions can be seen in the color bars. For the images in Figure~\ref{fig-activations-new-cifar}, the middle-scale and coarser-scale scale channels tend to be dominant, which is consistent with the idea that the GaussDerNet should select coarser scale levels as the size of the image structures increases; a trend already discussed in Section~\ref{sec-cifar10-scale-sel-hist}. In some of the activation maps, however, we observe that the contributions from scale channels at finer scales can occasionally be significant.

\begin{figure*}[htbp]
  \begin{center}
    \begin{tabular}{ccccccccccc}
      \raisebox{3\height}{Layer 6} &
      \hspace{-5mm}
      \includegraphics[width=0.1\textwidth]{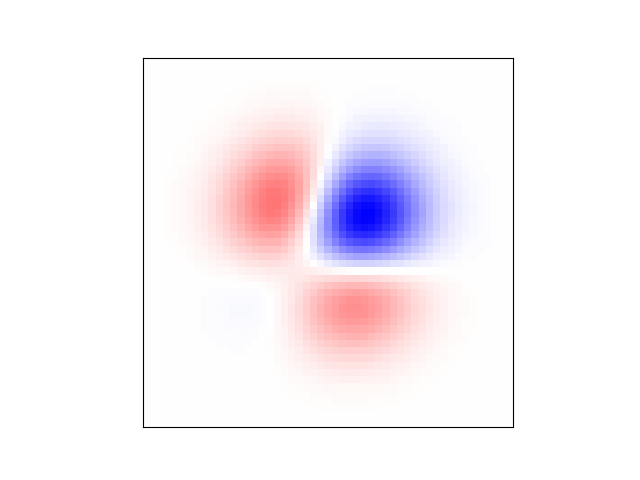} &
      \hspace{-7mm}
      \includegraphics[width=0.1\textwidth]{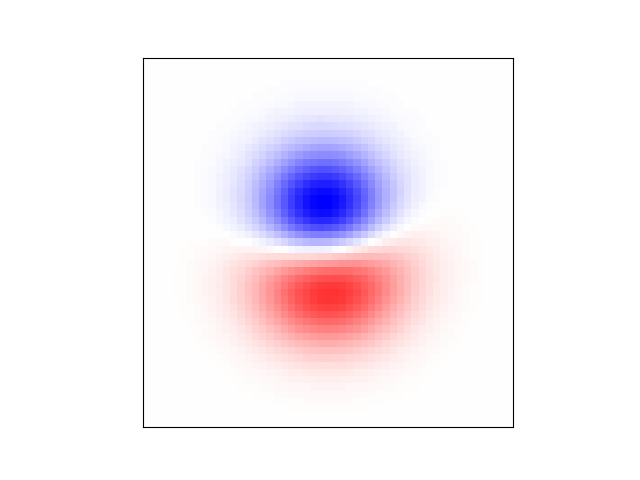} &
      \hspace{-7mm}
      \includegraphics[width=0.1\textwidth]{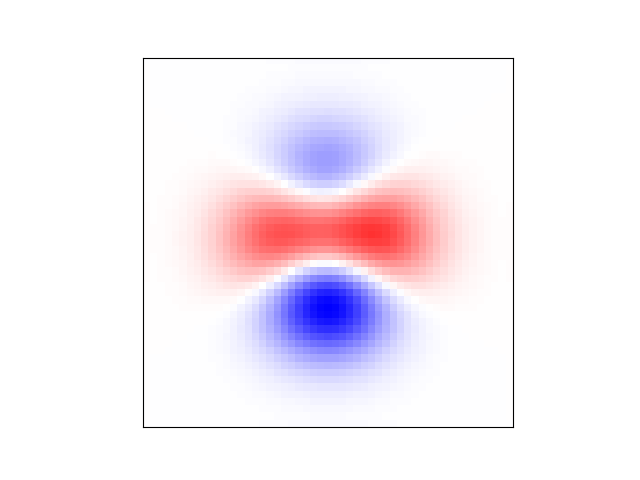} &
      \hspace{-7mm}
      \includegraphics[width=0.1\textwidth]{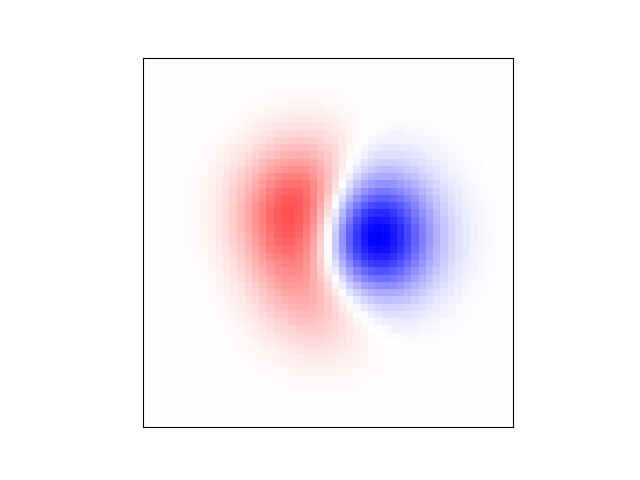} &
      \hspace{-7mm}
      \includegraphics[width=0.1\textwidth]{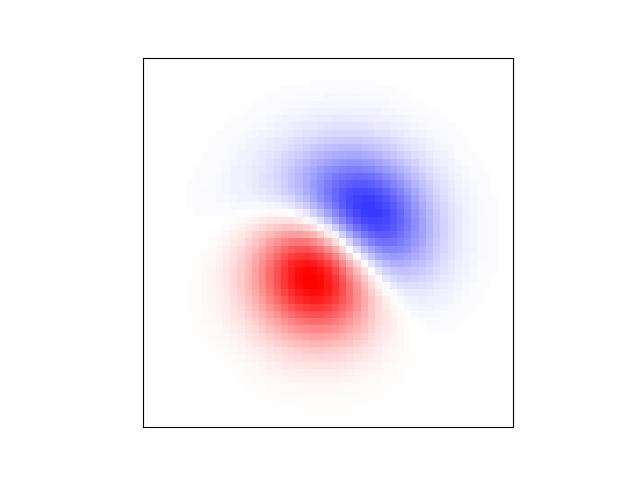} &
      \hspace{-7mm}
      \includegraphics[width=0.1\textwidth]{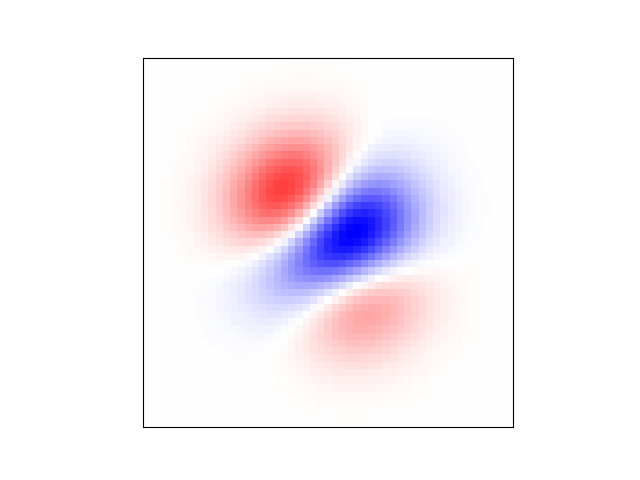} &
      \hspace{-7mm}
      \includegraphics[width=0.1\textwidth]{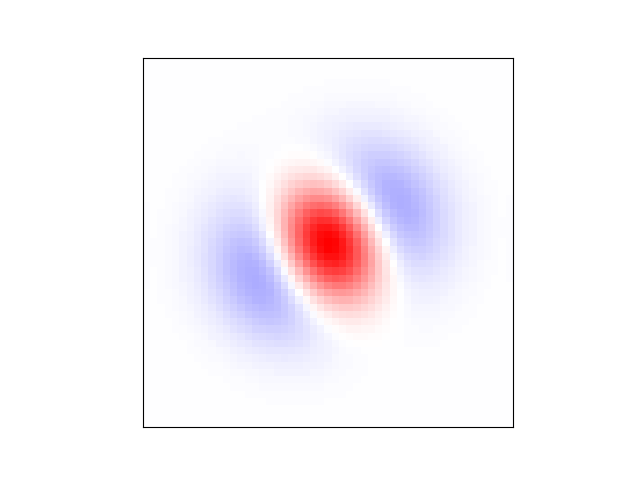} &
      \hspace{-7mm}
      \includegraphics[width=0.1\textwidth]{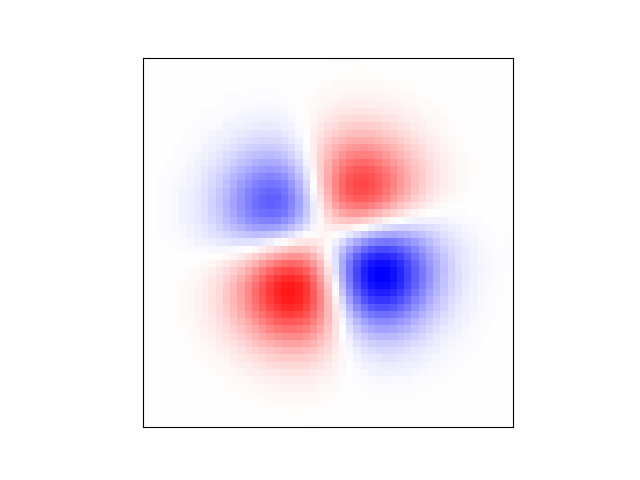} &
      \hspace{-7mm}
      \includegraphics[width=0.1\textwidth]{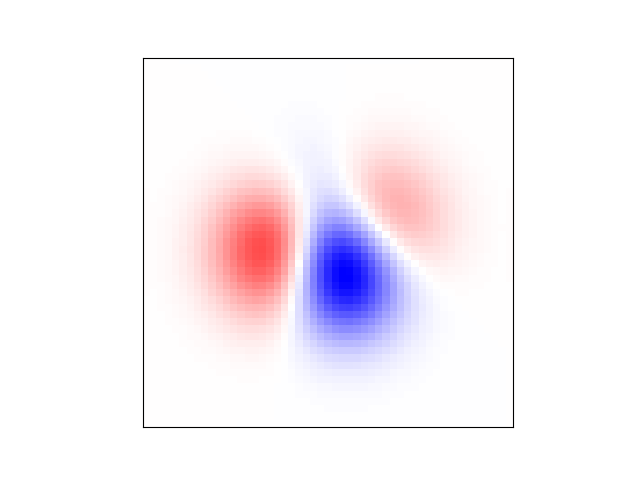} &
      \hspace{-7mm}
      \includegraphics[width=0.1\textwidth]{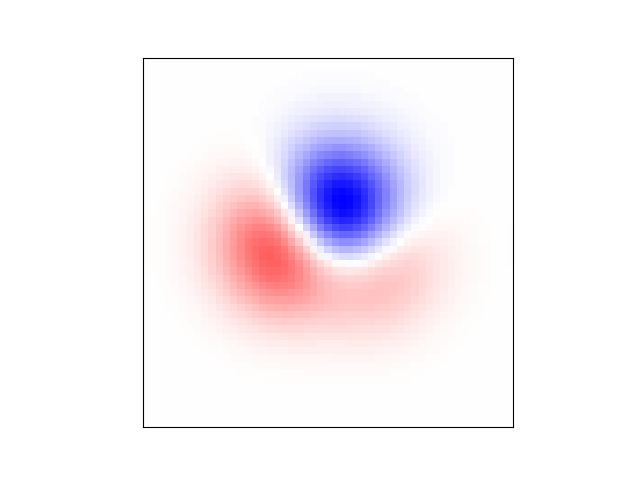}
      \\
    \end{tabular} 

  \vspace{-1mm}
  
    \begin{tabular}{ccccccccccc}
      \hline
      \raisebox{3\height}{Layer 5} &
      \hspace{-5mm}
      \includegraphics[width=0.1\textwidth]{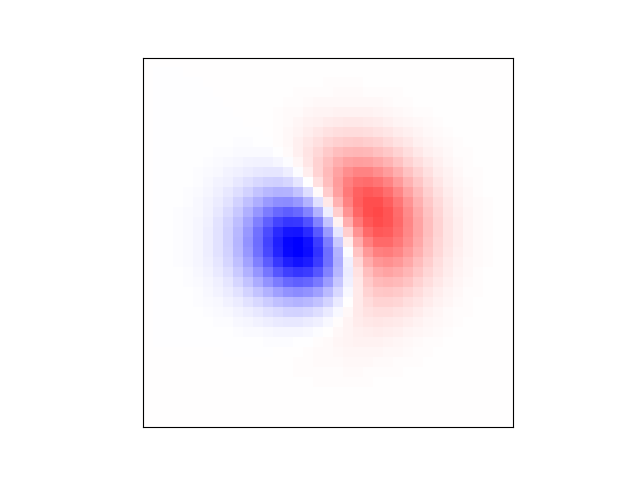} &
      \hspace{-7mm}
      \includegraphics[width=0.1\textwidth]{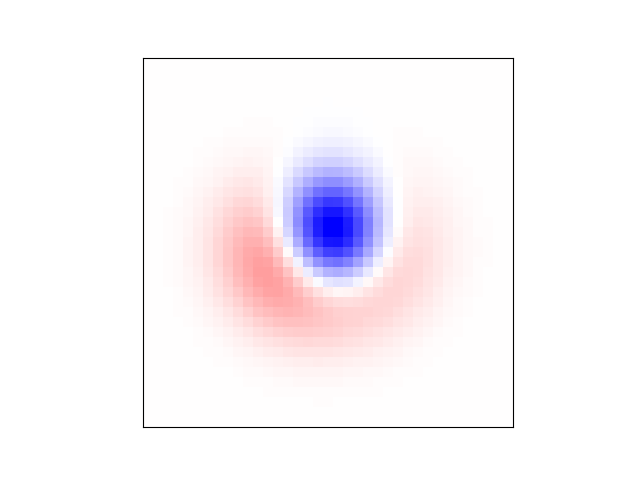} &
      \hspace{-7mm}
      \includegraphics[width=0.1\textwidth]{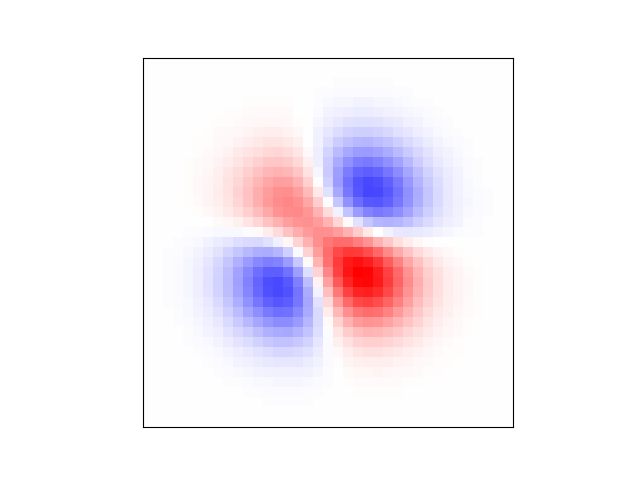} &
      \hspace{-7mm}
      \includegraphics[width=0.1\textwidth]{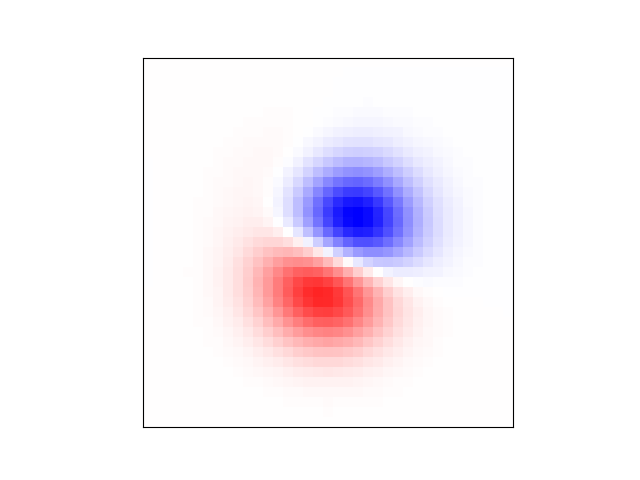} &
      \hspace{-7mm}
      \includegraphics[width=0.1\textwidth]{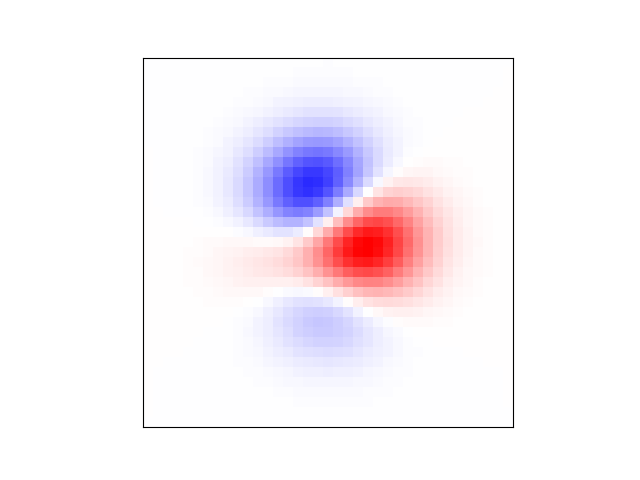} &
      \hspace{-7mm}
      \includegraphics[width=0.1\textwidth]{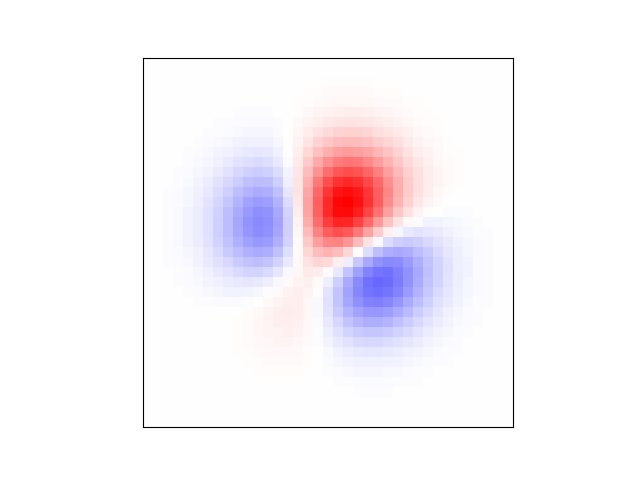} &
      \hspace{-7mm}
      \includegraphics[width=0.1\textwidth]{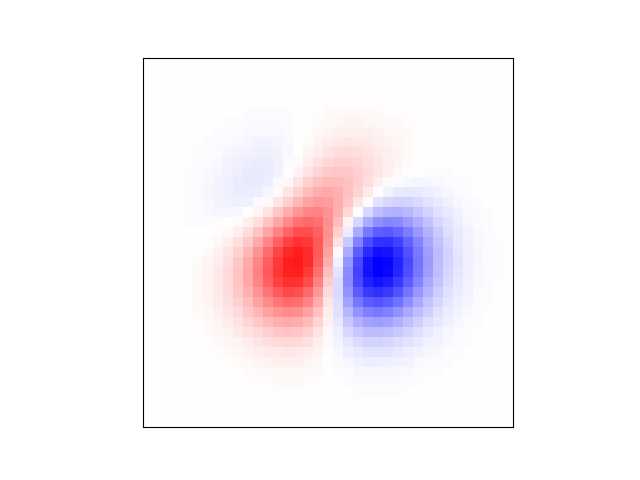} &
      \hspace{-7mm}
      \includegraphics[width=0.1\textwidth]{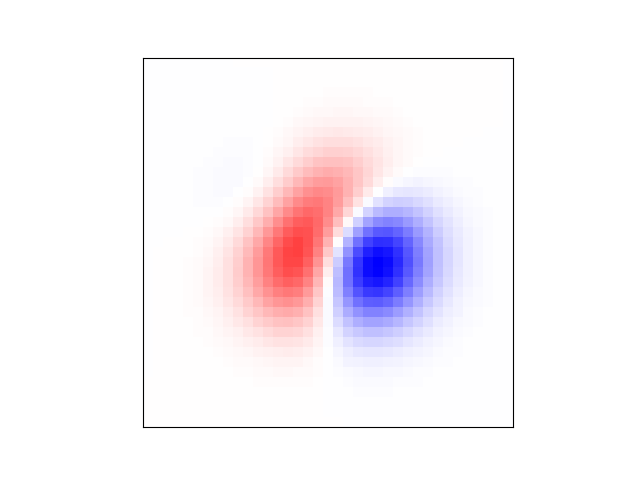} &
      \hspace{-7mm}
      \includegraphics[width=0.1\textwidth]{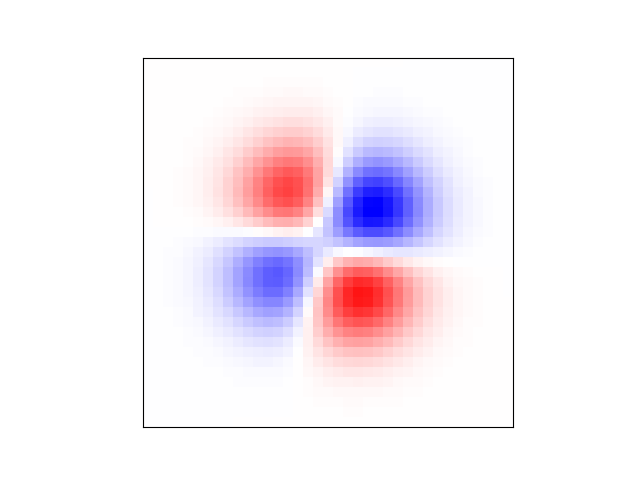} &
      \hspace{-7mm}
      \includegraphics[width=0.1\textwidth]{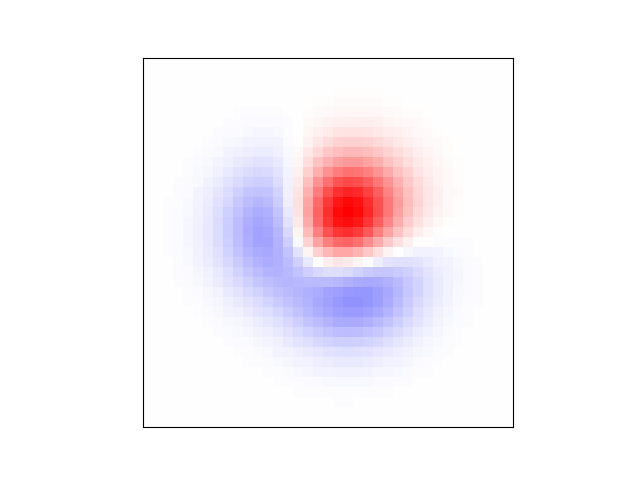}
      \\
    \end{tabular} 

  \vspace{-1mm}
  
    \begin{tabular}{ccccccccccc}
      \hline
      \raisebox{3\height}{Layer 4} &
      \hspace{-5mm}
      \includegraphics[width=0.1\textwidth]{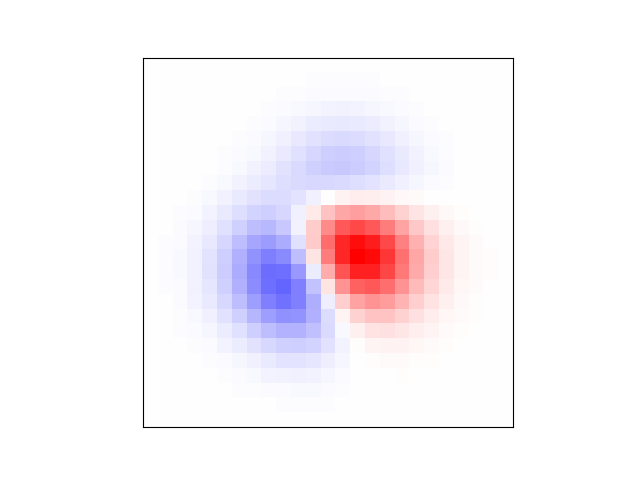} &
      \hspace{-7mm}
      \includegraphics[width=0.1\textwidth]{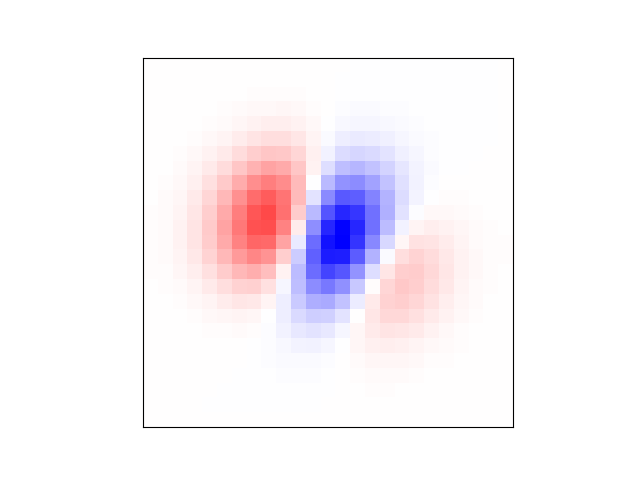} &
      \hspace{-7mm}
      \includegraphics[width=0.1\textwidth]{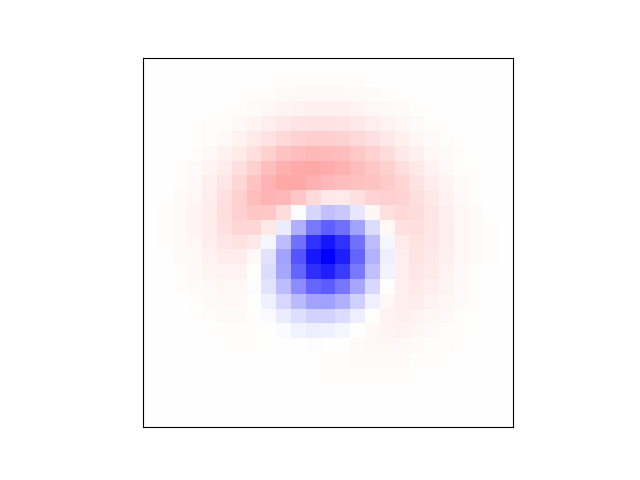} &
      \hspace{-7mm}
      \includegraphics[width=0.1\textwidth]{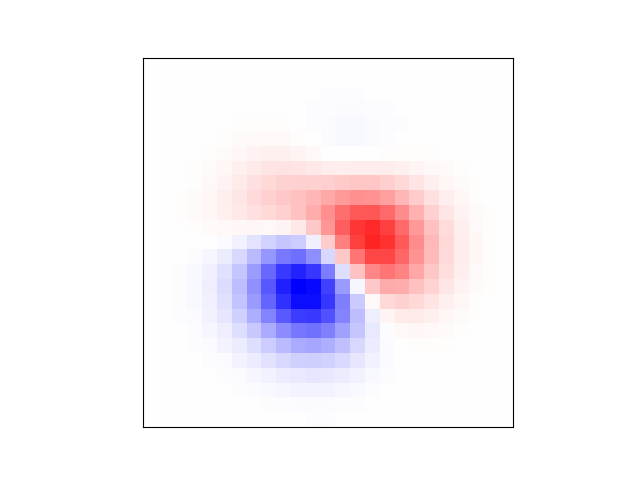} &
      \hspace{-7mm}
      \includegraphics[width=0.1\textwidth]{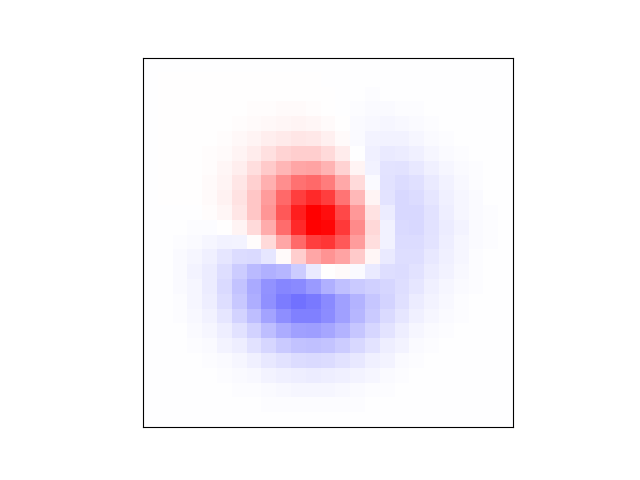} &
      \hspace{-7mm}
      \includegraphics[width=0.1\textwidth]{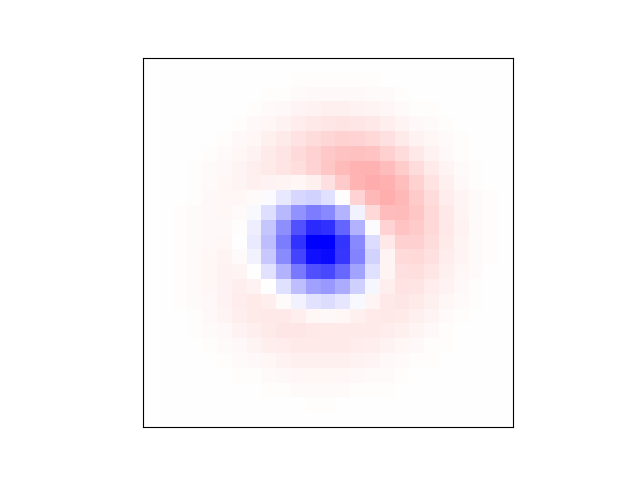} &
      \hspace{-7mm}
      \includegraphics[width=0.1\textwidth]{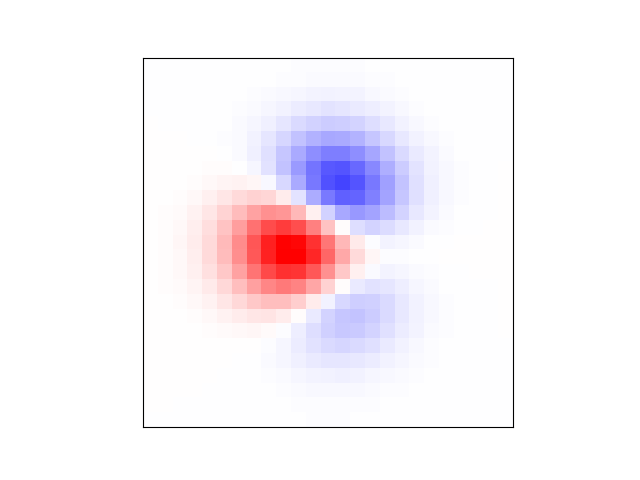} &
      \hspace{-7mm}
      \includegraphics[width=0.1\textwidth]{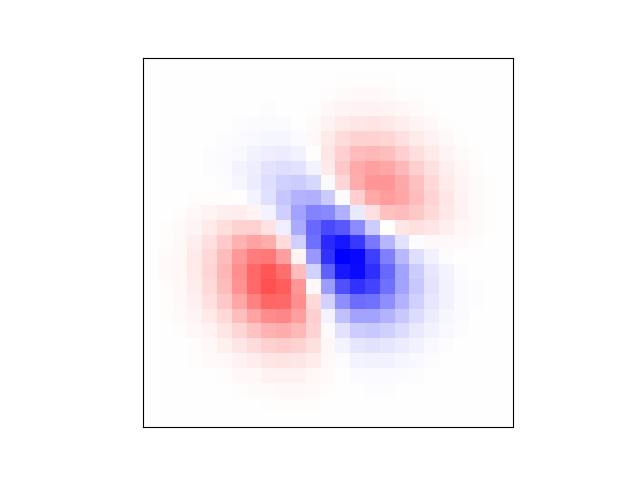} &
      \hspace{-7mm}
      \includegraphics[width=0.1\textwidth]{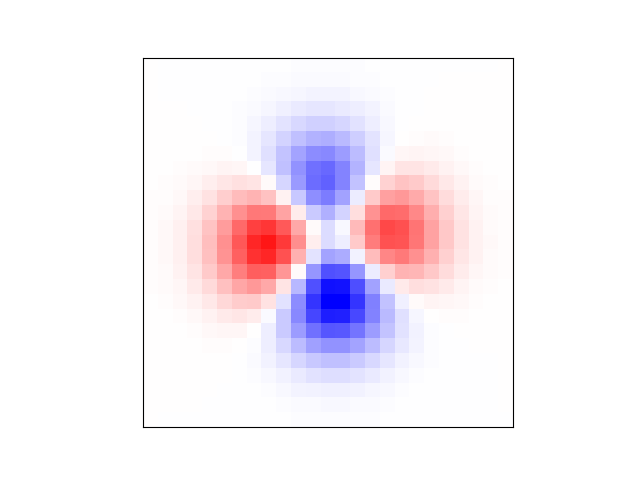} &
      \hspace{-7mm}
      \includegraphics[width=0.1\textwidth]{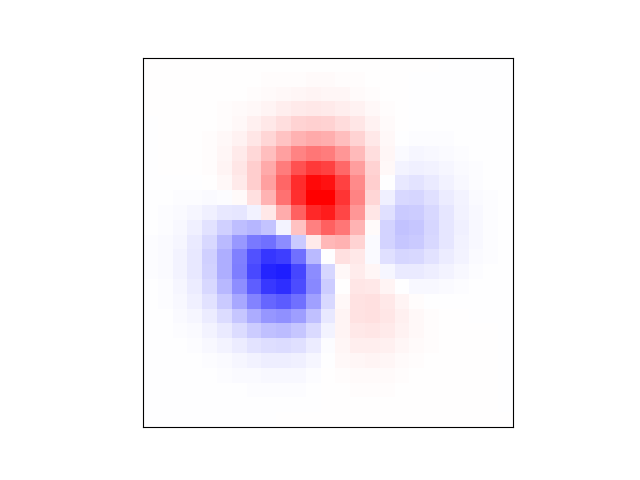}
      \\
    \end{tabular} 

  \vspace{-1mm}
  
    \begin{tabular}{ccccccccccc}
      \hline
      \raisebox{3\height}{Layer 3} &
      \hspace{-5mm}
      \includegraphics[width=0.1\textwidth]{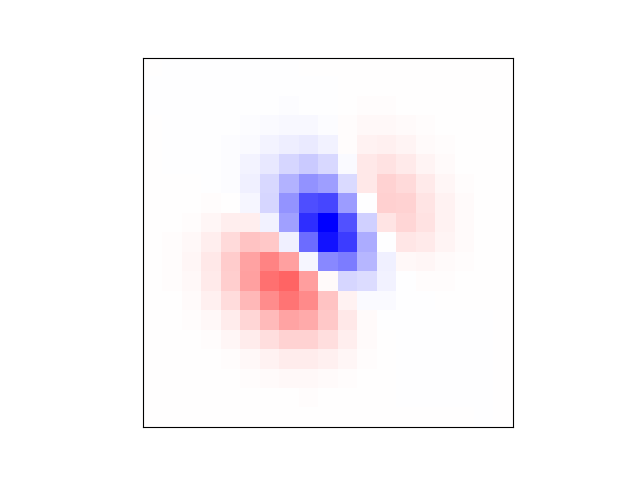} &
      \hspace{-7mm}
      \includegraphics[width=0.1\textwidth]{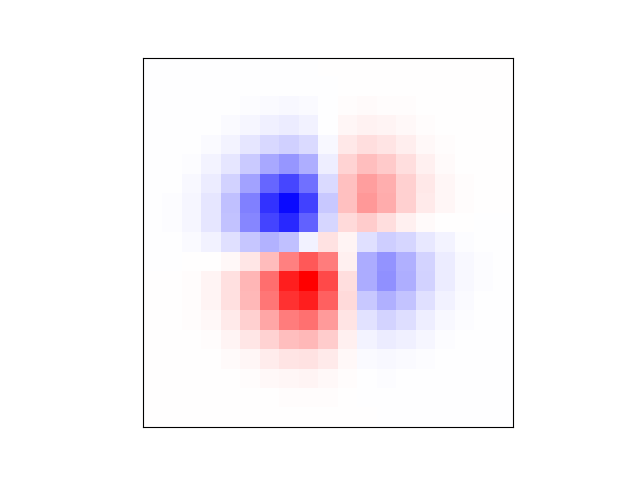} &
      \hspace{-7mm}
      \includegraphics[width=0.1\textwidth]{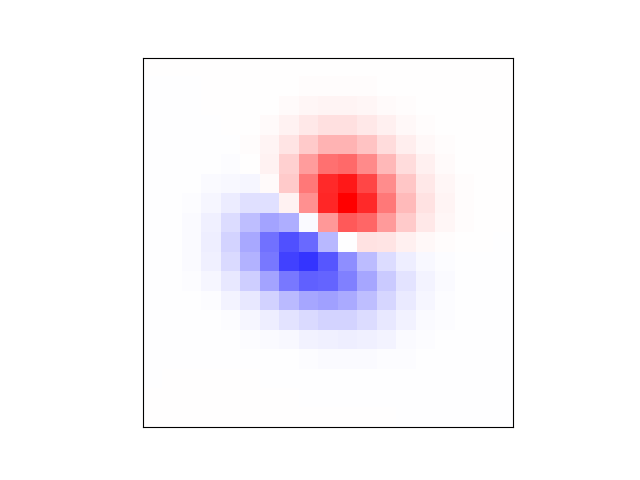} &
      \hspace{-7mm}
      \includegraphics[width=0.1\textwidth]{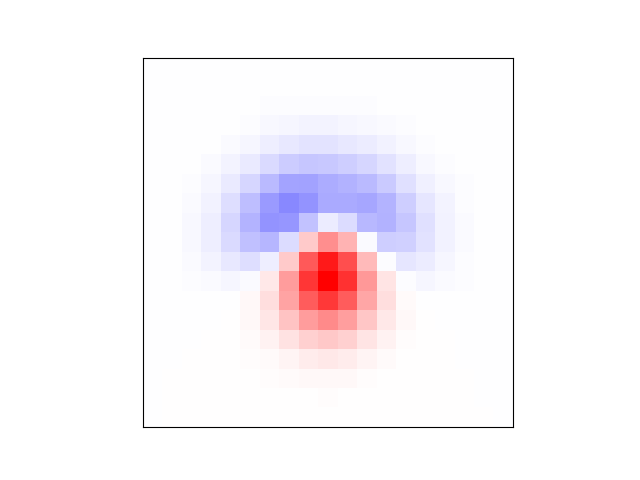} &
      \hspace{-7mm}
      \includegraphics[width=0.1\textwidth]{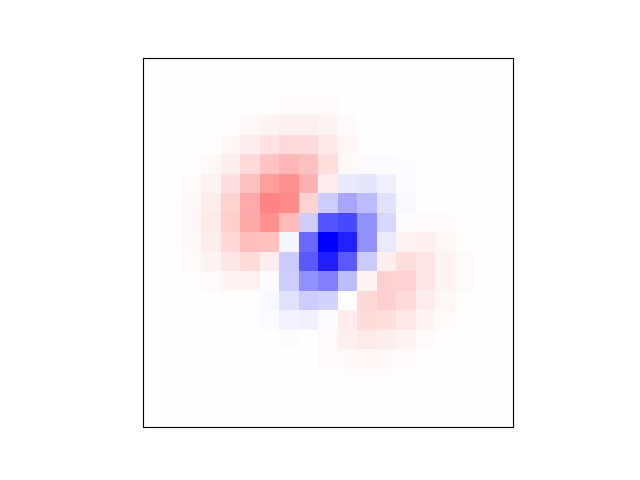} &
      \hspace{-7mm}
      \includegraphics[width=0.1\textwidth]{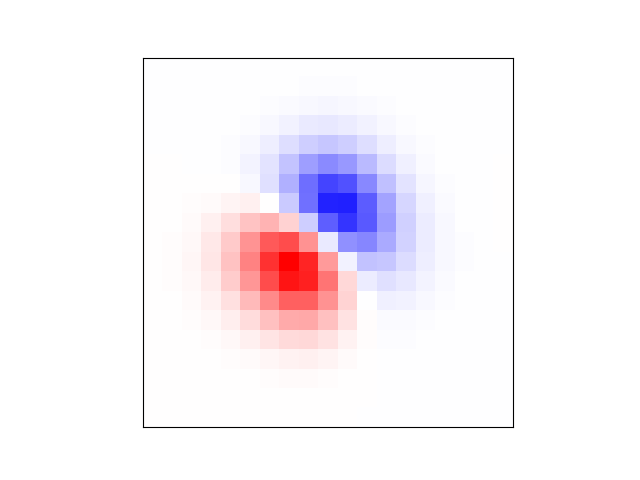} &
      \hspace{-7mm}
      \includegraphics[width=0.1\textwidth]{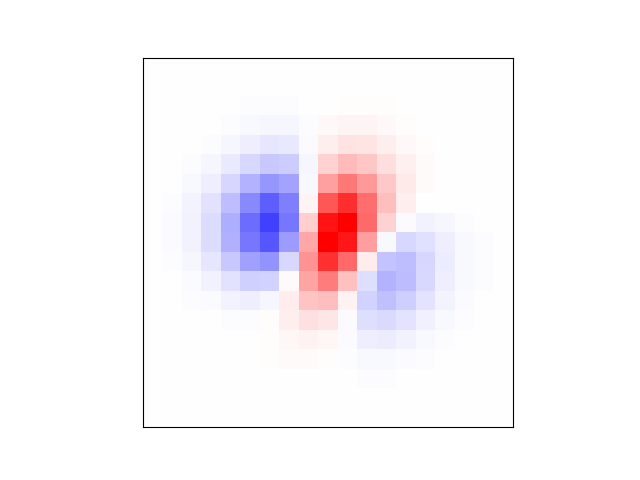} &
      \hspace{-7mm}
      \includegraphics[width=0.1\textwidth]{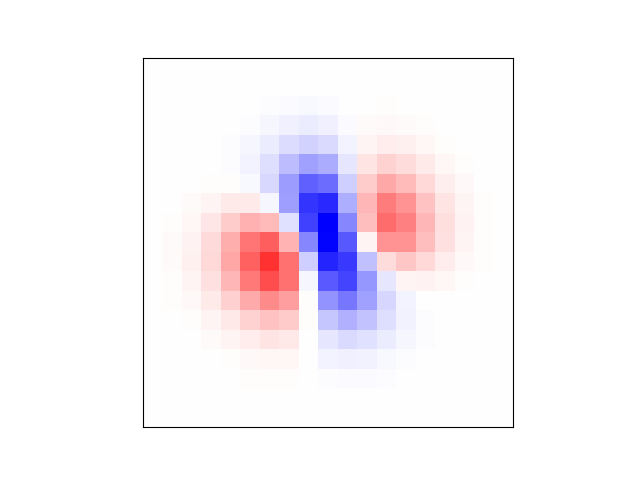} &
      \hspace{-7mm}
      \includegraphics[width=0.1\textwidth]{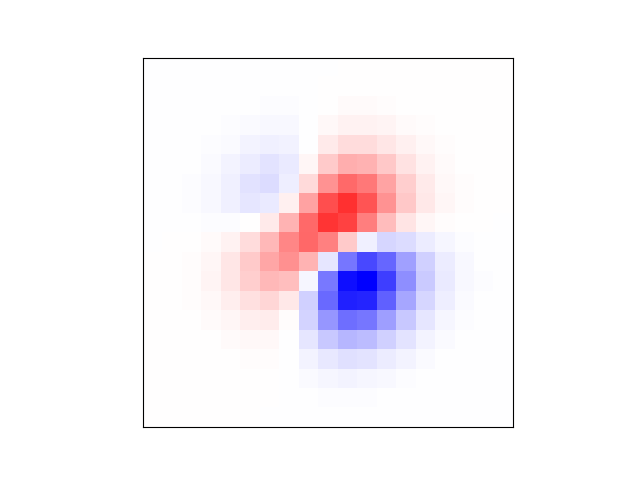} &
      \hspace{-7mm}
      \includegraphics[width=0.1\textwidth]{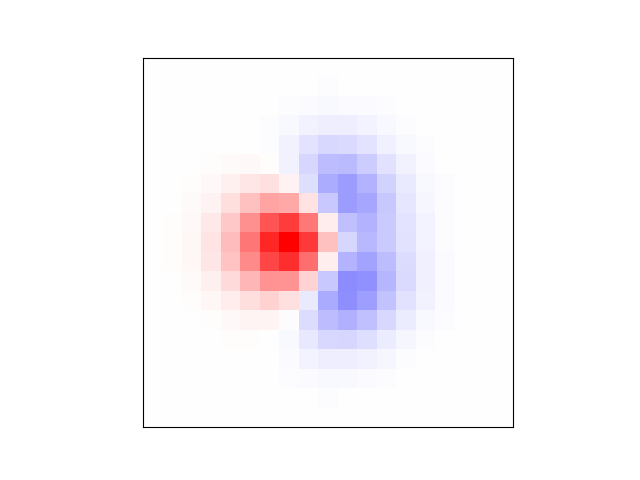}
      \\
    \end{tabular} 

  \vspace{-1mm}
  
    \begin{tabular}{ccccccccccc}
      \hline
      \raisebox{3\height}{Layer 2} &
      \hspace{-5mm}
      \includegraphics[width=0.1\textwidth]{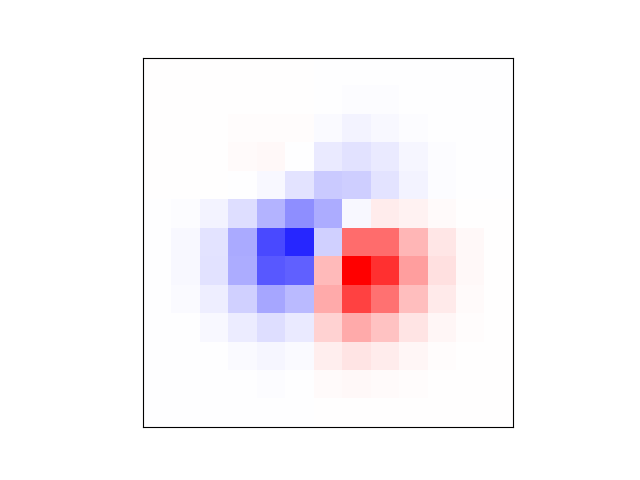} &
      \hspace{-7mm}
      \includegraphics[width=0.1\textwidth]{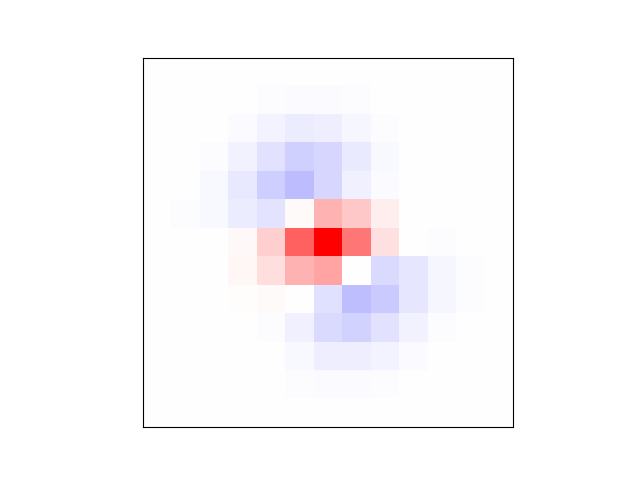} &
      \hspace{-7mm}
      \includegraphics[width=0.1\textwidth]{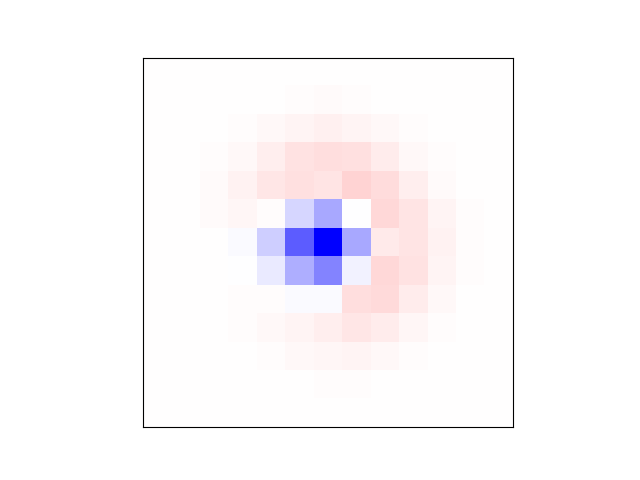} &
      \hspace{-7mm}
      \includegraphics[width=0.1\textwidth]{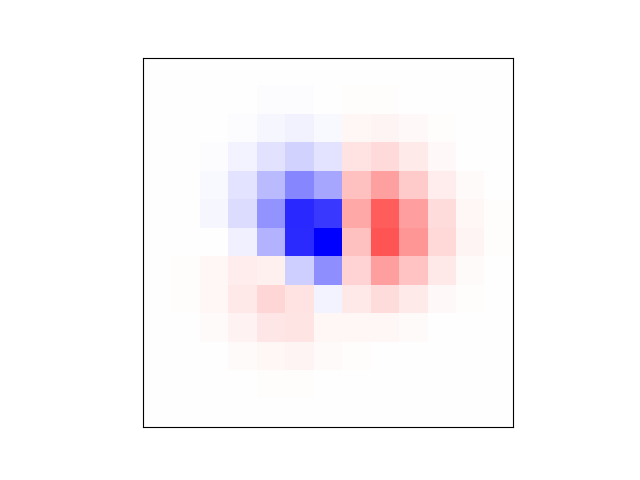} &
      \hspace{-7mm}
      \includegraphics[width=0.1\textwidth]{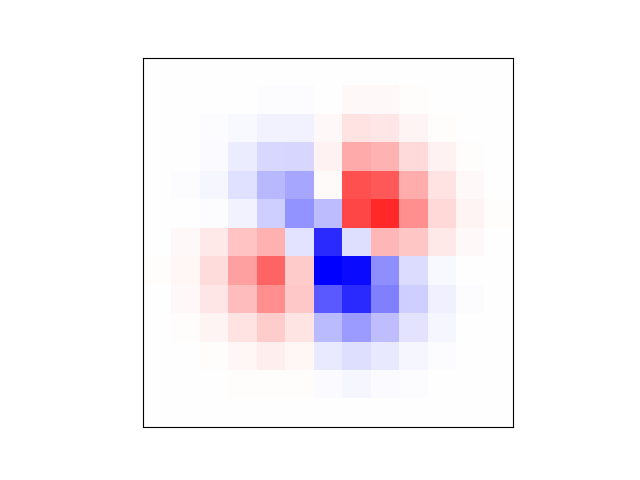} &
      \hspace{-7mm}
      \includegraphics[width=0.1\textwidth]{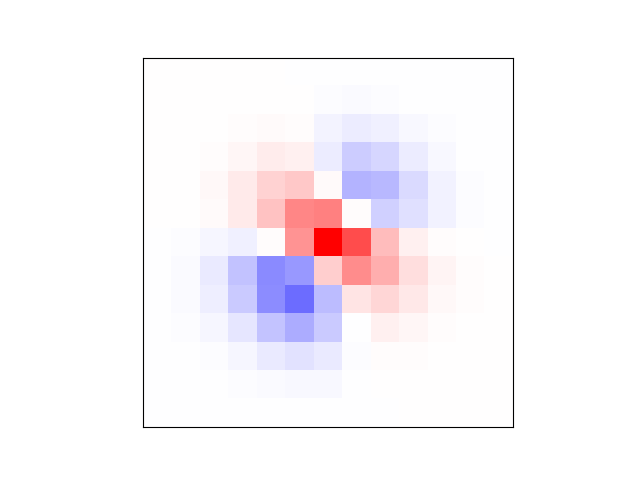} &
      \hspace{-7mm}
      \includegraphics[width=0.1\textwidth]{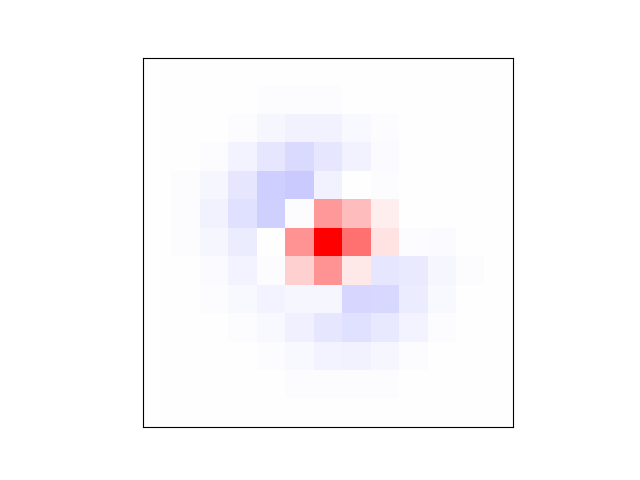} &
      \hspace{-7mm}
      \includegraphics[width=0.1\textwidth]{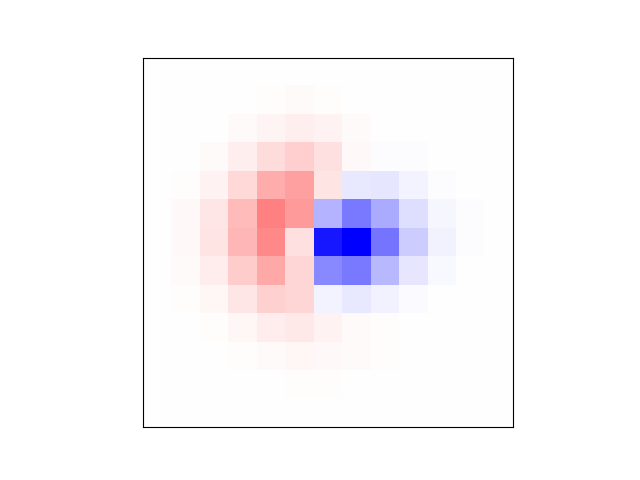} &
      \hspace{-7mm}
      \includegraphics[width=0.1\textwidth]{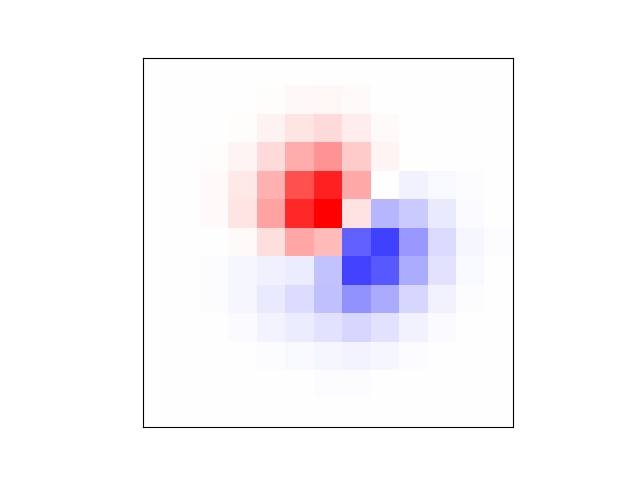} &
      \hspace{-7mm}
      \includegraphics[width=0.1\textwidth]{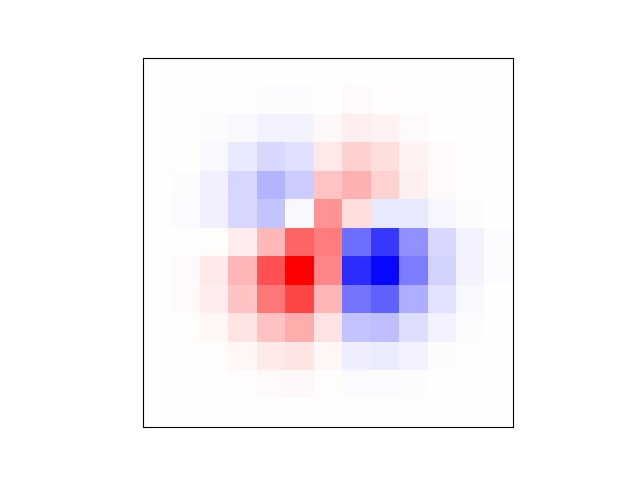}
      \\
    \end{tabular} 

  \vspace{-1mm}
  
    \begin{tabular}{ccccccccccc}
      \hline
      \raisebox{3\height}{Layer 1} &
      \hspace{-5mm}
      \includegraphics[width=0.1\textwidth]{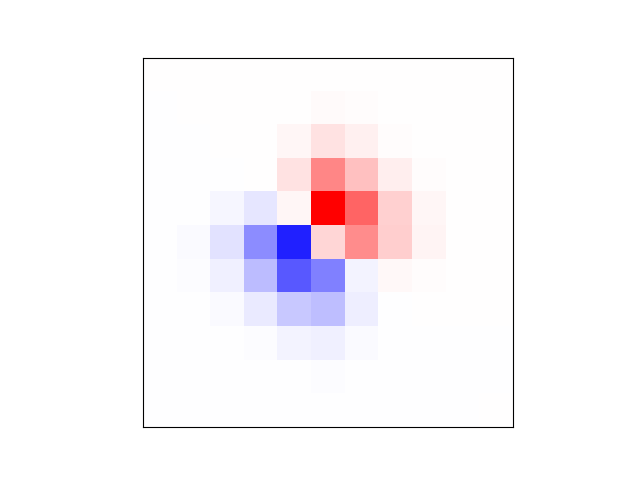} &
      \hspace{-7mm}
      \includegraphics[width=0.1\textwidth]{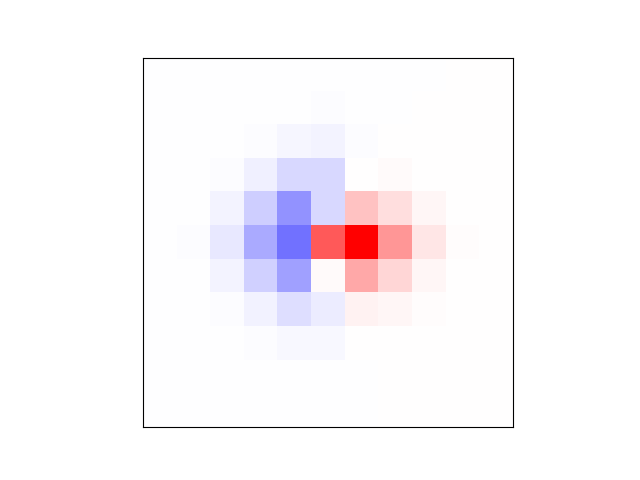} &
      \hspace{-7mm}
      \includegraphics[width=0.1\textwidth]{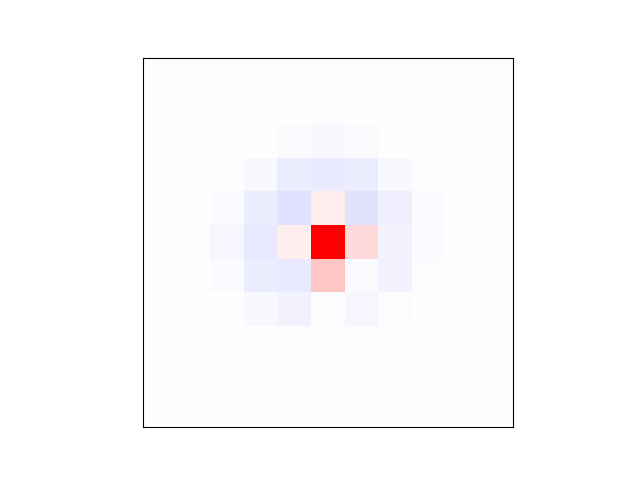} &
      \hspace{-7mm}
      \includegraphics[width=0.1\textwidth]{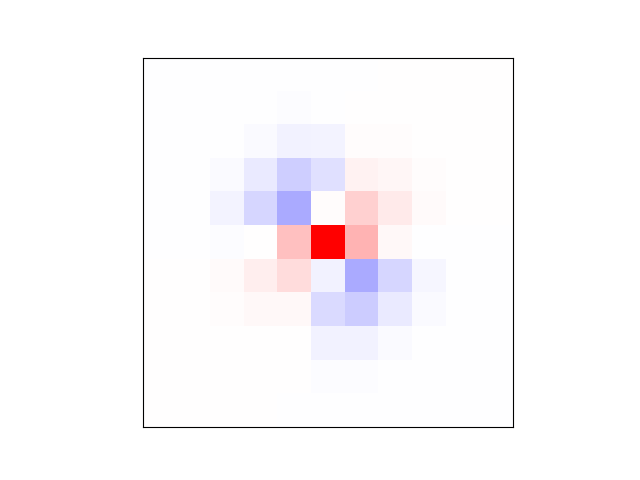} &
      \hspace{-7mm}
      \includegraphics[width=0.1\textwidth]{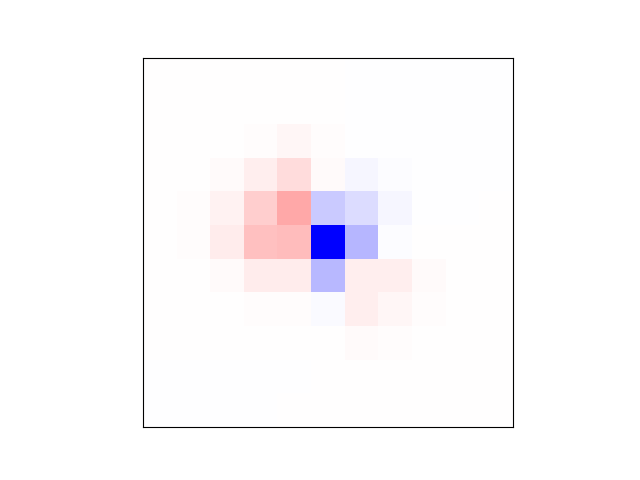} &
      \hspace{-7mm}
      \includegraphics[width=0.1\textwidth]{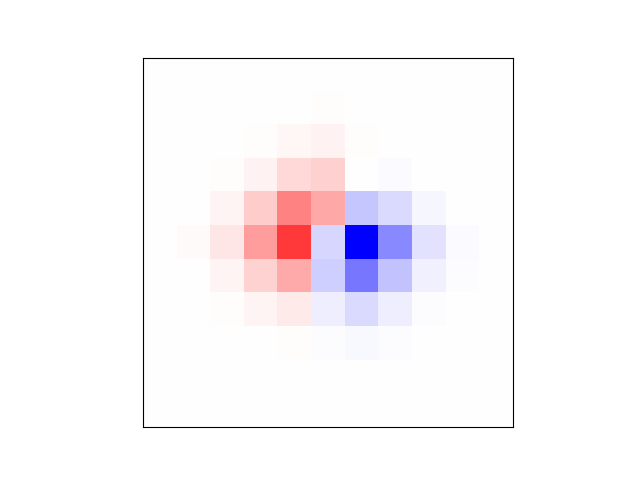} &
      \hspace{-7mm}
      \includegraphics[width=0.1\textwidth]{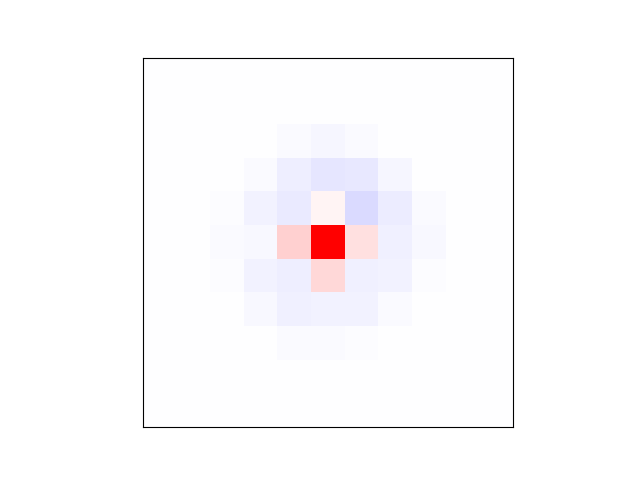} &
      \hspace{-7mm}
      \includegraphics[width=0.1\textwidth]{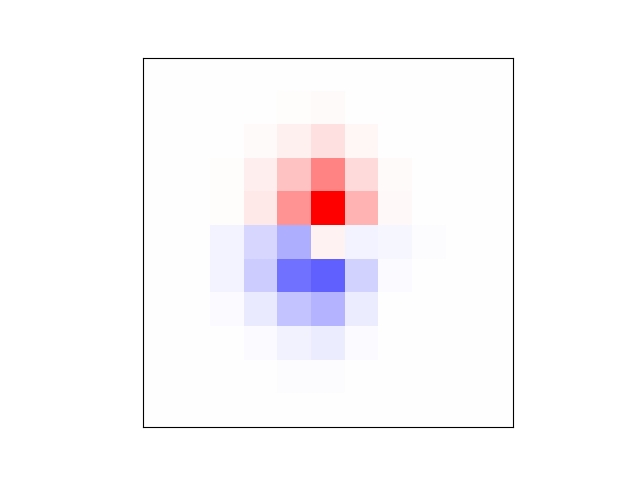} &
      \hspace{-7mm}
      \includegraphics[width=0.1\textwidth]{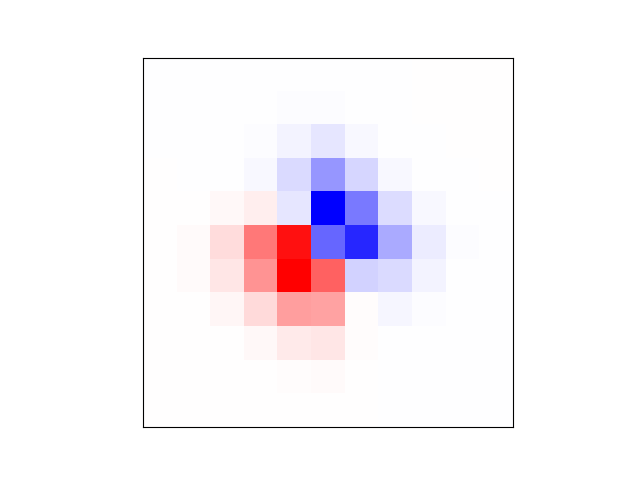} &
      \hspace{-7mm}
      \includegraphics[width=0.1\textwidth]{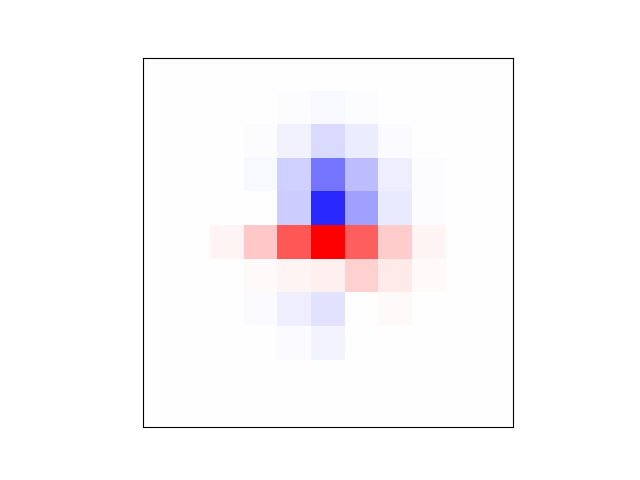}
    \end{tabular} 
  \end{center}
  \captionsetup{skip=-5pt}
  \caption{A selection of 10 learned effective filters from each of the layers of a single-scale-channel GaussDerNet, based on the 2-jet, and trained on the rescaled CIFAR-10 dataset (with image size $64 \times 64$ pixels). The color mapping has been set to map zero values to white, with positive values being mapped to red and negative values being mapped to blue. In each later layer, the scale parameter $\sigma$ becomes larger according to Equation~(\ref{eq:sigma-geometric}), thus increasing the size of the convolution kernels. Among the various learned linear combinations of Gaussian derivatives, we can note that some of the learned filters resemble directional derivatives or the Laplacian of Gaussian.}
  \label{fig-learned-filters_2_jet}
\end{figure*}
\begin{figure*}[htbp]
  \begin{center}
    \begin{tabular}{ccccccccccc}
      \raisebox{3\height}{Layer 6} &
      \hspace{-5mm}
      \includegraphics[width=0.1\textwidth]{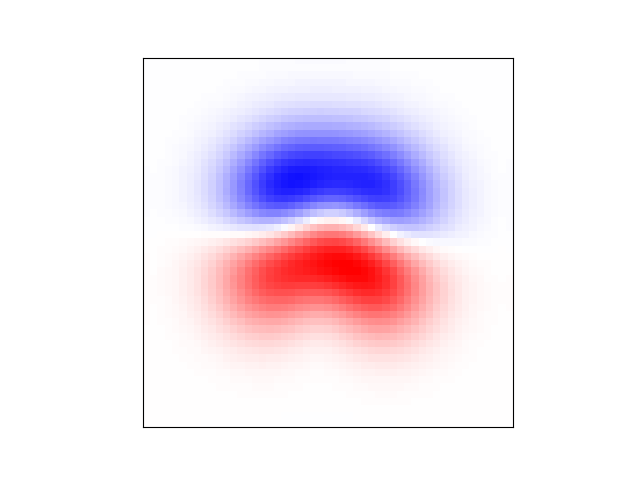} &
      \hspace{-7mm}
      \includegraphics[width=0.1\textwidth]{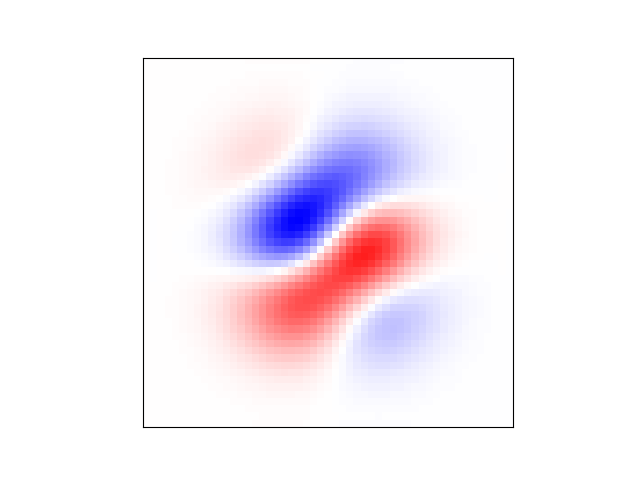} &
      \hspace{-7mm}
      \includegraphics[width=0.1\textwidth]{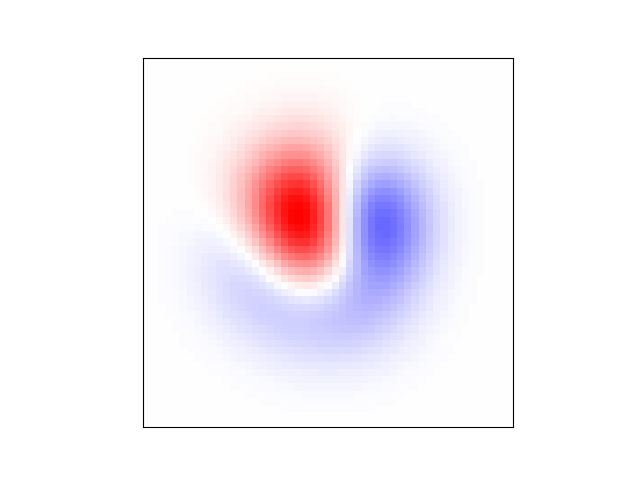} &
      \hspace{-7mm}
      \includegraphics[width=0.1\textwidth]{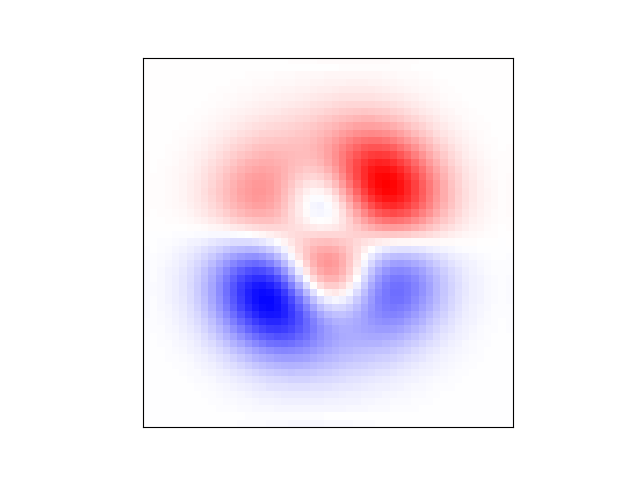} &
      \hspace{-7mm}
      \includegraphics[width=0.1\textwidth]{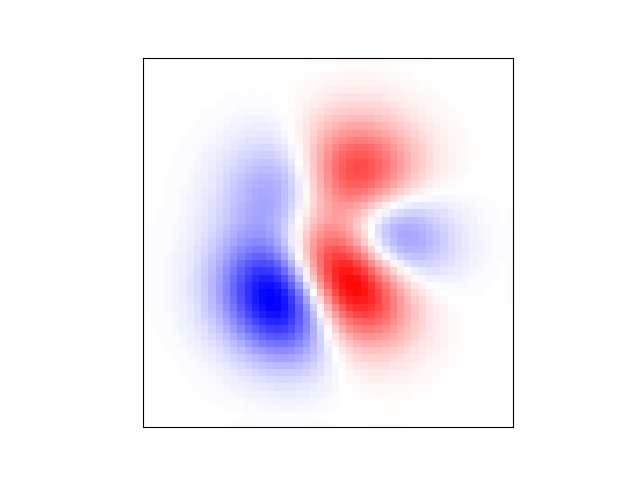} &
      \hspace{-7mm}
      \includegraphics[width=0.1\textwidth]{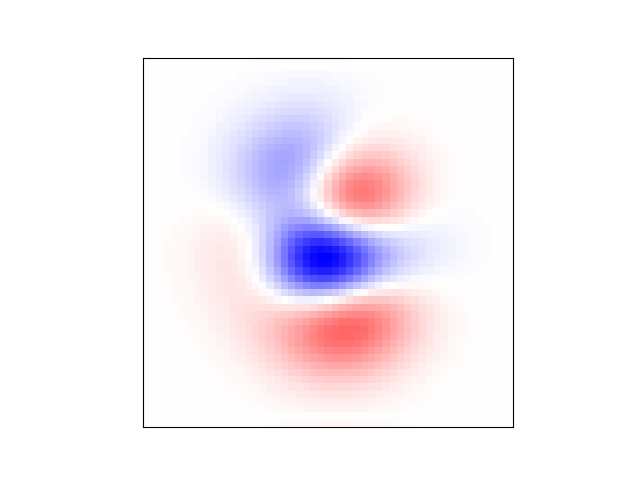} &
      \hspace{-7mm}
      \includegraphics[width=0.1\textwidth]{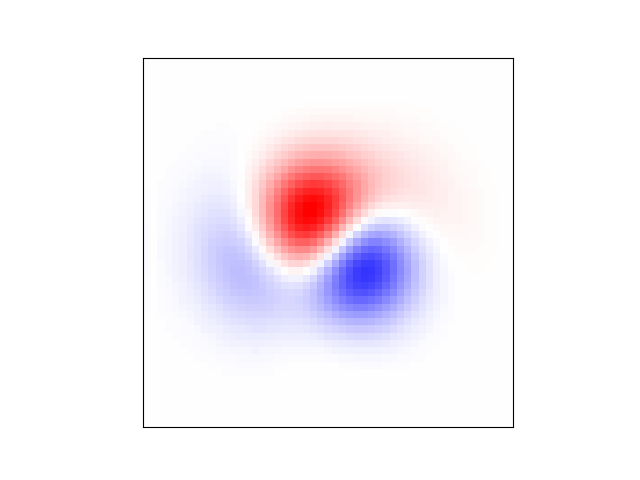} &
      \hspace{-7mm}
      \includegraphics[width=0.1\textwidth]{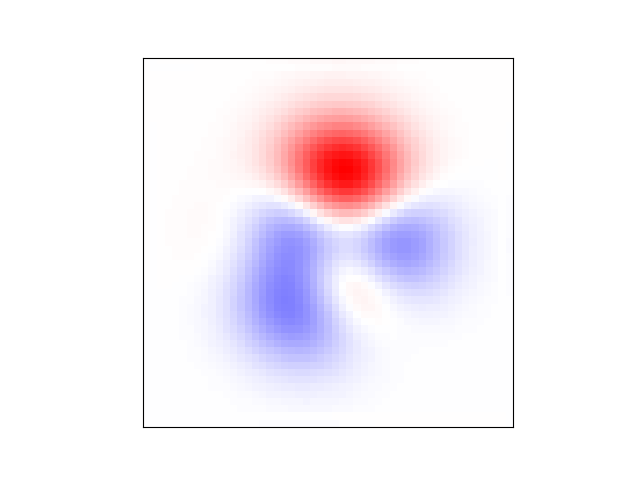} &
      \hspace{-7mm}
      \includegraphics[width=0.1\textwidth]{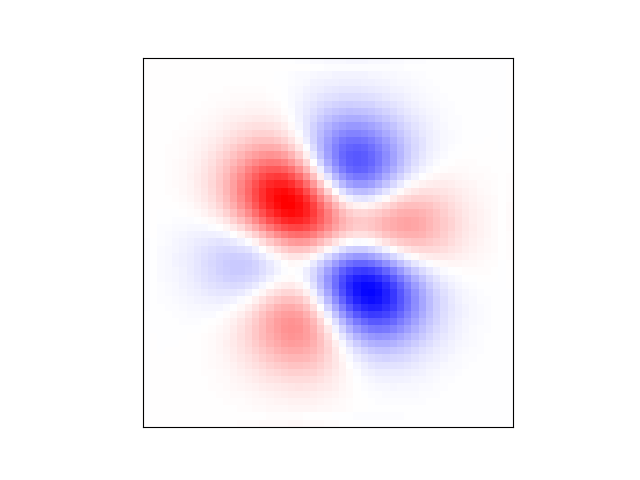} &
      \hspace{-7mm}
      \includegraphics[width=0.1\textwidth]{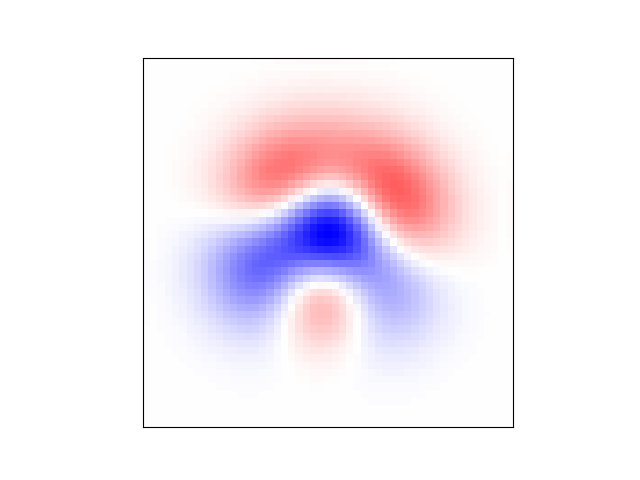}
      \\
    \end{tabular} 

  \vspace{-1mm}
  
    \begin{tabular}{ccccccccccc}
      \hline
      \raisebox{3\height}{Layer 5} &
      \hspace{-5mm}
      \includegraphics[width=0.1\textwidth]{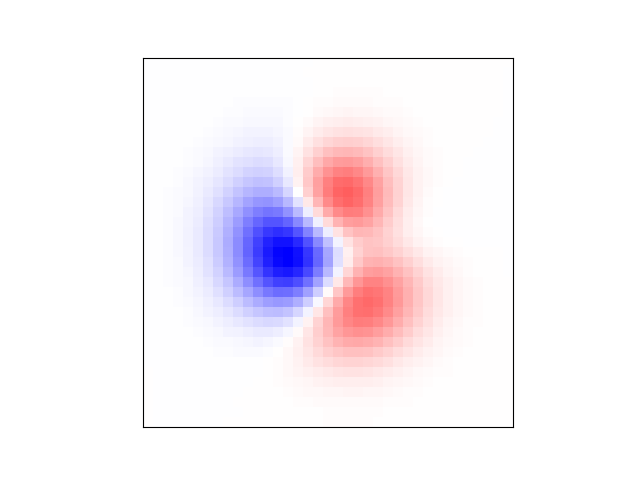} &
      \hspace{-7mm}
      \includegraphics[width=0.1\textwidth]{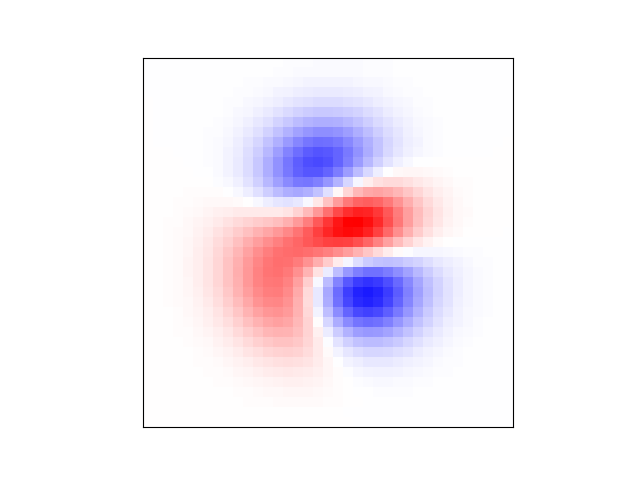} &
      \hspace{-7mm}
      \includegraphics[width=0.1\textwidth]{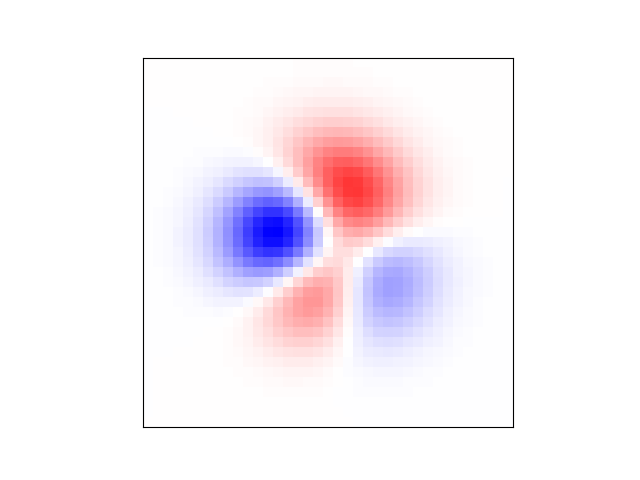} &
      \hspace{-7mm}
      \includegraphics[width=0.1\textwidth]{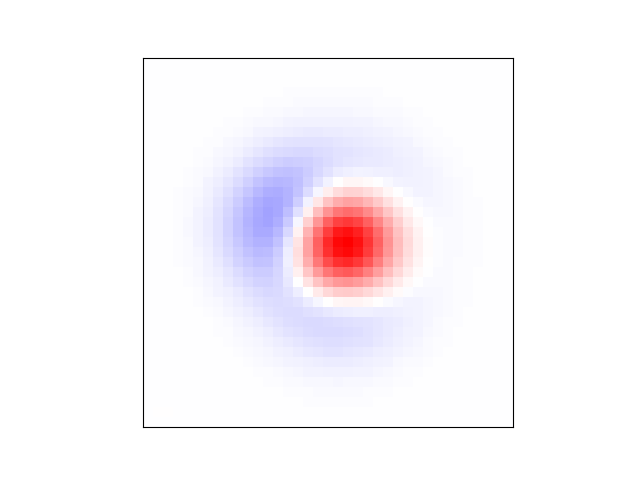} &
      \hspace{-7mm}
      \includegraphics[width=0.1\textwidth]{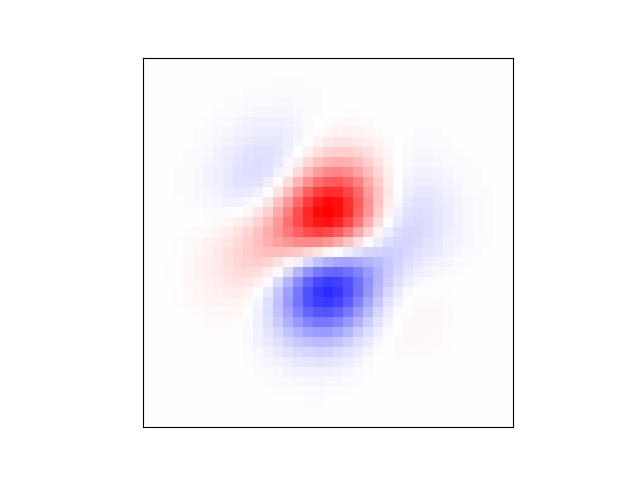} &
      \hspace{-7mm}
      \includegraphics[width=0.1\textwidth]{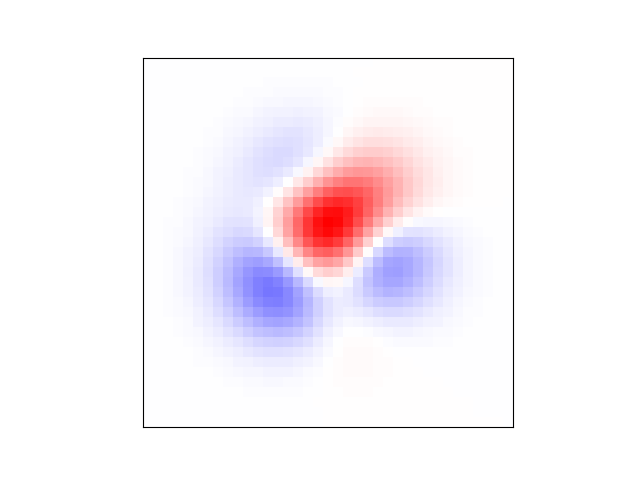} &
      \hspace{-7mm}
      \includegraphics[width=0.1\textwidth]{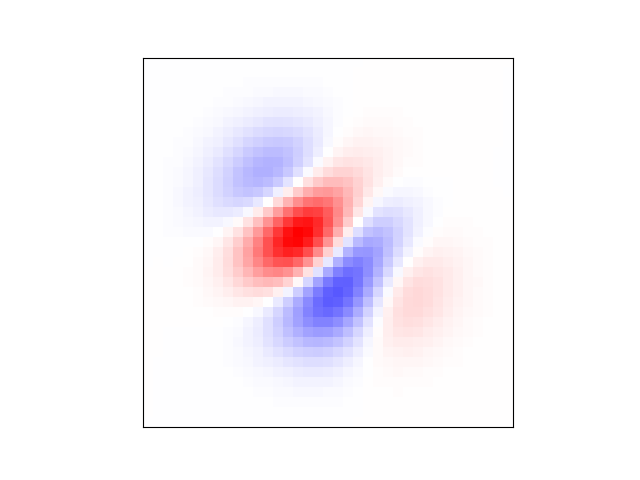} &
      \hspace{-7mm}
      \includegraphics[width=0.1\textwidth]{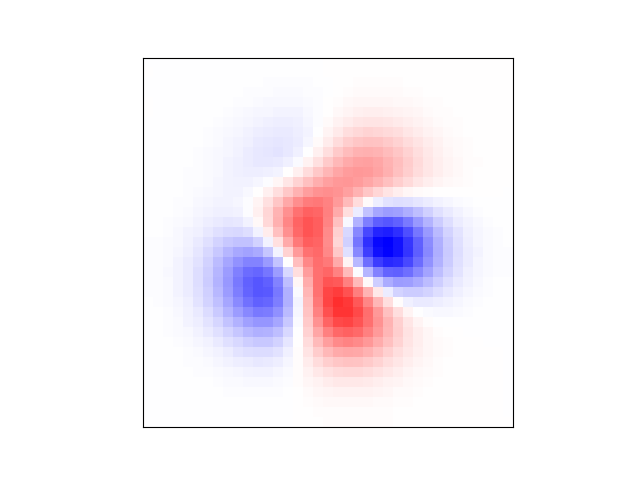} &
      \hspace{-7mm}
      \includegraphics[width=0.1\textwidth]{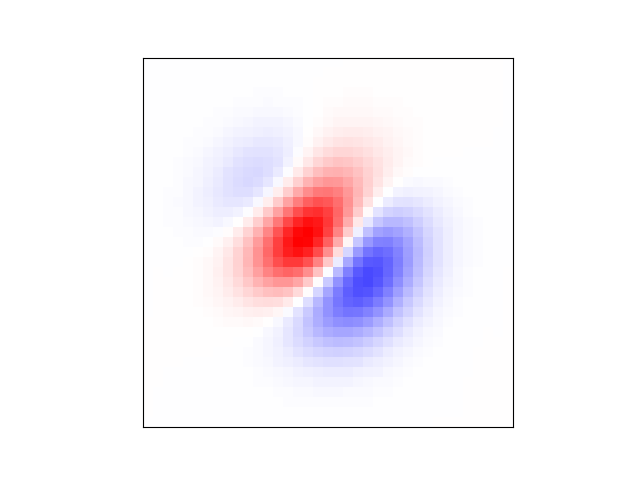} &
      \hspace{-7mm}
      \includegraphics[width=0.1\textwidth]{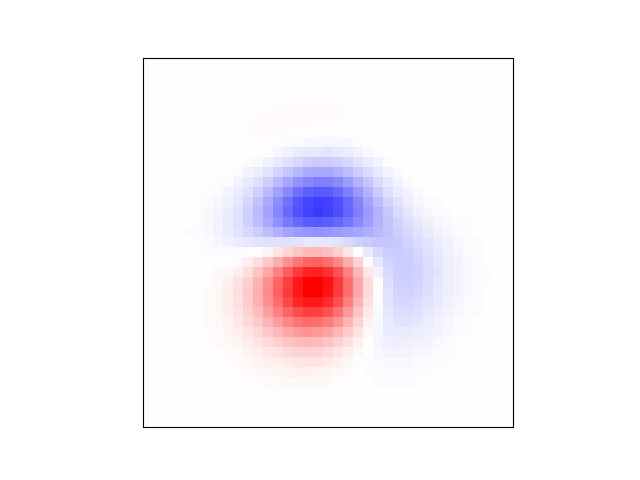}
      \\
    \end{tabular} 

  \vspace{-1mm}
  
    \begin{tabular}{ccccccccccc}
      \hline
      \raisebox{3\height}{Layer 4} &
      \hspace{-5mm}
      \includegraphics[width=0.1\textwidth]{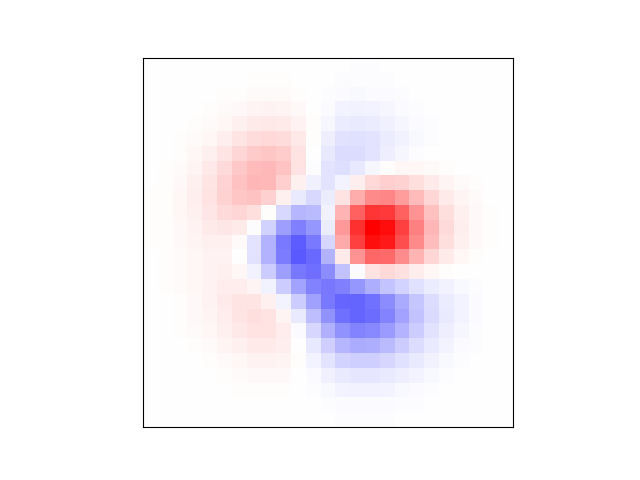} &
      \hspace{-7mm}
      \includegraphics[width=0.1\textwidth]{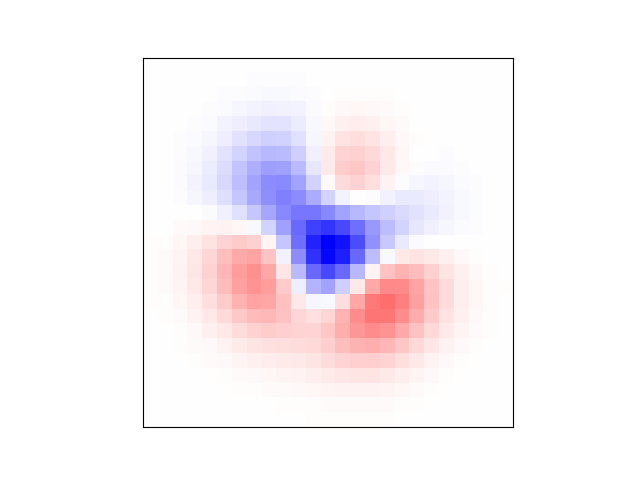} &
      \hspace{-7mm}
      \includegraphics[width=0.1\textwidth]{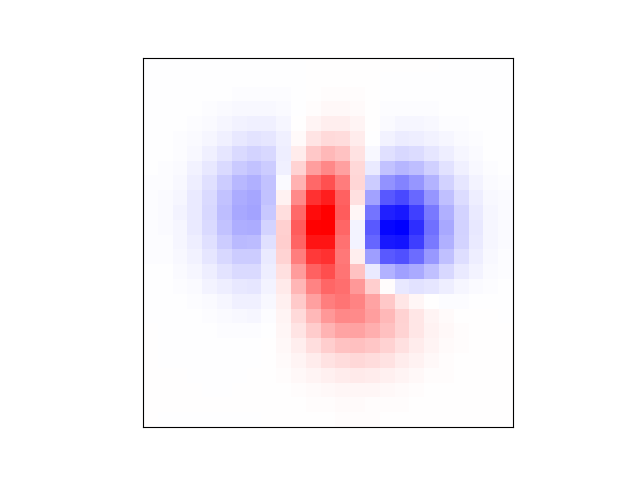} &
      \hspace{-7mm}
      \includegraphics[width=0.1\textwidth]{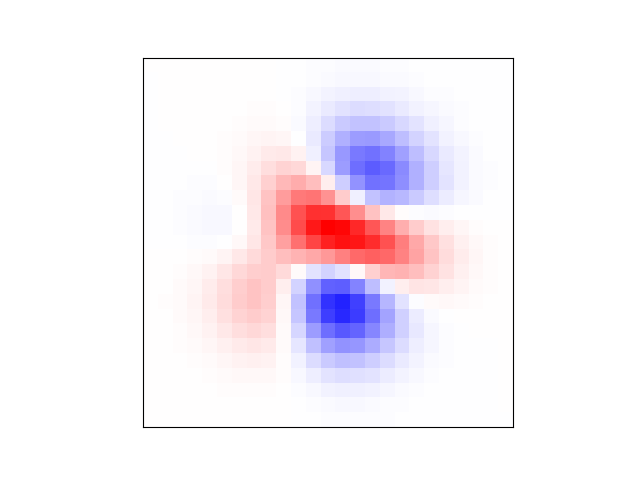} &
      \hspace{-7mm}
      \includegraphics[width=0.1\textwidth]{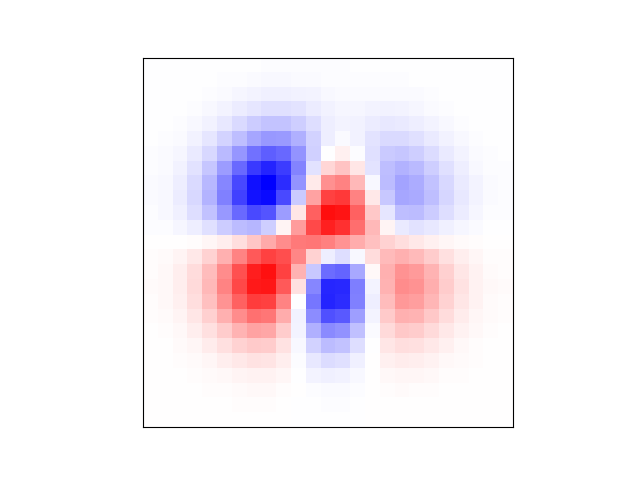} &
      \hspace{-7mm}
      \includegraphics[width=0.1\textwidth]{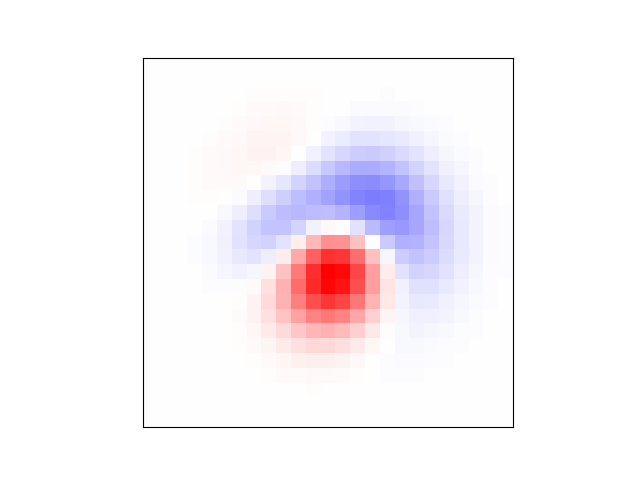} &
      \hspace{-7mm}
      \includegraphics[width=0.1\textwidth]{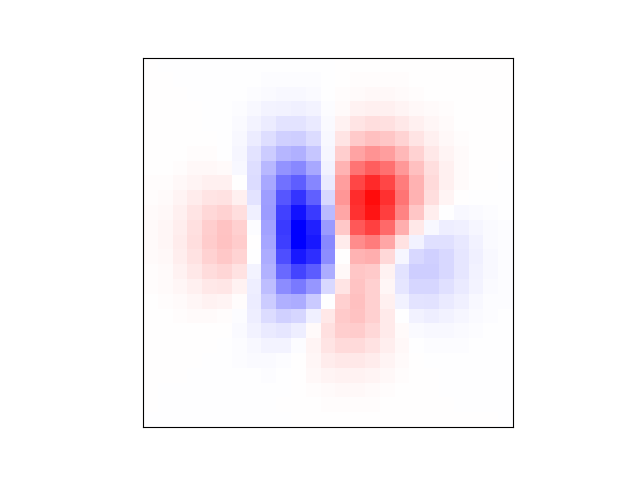} &
      \hspace{-7mm}
      \includegraphics[width=0.1\textwidth]{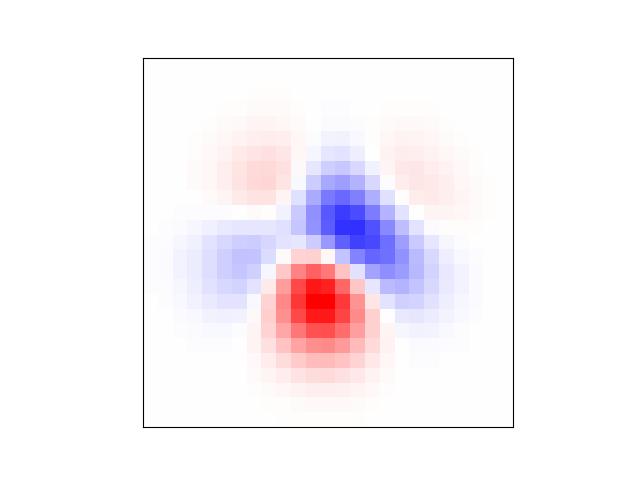} &
      \hspace{-7mm}
      \includegraphics[width=0.1\textwidth]{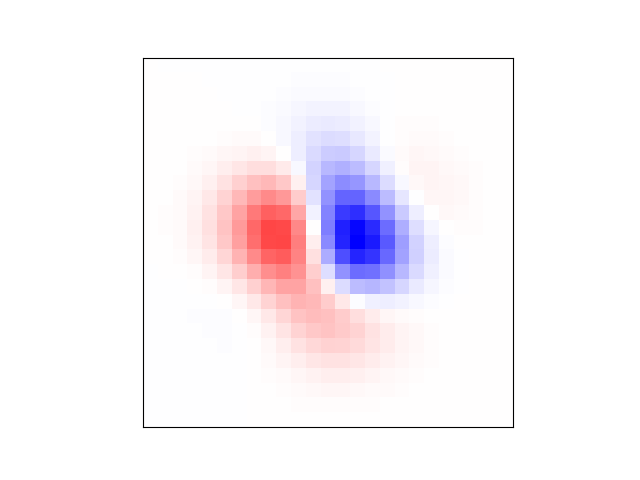} &
      \hspace{-7mm}
      \includegraphics[width=0.1\textwidth]{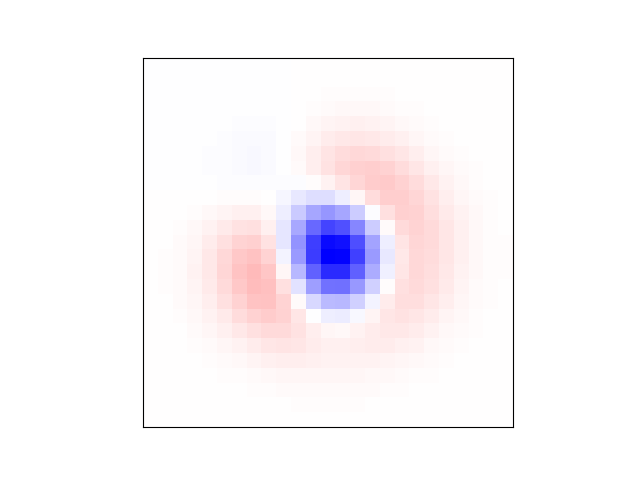}
      \\
    \end{tabular} 

  \vspace{-1mm}
  
    \begin{tabular}{ccccccccccc}
      \hline
      \raisebox{3\height}{Layer 3} &
      \hspace{-5mm}
      \includegraphics[width=0.1\textwidth]{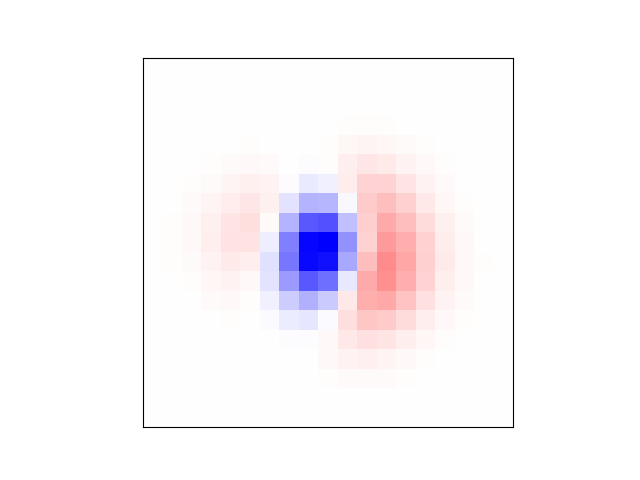} &
      \hspace{-7mm}
      \includegraphics[width=0.1\textwidth]{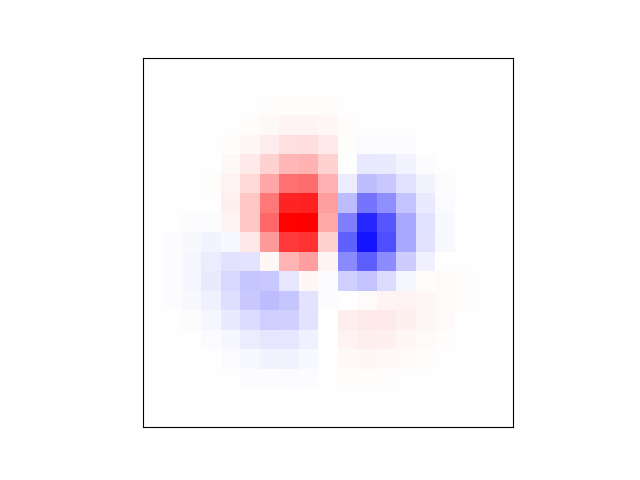} &
      \hspace{-7mm}
      \includegraphics[width=0.1\textwidth]{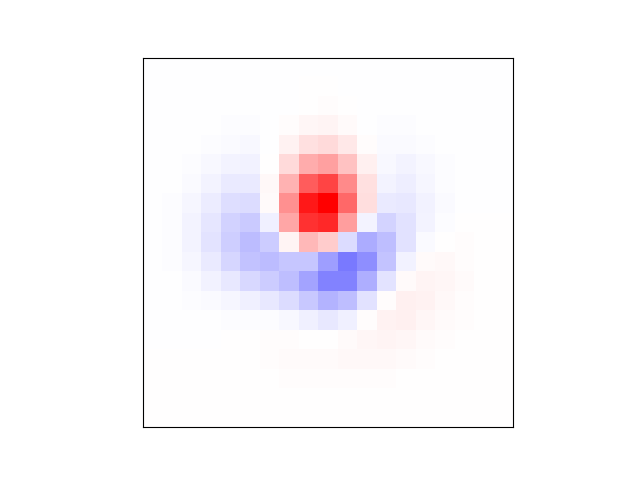} &
      \hspace{-7mm}
      \includegraphics[width=0.1\textwidth]{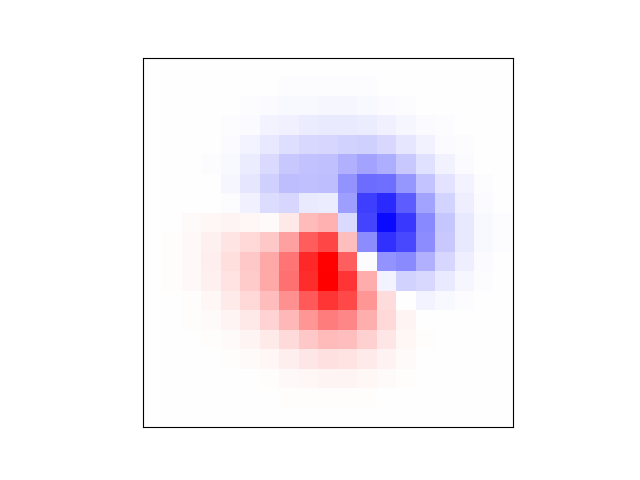} &
      \hspace{-7mm}
      \includegraphics[width=0.1\textwidth]{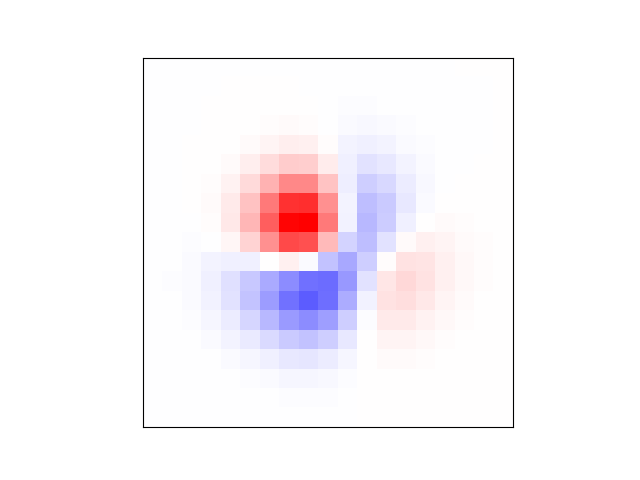} &
      \hspace{-7mm}
      \includegraphics[width=0.1\textwidth]{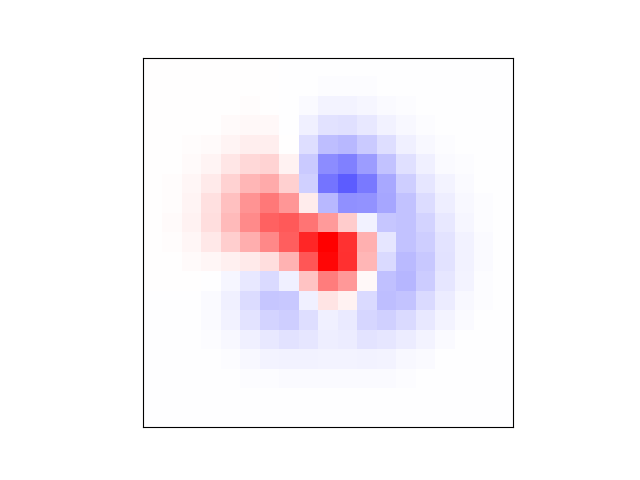} &
      \hspace{-7mm}
      \includegraphics[width=0.1\textwidth]{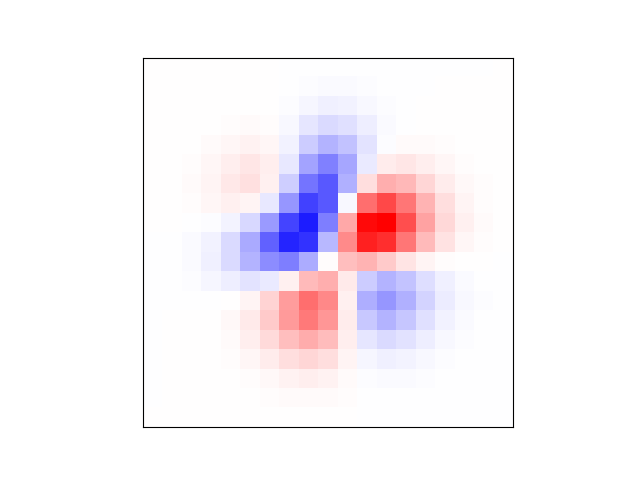} &
      \hspace{-7mm}
      \includegraphics[width=0.1\textwidth]{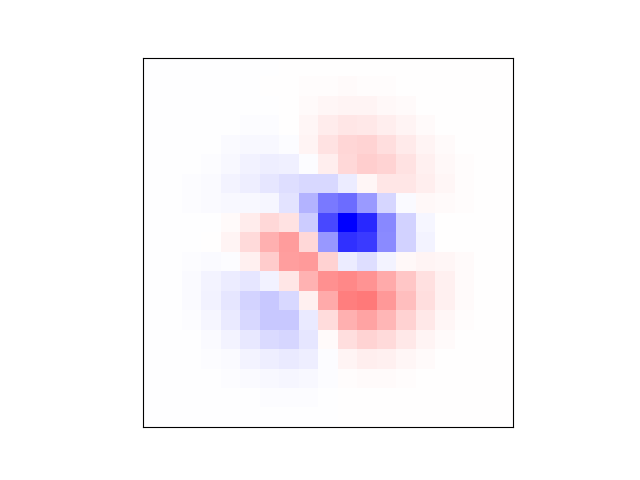} &
      \hspace{-7mm}
      \includegraphics[width=0.1\textwidth]{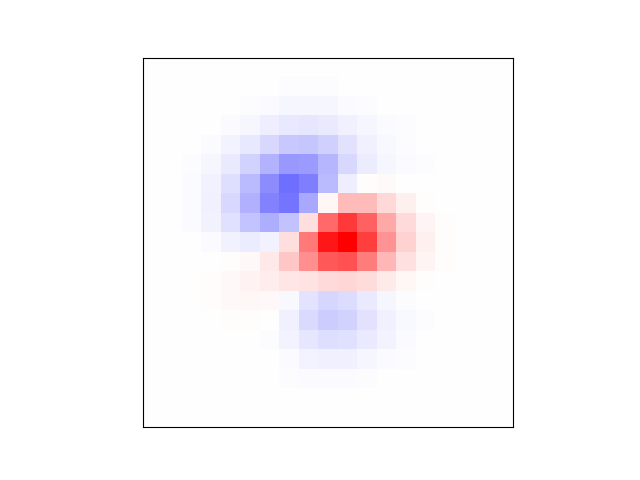} &
      \hspace{-7mm}
      \includegraphics[width=0.1\textwidth]{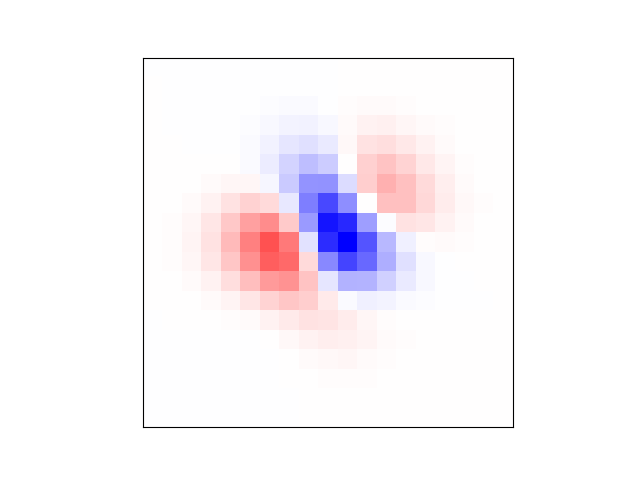}
      \\
    \end{tabular} 

  \vspace{-1mm}
  
    \begin{tabular}{ccccccccccc}
      \hline
      \raisebox{3\height}{Layer 2} &
      \hspace{-5mm}
      \includegraphics[width=0.1\textwidth]{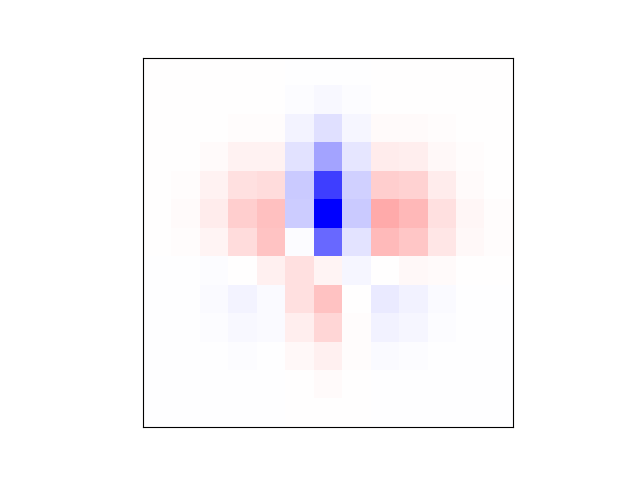} &
      \hspace{-7mm}
      \includegraphics[width=0.1\textwidth]{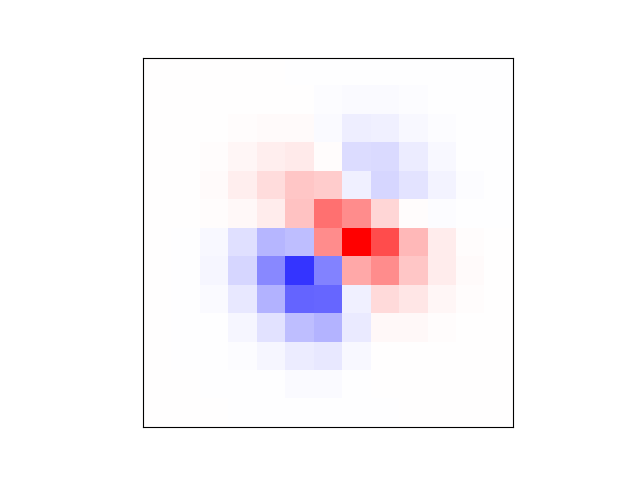} &
      \hspace{-7mm}
      \includegraphics[width=0.1\textwidth]{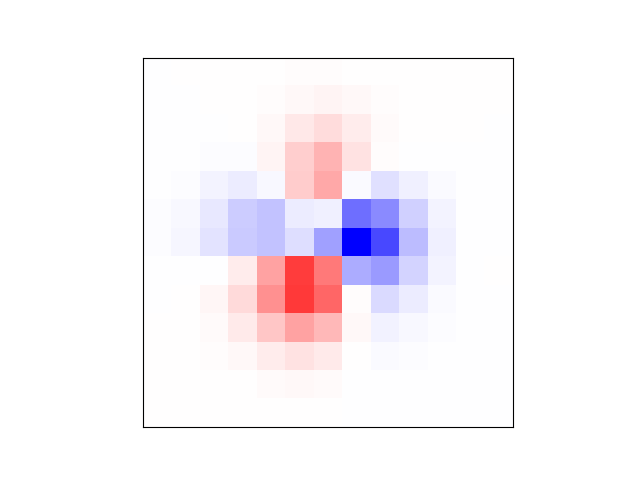} &
      \hspace{-7mm}
      \includegraphics[width=0.1\textwidth]{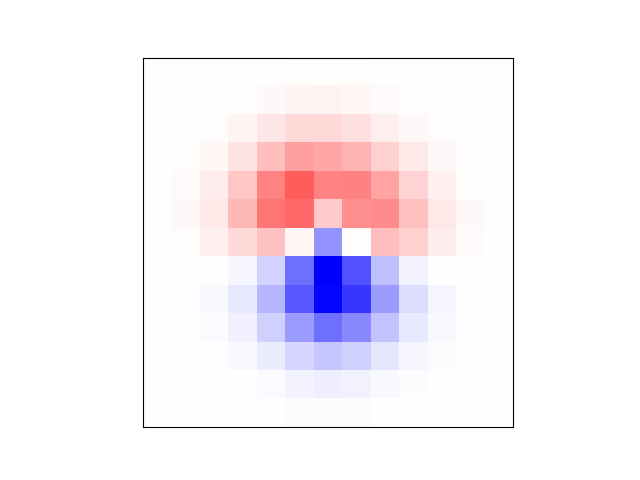} &
      \hspace{-7mm}
      \includegraphics[width=0.1\textwidth]{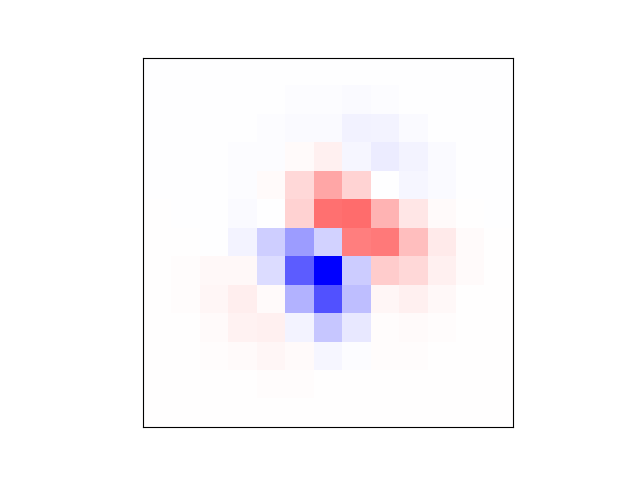} &
      \hspace{-7mm}
      \includegraphics[width=0.1\textwidth]{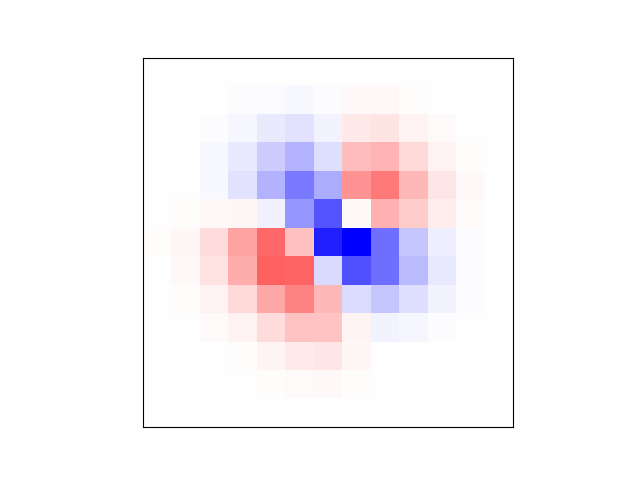} &
      \hspace{-7mm}
      \includegraphics[width=0.1\textwidth]{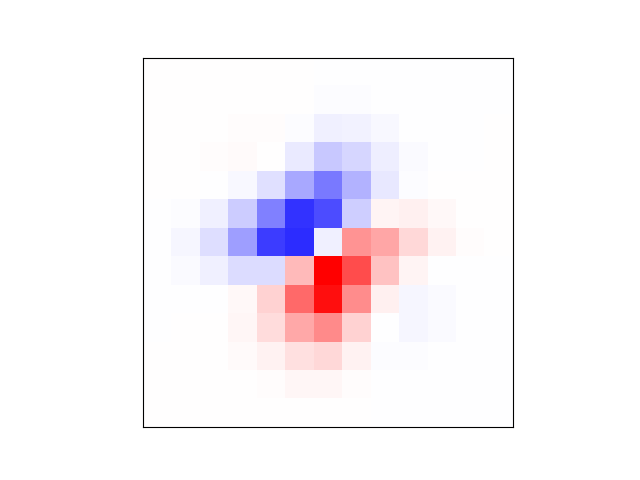} &
      \hspace{-7mm}
      \includegraphics[width=0.1\textwidth]{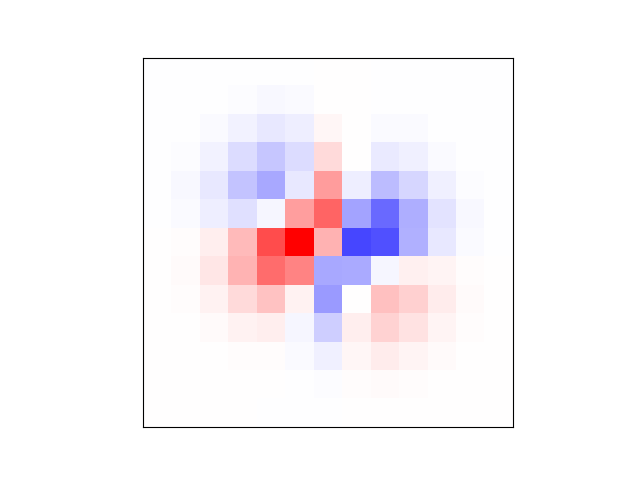} &
      \hspace{-7mm}
      \includegraphics[width=0.1\textwidth]{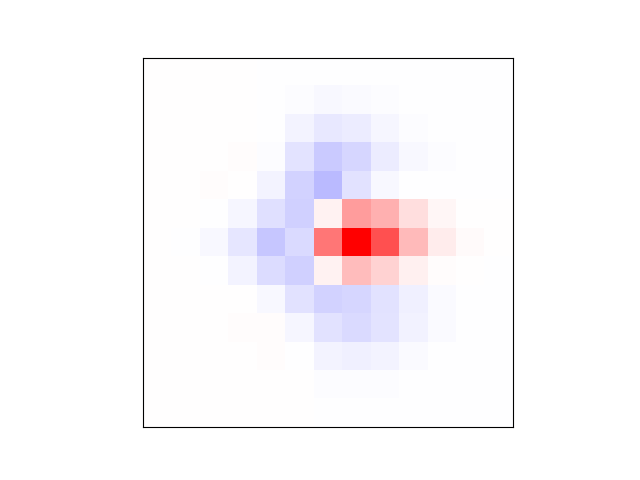} &
      \hspace{-7mm}
      \includegraphics[width=0.1\textwidth]{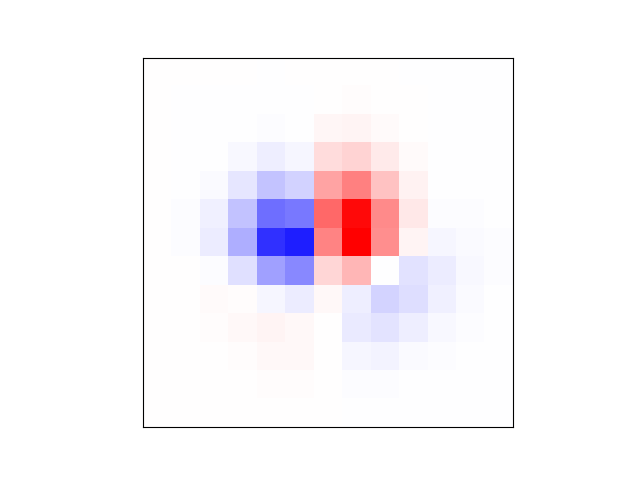}
      \\
    \end{tabular} 

  \vspace{-1mm}
  
    \begin{tabular}{ccccccccccc}
      \hline
      \raisebox{3\height}{Layer 1} &
      \hspace{-5mm}
      \includegraphics[width=0.1\textwidth]{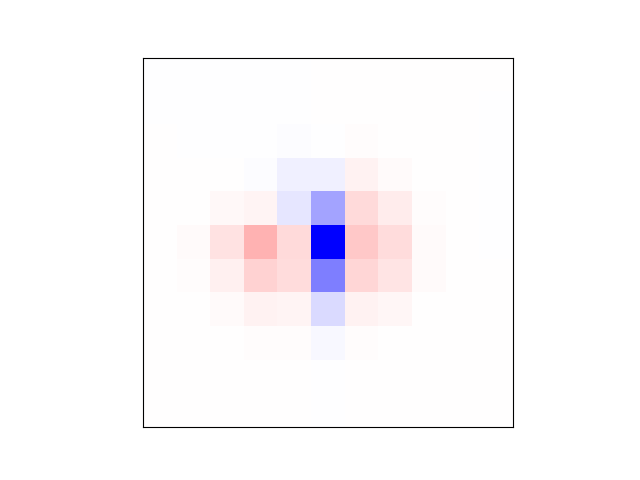} &
      \hspace{-7mm}
      \includegraphics[width=0.1\textwidth]{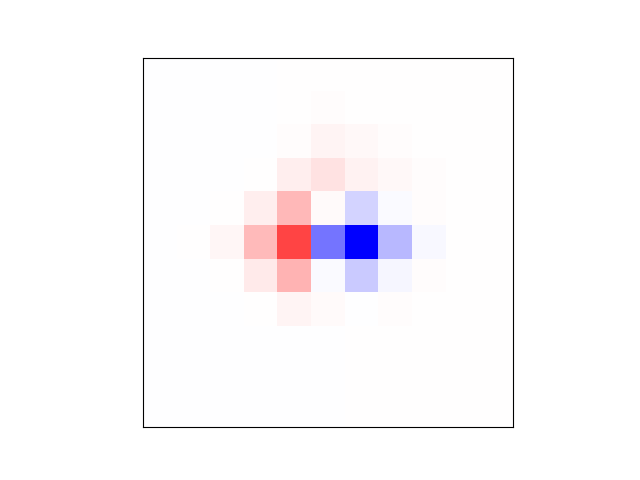} &
      \hspace{-7mm}
      \includegraphics[width=0.1\textwidth]{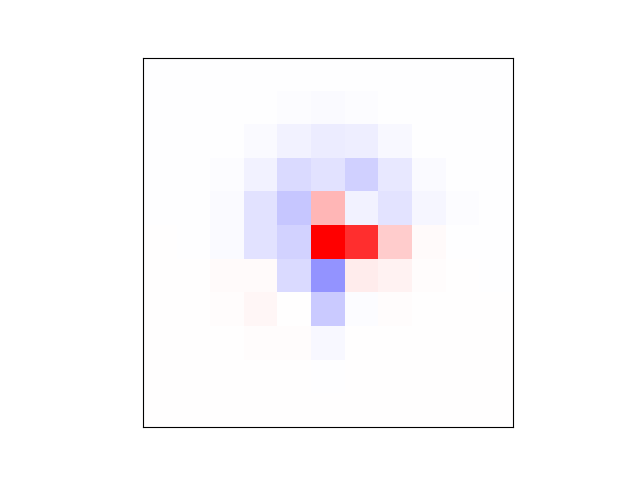} &
      \hspace{-7mm}
      \includegraphics[width=0.1\textwidth]{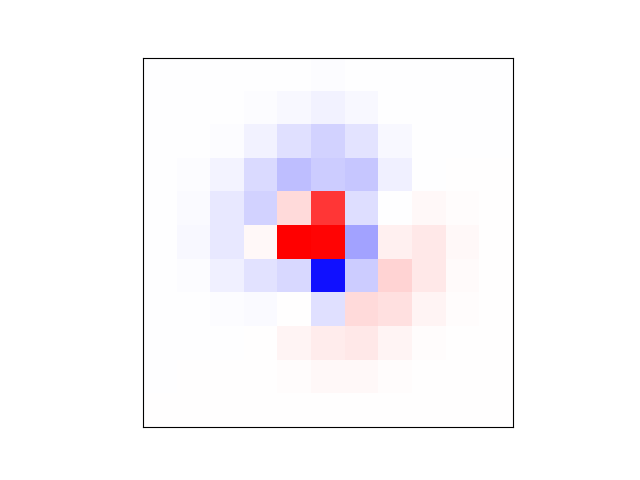} &
      \hspace{-7mm}
      \includegraphics[width=0.1\textwidth]{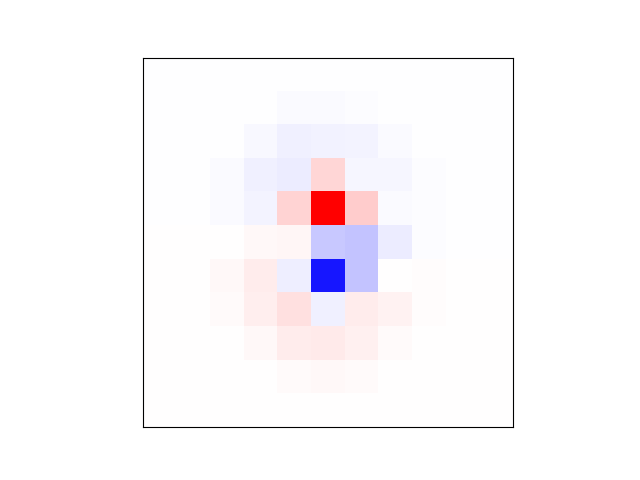} &
      \hspace{-7mm}
      \includegraphics[width=0.1\textwidth]{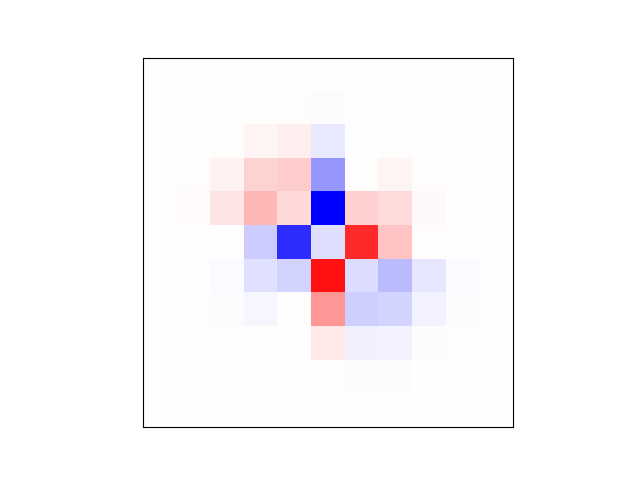} &
      \hspace{-7mm}
      \includegraphics[width=0.1\textwidth]{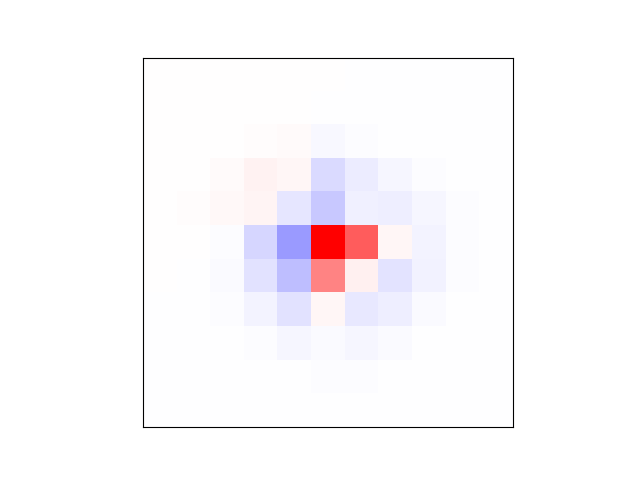} &
      \hspace{-7mm}
      \includegraphics[width=0.1\textwidth]{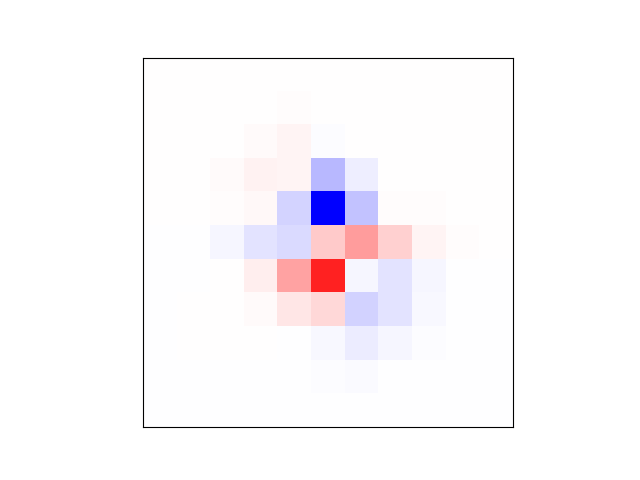} &
      \hspace{-7mm}
      \includegraphics[width=0.1\textwidth]{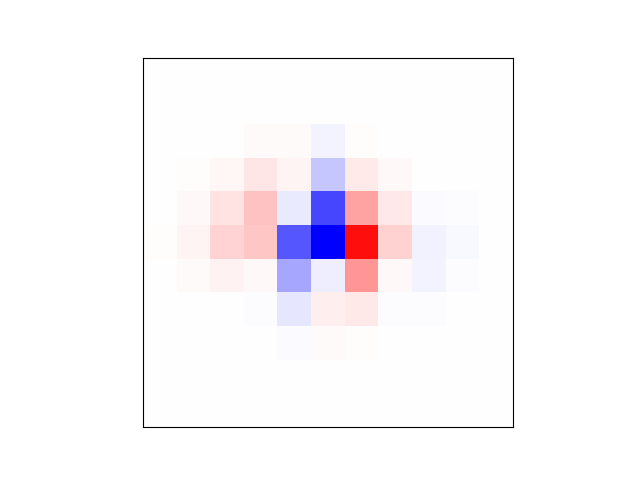} &
      \hspace{-7mm}
      \includegraphics[width=0.1\textwidth]{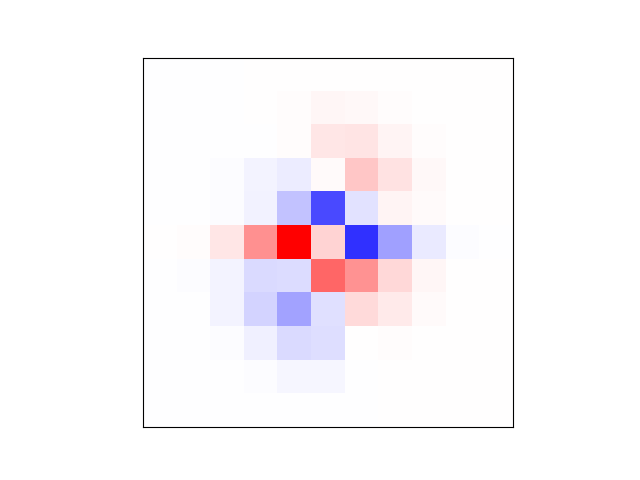}
    \end{tabular} 
  \end{center}
  \captionsetup{skip=-5pt}
  \caption{A selection of 10 learned effective filters from each of the layers of a single-scale-channel GaussDerNet, based on the 3-jet, and trained on the rescaled CIFAR-10 dataset (with image size $64 \times 64$ pixels). The color mapping has been set to map zero values to white, with positive values being mapped to red and negative values being mapped to blue. In each layer, the scale parameter $\sigma$ becomes larger according to Equation~(\ref{eq:sigma-geometric}), thus increasing the size of the convolution kernels. Note, in particular, that due to the higher orders of spatial differentiation, the learned kernels based on the 3-jet have more complex shapes than the kernels based on the 2-jet.}
  \label{fig-learned-filters_3_jet}
\end{figure*}

In these ways, these visualisations demonstrate that the architecture of the scale-covariant GaussDerNet is able to generalise the multi-scale notions of classical scale-space theory, based on linear Gaussian-derivative-based receptive fields or possibly non-linear polynomial combinations of such primitives into differential invariants, to multi-scale processing in deep networks, with intuitively very reasonable properties.

\subsection{Learned receptive fields for Gaussian derivative networks based on 2-jets or 3-jets}
\label{sec-jet-learned-receptive-field-comparison}
To demonstrate what types of receptive fields are learned in the GaussDerNets, as linear combinations of Gaussian derivatives up to order 2 or 3, according to Equations~(\ref{eq:2-Jet}) and~(\ref{eq:3-Jet}), we will in this section show visualisations of such receptive fields. For this purpose, we will use receptive fields learned from the rescaled CIFAR-10 dataset,\footnote{With regard to the visualisation of the receptive fields learned from the CIFAR-10 dataset, it should be noted that there is the possibility that the mirroring operation, performed to extend the size of the images for the size factor 1, could influence the receptive fields, because of a possible bias in the image data with regard to the horizontal and vertical mirroring operations. After performing a corresponding visualisation of receptive fields learned from the regular CIFAR-10 dataset, without any mirroring operations in that dataset, we have, however, not been able to find any significant differences between the distributions of the learned receptive fields. For this reason, we report the receptive fields learned by the multi-scale channel networks on the rescaled CIFAR-10, which constitutes one of the main use cases in this study.} because the images in this dataset come from a distribution closer to the distribution of natural images than for the two other datasets considered in this study.

Specifically, we will use multi-scale channel networks trained on the rescaled CIFAR-10 dataset, using central pixel extraction and average pooling over scale, with the discrete implementation based on the discrete analogue of the Gaussian kernel according to Equations~(\ref{eq:discrete-gaussian-kernel})-(\ref{eq:central-difference-operators-2}), and using the relative scale ratio $r = 1.45$, with scale-channel dropout with $q=0.3$ applied during training, and all the remaining parameters and training settings as previously described in Section~\ref{sec-train-proced}.

Figures~\ref{fig-learned-filters_2_jet} and~\ref{fig-learned-filters_3_jet} show 10 learned effective filters for each layer in the GaussDerNets, based on either the 2-jet or the 3-jet, respectively, and with the size of the receptive fields increasing from lower to higher layers, because of a corresponding increase in the scale parameter $\sigma$. As can be seen from the illustrations, a number of the receptive fields look very similar to directional derivatives of Gaussian kernels, as modelled according to Equations~(\ref{eq-1st-order-dir-der}) and~(\ref{eq-2nd-order-dir-der}), in particular for the lowest layers of the GaussDerNet based on the 2-jet.

In the higher layers for the GaussDerNet based on the 2-jet, somewhat different receptive field shapes than rather pure directional derivatives of Gaussian kernels are, however, learned. For the GaussDerNet based on the 3-jet, substantially more complex receptive field shapes are furthermore learned in the higher layers.

These visualisations do in this way demonstrate that the GaussDerNet based on the 3-jet learns substantially more complex receptive field shapes than the GaussDerNet based on the 2-jet. The presence of filters with such more complex shapes could thus serve as to explain why the Gaussian derivative network based on the 3-jet leads to better performance at coarser scales than the Gaussian derivative network based on the 2-jet, as previously observed in Section~\ref{sec-jet-comparison}. 

Additionally, some of the learned effective filters do also appear to be very similar to the Laplacian of the Gaussian $\nabla^2 L = L_{xx} + L_{yy}$, which has been previously used for modelling biological visual receptive fields in the retina and the lateral geniculate nucleus (see Lindeberg \citeyear{Lin21-Heliyon} Section~4.1). One example of such a filter can be found in layer 4, column 6 in Figure~\ref{fig-learned-filters_2_jet}. Directional derivatives of (affine) Gaussian kernels have, in turn, been previously used for modelling simple cells in the primary visual cortex (see Lindeberg \citeyear{Lin21-Heliyon} Section~4.3).

We can also see from these visualisations, that a portion of the learned effective filters in the first layers contain high values concentrated in the central region, resulting in relatively much fainter intensity in the area surrounding the centre, an example of which can be seen in layer 1 column 7 in Figure~\ref{fig-learned-filters_2_jet}. 

From these visualisations, we can thus see that the GaussDerNets, within the space spanned by the linear combinations of Gaussian derivatives, are able to learn both (i)~receptive fields that are very similar to directional derivatives of Gaussian kernels, (ii)~receptive fields that are very similar to the Laplacian of the Gaussian, found in both biological vision and classical computer vision algorithms, as well as (iii)~receptive fields with more complex shapes over the image domain.

\section{Summary}
\label{sec-conclusion}
We have presented an in-depth analysis of the scale generalisation properties of the scale-covariant and scale-invariant Gaussian derivative networks (GaussDerNets), on images with scaling variations in the testing data that are not spanned by the training data. Instead of explicitly training the network to handle scaling variations in the image data, the influence of the transformation properties on the receptive fields is provided as prior knowledge in the network architecture, and can in this way make it possible for the network to make predictions regarding the influence of the scaling transformations on the input data that have not been present in the training stage; a functionality that may be lacking in many other architectures for deep networks, despite the fact that differences in scale are common in natural image data. 

This is achieved by constructing the deep network by coupling layers, defined by trainable linear combinations of scale-normalised Gaussian derivative basis functions, in cascade, with weight sharing and pooling over multiple scale channels. We argue that deep networks with such built-in architectural attributes can learn more robust representations, and enable efficient use of training data during the training process, as the model shares information between the scale channels.

In this work, we have augmented these properties of GaussDerNets by adding a number of conceptual and algorithmic extensions, and showed the possibilities of what this design approach can achieve in terms of scale generalisation, by applying it to new datasets with spatial scale variations, containing substantially more complex types of image structures than the dataset used in the initial proof-of-concept in Lindeberg (\citeyear{Lin22-JMIV}). For the purpose of applying the networks to these datasets, we have first in Section~\ref{sec-alg-improv} complemented the previous notion of GaussDerNets with the following extensions:
\begin{itemize}
\item
Average pooling over scales as a scale selection mechanism, to make more efficient use of information over the different scale channels, compared to the previously used notion of max pooling over scales as the scale selection mechanism,
\item
Spatial max pooling as a focus-of-attention mechanism, to handle image data where the objects of interest are not necessarily centered in the image domain, and
\item
Complementary data augmentation and regularisation mechanisms, in terms of scale-channel dropout and cutout. 
\end{itemize}
Then, to evaluate the performance of the GaussDerNets under scaling variations, we have in Section~\ref{sec-generating-datasets} defined two new datasets with systematic spatial scaling variations over a factor of 4, which we refer to as the rescaled Fashion-MNIST dataset and the rescaled CIFAR-10 dataset. We have also defined a second variant of the rescaled Fashion-MNIST dataset, that in addition to the spatial scaling, applies random spatial translations of the objects within the image domain.
By performing training at only a single scale for these datasets, for the size factor 1, we have explored the sensitivity to scaling variations over the range of size factors $S \in [1/2, \,2]$. Additionally, we have also investigated the scale generalisation properties of the extended GaussDerNets on the previously existing STIR dataset, consisting of images containing rescaled objects grouped by size, as previously introduced by Altstidl et al. (\citeyear{AltNguSchKoeMutEskZan23-IJCNN}).

In our experimental characterisation of the scale generalisation properties for these networks in Section~\ref{sec-experiments}, we have found that due to their scale transformation properties, the multi-scale-channel GaussDerNets have the ability to classify image patterns at scales not spanned by the training data, on the rescaled Fashion-MNIST and the rescaled CIFAR-10 datasets. The networks defined using average pooling over scales may also sometimes lead to better scale generalisation results than max pooling over scales, with both methods obtaining rather flat scale generalisation curves on test data with relative size factors between 1/2 and 2, with the training process typically being more stable for networks using average pooling over scales. For both of these approaches, the scale levels selected during the inference stage were found to be linearly proportional to the object size in the testing data, with similarities to classical scale selection methods based on scale-normalised derivatives, and consistent with the behaviour observed in related previous work on the Large Scale MNIST dataset (Lindeberg \citeyear{Lin22-JMIV}, Jansson and Lindeberg \citeyear{JanLin22-JMIV}).

Additionally, while the focus of this work is not object detection, we experimentally showed that for a network trained on the rescaled Fashion-MNIST with translations dataset, that the spatial max pooling operation is capable of achieving good localisation of objects in images that contain objects that are not centred in the image domain. The scale generalisation performance and scale selection properties of this network were comparable to the results obtained on the rescaled Fashion-MNIST dataset, using a network with the previously used central pixel extraction mechanism.

Furthermore, we have found in our experiments on the STIR traffic sign and STIR aerial datasets, that by using a (global) spatial max pooling as a spatial selection operation, the GaussDerNet achieves better scale generalisation performance than the previously considered methods in Altstidl et al. (\citeyear{AltNguSchKoeMutEskZan23-IJCNN}) and Velasco-Forero (\citeyear{Velasco2023-ICGSI}), especially when trained with scale-channel dropout. Because the STIR dataset has rather few training samples, this demonstrates that the GaussDerNet is capable of effective learning, even when provided with rather limited amount of training data.

To characterise the properties of the extended GaussDerNet further, we have in Section~\ref{sec-ablation} performed a set of complementary ablation and comparative studies. 
First, we showed that using the data augmentation and regularisation techniques proposed in this work does improve the scale generalisation of the multi-scale-channel GaussDerNets. In particular, scale-channel dropout, which is an operation that during training applies dropout across the outputs from the scale channels before the permutation-invariant scale pooling stage, was shown to significantly improve the scale generalisation performance, especially for the networks based on max pooling over scales.

Based on the presented experimental results, from a comparative study on the effects of basing the GaussDerNet layers on third-order derivatives, it is clear that when trained on a dataset consisting of natural image data, the network based on 3-jet layers achieved improved scale generalisation performance compared to the 2-jet based network, for size factors $\geq$ 1. This suggests that defining the layers using higher order derivatives may result in better representation of the image structures. For finer-scale image data, the network based on Gaussian derivatives up to second order does, however, generalise better to scales not seen in the training data, likely due to a lower sensitivity to errors in the discrete approximations for spatial derivatives of lower order.
We have also experimentally demonstrated that the GaussDerNet architecture can be based on alternative permutation invariant pooling methods, such as logsumexp pooling, and achieve good scale generalisation performance.

In addition, we have experimentally investigated the effects of using different discrete approximations of the Gaussian derivative kernels, showing that using the discrete analogue of the Gaussian kernel in combination with central difference operators, in many of the studied cases gives the best or among the best results.
We have also investigated the effects of learning of the scale levels, demonstrating that the a priori choice of using a logarithmic distribution of the scale levels can be regarded as a very reasonable choice, and that networks with learned scale parameters usually provide only minor improvement in performance compared to networks using fixed scale parameters.
When learning the scale levels, we have specifically introduced a separate learning rate for the scale parameters during training, which we show is needed to be appropriately chosen to achieve better performance, as it influences the rate by which the scale parameters converge.

Finally, to demonstrate the properties of the GaussDerNets more visually, we have in Section~\ref{sec-network-function-visualisation} performed an in-depth study of the explainability properties of these networks in terms of activation maps, as well as visualised the receptive fields learned in the GaussDerNets of order either 2 or 3.
In particular, by our deliberate omission of any spatial subsampling operations in the networks, the visualisations of the activation maps in the final layer become visually highly interpretable, and show that the networks demonstrate a capability to identify key subsets of image structures required for object classification. 

The formulation of the GaussDerNets in terms of scale-space operations has also been shown to be well suited for the use of a spatial max pooling mechanism as a focus-of-attention mechanism, with structurally very close similarities to the use of strength measures in terms of scale-normalised differential expressions as a basis for feature detection algorithms over both space and scale. Our visualisations of the learned receptive fields show that the learned filters consist of directional derivatives of Gaussian kernels of different orders, and, most notably, include filters resembling the Laplacian of Gaussian operator, as well as more complex shapes.

We have also performed additional experiments and studies in the supplementary material. In Appendix~A.1 in the supplementary material, we experimentally showed that the scale-covariant properties of the GaussDerNets allow single-scale-channel networks to handle image structures at scales other than they were trained on, by applying a scaling transformation to the scale parameters in the networks. We have also experimentally demonstrated that this property enables reasonably good scale generalisation in multi-scale-channel networks with weights obtained by a weight transfer directly from a trained single-scale-channel network, which is possible because of the weight sharing. However, multi-scale-channel training is found to be essential to achieve really good scale generalisation properties.

In Appendix~A.2 in the supplementary material, to demonstrate the properties of scale parameter learning, we have performed an in-depth study of the learned scale levels for implementing GaussDerNets based on different discretisations of the Gaussian derivative operators. We demonstrated that in the first layer, the scale parameters converge to rather fine scale levels, and differ in values depending on which discretisations method is being used. However, because the discrete spatial extent of the Gaussian derivative kernels for a given scale level may differ between discretisation methods, these different learned scale values have been found to correspond to approximately rather similar amounts of spatial smoothing, as quantified by spatial spread measures.

In Appendix~A.3 in the supplementary material, we conducted statistical comparisons in order to determine which discretisation methods lead to the best performance of the GaussDerNets. For the networks trained using fixed scale parameters, the discrete analogue of the Gaussian kernel with central differences and the sampled Gaussian derivative kernel are found to perform significantly better than the hybrid sampled and normalised sampled discretisation approaches. For the networks trained using learned scale parameters, the sampled Gaussian derivative operators performed significantly better than the integrated Gaussian derivative kernels, as well as better than the hybrid integrated and sampled discretisation approaches.

To conclude, a major aim of this paper has been to demonstrate that the GaussDerNets can, with appropriate extensions, be successfully applied to datasets with spatial scaling variations, including datasets containing natural images, to classify objects at scales not seen in the training stage. We argue that such a functionality is essential for deep networks, to be able handle scaling variations in image data, as arise from the variabilities of image structures generated by natural image transformations.

\section*{Acknowledgements}

The experiments underlying this work were performed on GPU resources provided by the PDC Center for High Performance Computing at KTH Royal Institute of Technology in Sweden.

\appendix
\section{Appendix}

\begin{table*}[htbp]
  \centering
  \begin{tabular}{lcc}
       \hline
        \textbf{Network} & \textbf{Relative scale ratio} & \textbf{Initial scale values} \\ 
       \hline 
       Single-scale-channel & $r = 1.47$ & $\sigma_{0} = 1/2$\\ 
       Multi-scale-channel (max pooling) & $r = 1.47$ & $\sigma_{i,0} \in \{1/4, 1/(2\sqrt{2}), 1/2, 1/\sqrt{2}, 1, \sqrt{2}\}$ \\ 
       Multi-scale-channel (average pooling) & $r = 1.47$ & $\sigma_{i,0} \in \{1/4, 1/(2\sqrt{2}), 1/2, 1/\sqrt{2}, 1, \sqrt{2}\}$ \\ \hline

       Single-scale-channel & $r = 1.37$ & $\sigma_{0} = 1/\sqrt{2}$\\ 
       Multi-scale-channel (max pooling) & $r = 1.37$ & $\sigma_{i,0} \in \{1/4, 1/(2\sqrt{2}), 1/2, 1/\sqrt{2}, 1, \sqrt{2}, 2\}$ \\ 
       Multi-scale-channel (average pooling) & $r = 1.37$ & $\sigma_{i,0} \in \{1/4, 1/(2\sqrt{2}), 1/2, 1/\sqrt{2}, 1, \sqrt{2}, 2\}$ \\ \hline

       Single-scale-channel & $r = 1.28$ & $\sigma_{0} = 1$\\ 
       Multi-scale-channel (max pooling) & $r = 1.28$ & $\sigma_{i,0} \in \{1/(2\sqrt{2}), 1/2, 1/\sqrt{2}, 1, \sqrt{2}, 2, 2\sqrt{2} \}$ \\ 
       Multi-scale-channel (average pooling) & $r = 1.28$ & $\sigma_{i,0} \in \{1/(2\sqrt{2})$, $1/2, 1/\sqrt{2}, 1, \sqrt{2}, 2, 2\sqrt{2} \}$ \\ \hline
  \end{tabular}
  \caption{Parameter settings for the networks used for investigating the importance of multi-scale training in Appendix~\ref{sec-multi-chan-training-importance-appendix} on the rescaled Fashion-MNIST dataset. The value of relative scale ratio $r$ has been chosen such that all the models have approximately an equal $\sigma$-value in the final layer. The initial scale values of models in different row groups differ by a factor of $\sqrt{2}$, however $\sigma_{i,0}$-values below 1/4 are not used, as they are consider too small. The multi-scale-channel networks are then trained either in a genuine multi-scale-channel manner, or constructed by weight transfer by training just a single-scale-channel network on the training data, and then inserting those weights into a multi-scale-channel network.}
  \label{tab:adjusting-initial-scale-values-single-fashion}
\end{table*}

\begin{table*}[htbp]
  \centering
  \begin{tabular}{lcc}
       \hline
        \textbf{Network} & \textbf{Relative scale ratio} & \textbf{Initial scale values} \\ 
       \hline 
       Single-scale-channel & $r = 1.67$ & $\sigma_{0} = 1/2$\\ 
       Multi-scale-channel (max pooling) & $r = 1.67$ & $\sigma_{i,0} \in \{1/4, 1/(2\sqrt{2}), 1/2, 1/\sqrt{2}, 1\}$ \\ 
       Multi-scale-channel (average pooling) & $r = 1.67$ & $\sigma_{i,0} \in \{1/4, 1/(2\sqrt{2}), 1/2, 1/\sqrt{2}, 1\}$ \\ \hline

       Single-scale-channel & $r = 1.56$ & $\sigma_{0} = 1/\sqrt{2}$\\ 
       Multi-scale-channel (max pooling) & $r = 1.56$ & $\sigma_{i,0} \in \{1/4, 1/(2\sqrt{2}), 1/2, 1/\sqrt{2}, 1, \sqrt{2}\}$ \\ 
       Multi-scale-channel (average pooling) & $r = 1.56$ & $\sigma_{i,0} \in \{1/4, 1/(2\sqrt{2}), 1/2, 1/\sqrt{2}, 1, \sqrt{2}\}$ \\ \hline

       Single-scale-channel & $r = 1.45$ & $\sigma_{0} = 1$\\ 
       Multi-scale-channel (max pooling) & $r = 1.45$ & $\sigma_{i,0} \in \{1/(2\sqrt{2}), 1/2, 1/\sqrt{2}, 1, \sqrt{2}, 2\}$ \\ 
       Multi-scale-channel (average pooling) & $r = 1.45$ & $\sigma_{i,0} \in \{1/(2\sqrt{2})$, $1/2, 1/\sqrt{2}, 1, \sqrt{2}, 2\}$ \\ \hline
  \end{tabular}
  \caption{Parameter settings for the networks used for investigating the importance of multi-scale training in Appendix~\ref{sec-multi-chan-training-importance-appendix} on the rescaled CIFAR-10 dataset. The value of relative scale ratio $r$ has been chosen such that all the models have approximately an equal $\sigma$-value in the final layer. The initial scale values of models in different row groups differ by a factor of $\sqrt{2}$, however $\sigma_{i,0}$-values below 1/4 are not used, as they are consider too small. The multi-scale-channel networks are then trained either in a genuine multi-scale-channel manner, or constructed by weight transfer by training just a single-scale-channel network on the training data, and then inserting those weights into a multi-scale-channel network.}
  \label{tab:adjusting-initial-scale-values-single-cifar}
\end{table*}

\subsection{The importance of multi-scale-channel training}
\label{sec-multi-chan-training-importance-appendix}
Scale covariance in the multi-scale-channel Gaussian derivative network is achieved through weight sharing between multiple identical scale channels, each based on a different initial scale value $\sigma_{0}$. In view of this property, one could ask if it would be possible to handle image structures of different size in the image domain by first training a single-scale-network with initial scale level $\sigma_0$ on training data with a fixed object size $l_0$, and then transferring the learned weights to a single-scale-network with the initial scale value $S \, \sigma_{0}$, and applying that model to test data with a rescaled object size $S \, l_0$.

One could thus also ask, if it would be possible to take a short-cut when training a multi-scale-channel network, by instead training a single-scale-channel network on training data with a fixed object size, and then constructing the multi-scale-channel network by creating a set of scale channels for different values of the initial scale value $\sigma_0$, each using the weights of the single-scale-channel network, and then just coupling these transferred scale channels together by either max pooling over scales or average pooling over scales, an approach that would significantly reduce the computational cost of training.

In this section, we will experimentally investigate how weight transfer from a trained single-scale-channel network to a multi-scale-channel network, constructed by coupling a set of such transferred scale channels by either max pooling or average pooling over scales, affects the scale generalisation properties of the network, compared to genuine multi-scale-channel training, on either the rescaled Fashion-MNIST dataset or the rescaled CIFAR-10 dataset.

We will also perform an investigation of the scale covariance properties of the Gaussian-derivative-based network architecture, by applying a scaling transformation to the scale parameters of a single-scale-channel network trained on objects of a given size, which we here artificially achieve by manually adjusting the initial scale value of the model, and then applying this single-scale network with the same weights to testing data of a different size, scaled by the same spatial scaling factor as used for scaling the spatial receptive fields in the GaussDerNet.

Both of these studies will be repeated for three different choices of the initial scale value $\sigma_0 \in \{1/2, 1/\sqrt{2}, 1\}$, and with the value of the relative scale factor $r$ between adjacent scale levels adjusted in such a way that the spatial scale parameter $\sigma_{k}$ in the last layer is the same for the different choices of $\sigma_0$. In other words, the scale parameter in the last layer $k$ is determined such that $\sigma_{k} = \sigma_{0} \, r^{k-1} \approx D$, where $D$ is a constant having different values for different datasets.

When performing these experiments, we did thus for each dataset, and for each value of the initial scale level $\sigma_0$, first train a single-scale-channel network on size factor 1 training data. Then, we applied a transferred copy of the single-scale-channel with manually rescaled initial scale value $S \, \sigma_0$ to the rescaled image data in the dataset for the same spatial rescaling factor $S$. In other words, during evaluation we applied a proportional scaling transformation to the scale parameters by manually setting $\sigma_{j}' = S \, \sigma_{j}$ in each layer $j$ in the given single-scale-channel network, while the weights of the network remained fixed.


\begin{figure*}[htbp] 
  \begin{center}
    \begin{subfigure}{0.47\textwidth}
        \centering
        \textit{\hspace{0.6cm} Networks corresponding to $\sigma_{0}=1/2$}
        \includegraphics[width=\linewidth]{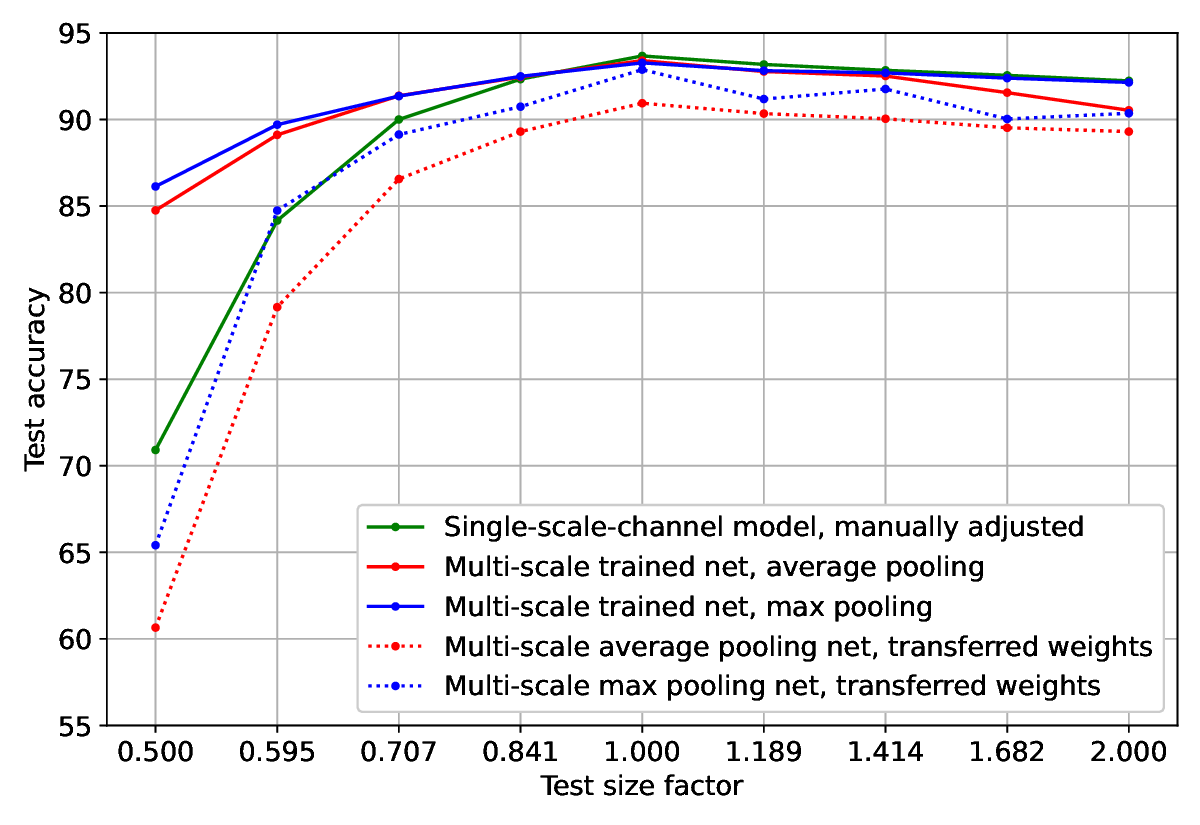}
        \label{fig:subfiga-fashion-05}
    \end{subfigure}
    \hfill
    \begin{subfigure}{0.42\textwidth}
        \centering
        \textit{\hspace{0.6cm} Scale selection histogram for $\sigma_{0}=1/2$}
        \includegraphics[width=\linewidth]{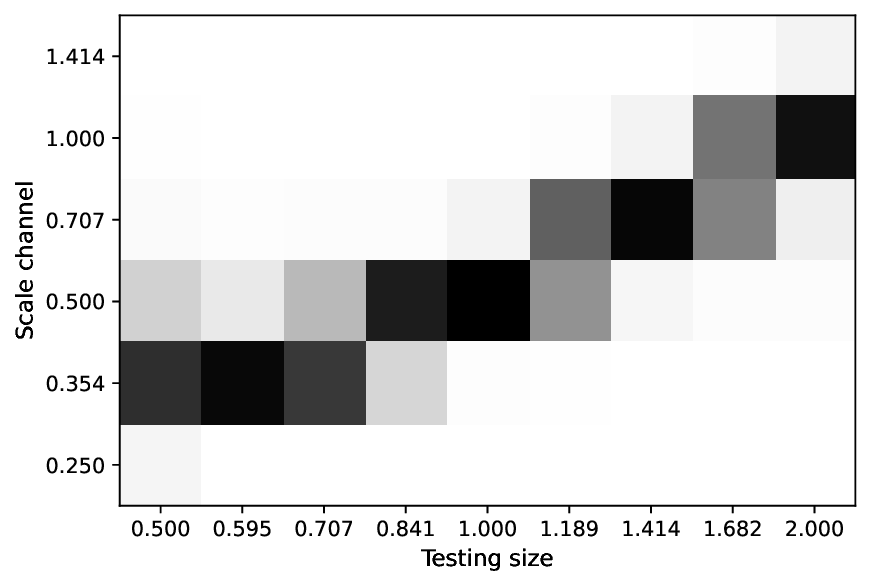}
        \label{fig:subfigb-fashion-05-histogram}
    \end{subfigure}

    \begin{subfigure}{0.47\textwidth}
        \centering
        \textit{\hspace{0.6cm} Networks corresponding to $\sigma_{0}=1/\sqrt{2}$}
        \includegraphics[width=\linewidth]{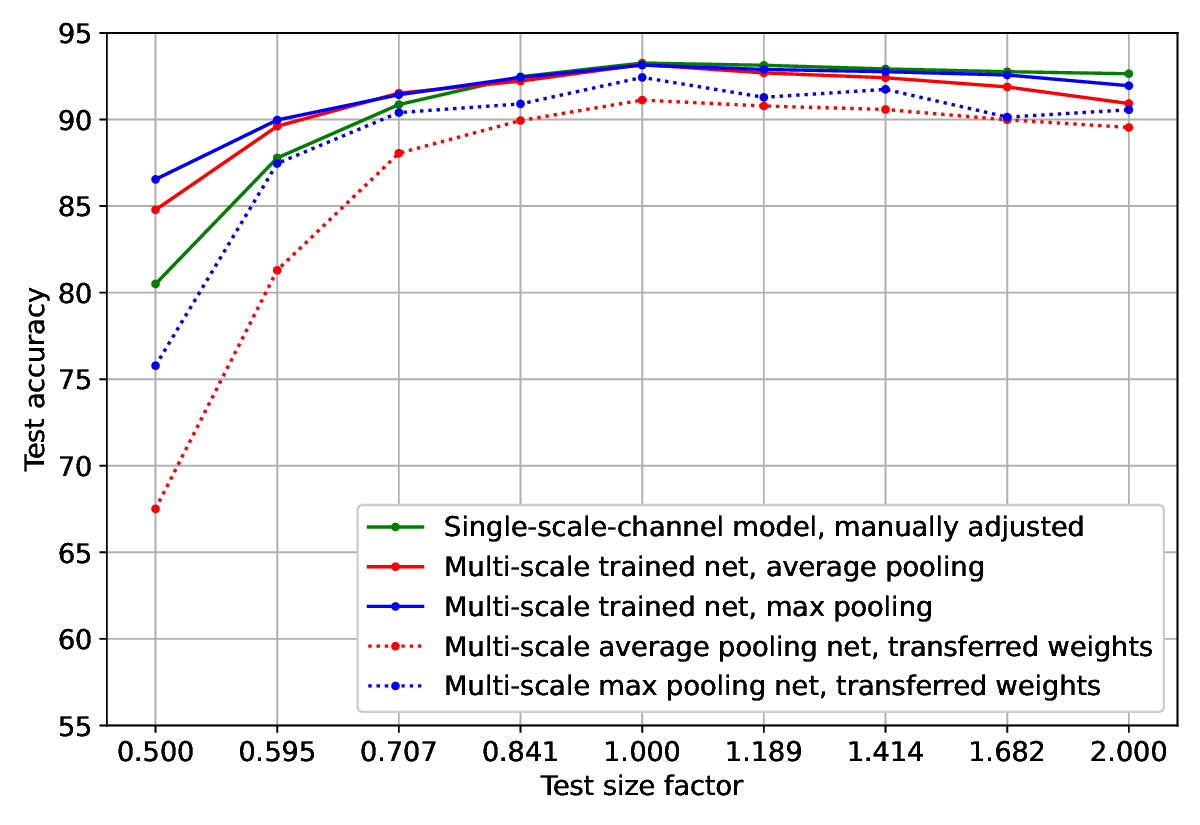}
        \label{fig:subfigb-fashion-0707}
    \end{subfigure}
    \hfill
    \begin{subfigure}{0.42\textwidth}
        \centering
        \textit{\hspace{0.7cm} Scale selection histogram for $\sigma_{0}=1/\sqrt{2}$}
        \includegraphics[width=\linewidth]{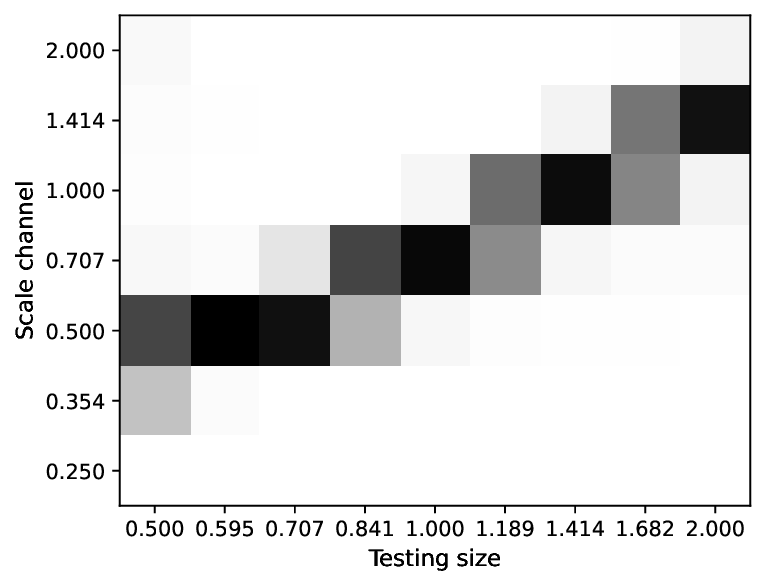}
        \label{fig:subfigb-fashion-0707-histogram}
    \end{subfigure}
    
    \begin{subfigure}{0.47\textwidth}
        \centering
        \textit{\hspace{0.6cm} Networks corresponding to $\sigma_{0}=1$}
        \includegraphics[width=\linewidth]{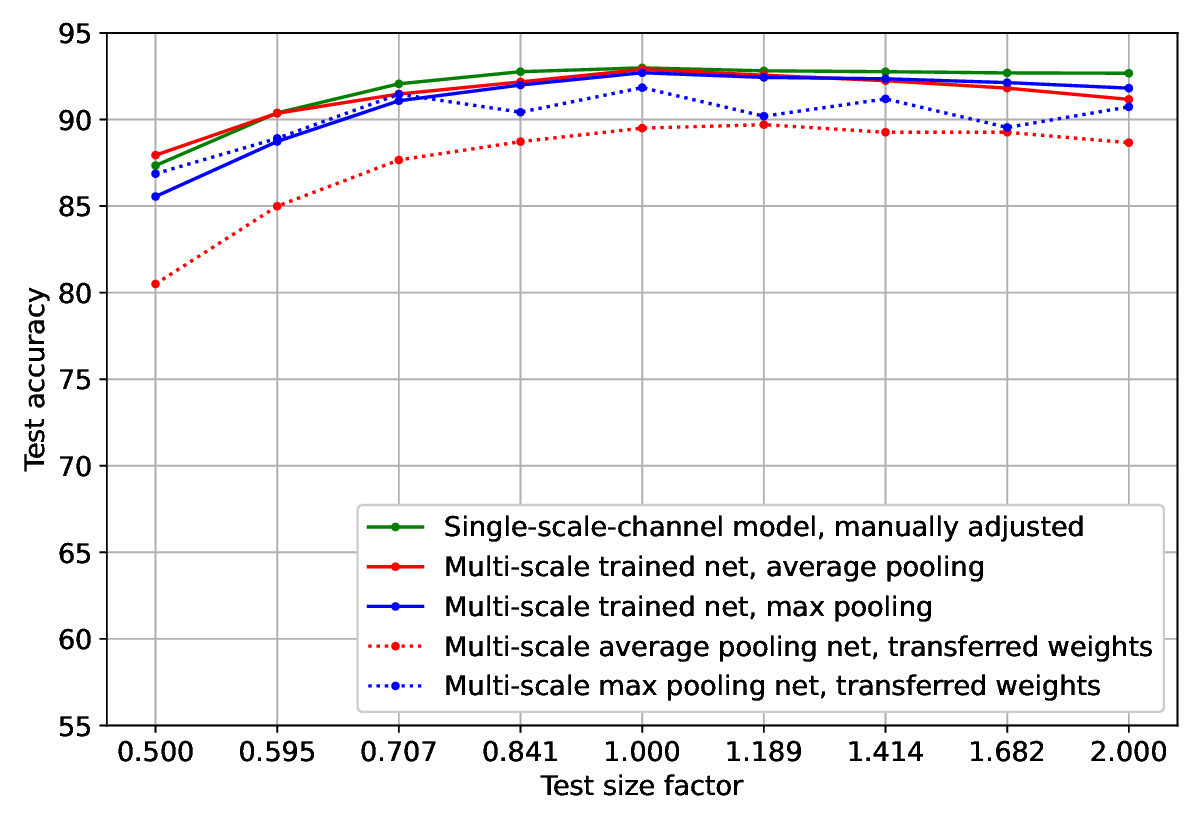}
        \label{fig:subfiga-fashion-1}
    \end{subfigure}
    \hfill
    \begin{subfigure}{0.42\textwidth}
        \centering
        \textit{\hspace{0.7cm} Scale selection histogram for $\sigma_{0}=1$}
        \includegraphics[width=\linewidth]{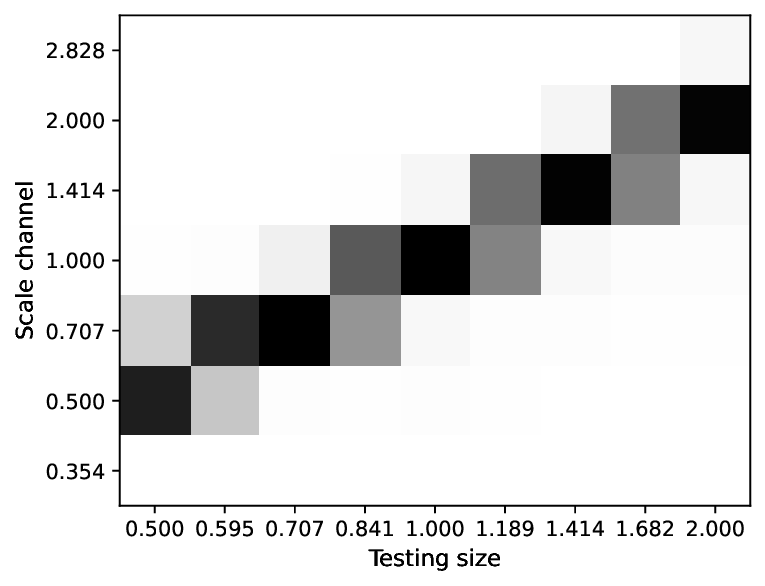}
        \label{fig:subfiga-fashion-1-histogram}
    \end{subfigure}
  \end{center}
\caption{Scale generalisation performance on the rescaled Fashion-MNIST dataset, for different initial scale values $\sigma_0$ and different ways of defining the multi-scale-channel networks, by either (i)~genuine multi-scale-channel training on the training data for the training data for size factor 1 (the solid red and solid blue curves), or instead (ii)~constructing the multi-scale-channel networks by weight transfer from a single-scale-channel network trained on the training data for the same size factor 1 (the dashed red and the dashed blue curves). For comparison, we also show the result of applying (iii)~the single-scale-channel network with manually transformed scale parameters on the rescaled testing data for the same scaling factors during evaluation, using a priori knowledge about the scaling transformation between the training data and the testing data. As can be seen from these graphs, genuine multi-scale-channel training leads to substantially better scale generalisation performance, compared to weight transfer from single-scale-channel learning. The single-scale-channel network with manually adjusted scale parameters (the solid green curves) does, however, lead to rather good performance under the spatial scaling transformations, in agreement with the provable scale-covariant properties of the underlying continuous Gaussian derivative network, that is approximated discretely in these implementations. Scale selection histograms for the multi-scale-channel networks using max pooling over scales with weights transferred from the corresponding single-scale-channel networks, represented as the dashed blue curve in the graphs, are shown on the right.} 
\label{fig-fashion-covariance-experiments}
\end{figure*}

\begin{figure*}[htbp] 
  \begin{center}
    \begin{subfigure}{0.47\textwidth}
        \centering
        \textit{\hspace{0.6cm} Networks corresponding to $\sigma_{0}=1/2$}
        \includegraphics[width=\linewidth]{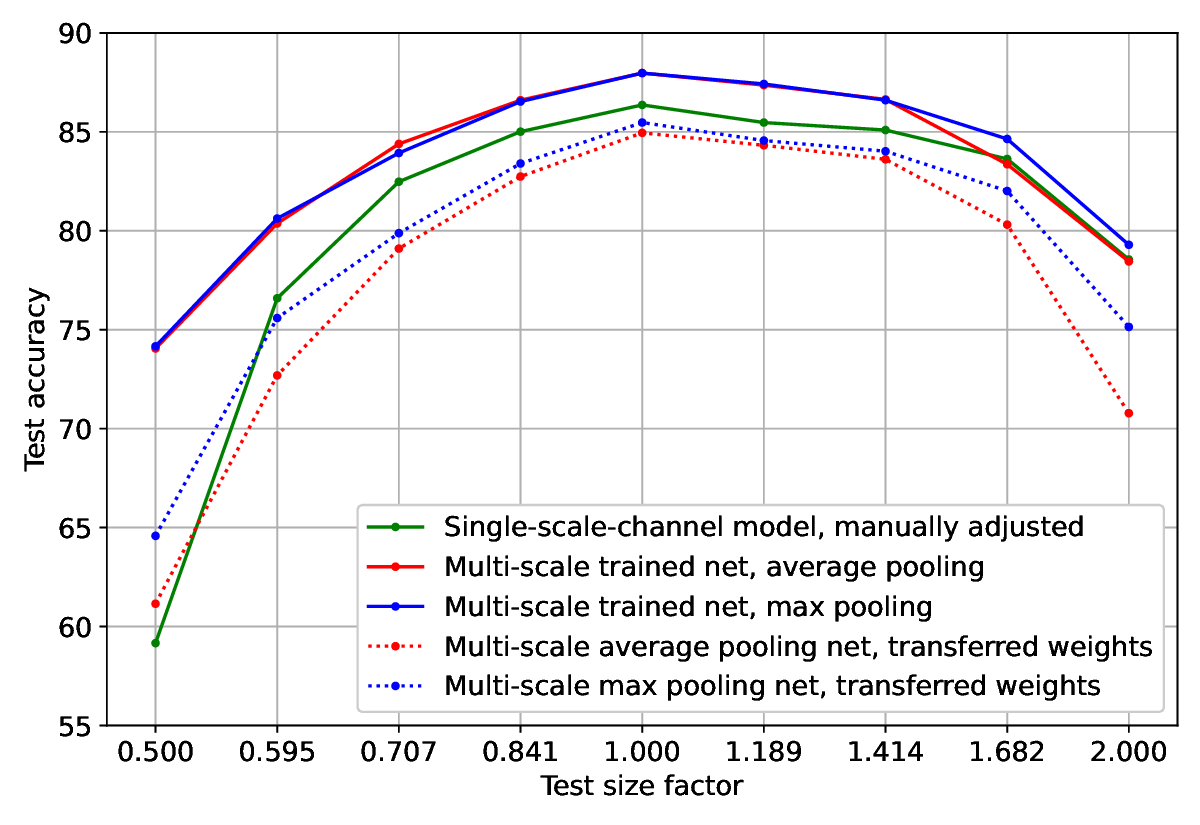}
        \label{fig:subfiga-cifar-05}
    \end{subfigure}
    \hfill
    \begin{subfigure}{0.45\textwidth}
        \centering
        \textit{\hspace{0.6cm} Scale selection histogram for $\sigma_{0}=1/2$}
        \includegraphics[width=\linewidth]{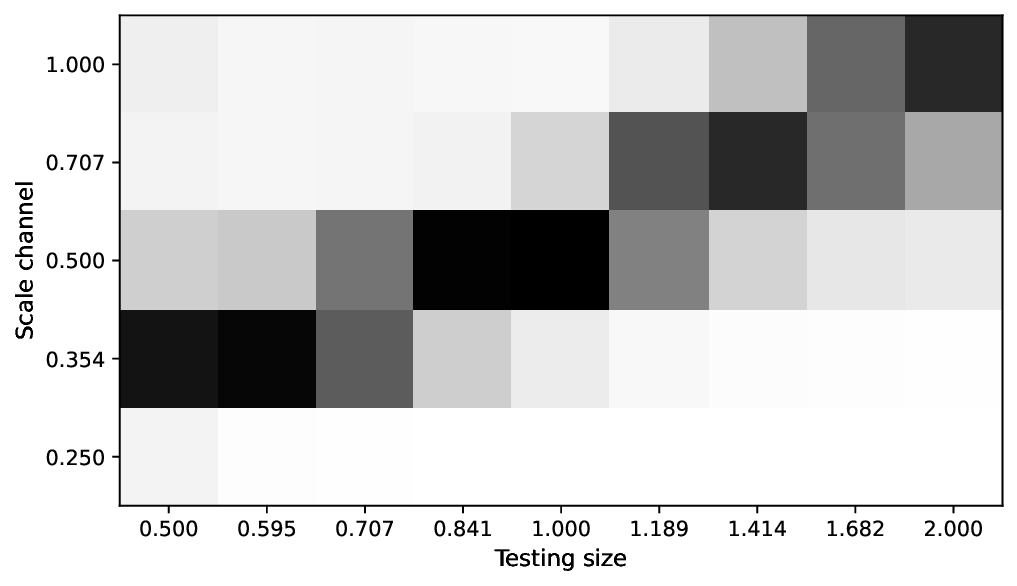}
        \label{fig:subfigb-cifar-05-histogram}
    \end{subfigure}

    \begin{subfigure}{0.47\textwidth}
        \centering
        \textit{\hspace{0.6cm} Networks corresponding to $\sigma_{0}=1/\sqrt{2}$}
        \includegraphics[width=\linewidth]{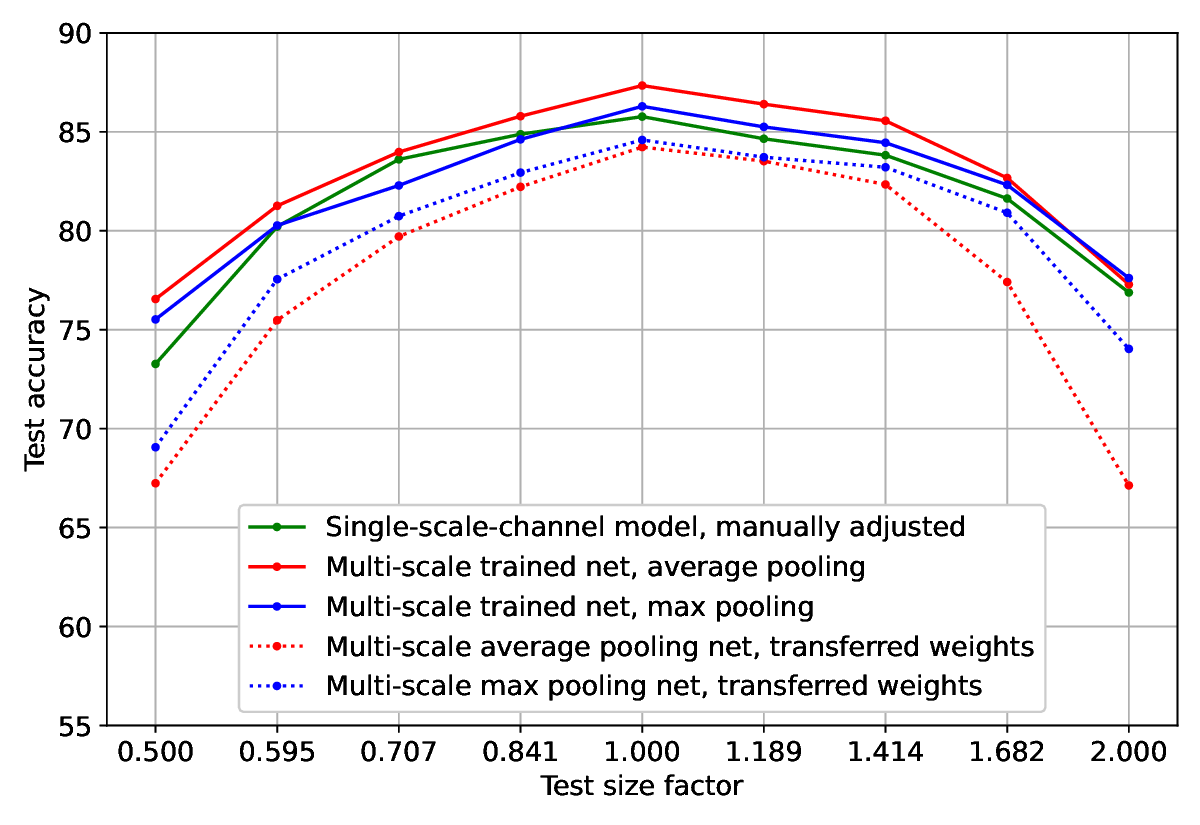}
        \label{fig:subfigb-cifar-0707}
    \end{subfigure}
    \hfill
    \begin{subfigure}{0.45\textwidth}
        \centering
        \textit{\hspace{0.6cm} Scale selection histogram for $\sigma_{0}=1/\sqrt{2}$}
        \includegraphics[width=\linewidth]{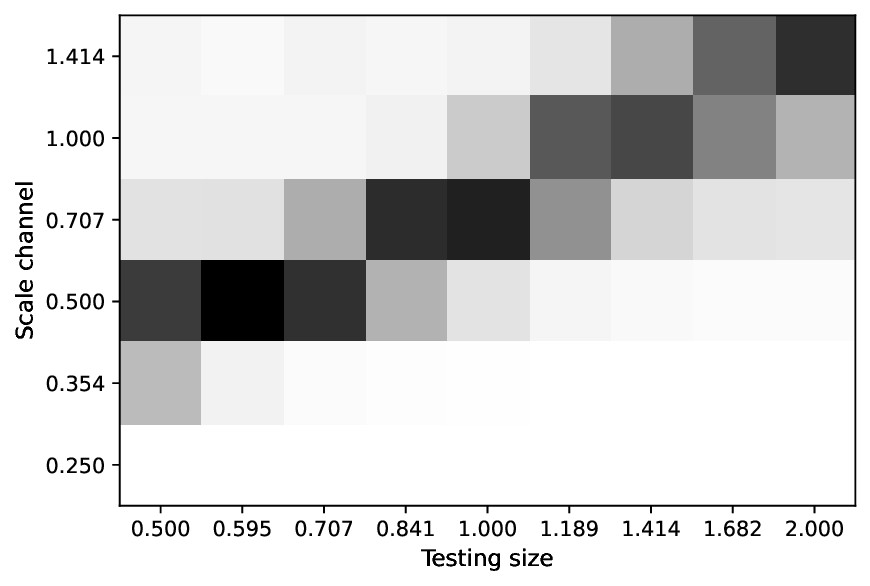}
        \label{fig:subfigb-cifar-0707-histogram}
    \end{subfigure}
    
    \begin{subfigure}{0.47\textwidth}
        \centering
        \textit{\hspace{0.6cm} Networks corresponding to $\sigma_{0}=1$}
        \includegraphics[width=\linewidth]{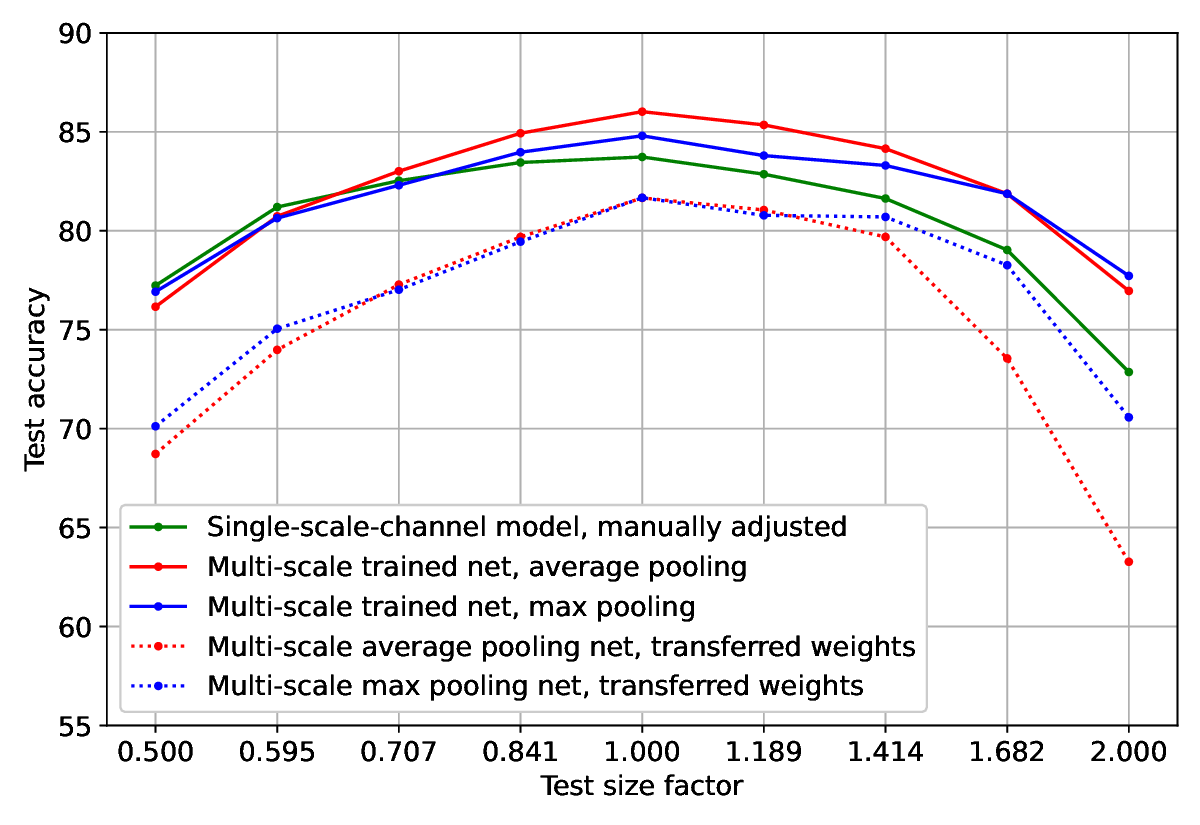}
        \label{fig:subfiga-cifar-1}
    \end{subfigure}
    \hfill
    \begin{subfigure}{0.45\textwidth}
        \centering
        \textit{\hspace{0.6cm} Scale selection histogram for $\sigma_{0}=1$}
        \includegraphics[width=\linewidth]{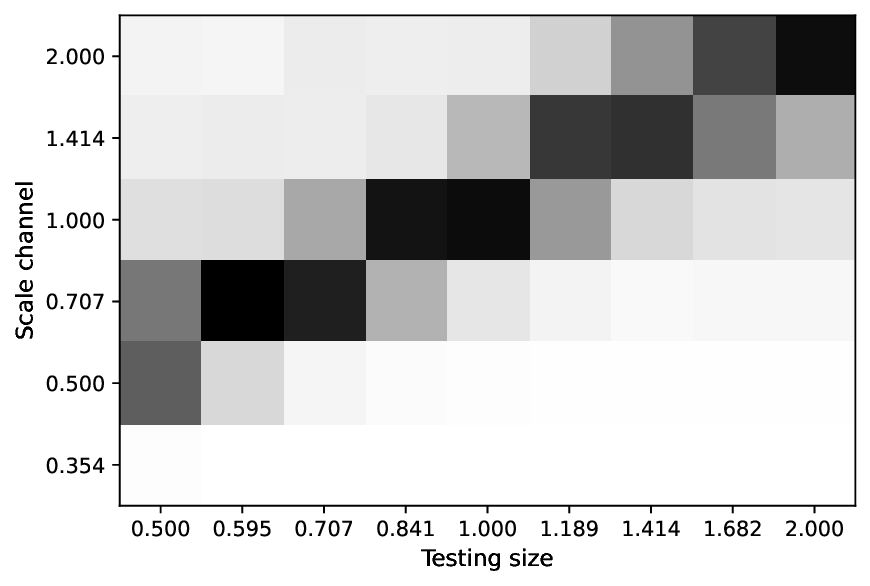}
        \label{fig:subfiga-cifar-1-histogram}
    \end{subfigure}
  \end{center}
\caption{Scale generalisation performance on the rescaled CIFAR-10 dataset, for different initial scale values $\sigma_0$ and different ways of defining the multi-scale-channel networks, by either (i)~genuine multi-scale-channel training on the training data for the training data for size factor 1 (the solid red and solid blue curves), or instead (ii)~constructing the multi-scale-channel networks by weight transfer from a single-scale-channel network trained on the training data for the same size factor 1 (the dashed red and the dashed blue curves). For comparison, we also show the result of applying (iii)~the single-scale-channel network with manually transformed scale parameters on the rescaled testing data for the same scaling factors, using a priori knowledge about the scaling transformation between the training data and the testing data. As can be seen from these graphs, genuine multi-scale-channel training leads to substantially better scale generalisation performance, compared to weight transfer from single-scale-channel learning. The single-scale-channel network with manually adjusted scale parameters (the solid green curves) does, however, lead to rather good performance under the spatial scaling transformations, in agreement with the provable scale-covariant properties of the underlying continuous Gaussian derivative network, that is approximated discretely in these implementations. Scale selection histograms for the multi-scale-channel networks using max pooling over scales with weights transferred from the corresponding single-scale-channel networks, represented as the dashed blue curve in the graphs, are shown on the right.}
\label{fig-cifar-covariance-experiments}
\end{figure*}


We did also construct two multi-scale-channel networks, using either max pooling over scale or average pooling over scale, by transferring the weights obtained from the training of the single-scale-channel network. For comparison, we did also construct two multi-scale-channel networks with identical model parameters, using either max pooling over scales or average pooling over scales, however, based on genuine multi-scale training of the networks. All the models were trained as described in Section~5 in the main paper, except for using different values of the initial scale values $\sigma_0$ and the relative scale factor $r$, as further specified in Tables~\ref{tab:adjusting-initial-scale-values-single-fashion} and~\ref{tab:adjusting-initial-scale-values-single-cifar}.

The results from these experiments are shown in Figures~\ref{fig-fashion-covariance-experiments} and~\ref{fig-cifar-covariance-experiments}, where all the networks were first trained on training data for the size factor 1, and then applied to testing data for all the size factors between 1/2 and 2 in the respective datasets. As can be seen from the figures, for both the rescaled Fashion-MNIST dataset and the rescaled CIFAR-10 dataset, the multi-scale-channel networks obtained by genuine multi-scale-channel training perform significantly better than the multi-scale-channel networks obtained by transferring the weights obtained by single-scale training to multi-scale-channel networks. 
Thus, we can conclude that multi-scale-channel training is really essential to obtain good scale generalisation properties for multi-scale-channel networks.

A possible explanation for this is that when a single-scale-channel network is trained on training data for a single size in the image domain, there is no feedback into the training process from the potential responses that would be obtained, if the single-scale-channel network is applied to testing data of a different size. Specifically, this implies that if the single-scale-channel would be trained to deliver the right classification class ”A” for training data belonging to the same class ”A”, while when exposed to testing data for a different size, there is no mechanism that would prevent the network from learning weights that would result in a wrong classification class ”B” for rescaled testing data belonging to the class ”A”.

This view could also be supported from the observation that for the networks obtained by transfer of a single-scale-channel network, the transferred multi-scale-channel networks based on max pooling over the scale channels often perform significantly better than the transferred multi-scale-channel networks based on average pooling over the scale channels. A possible explanation for this, is that when a multi-scale-channel network with average pooling over scales is constructed by mere training of a single-scale-network, it has not during the training process had any explicit feedback on how to combine the information from the different scale channels, while the multi-scale-channel network based on average pooling over scales with genuine multi-scale-training has had the possibility to learn how to combine the information from the different scale channels in the training process.

In a comparison of Riesz-transform-based networks to GaussDerNets, Barisin et al. (\citeyear{BarSchRed24-JMIV}) compare to a scale-transferred single-scale-channel GaussDerNet (see Section~4.2.2 in Barisin et al. (\citeyear{BarSchRed24-JMIV})). In view of the results presented here, it seems plausible that the performance of the GaussDerNets in that comparison could be improved by instead using multi-scale-channel GaussDerNets with genuine multi-scale training.

When applying the single-scale-channel network with manually adjusted initial scale value $S \, \sigma_0$ to rescaled testing data for a matching spatial scale factor $S$, the manually adjusted single-scale-channel network does perform quite well, in agreement with the underlying, in the continuous case, provable scale-covariant properties of the GaussDerNets, as described in Section~2.4 in the main paper. However, when transferring these weights to multi-scale-channel models, a slight drop in scale generalisation can be seen, compared to manually adjusted single-scale-channel networks, suggesting that the scale selection properties in the multi-scale-channel networks are affected by the discretisation of the model architecture, as caused by the use of a finite amount of scale channels in the model.

The scale selection histograms in Figures~\ref{fig-fashion-covariance-experiments} and~\ref{fig-cifar-covariance-experiments}, which correspond to multi-scale-channel networks using max pooling over scales with weights transferred from corresponding single-scale-channel networks, show a clear linear trend for all the models, demonstrating that the weight transfer does preserve reasonable scale selection properties. However, certain approximation issues begin to appear for scale channels using initial scale values $\sigma_{i,0} \leq 1/(2\sqrt{2})$, which can be seen in the corresponding scale generalisation graphs, with scale covariance not holding fully for smaller size factors. This manifests itself in the scale selection histograms by the scale channels using initial scale values below this threshold being avoided by the scale selection mechanism; the neighbouring scale channels covering a range of several size factors instead.
The scale generalisation curves for the networks corresponding to $\sigma_{0}=1/2$ in Figure~\ref{fig-cifar-covariance-experiments} show how in extreme cases this can lead to unusual phenomena; as we can see from the graph that the multi-scale-channel model with transferred weights outperformed the single-scale-channel network on size factor 1/2 test data, due to the scale selection mechanism compensating for the approximation issues by making use of coarser scale channels.

Finally, in connection with the aforementioned observations, from Figures~\ref{fig-fashion-covariance-experiments} and~\ref{fig-cifar-covariance-experiments}, we can see that using smaller values of the initial scale value $\sigma_0$ and therefore larger values of the relative scale ratio $r$ results in a drop in performance on the smaller size factors, while being beneficial for the medium and the large size factors. This need for balancing between generalisation on small and large size factors has been consistently noted in our experiments, with the choice of the $\sigma_0\text{-}$ and $r\text{-}$values being important in achieving successful generalisation over the entire scale range.

\subsection{Analysis of the scale levels learned by the scale learning methodology}
\label{sec-scale-levels-analysis-appendix}
In this section, which builds on the previous treatment about learning of the scale parameters experiments in Section~7.5 in the main paper, we will analyse the scale levels that are learned by the scale learning methodology, for the different methods for discretising the Gaussian derivative operators in the GaussDerNets. We will specifically also compare these scale values in terms of the spatial extent for the discrete kernels that they give rise to, for the corresponding discretisation methods.

\begin{table*}[htbp]
  \centering
  \begin{tabular}{lccc}
       \hline
       \textbf{Type of kernel} & $\eta_{\sigma} = 0.0001$ & $\eta_{\sigma} = 0.001$ & $\eta_{\sigma} = 0.01$ \\ 
       \hline 
       Sampled Gaussian derivatives & $0.659 \pm~0.003$ & $0.660 \pm~0.003$ & $0.664 \pm~0.002$ \\ 
       Integrated Gaussian derivatives & $0.524 \pm~0.008$ & $0.422 \pm~0.005$ & $0.397 \pm~0.008$ \\
       Sampled Gaussian derivatives + no BN final layer & $0.659 \pm~0.002$ & $0.660 \pm~0.004$ & $0.656 \pm~0.008$ \\
       Integrated Gaussian derivatives + no BN final layer & $0.524 \pm~0.006$ & $0.396 \pm~0.003$ & $0.381 \pm~0.010$ \\
       \noalign{\vskip 0.8mm}
       \hline
       \noalign{\vskip 0.9mm}
       Sampled Gaussian + central differences & $0.467 \pm~0.008$ & $0.390 \pm~0.009$ & $0.372 \pm~0.012$ \\
       Integrated Gaussian + central differences & $0.404 \pm~0.010$ & $0.308 \pm~0.007$ & $0.298 \pm~0.004$ \\
       Sampled Gaussian + central differences and no BN final layer & $0.469 \pm~0.004$ & $0.384 \pm~0.002$ & $0.370 \pm~0.007$ \\
       Integrated Gaussian + central differences and no BN final layer & $0.407 \pm~0.003$ & $0.306 \pm~0.007$ & $0.301 \pm~0.006$ \\ \hline
  \end{tabular}
  \caption{Mean and unbiased standard deviation of the learned scale parameter value $\sigma_{1}$ in the first layer, over 5 runs, obtained with {\em learning of the scale-parameters\/} for the different layers in single-scale-channel networks, applied to the {\em regular\/} Fashion-MNIST dataset (with image size $28 \times 28$ pixels). Here, the networks have been trained for different learning rates $\eta_{\sigma} \in \{0.0001, 0.001, 0.01 \}$ for the scale parameters $\sigma_k$ in the layers, while keeping the learning rate for the weights $C_0$, $C_x$, $C_y$, $C_{xx}$, $C_{xy}$ and $C_{xx}$ fixed to $\eta_C = 0.01$. The scale parameters $\sigma_k$ in the layers have, in turn, been initialised to a similar geometric distribution according to Equation~(8) in the main paper, for $\sigma_0 = 1$ and $r = 1.28$, similar to values used for corresponding fixed-scale networks.}
  \label{tab:sigma_learning_learned_intial_values_fashion}
\end{table*}

\begin{table*}[htbp]
  \centering
  \begin{tabular}{lccc}
       \hline
       \textbf{Type of kernel} & $\eta_{\sigma} = 0.0001$ & $\eta_{\sigma} = 0.001$ & $\eta_{\sigma} = 0.01$ \\ 
       \hline 
       Sampled Gaussian derivatives & $0.505 \pm~0.006$ & $0.514 \pm~0.004$ & $0.513 \pm~0.002$ \\ 
       Integrated Gaussian derivatives & $0.339 \pm~0.004$ & $0.364 \pm~0.028$ & $0.394 \pm~0.004$ \\
       Sampled Gaussian derivatives + no BN final layer & $0.541 \pm~0.008$ & $0.511 \pm~0.002$ & $0.497 \pm~0.001$ \\
       Integrated Gaussian derivatives + no BN final layer & $0.336 \pm~0.003$ & $0.375 \pm~0.006$ & $0.390^{*} \pm~0.005$ \\
       \noalign{\vskip 0.8mm}
       \hline
       \noalign{\vskip 0.9mm}
       Sampled Gaussian + central differences & $0.304 \pm~0.002$ & $0.302 \pm~0.005$ & $0.325 \pm~0.019$ \\
       Integrated Gaussian + central differences & $0.214 \pm~0.004$ & $0.206 \pm~0.009$ & $0.227 \pm~0.021$ \\
       Sampled Gaussian + central differences and no BN final layer & $0.304 \pm~0.001$ & $0.292 \pm~0.005$ & $0.338 \pm~0.026$\\
       Integrated Gaussian + central differences and no BN final layer & $0.212 \pm~0.002$ & $0.207 \pm~0.004$ & $0.246 \pm~0.008$ \\ \hline
  \end{tabular}
  \caption{Mean and unbiased standard deviation of the learned scale parameter value $\sigma_{1}$ in the first layer, over 5 runs, obtained with {\em learning of the scale-parameters\/} for the different layers in single-scale-channel networks, applied to the {\em regular\/} CIFAR-10 dataset (with image size $32 \times 32$ pixels). Here, the networks have been trained for different learning rates $\eta_{\sigma} \in \{0.0001, 0.001, 0.01 \}$ for the scale parameters $\sigma_k$ in the layers, while keeping the learning rate for the weights $C_0$, $C_x$, $C_y$, $C_{xx}$, $C_{xy}$ and $C_{xx}$ fixed to $\eta_C = 0.01$. The scale parameters $\sigma_k$ in the layers have, in turn, been initialised to a similar geometric distribution according to Equation~(8) in the main paper, for $\sigma_0 = 1$ and $r = 1.4$, similar to values used for corresponding fixed-scale networks. The result marked with a * has been initialised using $r = 1.48$ instead, and had the final layer scale parameter fixed during training, to prevent the corresponding kernel from exceeding the image size.}
  \label{tab:sigma_learning_learned_intial_values_cifar}
\end{table*}

\subsubsection{Learned scale values for the regular Fashion-MNIST and CIFAR-10 datasets}
Tables~\ref{tab:sigma_learning_learned_intial_values_fashion} and~\ref{tab:sigma_learning_learned_intial_values_cifar} give an overview of the scale values that are learned in the first layer, when applying learning of the scale values to a single-scale-channel GaussDerNet to the regular Fashion-MNIST and CIFAR-10 datasets. The results are shown as the average value of the first scale value $\sigma_{1}$ over 5 runs, accompanied by the unbiased standard deviation. The scale parameter in the first layer was chosen for this comparison, because any difference in the learned distribution of the scale parameters between the various discretisation methods would be most evident at fine scales, due to the approximation effects described in Section~3.6 in the main paper.

As can be seen from these results, the learned values of the first scale parameter $\sigma_{1}$ tend to converge to very fine scale levels, especially with finer scale levels obtained for the CIFAR-10 dataset than for the Fashion-MNIST dataset. We can also note that the learned scale values differ significantly, depending on which discretisation method is used for the Gaussian derivative operators.

With regard to the different types of discretisation methods for the Gaussian derivative operators, we can, in particular, note the following general tendencies:
\begin{itemize}
\item
  The methods based on pure sampled Gaussian derivatives lead to selections of comparably coarser scale values. This can be understood from the analysis in (Lindeberg \citeyear{Lin24-JMIV-discgaussder} Sections~3.3 and~3.8.2), which shows that sampled Gaussian derivative kernels may have a more narrow shape than as specified by the scale parameter. In this case, that narrowness property of sampled Gaussian derivative kernels at very fine scales appears to drive the scale learning algorithm to compensate for the narrowness property, by choosing somewhat larger values of the scale parameter for the methods based on sampled Gaussian derivative kernels.
\item
The methods based on pure integrated Gaussian derivatives lead to selections of comparably finer scale values. This can be understood from the analysis in (Lindeberg \citeyear{Lin24-JMIV-discgaussder} Sections~3.4 and~3.8.2), which shows that the box integration step in the definition of the discrete integrated Gaussian derivative kernels from the continuous Gaussian derivatives introduces an additional amount of spatial smoothing. In the case of scale learning, the scale learning method appears to compensate for this property, by choosing lower values of the scale parameter for the methods based on integrated Gaussian derivative kernels.
\item
The hybrid discretisation methods, based on using either smoothing with the sampled Gaussian kernel or the integrated Gaussian kernel followed by the application of central difference operators, lead to selections of finer scale values than the corresponding non-hybrid methods based on convolutions with either sampled Gaussian derivative kernels or integrated Gaussian derivative kernels. 

This can be understood from the analysis in (Lindeberg \citeyear{Lin25-FrontSignProc} Section~3), which shows that spatial spread measures computed from the equivalent hybrid discretisation kernels lead to larger values of the measure of the spatial extent of the kernels than for the corresponding non-hybrid methods. In the case of scale learning, the method appears to compensate for this property, by choosing lower values of the scale parameter for the methods based on the hybrid discretisation methods than for the non-hybrid discretisation methods. 

Additionally, the selected values of the scale parameter are higher for the hybrid methods based on convolutions with sampled Gaussian kernels as the spatial smoothing step, than for the hybrid methods based on convolutions with integrated Gaussian kernels as the spatial smoothing step.
This relationship can be understood from the analysis in (Lindeberg \citeyear{Lin24-JMIV-discgaussder} Sections~2.3-2.5 and~2.7), which shows that the sampled Gaussian kernels may have too narrow shape for small values of the scale parameter, while the box integration step in the definition of the integrated Gaussian kernels from the continuous Gaussian kernel introduces an additional amount of spatial smoothing. In the case of scale learning, the scale learning method appears to compensate for these phenomena, by choosing larger values of the scale parameter for the hybrid methods based on convolutions with sampled Gaussian kernels than for the hybrid methods based on convolutions with integrated Gaussian kernels.
\end{itemize}
In these ways, we can thus from the theoretical analysis of the different discretisation methods for Gaussian derivative kernels in (Lindeberg \citeyear{Lin24-JMIV-discgaussder, Lin25-FrontSignProc}) explain why we obtain the observed differences in the learned scale values for the different discretisation methods.

Additionally, from the learned scale parameter values in Tables~\ref{tab:sigma_learning_learned_intial_values_fashion} and~\ref{tab:sigma_learning_learned_intial_values_cifar}, we can see that for many of the discretisation methods, the learned values of $\sigma_1$ may also vary depending on the choice of the learning rate $\eta_{\sigma}$ of the scale parameter. For the Fashion-MNIST dataset, the difference in the learned values of $\sigma_1$ is particularly large for the networks trained with the smallest learning rate, $\eta_{\sigma} = 0.0001$, with the exception of the approach with sampled Gaussian derivatives, for which there is very little variation. For the CIFAR-10 dataset, the learned values of $\sigma_1$ are considerably less dependent on the choice of the learning rate of the scale parameter.

For the Fashion-MNIST dataset, the differences between the learned scale values are larger than for the CIFAR-10 data. A possible explanation for this could be that the Fashion-MNIST dataset contains much fewer textures, which may then lead to a lower sensitivity to the choice of the scale levels, which is why different choices of the learning rate $\eta_{\sigma}$ of the scale parameter may lead the network to converge to different sets of scale parameters, because the feedback force with regard to the choice of scale levels could then be significantly weaker for the Fashion-MNIST dataset than for the CIFAR-10 dataset.

\begin{table*}[htbp]
  \centering
  \begin{tabular}{l|cc|cc|cc}
       \hline
         & \multicolumn{2}{|c|}{$\eta_{\sigma} = 0.0001$} & \multicolumn{2}{|c|}{$\eta_{\sigma} = 0.001$} & \multicolumn{2}{|c}{$\eta_{\sigma} = 0.01$} \\ 
        \textbf{Type of kernel} & ${\cal S}_1$ & ${\cal S}_2$ & ${\cal S}_1$ & ${\cal S}_2$ & ${\cal S}_1$ & ${\cal S}_2$ \\
       \hline
       Sampled Gaussian derivatives & 1.087 & 0.866 & 1.087 & 0.868 & 1.091 & 0.874 \\ 
       Integrated Gaussian derivatives & 1.039 & 0.786 & 1.005 & 0.719 & 1.003 & 0.713 \\
       Sampled Gaussian derivatives + no BN final layer & 1.087 & 0.866 & 1.087 & 0.868 & 1.084 & 0.861 \\
       Integrated Gaussian derivatives + no BN final layer & 1.039 & 0.786 & 1.003 & 0.712 & 1.002 & 0.710 \\
       \noalign{\vskip 0.8mm}
       \hline
       \noalign{\vskip 0.9mm}
       Sampled Gaussian + central differences & 1.130 & 0.818 & 1.053 & 0.747 & 1.039 & 0.736 \\
       Integrated Gaussian + central differences & 1.168 & 0.860 & 1.080 & 0.770 & 1.071 & 0.763 \\
       Sampled Gaussian + central differences and no BN final layer & 1.132 & 0.820 & 1.048 & 0.743 & 1.037 & 0.735 \\
       Integrated Gaussian + central differences and no BN final layer & 1.171 & 0.863 & 1.078 & 0.768 & 1.073 & 0.765 \\ \hline
  \end{tabular}
  \caption{Spatial spread measures for the different discrete approximations of the first- and second-order Gaussian derivative kernels, computed for the mean over 5 runs of the learned scale parameter value $\sigma_{1}$ in the first layer from Table~\ref{tab:sigma_learning_learned_intial_values_fashion}, obtained with {\em learning of the scale-parameters\/} for the different layers in single-scale-channel networks, applied to the {\em regular\/} Fashion-MNIST dataset (with image size $28 \times 28$ pixels). Here, the networks have been trained for different learning rates $\eta_{\sigma} \in \{0.0001, 0.001, 0.01 \}$ for the scale parameters $\sigma_k$ in the layers, while keeping the learning rate for the weights $C_0$, $C_x$, $C_y$, $C_{xx}$, $C_{xy}$ and $C_{xx}$ fixed to $\eta_C = 0.01$. The scale parameters $\sigma_k$ in the layers have, in turn, been initialised to a similar geometric distribution according to Equation~(8) in the main paper, for $\sigma_0 = 1$ and $r = 1.28$, similar to values used for corresponding fixed-scale networks.}
  \label{tab:sigma_learning_both_order_spread_fashion}
\end{table*}

\begin{table*}[htbp]
  \centering
  \begin{tabular}{l|cc|cc|cc}
       \hline
         & \multicolumn{2}{|c|}{$\eta_{\sigma} = 0.0001$} & \multicolumn{2}{|c|}{$\eta_{\sigma} = 0.001$} & \multicolumn{2}{|c}{$\eta_{\sigma} = 0.01$} \\ 
        \textbf{Type of kernel} & ${\cal S}_1$ & ${\cal S}_2$ & ${\cal S}_1$ & ${\cal S}_2$ & ${\cal S}_1$ & ${\cal S}_2$ \\
       \hline
       Sampled Gaussian derivatives & 1.008 & 0.688 & 1.010 & 0.696 & 1.010 & 0.695 \\ 
       Integrated Gaussian derivatives & 1.000 & 0.708 & 1.001 & 0.709 & 1.002 & 0.712 \\
       Sampled Gaussian derivatives + no BN final layer & 1.018 & 0.719 & 1.010 & 0.693 & 1.007 & 0.681 \\
       Integrated Gaussian derivatives + no BN final layer & 1.000 & 0.708 & 1.001 & 0.710 & $1.002^{*}$ & $0.712^{*}$ \\
       \noalign{\vskip 0.8mm}
       \hline
       \noalign{\vskip 0.9mm}
       Sampled Gaussian + central differences & 1.007 & 0.712 & 1.006 & 0.712 & 1.013 & 0.717 \\
       Integrated Gaussian + central differences & 1.015 & 0.718 & 1.011 & 0.715 & 1.021 & 0.722 \\
       Sampled Gaussian + central differences and no BN final layer & 1.007 & 0.712 & 1.004 & 0.710 & 1.019 & 0.721\\
       Integrated Gaussian + central differences and no BN final layer & 1.014 & 0.717 & 1.012 & 0.716 & 1.032 & 0.731 \\ \hline
  \end{tabular}
  \caption{Spatial spread measures for the different discrete approximations of the first- and second-order Gaussian derivative kernels, computed for the mean over 5 runs of the learned scale parameter value $\sigma_{1}$ in the first layer from Table~\ref{tab:sigma_learning_learned_intial_values_cifar}, obtained with {\em learning of the scale-parameters\/} for the different layers in single-scale-channel networks, applied to the {\em regular\/} CIFAR-10 dataset (with image size $32 \times 32$ pixels). Here, the networks have been trained for different learning rates $\eta_{\sigma} \in \{0.0001, 0.001, 0.01 \}$ for the scale parameters $\sigma_k$ in the layers, while keeping the learning rate for the weights $C_0$, $C_x$, $C_y$, $C_{xx}$, $C_{xy}$ and $C_{xx}$ fixed to $\eta_C = 0.01$. The scale parameters $\sigma_k$ in the layers have, in turn, been initialised to a similar geometric distribution according to Equation~(8) in the main paper, for $\sigma_0 = 1$ and $r = 1.4$, similar to values used for corresponding fixed-scale networks. The result marked with a * has been initialised using $r = 1.48$ instead, and had the final layer scale parameter fixed during training, to prevent the corresponding kernel from exceeding the image size.}
  \label{tab:sigma_learning_both_order_spread_cifar}
\end{table*}

\subsubsection{Spatial spread measures for the discrete derivative approximation kernels for the learned scale parameter values}
\label{sec-spatial-spread-measure-comparison-appendix}
Beyond analysing the actual values of the learned scale parameters for the different discretisation methods for the Gaussian derivative operators, given that the different discretisation methods may have significantly different properties at very fine scale levels, it is useful to additionally analyse the spatial extent of the discrete kernels.

In (Lindeberg \citeyear{Lin24-JMIV-discgaussder}), the spatial spread of a discrete approximation $T_{x^{\alpha}}(n;\; s)$ of a Gaussian derivative operator 
$g_{x^{\alpha}}(x;\; s)$ is measured by the standard deviation of the absolute value of the discrete derivative approximation kernel for differentiation order $\alpha$ according to
(Lindeberg \citeyear{Lin24-JMIV-discgaussder} Equation~(52))
\begin{equation}
  {\cal S}_{\alpha} = \sqrt{V(|T_{x^{\alpha}}(\cdot;\; s)|)}
\end{equation}
where the discrete variance of the kernel is defined as
\begin{multline}
  V(|T_{x^{\alpha}}(\cdot;\; s)|) = \\
  = \frac{\sum_{n \in \bbbz} n^2 \, |T_{x^{\alpha}}(n;\; s)|}{\sum_{n \in \bbbz} |T_{x^{\alpha}}(n;\; s)|}
  - \left(
    \frac{\sum_{n \in \bbbz} n \, |T_{x^{\alpha}}(n;\; s)|}{\sum_{n \in \bbbz} |T_{x^{\alpha}}(n;\; s)|}
  \right)^2.
\end{multline}
Tables~\ref{tab:sigma_learning_both_order_spread_fashion} and~\ref{tab:sigma_learning_both_order_spread_cifar} show such spatial spread measures ${\cal S}_1$ and ${\cal S}_2$ for the first-order and second-order derivatives respectively, computed from the first layers in GaussDerNets with scale learning applied to the regular Fashion-MNIST and CIFAR-10 datasets. Again, the results are shown for the different choices of the discretisation methods for the Gaussian derivative operators, as well as for different learning rates of the scale parameter.

As can be seen from these tables, within each class of experiments for the regular Fashion-MNIST dataset vs.\ the regular CIFAR-10 dataset, the first-order spatial spread measures ${\cal S}_1$ as well as the second-order spatial spread measures ${\cal S}_2$ are rather similar between the different discretisation methods within each class of spread measure and within each type of dataset. For the Fashion-MNIST dataset, there are, however, some differences depending on the learning rate $\eta_{\sigma}$. In agreement with the previous results regarding the difference in the learned scale values for the Fashion-MNIST vs.\ the CIFAR-10 datasets, the spatial spread measures are, however, somewhat lower for the CIFAR-10 dataset than for the Fashion-MNIST dataset.

Notably, the first-order spread measures are somewhat more similar than the second-order spread measures. The first-order spread measures for the CIFAR-10 dataset are indeed very similar. The second-order spatial spread measures for the CIFAR-10 dataset are also very similar, with the exception of the pure sampled Gaussian derivative kernels.
In comparison, the spread measures differ more between the different discretisation methods for the Fashion-MNIST dataset, possibly with a coupling to the previously obtained results regarding scale learning, for which the results appear to be less sensitive to the choice of scale levels, which in turn may cause a lower feedback force to choose the same effective scale levels for the GaussDerNets applied to the Fashion-MNIST dataset than for the GaussDerNets applied to the CIFAR-10 dataset.

When interpreting these results, it should be noted that for all the hybrid discretisation methods, the first-order spread measure is delimited from below by the value ${\cal S}_{1,\min} = \sqrt{V(|\delta_x|)} = 1$, while the second-order spread measure is delimited from below by the value ${\cal S}_{2,\min} = \sqrt{V(|\delta_{xx}|)} = 1/\sqrt{2}$. Additionally, the spread measure for the first-order sampled Gaussian derivatives and the first-order integrated Gaussian derivatives are also delimited from below by the value ${\cal S}_{1,\min} = \sqrt{V(|\delta_x|)} = 1$. The spread measure for the second-order integrated Gaussian derivative kernel is also delimited from below by the value ${\cal S}_{2,\min} = \sqrt{V(|\delta_{xx}|)} = 1/\sqrt{2}$, while for the second-order sampled Gaussian derivative kernel there is no such lower bound on its spatial spread measure towards lower scales. In this respect, the second-order sampled Gaussian derivative approach differs fundamentally from the other discretisation methods. 
As can be seen from the spatial spread measures in Tables~\ref{tab:sigma_learning_both_order_spread_fashion} and~\ref{tab:sigma_learning_both_order_spread_cifar}, all the experimentally obtained spread measures are in agreement with these lower bounds, except for the second-order sampled Gaussian derivative kernels, for which, as previously explained, there is no lower bound on their spatial spread measure.

To summarise, when interpreting these results concerning the spatial spread measures computed using the different discretisation approaches for the Gaussian derivative operators, we should thus, be careful to not draw too strong implications from the comparably larger variability in the spatial spread measures for the Fashion-MNIST dataset. This dataset is comparably simpler, with a very low amount of surface texture, which is why the influence of the spatial smoothing operations is comparably lower for this dataset, than for more natural images that may contain a substantially larger amount of surface textures. For the CIFAR-10 dataset, where the influence of the spatial smoothing operations in the receptive fields may have a substantial effect on the accuracy of the classification results, we can, however, see that by integrating a scale learning mechanism into the GaussDerNets, the networks appear to converge to learned scale values that lead to comparably similar amounts of spatial smoothing in the receptive fields, as quantitatively measured in terms of the spatial spread measures.

\subsection{Statistical analysis of the performance comparisons between the different discretisation methods}
\label{sec-$p$-values-comparison-appendix}

In this section, which builds on the previous treatments about experiments with different discrete approximations of Gaussian derivative kernels in Section~7.4 in the main paper and learning of the scale parameters experiments in Section~7.5 in the main paper, we will statistically compare the differences in accuracies of GaussDerNets for the different discretisation methods for the Gaussian derivative operators, with either fixed or learned scale parameters, across the same 5 training runs for each of the 3 parameter settings, as summarised in Tables~6 and~8 in the main paper.

Given these test accuracies and the corresponding unbiased standard deviations for the networks for the different discretisation methods, we will first perform pairwise statistical comparisons to get a rough idea of the degree to which the differences in performance could generally be considered significant; the aim not being to define definite rigorous statistical differences. Despite the limitation of the training being done for only 5 runs per discretisation method, in order to limit the training cost, we could still ask what happens when statistical tests are applied to these accuracies, and draw conclusions about which design options of the GaussDerNets could be regarded as better. 

Next, we will perform multi-comparison statistical tests, with the aim to provide a somewhat more rigorous and conservative estimate of the degree to which the differences in performance, across all the different parameter settings, could generally be considered significant.

Our analysis will focus only on the CIFAR-10 dataset results here, since it is the dataset with properties closest to natural image data among the two larger types of datasets considered in this work, and due to results on the Fashion-MNIST dataset showing lower variation in test accuracy between the different discretisation methods and the experimental settings.

\subsubsection{Statistical measures for comparing the different types of discretisation methods}
The notion of effect size can be defined as the magnitude of the difference or relationship observed between different groups, and is used to complement the statistical hypothesis testing, in order to quantify the size of a given significant effect, see Sullivan and Feinn (\citeyear{SullFein2012-JGME}) for a detailed overview. To give a coarse estimate of how large the effect size is between the performance of two methods, one can simply measure the mean difference in accuracy $m = \bar{x} - \bar{y}$ relative to the corresponding standard deviation of the data. A standard way to do this is in terms of Cohen’s d-value, see Cohen (\citeyear{Cohen1988-book}), usually expressed as
\begin{equation}
  d = \frac{m}{\sigma^{*}} = \frac{m}{\sqrt{\frac{(n_1 - 1)\sigma_1^2 + (n_2 - 1)\sigma_2^2}{n_1 + n_2 - 2}}},
\end{equation}
where $\sigma^{*}$ is the pooled standard deviation of the two samples, with $n$ representing the sample size, and $\sigma$ representing the standard deviation of the sample.

When the sample size is small, a more appropriate alternative can, however, be to instead use the Hedge's g-value proposed in Hedges (\citeyear{Hedges1981-JES}), defined as a bias-corrected\footnote{When using $n_1 = n_2 = 5$ samples, as will be the case for
the individual pairwise comparisons of different discretisation methods below,
the numerical value for this correction factor is $g/d \approx 0.903$,
whereas when using $n_1 = n_2 = 15$ samples, as will be the case
for the pooled pairwise comparisons of different discretisation methods below,
the numerical value for this correction factor is $g/d \approx 0.973$.} Cohen’s d-value according to 
\begin{equation}
  \label{eq:hedges-g-definition}
  g = d \left( \frac{\Gamma\left(\frac{\alpha}{2} \right)}{\sqrt{\frac{\alpha}{2}} \cdot \Gamma\left(\frac{\alpha-1}{2})\right)} \right) \approx d \left(1 - \frac{3}{4\alpha - 1}\right),
\end{equation}
where $\Gamma$ is the Gamma function and $\alpha = n_{1} + n_{2} - 2$. A Hedge's g-value of 0.8 is typically considered to signify a large effect size. In the context of only 5 samples with a large difference in means relative to variability, this is, however, not a clear-cut rule, and much larger effect size values become more common in such cases.

Hedge's g-value does, however, not provide the actual significance level of the effect. For this purpose, in order to pairwise statistically compare the differences in accuracies of GaussDerNets for the different discretisation methods for the Gaussian derivative operators, we make use of the non-parametric Wilcoxon rank-sum test, also known as the Mann–Whitney U test, introduced in Mann and Whitney (\citeyear{MannWhit47-TAMS}). 

This test was chosen, because we consider networks based on different discretisation methods as distinct models, each using separate weight initialisations, and we therefore assume that the training runs are independent, and not paired, although all training runs are using the same training data. The Wilcoxon rank-sum test tests the null hypothesis that for randomly selected values from given two populations, the samples are from continuous distributions with equal medians.\footnote{It should be noted, that for very low number of samples, which is the case for our comparisons, the $p$-values for the Wilcoxon rank-sum test are quantised, which can also lead to a small, but non-zero lower bound on the $p$-value.}

To give an estimate of what $p$-value we would get if we would assume that the performance values would be normally distributed under variations of the random seed for the training stage, we will also perform the two-sample t-test. This test evaluates the null hypothesis that the data comes from two independent random samples from normal distributions that have equal means, against the alternative that they are unequal. However, with so few samples, there is a possibility of getting somewhat skewed conclusions following such an approach, which is why having a non-parametric test like the Wilcoxon rank-sum test as an additional comparison approach is desirable. For a detailed analysis of the validity and interpretation of using either the t-test or the Wilcoxon rank-sum test under different sets of assumptions, see Fay and Proschan (\citeyear{FayPro10-SS}).

\begin{table*}[htbp]
  \centering
  \begin{tabular}{l|ccc|ccc|ccc}
       \hline
         & \multicolumn{3}{|c|}{$\sigma_0 = 1/2$} & \multicolumn{3}{|c|}{$\sigma_0 = 1/\sqrt{2}$} & \multicolumn{3}{|c}{$\sigma_0 = 1$} \\ 
        \textbf{Compared kernels} & Hedge & Wilcoxon & t-test & Hedge & Wilcoxon & t-test & Hedge & Wilcoxon & t-test \\
       \hline
       discgaussdiff vs. samplgaussder & -0.69  &  0.103  &  0.259  &  \textbf{2.42} &   \textbf{0.016}  &  \textbf{0.003}  &  1.40  &  0.151  & 0.040  \\ 
       discgaussdiff vs. samplgaussdiff & \textbf{3.69}  &  \textbf{0.008}   & \textbf{0.000}  &  \textbf{5.26}  &  \textbf{0.008}  &  \textbf{0.000}  &  \textbf{7.90}   & \textbf{0.008}   & \textbf{0.000} \\
       discgaussdiff vs. normsamplgaussdiff & \textbf{4.75}  &  \textbf{0.008}   & \textbf{0.000} &   \textbf{5.27} &   \textbf{0.008}  &  \textbf{0.000}  &  \textbf{7.94}  &  \textbf{0.008}  & \textbf{0.000} \\
       samplgaussder vs. samplgaussdiff & \textbf{4.07}  &  \textbf{0.008}  &  \textbf{0.000} &   \textbf{3.44}  & \textbf{0.008}  &  \textbf{0.000}  &  \textbf{2.64}  &  \textbf{0.008}  &  \textbf{0.002} \\
       samplgaussder vs. normsamplgaussdiff & \textbf{5.01}  &  \textbf{0.008}  & \textbf{0.000}  &  \textbf{3.32}  &  \textbf{0.008}   & \textbf{0.001}  &  \textbf{1.90} &   \textbf{0.008}  &  \textbf{0.011} \\
       normsamplgaussdiff vs. samplgaussdiff & 0.90  &  0.135  &  0.153 &  0.25 &   0.691  &  0.672 &  \textbf{1.86}  &  \textbf{0.016}  &  \textbf{0.012} \\ \hline
  \end{tabular}
  \caption{Statistical measures of the difference in performance of Gaussian derivative networks using different discrete derivative approximation methods with fixed scale parameters, across 5 runs, trained on the {\em regular\/} CIFAR-10 dataset. This is done for networks trained using different initial scale value settings $\sigma_0 \in \{1/2, 1/\sqrt{2}, 1\}$, with the relative scale ratios correspondingly set to be $r \in \{1.47, 1.37 ,1.28\}$. For each pairwise comparison, the table shows (i)~the Hedge's g effect size measure, (ii)~the $p$-value obtained from the Wilcoxon rank-sum test, and (iii)~the $p$-value obtained from the two-sample t-test. The large effect size values are due to the mean difference being substantially larger than the standard deviation, with little or no overlap between the populations. Significant values are highlighted in bold.}
  \label{tab:statistics_table_disc_cifar}
\end{table*}

\begin{table*}[htbp]
  \centering
  \begin{tabular}{lccc}
       \hline
       \textbf{Compared kernels} &  \textbf{Hedge's g pooled} & \textbf{Friedman Dunn-\v{S}id\'{a}k} & \textbf{ANOVA Tukey} \\
       \hline
       discgaussdiff vs. samplgaussder & 0.53 & 0.999 & 0.118  \\ 
       discgaussdiff vs. samplgaussdiff & \textbf{4.03} & \textbf{0.000} & \textbf{0.000}  \\ 
       discgaussdiff vs. normsamplgaussdiff & \textbf{3.91} & \textbf{0.001} & \textbf{0.000}  \\ 
       samplgaussder vs. samplgaussdiff & \textbf{2.15} & \textbf{0.000} & \textbf{0.000}  \\ 
       samplgaussder vs. normsamplgaussdiff & \textbf{2.33} & \textbf{0.007} & \textbf{0.000}  \\ 
       normsamplgaussdiff vs. samplgaussdiff & 0.55 & 0.535 & 0.160  \\  \hline
  \end{tabular}
  \caption{Statistical measures from multi-comparison methods, for comparing the differences in performance between Gaussian derivative networks using different discrete derivative approximation methods with fixed scale parameters, across 5 runs, trained on the {\em regular\/} CIFAR-10 dataset. This is done for networks trained using different initial scale value settings $\sigma_0 \in \{1/2, 1/\sqrt{2}, 1\}$, with the relative scale ratios correspondingly set to be $r \in \{1.47, 1.37 ,1.28\}$. For each comparison method, the results across the different parameter settings are pooled (15 runs in total). The comparison methods include (i)~the Hedge's g effect size measure, computed for the pairwise differences in performance, (ii)~the $p$-value for the Friedman test combined with the Dunn-\v{S}id\'{a}k's post hoc test, and (iii)~the $p$-value for the ANOVA with the Tukey post hoc test. Significant values are highlighted in bold.}
  \label{tab:statistics_table_disc_cifar_friedman}
\end{table*}

\subsubsection{Statistical tests for pooled comparisons over multiple parameter settings}
Comparing multiple networks based on different Gaussian derivative discretisation methods, by performing multiple pairwise comparison tests, does, however, not correct for the multiple testing problem. While this issue could in part be addressed by application of certain correction methods or requiring a stricter significance threshold in the previously proposed statistical tests, to remedy this, we will instead perform additional multi-comparison tests, to obtain a statistical comparison verdict that better accounts for multiple comparisons, to investigate if some performance trend exists in general across all parameter settings for a given experiment. 

Since we in our experiments only use five runs for each method, while also performing experiments for different parameter values, such as different initial scale values for the learning at fixed scale levels, or different learning rates for the learning of the scale levels, we will perform these complementary statistical analysis by pooling the results across the three different parameter settings, into a combined sample of size 15.

For that purpose, we will (i)~replace the use of the Hedge g-value by Hedge’s g-value computed over the samples from the different parameter settings pooled together, computed by applying Equation~(\ref{eq:hedges-g-definition}) to the difference in performances between corresponding runs, which, for simplicity, we now assume to be paired within their "subpopulations", with $\sigma^{*} = \sigma_{1}=\sigma_{2}$ being the standard deviation over all of these measurements. Additionally, we will (ii)~replace the Wilcoxon test by the non-parametric Friedman test (Friedman \citeyear{friedman1937-JASA, friedman1940-AMS}) with the Dunn-\v{S}id\'{a}k's post hoc test, and (iii)~replace the t-test by the two-way analysis of variance (ANOVA) (Fisher \citeyear{fisher1925-statistical, fisher-1935-statistical}) with the Tukey post hoc test.

The Friedman test is a multi-comparison test, that is used to find out if the column effects are all the same after adjusting for possible row effects, across multiple test attempts, which in our case means finding out if all the networks perform similarly across the different parameter settings. This test involves ranking the data within each block (row), which in our case represents data paired within the corresponding parameter settings, and tests for a difference across the columns.
The two-way ANOVA test could be considered to, in a way, correspond to the Friedman test, but being a parametric test that compares the means of columns and rows.

The Dunn-\v{S}id\'{a}k's test and the Tukey test are post hoc tests, designed to produce pairwise comparison results after a multiple-comparison test has been found to be significant. The Dunn-\v{S}id\'{a}k's and Tukey tests adjust for multiple comparisons to more rigorously control the family-wise error rate, making them rather conservative post hoc tests.

In our case, the GaussDerNets based on different discrete approximations of Gaussian derivative kernels are the treatments under study, representing the columns in the sample matrix, and the paired runs within the corresponding parameter settings for the networks are representing the rows in the sample matrix, and are considered as different experimental environments.

\begin{table*}[htbp]
  \centering
  \begin{tabular}{l|ccc|ccc|ccc}
       \hline
         & \multicolumn{3}{|c|}{$\eta_{\sigma} = 0.0001$} & \multicolumn{3}{|c|}{$\eta_{\sigma} = 0.001$} & \multicolumn{3}{|c}{$\eta_{\sigma} = 0.01$} \\ 
        \textbf{Compared kernels} & Hedge & Wilcoxon & t-test & Hedge & Wilcoxon & t-test & Hedge & Wilcoxon & t-test \\
       \hline
       samplgaussder vs.\ intgaussder & \textbf{4.07} & \textbf{0.008} & \textbf{0.000} & \textbf{2.74} & \textbf{0.008} &  \textbf{0.001}  & \textbf{2.34}  & \textbf{0.008}  & \textbf{0.004}  \\ 
       samplgaussder vs.\ samplgaussdiff & \textbf{3.00}  & \textbf{0.008}  & \textbf{0.001} &  \textbf{2.80} & \textbf{0.008}  & \textbf{0.001}  & \textbf{1.81}  & \textbf{0.032}  & \textbf{0.013} \\
       samplgaussder vs.\ intgaussdiff & \textbf{3.80}  & \textbf{0.008}  & \textbf{0.000}  & \textbf{2.21}  &  \textbf{0.016}  & \textbf{0.005}  & \textbf{1.65} & \textbf{0.032}  & \textbf{0.021} \\
       intgaussder vs.\ samplgaussdiff & 0.15 & 0.691 & 0.804 & -0.25 & 0.691 & 0.670  & -0.75  &  0.310  &  0.225 \\
       intgaussder vs.\ intgaussdiff & 0.85 & 0.071 &  0.176 & -0.32 & 0.691 &  0.590  & -0.53  &  0.691   & 0.382 \\
       samplgaussdiff vs.\ intgaussdiff & 0.53  &  0.413  &  0.381  & -0.11  &  1.000  &  0.848  &  0.09  &  0.691  &  0.882 \\ \hline
  \end{tabular}
  \caption{Statistical measures of the difference in performance of Gaussian derivative networks using different discrete derivative approximation methods and scale parameter learning, across 5 runs, trained on the {\em regular\/} CIFAR-10 dataset. This is done for networks trained using different learning rates $\eta_{\sigma} \in \{0.0001, 0.001, 0.01 \}$ for the scale parameters $\sigma_k$ in the layers. For each pairwise comparison, the table shows (i)~the Hedge's g effect size measure, (ii)~the $p$-value obtained from the Wilcoxon rank-sum test, and (iii)~the $p$-value obtained from the two-sample t-test. Only the networks with an architecture containing batch normalisation after the final layer are compared. The large effect size values are due to the mean difference being substantially larger than the standard deviation, with little or no overlap between the populations. Significant values are highlighted in bold.}
  \label{tab:statistics_table_siglr_cifar10}
\end{table*}

\begin{table*}[htbp]
  \centering
  \begin{tabular}{lccc}
       \hline
       \textbf{Compared kernels} &  \textbf{Hedge's g pooled} & \textbf{Friedman Dunn-\v{S}id\'{a}k} & \textbf{ANOVA Tukey} \\
       \hline
       samplgaussder vs.\ intgaussder & \textbf{2.98} & \textbf{0.000} & \textbf{0.000}  \\ 
       samplgaussder vs.\ samplgaussdiff & \textbf{1.78} & \textbf{0.004} & \textbf{0.000} \\
       samplgaussder vs.\ intgaussdiff & \textbf{1.87} & \textbf{0.001} & \textbf{0.000} \\
       intgaussder vs.\ samplgaussdiff & -0.20 & 0.952 & 0.882 \\
       intgaussder vs.\ intgaussdiff & -0.06 & 0.999 & 0.991 \\
       samplgaussdiff vs.\ intgaussdiff & 0.11 & 0.994 & 0.971 \\ \hline
  \end{tabular}
  \caption{Statistical measures from multi-comparison methods, for comparing the differences in performance between Gaussian derivative networks using different discrete derivative approximation methods with learned scale parameters, across 5 runs, trained on the {\em regular\/} CIFAR-10 dataset. This is done for networks trained using different learning rates $\eta_{\sigma} \in \{0.0001, 0.001, 0.01 \}$ for the scale parameters $\sigma_k$ in the layers. For each comparison method, the results across the different parameter settings are pooled (15 runs in total). The comparison methods include (i)~the Hedge's g effect size measure, computed for the pairwise differences in performance, (ii)~the $p$-value for the Friedman test combined with the Dunn-\v{S}id\'{a}k's post hoc test, and (iii)~the $p$-value for the ANOVA with the Tukey post hoc test. Only the networks with an architecture containing batch normalisation after the final layer are compared. Significant values are highlighted in bold.}
  \label{tab:statistics_table_siglr_cifar_friedman}
\end{table*}

\subsubsection{Statistical analysis of the experiments at fixed scales}
\label{sec-statistical-comparison-fixed-appendix}

Table~\ref{tab:statistics_table_disc_cifar} shows the result of comparing the performance between different selected pairs of discretisations methods at fixed scales,\footnote{The approaches with the integrated Gaussian kernel and the integrated Gaussian derivative operators involve a certain scale offset, that depends upon the scale level, while simultaneously the accuracy decreases with increasing scale levels over the considered scale interval, because of properties of the dataset. It is because of this property not fully appropriate to compare the integration-based discretisation methods, which imply a non-zero scale offset, to the other discretisation methods, that do not imply any scale offset.} for different individual settings of the initial scale value $\sigma_0$, and with the significance threshold for the statistical tests set to the conventional 5\%. As can be seen from these results:
\begin{itemize}
\item
The difference in performance between using the discrete analogue of the Gaussian kernel with central differences is, with the chosen bounds on the statistical tests, significantly better than using either the sampled Gaussian kernel or the normalised sampled Gaussian kernel combined with central differences, with a very large effect size. 
\item
The difference in performance between using the sampled Gaussian derivative kernel is, with the chosen bounds on the statistical tests, significantly better than using either the sampled Gaussian kernel or the normalised sampled Gaussian kernel combined with central differences, with a very large effect size.
\item
There is, however, not a generally valid statistical difference between the results of using the discrete analogue of the Gaussian kernel with central differences, compared to using sampled Gaussian derivative kernels.
\end{itemize}
Given that the discrete derivative approximations based on the discrete analogue of the Gaussian kernel with central differences can be computed much more efficiently than using sampled Gaussian derivative kernel, the approach with the discrete analogue of the Gaussian kernel therefore stands out as a very good implementation method when using fixed scale levels.

Table~\ref{tab:statistics_table_disc_cifar_friedman} shows corresponding results obtained by pooling the results for all the individual initial scale levels into a combined contrast. From these multiple comparisons, we can see that similar to the pairwise comparisons above, the discrete analogue of the Gaussian kernel with central differences and the sampled Gaussian derivative kernel approximations are found to perform much better than the hybrid sampled and normalised sampled approximation methods, with a high degree of statistical significance given the chosen bounds on the statistical tests, and with relatively large effect sizes.

\subsubsection{Statistical analysis of the experiments with learned scale levels}
\label{sec-statistical-comparison-learned-appendix}

When performing learning of the scale levels, we cannot, as previously explained, make use of the discrete analogue of the Gaussian kernel, because the underlying mathematical primitives in terms of the modified Bessel functions of integer order are not available as built-in primitives in PyTorch.
Since the learning-based approach for the scale levels can, however, be expected to compensate for the scale offsets in the discretisation approaches based on the integrated Gaussian kernel or the integrated Gaussian derivatives, we can, however, meaningfully include those discretisation approaches when comparing different discretisation approaches in a context of learning of the scale levels. We will, however, only compare test accuracies of networks that use an architecture with a batch normalisation layer after the final layer.

Table~\ref{tab:statistics_table_siglr_cifar10} shows a selection of pairwise comparisons with different approaches for learning of the scale levels. As can be seen from these results:
\begin{itemize}
\item
The approach based on sampled Gaussian derivative operators leads to significantly better results than the approach based on integrated Gaussian derivatives, with a large effect size.
\item
The approach based on sampled Gaussian derivative operators leads to significantly better results than either of the approaches of first smoothing with the sampled Gaussian kernel or the integrated Gaussian kernel, and then applying central difference operators. The effect size of this difference is large.
\end{itemize}
Since the approach based on the sampled Gaussian derivative operators implies significantly much more computational work compared to first smoothing with the sampled Gaussian kernel or the integrated Gaussian kernel and then applying central difference operators, the choice of discretisation method does in this respect imply a trade-off problem between requirements of accuracy and computational efficiency.

Table~\ref{tab:statistics_table_siglr_cifar_friedman} shows the results of corresponding comparisons of test accuracies pooled across all the different settings of the learning rates $\eta_{\sigma}$ for the scale parameters. As can be seen from the results, the differences in performance between the sampled Gaussian derivative kernels and the approaches based on the integrated Gaussian derivative kernels, as well as the hybrid integrated Gaussian kernel, are statistically significant for both the parametric and non-parametric comparisons. With the effect size being large, this suggests that the statistical results strongly suggest a trend, where the sampled Gaussian derivative kernels perform better than the integrated Gaussian derivative kernels or the hybrid methods.

\subsubsection{Complementary remark}
Let us finally stress that numerical estimates of the degree of statistical
significance in pairwise comparisons between different discretisation
methods for the Gaussian derivative operators, that we have computed 
in this section, are not necessarily aimed at constituting quantitatively highly accurate statistical characterisations, due to the low number of runs (5 runs for the individual comparisons or $3 \times 5 = 15$ runs in the pooled comparisons). For such a purpose, more runs would be needed for each discretisation method, which would, however, imply a substantial amount of additional computational work, since each measurement involves training and testing of a new instance of a deep network.

Instead, the presented estimates are intended to reflect the order of magnitude in the statistical significance measures that are obtained from a selected set of statistical measures, to in more detail constitute a guide
concerning what differences in performance, for a set of, from a theoretical
background motivated, contrasts between discretisation methods, 
in the results summarised in Tables~6 and~8 in the main paper,
could be regarded as statistically significant.

{\footnotesize
\bibliographystyle{Frontiers-Harvard}
\bibliography{defs,tlmac}}

\end{document}